\documentclass[12pt]{cmuthesis}

\usepackage[authoryear]{natbib}
\usepackage{baskervald}
\usepackage[T1]{fontenc}
\PassOptionsToPackage{svgnames,dvipsnames}{xcolor}
\usepackage{soul} % for highlight
\usepackage{fullpage}
\usepackage{graphicx}
\definecolor{link_color}{RGB}{0,128,255}

\usepackage[%
colorlinks=true,allcolors=link_color,pageanchor=true,%
plainpages=false,pdfpagelabels,bookmarks,bookmarksnumbered,
pdfborder={0 0 0}%
]{hyperref}

\usepackage[letterpaper,twoside,vscale=.8,hscale=.75,nomarginpar,hmarginratio=1:1]{geometry}

\usepackage{graphicx} % more modern
\usepackage{subfigure}
\usepackage{todonotes}

\usepackage{wrapfig}
\definecolor{lightgray}{gray}{0.95} % 10%
\usepackage{hyperref}

\usepackage{algorithm}
\usepackage{algorithmic}
\usepackage{easytable}
\usepackage{dialogue} %For the dialogue in the last chapter

\usepackage{url}            % simple URL typesetting
\usepackage{booktabs}       % professional-quality tables
\usepackage{amsfonts}       % blackboard math symbols
\usepackage{nicefrac}       % compact symbols for 1/2, etc.
\usepackage{microtype}      % microtypography
\usepackage{amsmath}
\usepackage{amssymb}
\usepackage{amsthm}
\usepackage{xifthen} % for newcommands where some arguments can be empty
\usepackage{relsize} % to resize symbols
\usepackage{amsbsy}
\usepackage{xcolor} 	% for my todo command
\usepackage{upgreek}  % for upright greek symbols
\usepackage{thmtools}
\usepackage{thm-restate} % To restate some theorems
\usepackage{graphicx}
\usepackage{float}
\usepackage{mathtools} % For short intertexts
\usepackage{mathrsfs}
\usepackage[export]{adjustbox} %For trimming figures
\usepackage{thmtools}
\usepackage{thm-restate}
\usepackage{mathrsfs}
\usepackage[shortlabels]{enumitem}
\usepackage{array}
\newcolumntype{M}[1]{>{\centering\arraybackslash}m{#1}}
\newcolumntype{N}{@{}m{0pt}@{}}

 \usepackage[bb=boondox]{mathalfa}

\allowdisplaybreaks

\usepackage{etoolbox} % To change spacing

\makeatletter
\pretocmd{\chapter}{\addtocontents{toc}{\protect\addvspace{10\p@}}}{}{}
\pretocmd{\section}{\addtocontents{toc}{\protect\addvspace{3\p@}}}{}{}
\makeatother

\makeatletter
\def\thm@space@setup{%
  \thm@preskip=10pt
  \thm@postskip=\thm@preskip % or whatever, if you don't want them to be equal
}
\makeatother

\newcommand{\jacobian}[5]{\vecJ^{#3/#2}(#5; #1)\ifthenelse{\isempty{#4}}{}{[#4]}}
\newcommand{\compjacobian}[5]{\breve{\vecJ}^{#3/#2}(#5; #1)\ifthenelse{\isempty{#4}}{}{[#4]}}

\newcommand{\verbaljacobian}[2]{#2/#1}
\newcommand{\act}[3]{A_{\ifthenelse{\isempty{#1}}{}{\mathsmaller{#1},}#2}\ifthenelse{\isempty{#3}}{}{^{#3}}
}
\newcommand{\frob}[1]{\left\|{#1} \right\|_F}

\newcommand{\spec}[1]{\left\|{#1} \right\|_2}
\newcommand{\sq}[1]{ \left(#1\right)^2}

\newcommand{\Z}{\mathcal{Z}}

\newcommand{\focus}[2][yellow]{\mathpalette{\focuswithstyle[#1]}{#2}}
\newcommand{\focuswithstyle}[3][red!50]{
  \begingroup                         %% <- limit scope of \box0 and \fboxsep assignment
    \sbox0{$\mathsurround 0pt #2#3$}% %% <- typeset content in box 0
    \setlength{\fboxsep}{.5pt}        %% <- set (smaller) framebox margins
    \sbox2{\hspace{-.5pt}%            %% <- create box 2, undo margin
      \colorbox{#1}{\usebox0}%        %% <- print the contents of box 0 in a \colorbox
    }%
    \dp2=\dp0 \ht2=\ht0 \wd2=\wd0     %% <- set dimensions of box 2 to match box 0
    \box2                             %% <- print box 2
  \endgroup                           %% <- revert old definitions of the boxes and \fboxsep
}

\newcommand{\nn}[4]{ f\ifthenelse{\isempty{#2}}{}{^{#2}}\left(#4 {\ifthenelse{\isempty{#1}}{}{;#1}}\right)\ifthenelse{\isempty{#3}}{}{[#3]}}
\newcommand{\prenn}[4]{ g\ifthenelse{\isempty{#2}}{}{^{#2}}\left(#4; {{#1}}\right)\ifthenelse{\isempty{#3}}{}{[#3]}}
\newcommand{\compnn}[4]{ \breve{f}\ifthenelse{\isempty{#2}}{}{^{#2}}\left(#4; {{#1}}\right)\ifthenelse{\isempty{#3}}{}{[#3]}}
\newcommand{\compprenn}[4]{ \breve{g}\ifthenelse{\isempty{#2}}{}{^{#2}}\left(#4; {{#1}}\right)\ifthenelse{\isempty{#3}}{}{[#3]}}
\newcommand{\comppreprenn}[4]{ \breve{h}\ifthenelse{\isempty{#2}}{}{^{#2}}\left(#4; {{#1}}\right)\ifthenelse{\isempty{#3}}{}{[#3]}}

\newcommand{\classmargin}{\gamma_{\text{class}}}

\newcommand{\tolset}{\hat{\mathscr{C}}}
\newcommand{\tolspec}[2]{\hat{\psi}_{#2/#1}}
\newcommand{\toljacob}[2]{\hat{\zeta}_{#2/#1}}
\newcommand{\toloutput}[1]{\hat{\alpha}_{#1}}
\newcommand{\tolpreact}[1]{\hat{\gamma}_{#1}}
\newcommand{\toldelta}{\hat{\delta}}

\newcommand{\trainset}{{\mathscr{C}}^\star}
\newcommand{\trainjacob}[2]{{\zeta}^\star_{#2/#1}}
\newcommand{\trainspec}[2]{{\psi}^\star_{#2/#1}}
\newcommand{\trainoutput}[1]{{\alpha}^\star_{#1}}
\newcommand{\trainpreact}[1]{{\gamma}^\star_{#1}}
\newcommand{\trainmargin}{{\Delta}^\star}

\newcommand{\testset}{{\mathscr{C}}^\dagger}
\newcommand{\testjacob}[2]{{\zeta}^\dagger_{#2/#1}}
\newcommand{\testspec}[2]{{\psi}^\dagger_{#2/#1}}
\newcommand{\testoutput}[1]{{\alpha}^\dagger_{#1}}
\newcommand{\testpreact}[1]{{\gamma}^\dagger_{#1}}

\newcommand{\marginset}{\mathscr{C}^{\smalltriangleup}}
\newcommand{\marginjacob}[2]{{\zeta}^{\smalltriangleup}_{#2/#1}}
\newcommand{\marginspec}[2]{{\psi}^{\smalltriangleup}_{#2/#1}}
\newcommand{\marginoutput}[1]{{\alpha}^{\smalltriangleup}_{#1}}
\newcommand{\marginpreact}[1]{{\gamma}^{\smalltriangleup}_{#1}}

\newcommand{\tempset}{\mathscr{C}}
\newcommand{\tempspec}[2]{{\psi}_{#2/#1}}
\newcommand{\tempjacob}[2]{{\zeta}_{#2/#1}}
\newcommand{\tempoutput}[1]{{\alpha}_{#1}}
\newcommand{\temppreact}[1]{{\gamma}_{#1}}

\newcommand{\prev}[1]{#1_{\text{prev}}}

\newcommand{\boundednormevent}{{ \textrm{\textsc{\textsf{norm-bound}}}}}
\newcommand{\boundedperturbationevent}{{ \textrm{\textsc{\textsf{pert-bound}}}}}
\newcommand{\unchangedacts}{{ \textrm{\textsc{\textsf{unchanged-acts}}}}}
\newcommand{\nr}{\mathfrak{N}}

\newcommand{\boundspec}{\mathcal{B}_{\textrm{jac-spec}}}
\newcommand{\boundjacob}{\mathcal{B}_{\textrm{jac-row-}\ell_2}}
\newcommand{\boundoutput}{\mathcal{B}_{\textrm{layer-}\ell_2}}

\newcommand{\boundhiddenpreact}{\mathcal{B}_{\textrm{preact}}}
\newcommand{\boundoutputpreact}{\mathcal{B}_{\textrm{output}}}

\newcommand\numberthis{\addtocounter{equation}{1}\tag{\theequation}}

\makeatletter
\newcommand{\thickhline}{%
    \noalign {\ifnum 0=`}\fi \hrule height 1pt
    \futurelet \reserved@a \@xhline
}

\renewcommand{\vec}[1]{\boldsymbol{\mathbf{#1}}}
\newcommand{\N}{ \mathcal{N} }
\newcommand{\E}{ \mathbb{E} } % dont add argument to this bec sometimes we need subscript
\newcommand{\Esub}[2]{ \mathbb{E}_{#1}\left[ {#2}\right] } % dont add argument to this bec sometimes we need subscript
\newcommand{\pr}[1]{ \mathbb{P} \left[ {#1}\right] }
\newcommand{\prsub}[2]{ \mathbb{P}_{#1}\left[ {#2}\right] }
\newcommand{\relu}[1]{ \Phi_{\text{\tiny RELU}}\ifthenelse{\isempty{#1}}{}{\left(#1\right)}}

\newcommand{\U}{ \mathcal{U}}
\newcommand{\W}{\mathcal{W}}

\renewcommand{\omega}[1]{\Omega\left(#1\right)}
\newcommand{\bigoh}[1]{\mathcal{O}\left(#1\right)}

\newcommand{\ellone}[1]{\left|{#1} \right|}

\newcommand{\elltwo}[1]{\left\|{#1} \right\|}
\newcommand{\matrixnorm}[2]{\left\|{#1} \right\|_{#2}}

\newtheorem{theorem}{Theorem}[section]
\newtheorem*{theorem*}{Theorem}

\newtheorem{corollary}{Corollary}[theorem]
\newtheorem{lemma}[theorem]{Lemma}
\newtheorem{proposition}{Proposition}[section]
\newtheorem{fact}{Fact}[section]
\newtheorem{definition}{Definition}[section]

\newtheorem{remark}{Remark}[section]

\DeclareFontFamily{U} {MnSymbolC}{}

\DeclareFontShape{U}{MnSymbolC}{m}{n}{
  <-6> MnSymbolC5
  <6-7> MnSymbolC6
  <7-8> MnSymbolC7
  <8-9> MnSymbolC8
  <9-10> MnSymbolC9
  <10-12> MnSymbolC10
  <12-> MnSymbolC12}{}
\DeclareFontShape{U}{MnSymbolC}{b}{n}{
  <-6> MnSymbolC-Bold5
  <6-7> MnSymbolC-Bold6
  <7-8> MnSymbolC-Bold7
  <8-9> MnSymbolC-Bold8
  <9-10> MnSymbolC-Bold9
  <10-12> MnSymbolC-Bold10
  <12-> MnSymbolC-Bold12}{}

 \DeclareSymbolFont{MnSyC} {U} {MnSymbolC}{m}{n}

\DeclareMathSymbol{\smalltriangleup}{\mathrel}{MnSyC}{73}

\newcommand{\calA}{\mathcal{A}}
\newcommand{\calB}{\mathcal{B}}
\newcommand{\calC}{\mathcal{C}}
\newcommand{\calD}{\mathcal{D}}

\newcommand{\calF}{\mathcal{F}}

\newcommand{\calH}{\mathcal{H}}

\newcommand{\calR}{\mathcal{R}}
\newcommand{\calS}{\mathcal{S}}

\newcommand{\calU}{\mathcal{U}}

\newcommand{\calW}{\mathcal{W}}
\newcommand{\calX}{\mathcal{X}}
\newcommand{\calY}{\mathcal{Y}}
\newcommand{\calZ}{\mathcal{Z}}

\newcommand{\scrD}{\mathscr{D}}

\newcommand{\scrH}{\mathscr{H}}

\newcommand{\scrL}{\mathscr{L}}

\newcommand{\vecI}{\mathbf{I}}
\newcommand{\vecJ}{\mathbf{J}}

\newcommand{\vecU}{\mathbf{U}}

\newcommand{\vecW}{\mathbf{W}}
\newcommand{\vecX}{\mathbf{X}}
\newcommand{\vecY}{\mathbf{Y}}
\newcommand{\vecZ}{\mathbf{Z}}

\newcommand{\vecDelta}{\boldsymbol{\Delta}}

\newcommand{\vecrho}{\boldsymbol{\rho}}

\newcommand{\vecxi}{\boldsymbol{\xi}}

\newcommand{\vecq}{\mathbf{q}}

\newcommand{\vecu}{\mathbf{u}}

\newcommand{\vecw}{\mathbf{w}}
\newcommand{\vecx}{\mathbf{x}}

\newcommand{\tildeh}{\tilde{h}}

\newcommand{\removed}[1]{}

\newcommand{\argmax}[1]{\underset{#1}{\mathrm{argmax}} \:}

\newcommand{\testerr}{\texttt{TestErr}}
\newcommand{\disag}{\texttt{Dis}}
\defcitealias{nakkiran20distributional}{N\&B'20}

\newcommand{\D}{\scrD}

\begin {document} 
\frontmatter

%initialize page style, so contents come out right (see bot) -mjz
\pagestyle{empty}

\title{ %% {\it \huge Thesis Proposal}\\
{\bf Explaining generalization in deep learning: progress and fundamental limits}}
\author{\textbf{Vaishnavh Nagarajan}}
\date{August 2021}
\Year{2021}
\trnumber{CMU-CS-21-122}

\committee{
\begin{tabular}{rl}
& \\
J. Zico Kolter, Chair & \textit{Carnegie Mellon University} \\
Andrej Risteski & \textit{Carnegie Mellon University} \\
Ameet Talwalkar & \textit{Carnegie Mellon University} \\
Nathan Srebro & \textit{Toyota Technological Institute at Chicago} \\
\end{tabular}
}

\support{This research was sponsored by Robert Bosch GMBH award 0087016732-PCR; by National Science Foundation award: CCF1525971; by United States Air Force Research Laboratory awards FA87501720152 and FA87501720027; and by Defense Advanced Research Project Agency award N660011714036. 
}

\disclaimer{The views and conclusions contained in this document are those of the author and should not be interpreted as representing the official policies, either expressed or implied, of any sponsoring institution, the U.S. government or any other entity.}

% copyright notice generated automatically from Year and author.
% permission added if \permission{} given.

\keywords{machine learning theory, deep learning theory, generalization puzzle, overparameterization,  stochastic gradient descent, uniform convergence.}

\maketitle

\begin{dedication}
To all the students who had to discontinue their PhD because of toxic work environments, \\
and to all the kind and humble researchers who are striving to make academia a better place.
\end{dedication}

\pagestyle{plain} % for toc, was empty

%% Obviously, it's probably a good idea to break the various sections of your thesis
%% into different files and input them into this file...

\begin{abstract}
This dissertation studies a fundamental open challenge in deep learning theory: why do deep networks generalize well even while being overparameterized, unregularized and fitting the training data to zero error? 

In the first part of the thesis, we will empirically study how training deep networks via stochastic gradient descent implicitly controls the networks' capacity. Subsequently, to show how this leads to better generalization, we will derive {\em data-dependent} {\em uniform-convergence-based} generalization bounds with improved dependencies on the parameter count.

Uniform convergence has in fact been the most widely used tool in deep learning literature, thanks to its simplicity and generality. Given its popularity, in this thesis, we will also take a step back to identify the fundamental limits of uniform convergence as a tool to explain generalization. In particular, we will show that in some example overparameterized settings, {\em any} uniform convergence bound will provide only a vacuous generalization bound. 

With this realization in mind, in the last part of the thesis, we will change course and introduce an {\em empirical} technique to estimate generalization using unlabeled data. Our technique does not rely on any notion of uniform-convergece-based complexity and is remarkably precise. We will theoretically show why our technique enjoys such precision.

We will conclude by discussing how future work could explore novel ways to incorporate distributional assumptions in generalization bounds (such as in the form of unlabeled data) and explore other tools to derive bounds, perhaps by modifying uniform convergence or by developing completely new tools altogether.
\end{abstract}

% \begin{postabstract}

% \end{postabstract}

\begin{acknowledgments}

%TODO
% Reword
% Add more specific instances
% Reorder

%\vaishnavh{unrelenting, unyielding}

I've been unimaginably lucky (arguably to an unfair extent) to have enjoyed the support of many friends, colleagues, mentors and role models during my PhD. 
What follows is a heavily abridged account of the invaluable role they have all played in building this thesis and in building me as a researcher.   \\

Around the time I joined my advisor Zico Kolter's lab, like many other junior PhD students, I was filled with not just excitement but also self-doubt. Zico however welcomed me with optimism, and worked towards finding something that I enjoyed working on. When we started our first collaboration, I was amazed by how thrilling research can get when working with him. Needless to say, I left every meeting (which happened almost every other day) re-energized with ideas and looking forward to the next. I've fondly held on to the memories of this experience as they have helped me sail through the ups and downs of graduate school. Zico has also been instrumental in introducing me to the world of deep learning and helping me shape my theoretical interests in a way that is more mindful of practice. He has also invested a lot of time and effort in teaching me the art of organizing ideas and communicating them with clarity. Thanks to him, I have grown to take pleasure in not just communicating research but also in teaching and making technical ideas as accessible as I can. 
 
I cannot emphasize enough how much effort Zico has taken to set up a healthy work environment for me to thrive in. In each and every one of the interactions I have had with him, he has listened to my ideas with infinite patience before conveying his thoughts. He also placed his trust in me by giving me the space to stumble, learn and grow. There was never a point where I felt guilty about having made a mistake or taking my own pace to complete a goal. This has helped me nourish the creative and explorative side of me during my PhD. Importantly, Zico has also always been forthcoming in talking about his own current and past shortcomings and struggles. This has helped me in ignoring the nagging voice of the imposter within me.\\

This thesis would not have been possible without the valuable feedback, suggestions and enthusiasm from my committee members: Andrej Risteski, Nathan Srebro, and Ameet Talwalkar.

I remember meeting Andrej when he politely stopped by my poster at ICLR '19 to listen intently to whatever I was rambling on about. We met after the poster session, and the first thing I noticed was how approachable and humble he was --- CMU was lucky to have him join as faculty! I'm glad I have had the pleasure of having him on my committee. 

The first time I met Nati was at a workshop in Princeton in 2019, where he was kind enough to 
highlight some of the work from this thesis during a panel discussion. I'm grateful for those words of appreciation!

 I have also been incredibly fortunate to have found a mentor and collaborator in Ameet. Ameet's positive and encouraging attitude in meetings is something I've always looked forward to. I have also learned a lot from him and his students both in terms of research style and in terms of technical ideas from a field different from my dissertation. I also cannot thank Ameet enough for treating me like I was a part of his lab. He has spent a lot of time helping me with planning my career, and has often reached out {\em on his own} to check in on me. His advice and constant reassurance has helped me stay afloat during some of the stressful times in the last year of my PhD. \\

Next, I would like to thank all my other collaborators, who have broadened my approach towards picking problems, and have also made research a much livelier endeavor. My heartfelt thanks to 
Adarsh Prasad,
Anders Andreassen, 
Arun Sai Suggala,
Avrim Blum,
 Behnam Neyshabur, 
 Colin Raffel,
  Colin White,
 Christina Baek,
 Ellen Vitercik,
Gregory Plumb,
Hanie Sedghi,
 Ian Goodfellow,
 Jay Mohta,
 Jeffrey Li,  
Melrose Roderick,
Nina Balcan,
Thao Nguyen, and
 Yiding Jiang.
I also want to thank Nina Balcan and Avrim Blum for their well-designed and well-taught course on learning theory which I believe laid a strong foundation for all of my research at CMU.\\

Some of my most favorite memories from my PhD are from internships.  These internships were a refreshing and necessary break from my thesis research, giving me the space to explore new topics and new places before coming back to my thesis with a fresh perspective. In the next few paragraphs, I would like to thank my internship hosts for these opportunities. First, I would like to extend my deep gratitude to Colin Raffel and Ian Goodfellow for an enjoyable collaboration at Google Brain in 2018. It was an absolute pleasure working with Colin, and I want to thank him specifically for finding opportunities for subsequent brainstorming sessions and collaborations on bridging theory and practice.  I'd also like to thank Ian for inviting me to many one-on-one lunches during the internship, during which I learned profound insights from him about the field. I also learned many other miscellaneous ideas including a ten-minute tutorial on "measure theory for dummies" which I regret not writing down! In each of our meetings, I witnessed Ian's humility and curiousity which continue to inspire me. 

In the following year, I interned at Bosch Germany, and I'd like to thank David Reeb for
providing an opportunity to create many memories that I'm now nostalgic for. 

In the summer of 2020, I interned at Google and was hosted by Behnam Neyshabur who was crucial in making the internship productive and smooth even though it was remote. I've immensely benefitted from the many engaging meetings we've had. Behnam introduced me to problems that have since significantly shaped my research interests beyond this thesis. 
He was also always quick to grasp my ideas and provide his unique perspective on how they could be made more useful from a practical viewpoint.

Even beyond that internship, I'm indebted to Behnam in many other ways that I don't have the words to describe. My first interaction with him was when I had cold-emailed him during the fourth year of my PhD. Truth be told, I did not expect any response as at that point, I was an introverted student who barely knew other researchers in the theory community. Behnam, however, responded warmly and even agreed to meet with me one-on-one for lunch during NeurIPS 2018. To this day, I cannot forget the intellectually stimulating conversation we had about the subject of this dissertation during that meeting. Behnam left me inspired and also more confident about reaching out to other researchers.  Since  then, he has played a significant role in my PhD as a mentor, spending {\em tremendous} amounts of time helping me with my career. Everytime I need advice, I know I can reach out to him to get a prompt response (even when he is out hiking on some other planet). Behnam has also connected me with many other researchers and he is the main reason I feel at home in the deep learning theory research community. \\

Besides my collaborators, there are many other researchers with whom I've had fruitful technical discussions related to this thesis. Thanks to Jason Lee, Daniel Roy, Gintare Karolina Dziugaite, Jeffrey Negrea, Vitaly Feldman, Phil Long, Peter Bartlett, Tengyu Ma, Matus Telgarsky,  Aditi Raghunathan $\hdots$ --- I'm absolutely confident that I'm forgetting a lot of people in this list. Thanks to Jason for being kind enough to invite me for a one-on-one lunch at ICLR 2019. Since then, I've never hesitated to message him if I've had any questions about the most recent advances in deep learning research.   I also want to thank Matus for e-mailing me as an area chair a couple of years ago (when I was a reviewer). That was a starting point for me to feel comfortable about reaching out to him with many other technical questions about learning theory. My heartfelt thanks to him for his career-related advice and for his many warm e-mails checking in on me!

Many other researchers have taken the time and effort to share crucial advice along the way that helped me with my research career, especially in the last leg of my PhD. I want to extend my sincerest thanks to all of them: Jonathan Frankle, Sarath Chandar, Kunal Talwar, Suriya Gunasekar, Hanie Sedghi $\hdots$ (again, this is certainly an incomplete list). I want to express my gratitude to Hanie in particular for her advice and support in my last year of PhD. \\

Let me now rewind a bit further back to the past. Earlier in my PhD, I greatly benefitted from the advice of other researchers (especially students) who shared their own PhD experiences with me: Jing Xiang, Kirthevasan Kandaswamy, Nika Haghtalab, Alnur Ali, Manzil Zaheer, Lev Reyzin $\hdots$ I'm indebted to them for helping me navigate some of my toughest times as a junior PhD student, and 
for providing the moral support I needed to continue with my PhD.
 
 Especially, I cannot thank Jing enough for taking up the role of a senior student mentor in the second year of my PhD. Jing was the sole reason I had the confidence to continue with my PhD program at that point. She helped me rationally assess my options and find the right people and the right resources to empower me. I also continue to be inspired by the grit she has shown in her own academic life. She taught me the importance of finding student mentors, and also the importance of giving back to the student community. \\

All of my work has been through rigorous rounds of feedback and suggestions from the rest of LocusLab. For that, I'd like to especially thank 
 Alnur Ali,
 Brandon Amos, 
  Chun Kai Ling,
 Eric Wong,
 Ezra Winston,
  Filipe de Avila Belbute-Peres,
 Gaurav Manek,
 Jeremy Cohen,
 Josh Williams,
  Leslie Rice,
 Mel Roderick,
Po-Wei Wang, 
Priya Donti,
Rizal Fathony,
 Shaojie Bai, 
 Swaminathan Gurumurthy and Yiding Jiang.
Special thanks to Gaurav for spending a lot of time engineering and managing the cluster for us to run our experiments smoothly. 

Outside of LocusLab, I want to thank Ellen Vitercik, Colin White and Travis Dick for providing me support, feedback, and company during the first two years of my PhD. I have learned from Ellen's attention to detail and her ability to communicate clearly --- I  vividly remember and follow her presentation tips to this day (``objects on a slide must appear in a linear order rather than haphazardly!''). I also want to particularly thank Colin for providing moral support especially in my second year, and for all the subsequent encouragement he has given me over these years.  

I also want to thank my writing skills committee members --- David Woodruff, Pradeep Ravikumar, Ellen Vitercik --- and my speaking skills commitee members --- Danny Sleator, Tai-Sing Lee, Noam Brown --- for their feedback. 

My graduate school life would not have been as smooth as it was without the timely assistance of Deborah Cavlovich and Ann Stetser. Right from the day I received my admit to my final days at CMU when I've been scrambling to figure out my health insurance after graduation, Deb has swiftly resolved every issue that would have otherwise taken so much time away from my thesis work. Ann too has been prompt in booking rooms for countless practice talks and for organizing my trips to conferences. \\

This acknowledgment would not be complete without expressing my sincerest thanks to my 
 undergraduate thesis advisor, Balaraman Ravindran Sir for introducing me to research, and for encouraging me to apply to PhD programs. I've taken many of his courses in machine learning, which imbued me with a fascination for the role of learning in AI. He also introduced me to theoretical problems which later inspired me to pursue research in learning theory. Further down the memory lane, I want to thank KK Anand Sir who did a brilliant job at teaching me olympiad-level problem-solving in maths and physics, which was instrumental in honing my critical thinking skills.\\

 If you have come this far, thanks to you too! Remember to hydrate yourself. The acknowledgment is far from being over! \\

Moving on to my personal life, I have been incredibly privileged to be surrounded by a large group of thoughtful and talented friends in Pittsburgh who have literally become my family here:
Abhishek Ravi,
Abhishikta Pal,
Aditya Menon,
Ajay Pisat,
Anand Sankar,
Annesha Ganguly,
Anuva Kulkarni,
Archana Ravi,
Arnab Debnath,
Arushi Vyas,
Dipan Pal,
Deepanjana Gupta,
Devansh Zurale,
Hridya Ravimohan,
Harshad Shirwadkar,
Ishani Chatterjee,
Prithvi Shankar,
Purvasha Chakravarti,
Micah Corah,
Raksha Mahalinkam, 
Rithisha Padmanabh,
Satwik Kottur,
Saurabh Kadekodi,
Shounak Joshi,
Shweta Jain,
Siddharth Singh,
Srujana Rao,
Suvidha Menon,
Sudharshan Suresh and
Tushar Kusnur.

I've had innumerable philosophical debates and conversations with them which have helped me discover and shape my beliefs and values over these years. These conversations have also helped me articulate my thoughts with more clarity, which of course, has come in handy in research. My friends have also diligently attended all my practice talks and provided feedback that has been crucial in making my talks clearer and more accessible.

My friends have also always had their doors open (literally speaking) --- even during ungodly hours --- whenever I needed a break from my work or whenever I needed someone to pour my heart out to. I'm deeply indebted to Saurabh, Sidharth and Purvasha for the many, many times they invited me (or allowed me to invite myself) to their places especially during some of the roughest patches of my PhD.  I want to thank Saurabh for intoxicating me with gallons of spicy chai, for making me discover so much about myself through his cleverly-posed thought-provoking questions, and for always listening to me with unending enthusiasm. Saurabh is only second to my mother when it comes to providing unconditional positive affirmations about me and my work. Thanks to Siddharth for spending what might be thousands of hours in laboriously preparing extra-strong South Indian filter coffee for us to enjoy during our cryptic crossword sessions. I cherish the many uncanny intellectual similarities we had. Thanks to Abhishek for inviting me over for Only Connect sessions, for enriching my life with his eclectic sense of humor and for providing free pop culture education. Thanks to Tushar and Sudharshan for the jamming sessions,  and to Tushar in particular for the heart-to-heart conversations during tea walks around Squirrel Hill. I'm also grateful to Dipan, Arnab, Devdutta, Srujana, Aditya and Suvidha for graciously hosting many other memorable hangouts at their respective places.

The list is not over yet. Thanks to Arushi for spreading endless amounts of cheer and dance --- it's a mystery where she gets all that energy from. Thanks to Annesha and Anuva for keeping me well-fed with expertly baked cakes. Thanks to Ishani for an honest and open friendship where we could be both kids and adults at the same time. Thanks to Satwik for being an amazingly co-operative and responsible roommate. The discipline he has shown in his academic life has always inspired me. Thanks to Dipan for introducing me to all these friends in the first place, and for tirelessly taking us on so many road trips across the East Coast. Thanks to Abhishikta for helping me discover the courage in me to be myself. Thanks to Deepanjana for always affectionately taking care of me like family. Thanks to Deepanjana, Rithisha and Harshad for being wonderful hosts during my internship in the Bay Area. Thanks to Rithisha and Raksha for the many soul-stirring conversations we have had about the beauty of music, and to Rithisha in particular for encouraging me to pursue singing, which has become my refuge from stress. 
 Thanks to Archana for constantly challenging me to be a more socially conscious person, for caring for me, and for being relatable in so many ways. Thanks to Devdutta and Prithvi for taking the lead on organizing unforgettable annual trips across the US (which is no mean feat considering the number of people involved). Thanks to Srujana for always being ready to help others (and making everyone else in the group look selfish in comparison). \\

There are friends outside of this group too who have played equally major roles along the way. I want to begin by expressing my deepest gratitude Aditi Raghunathan for taking so much effort to keep in touch with me through spontaneous hour-long calls over the course of my PhD. I cannot emphasize enough how much her emotional and intellectual support and constructive feedback has helped me take important strategic decisions in my research. She was also helpful in connecting me with many other people in the field.

 Thanks to Octavio Mesner for his support and love, and for constantly cheering me up with his wit. Thanks to Priya Donti for providing insightful feedback on all of my work, and for always being there to talk to, and for all the funny side conversations on Slack during group meetings. Thanks to Surbhi Goel for always being ready to help me, and for baking a cake for my birthday even when she barely knew me. Thanks to Dhivya Eswaran for helping me strategize my advisor search process. Thanks to Rahul Ladhania for his delightful company and especially for his timely support when I had troubles with the Canadian visa process before NeurIPS '18. Thanks to Ashwini Pokle for making me feel at home in Pittsburgh by bringing Indian festivals, food and sweets to my doorstep. Special thanks to Akhilesh Godi, my friend from my undergraduate years, without whom I would not be where I am today. \\

Words can't adequately describe my profoundest gratitude to Ahmet Oguz Atli for his unwavering companionship over the last three years. He has brought me the much-needed level-headedness and maturity required to deal with the adversities of a PhD. This thesis would not have been possible without his efforts at meticulously and tirelessly taking care of me like I was a delicate house-plant.

I also want to thank my family -- my Mom, Dad, sister, brother-in-law and my grandparents --- for their blessings and love. I want to especially thank my parents for realizing the value of education and doing all they could to give us access to good education even in the face of monetary constraints. \\

As I end this section, I must remind the reader that this acknowledgment is far from being an exhaustive record of the ways in which many kind people have helped me. Each and every one of them has inspired me and I hope I can pay forward at least a tiny fraction of that kindness.

\end{acknowledgments}	
\setcounter{tocdepth}{1}
\tableofcontents
\listoffigures
\listoftables

\mainmatter

%% Double space document for easy review:
% \renewcommand{\baselinestretch}{1.66}\normalsize

% The other requirements Catherine has:
%
%  - avoid large margins.  She wants the thesis to use fewer pages, 
%    especially if it requires colour printing.
%
%  - The thesis should be formatted for double-sided printing.  This
%    means that all chapters, acknowledgements, table of contents, etc.
%    should start on odd numbered (right facing) pages.
%
%  - You need to use the department standard tech report title page.  I
%    have tried to ensure that the title page here conforms to this
%    standard.
%
%  - Use a nice serif font, such as Times Roman.  Sans serif looks bad.
%
% Other than that, just make it look good...

 %, and yet, in the deep learning world, models with millions of more parameters than training datapoints have achieved state-of-the-art test accuracy.
% Deep networks that generalize well are typically massively overparameterized i.e., they have millions of more parameters than training data.  This fact however flies in the face of wisdom from classical learning theory which suggests that such high model complexity is detrimental to generalization. % Explaining this apparent paradox has become a grand open challenge in deep learning theory over the last few years. 

\chapter{Introduction: The Generalization Puzzle}
\label{chap:intro}
\section[Deep learning]{A gentle introduction to deep learning}

It almost seems impossible that one can represent complex relationships between complex real-world variables
 using familiar mathematical functions. Yet, deep learning has proven that it is not only possible to {\em represent} such relationships, but also possible to automatically {\em learn} them.

Deep learning, or more broadly, Machine Learning (ML), is aimed at building machines that can use data to learn relationships between variables, such as say the variable ``the pixels of a photograph'' and the variable ``is there a cat in that photo?''. There are many flavors to how these learning problems are cast, but in this thesis we will focus on {\em supervised learning}. In the context of our example, we would supply to the machine a set $S = \{(\vecx_i, y_i) \}_{i=1}^{m}$ of example photos (the $\vecx$'s) labeled as cat or otherwise (the $y$'s, which in this case is either $+1$ or $-1$), with the assumption that all the examples are independently drawn from an underlying distribution $\scrD$. Based on the data $S$, the machine would produce a function (or typically, the {\em parameters} $\calW$ of a function $f_{\calW}$) that maps any given vector of pixels to a boolean value. The hope is that the learned function would have captured implicit patterns in the {\em training data} that {\em generalize} to most unseen {\em test data} drawn from the same distribution $\scrD$. If the model generalizes successfully, we can expect that it can accurately tell whether even new pictures of cats (or otherwise) drawn from $\scrD$ are indeed cats (or not). \\

The deep learning approach to this learning problem is defined by multiple key design choices. First, we model the function $f_{\calW}$ as a composition of many functions, each with its own set of parameters. Each {\em layer} in such a {\em a deep network} would take as input a vector of outputs from the previous layer and apply a parameterized transformation to produce a new vector of outputs. Notationally, we can think of this as $f_{\calW}(\vecx) := \vecW_D \Phi(\vecW_{D-1} \Phi(\hdots \vecW_2 \Phi(\vecW_1 \vecx)) )$ where $\calW = \{\vecW_d\}_{d=1}^{D}$ are parameters to be learned.
 Crucially, the transformations $\Phi$ must be {\em non-linear} --- a composition of linear transformations would boil down to a boring linear function but a composition of non-linear transformations could represent many interesting real-world functions that involve a hierarchy of representations. Indeed, in order to identify whether an image is that of a cat, it seems reasonable that the machine would first have to identify low-level, local features like the curves and edges in the photograph, and then identify more global features like shapes that emerge from those curves, before identifying which category the shape belongs to.
 %\footnote{This is just an informal motivating statement. Deep learning does not always work this way! \vaishnavh{fill this}}. 

The second key design choice in deep learning is to learn the parameters of the function via a simple greedy procedure: {\em gradient descent} (GD). We randomly initialize the parameters of the network and improve the parameters by taking a step against the gradient of some loss that is being minimized on average over datapoints in the training set (such as the cross-entropy loss):
\begin{equation}
 \calW \gets \calW - \frac{1}{m}\sum_{i=1}^{m} \nabla_{\calW} \scrL_{\text{ce}}(f_{\calW}(x_i),y_i).
\end{equation}

In {\em stochastic gradient descent} (SGD), in each step, the average of the loss is taken only over a smaller batch of the dataset. 

The third design choice that is often made is that of {\em overparameterization}, meaning that the total number of parameters (the dimensionality of $\calW$, which we will denote by $p$) in the model exceeds the number of training data points ($m$). Intuitively, greater the parameter count, the richer the set of functions that the network can realize, hopefully encompassing many real-world functions. \\

Together, these fundamental design choices --- and many other sophisticated improvements upon those --- have led deep learning to achieve state-of-the-art generalization, thereby revolutionizing the field of Artificial Intelligence (AI).
Deep networks have become the go-to blackbox approach to learn useful representations of the real-world that can be later used to solve many downstream tasks in problems like reinforcement learning, natural language processing and computer vision.

\section{The Generalization Puzzle}

Notwithstanding all the empirical successes of deep learning, there is a disturbing truth underlying it: we do not, in many ways, understand how deep learing works. How do we go about building an understanding of such a complex system? To do that, we need an abstract theoretical model that can explain some of the fundamental ways in which the system behaves in reality. For example, to understand the solar system, we need a theoretical model that can estimate lengths of shadows or the occurences of solar eclipses in a manner consistent with what is observed in reality. In our attempts to form a coherent theory of such a system, we would often find ourselves wrestling with some phenomena that run counter to existing theoretical intuition. By successfully reinventing the theory to accommodate such counter-intuitive phenomena, we can hope to discover profound insights into the workings of the complex system.\\

The Generalization Puzzle is one such counter-intuitive phenomenon that has taken center stage in {deep learning} theory. Existing intuition from classical learning
theory suggests that complex and massive deep networks should {\em not} generalize well at all in the first place. Formally, complexity is usually thought of in terms of the number of parameters. For example, if we let $\hat{\scrL}_{S}(f_{\calW}) := \frac{1}{m} \sum_{i=1}^{m} \mathbb{1}[f_{\calW}(x_i) \cdot y_i < 0]$ denote the error on the training set $S$, and ${\scrL}_{\scrD}(f_{\calW}) := \Esub{(x,y) \sim \scrD}{f_{\calW}(x) \cdot y < 0}$ denote the error on the test set and $\hat{\calW}$ the weights (i.e., the parameters) learned on $S$, a standard {\em generalization bound} would be of the form: with high probability over draws $S \sim \scrD^m$,

\begin{equation}
\scrL_{\scrD}(f_{\hat{\calW}}) \leq \hat{\scrL}_{S}(f_{\hat{\calW}}) + O\left(\sqrt{\frac{p}{m}}\right),
\end{equation}
where $p$ is the parameter count and $m$ is the number of training datapoints. Stated a bit differently, this tells us that the {\em generalization gap}, namely the difference between the test and training error is bounded by $\sqrt{\frac{p}{m}}$. Observe that when $p \geq m$, the bound becomes vacuous as all it tells us is the obvious fact that the gap cannot exceed $1$.\\

 Intuitively, when the model is overparameterized, and when we desire a function $f_{\calW}$ that fits the training data in that $f_{\calW}(x_i) = y_i$, we essentially have an underspecified system of equations for finding $\calW$. The set of all possible solutions to $f_{\calW}$ could include not only the ground truth function, but also functions that simply ``memorize'' the training data. Such functions would latch onto complex and obscure patterns that are specific to the training dataset and have poor test error on unseen data. To steer clear of such absurd functions, the traditional workaround is to explicitly control the model complexity. We could do this in one of many ways:

\begin{enumerate}
	\item Choose an underparameterized model to begin with.
	\item Add some kind of regularizer to the loss (e.g., $\frac{1}{m}\sum_{i=1}^{m} \scrL_{\text{ce}}(f_{\calW}(x_i),y_i) + \| \calW \|_2^2$) that biases gradient descent towards specific kinds of parameters (like ones with small norms).
	\item Avoid fitting the training set perfectly (e.g., by stopping gradient descent before the loss reaches a particular threshold).
	\end{enumerate}
These were some of the fundamental guiding principles for machine learning practice for a few decades. \\

Each of these principles, however, has been contradicted in deep learning. First, state-of-the-art deep network models are {\em massively} overparameterized, typically with millions of more parameters than training datapoints. In fact, larger models outdo smaller ones in terms of generalization. Furthermore, while these models are generally subject to different regularization techniques, even networks that are not explicitly regularized find reasonably good solutions. Finally, it is standard practice in deep learning to train the model zero error (which is often called as {\em interpolation} \footnote{Although interpolation was originally defined as achieving zero squared error loss, and not zero classification error \citep{belkin18kernel}.}), and even for many steps beyond that (which can further improve generalization)! This apparent paradox between reality and our existing intuition forms the crux of the Generalization Puzzle. To summarize the puzzle informally: 

\begin{quote}
{\em Why do deep networks generalize well even when being overparameterized, while not being explicitly regularized and while being trained to zero error and beyond?}
\end{quote}

As we will note later, there are also stronger and broader versions of this puzzle that can be posed for a variety of overparameterized models, and for noisy learning tasks.

\subsection{The history of the generalization puzzle}

The puzzle was brought to light much before the deep learning revolution more than two decades ago in \citet{bartlett1998sample} and \citet{breiman2018reflections} and then gained popularity a few years ago due to \citet{neyshabur2014search} and \citet{zhang17generalization}. \citet{zhang17generalization} in particular demonstrated a popular experimental illustration of this puzzle. Consider an overparameterized network trained via SGD to zero error on a dataset like CIFAR-10 \citep{krizhevsky2009learning}. One would observe this network to generalize well, i.e., its 0-1 error (the number of misclassifications) on the test set would be much smaller than a random classifier. Now train the same model from its random initialization on a {\em corruped} CIFAR-10 dataset where every training point is {\em randomly} assigned a label. One would observe that SGD is still able to find parameters that fit this meaningless dataset to zero error. Since there is no pattern whatsoever to the labels in this dataset, the network must clearly have had enough capacity to simply memorize those labels --- indeed such a network has performance equal to that of a random classifier on the test set. Why does the same network, trained via the same algorithm, somehow do something more meaningful on the original CIFAR-10 dataset?

It turns out that this sort of a puzzle is not too unique to deep learning, but similarly applies to overparameterized models at large \citep{hsu20research}. 
%\footnote{Although as we will discuss later there {\em are} interesting sub-questions that are unique to deep learning}.  
This includes boosting \citep{schapire97boosting}, kernel machines \citep{belkin18kernel} and even high-dimensional linear regression \citep{hastie19surprises}. However, the puzzle becomes much more challenging under the non-linearity and non-convexity of deep learning as we will later see.

\section[Direct and indirect approaches]{Two approaches to a theory of generalization} 
A key step towards resolving the Generalization Puzzle is the realization that the parameter-count-based analysis is ignorant of the training algorithm or the  dataset. We would want to perform an analysis of the model that cleverly incorporates properties of the data and the algorithm, or more specifically, {\em how the algorithm behaves on the data}. This idea has materialized into two different approaches.

\subsection{The Direct Approach}
\label{sec:direct-approach}
 One class of works consider the relatively tractable high-dimensional linear regression setting, where one can write down a closed-form solution to the parameters learned by gradient descent and {\em directly} analyze its error \citep{hastie19surprises,bartlett19benign,belkin20two,muthukumar20harmless,mei2020generalization,tsigler2020benign}. These analyses consider specific classes of distributions, typically sub-Gaussian in nature, and derive fairly precise bounds on the test error that usually depend on quantities like the $\ell_2$ norm of the solution and also properties of the distribution such as its covariance matrix. It is also worth noting that these studies often focus on a stronger form of the generalization puzzle --- why do overparameterized models generalize well even when there is noise in the labels? Furthermore, some recent follow-ups of these works have extended these ideas to the (noisy) linear max-margin classication setting again under sub-Gaussian assumptions on the data \citep{montanari2020generalization,wang21benignmulti,wang21benign,muthukumar20classification,cao21risk,chatterji20finite,deng20model}. Other works \citep{liang20multiple,ghorbani21linearized,li21towards} have further developed these results to apply to features that are produced by a neural network. However, these features are computed by a randomly initialized neural network rather than a trained network.

\subsection{The Indirect Approach}
\label{sec:indirect-approach}
A second distinct style of research --- the one this thesis will take up --- tackles this problem via a more abstract, indirect analysis which would apply to almost all distributions, and where the model is a deep network itself. Here, unlike in the linear settings above, the optimization problem is {\em non-convex}. As a result, it is hard to write down a neat, analyzable closed form expression for the parameters found by running gradient descent. There is only one viable alternative: literally run SGD on the deep network. Then the idea is to examine the function or the parameters learned for any ``desirable properties'' it may satisfy. Subsequently, if we can show that those ``desirable properties``  imply good generalization, our story is complete: \textit{the deep network generalizes because SGD happens to find weights satisfying certain ``desirable properties'' in practice}. 

What are these ``desirable properties''? Broadly, we must investigate how SGD training has {\em implicitly controlled} the representational capacity\footnote{We will use the term ``implicit bias'', ``implicit regularization'' and ``implicit capacity control'' interchangeably. Similarly, we will use the terms ``capacity'' and ``complexity'' interchangeably.} of the network in practice. For example, one might observe in the experiment of \citet{zhang17generalization} that when training on the noisy CIFAR-10 data, the $\ell_2$ norms of the network tend to be quite large, and when training on the clean CIFAR-10 data, the norms are relatively small. Hence, one could surmise that the SGD is biased towards smaller $\ell_2$ norm solutions when the data is simple.

Once we identify some sort of norm that is implicitly controlled by SGD in practice, the next step is to theoretically derive a generalization bound that takes advantage of this insight. Such a bound would quantify complexity via {\em norms} that are adaptively controlled by the algorithm depending on the data, rather than the rigid parameter count. Indeed, such norm-based bounds have been popular in the context of generalization bounds for Support Vector Machines (SVMs) \citep{cortes1995support}. 
 Consider an SVM with weight vector $\vecw$. 
 %For a threshold $\gamma > 0$, let $\hat{\scrL}^{(\gamma)}_S(\calW) := \frac{1}{m} \sum_{i=1}^{m} \mathbb{1}[f_{\vecw}(x_i) \cdot y < \gamma ]$ denote the {\em margin-based} error of the SVM i.e., if the datapoint is not classified by a sufficiently large margin, it is counted as a misclassification.  
 Then, very roughly\footnote{Note that this is a highly imprecise bound, but we will see the precise version of this in the next chapter.}, one can write bounds of the following form which captures the $\ell_2$ norm and does not involve the parameter count in any way (the SVM could even have infinitely many parameters!):
\begin{equation}
\scrL_{\scrD}(f_{\hat{\vecw}}) \leq \hat{\scrL}_{S}(f_{\hat{\vecw}}) + \sqrt{\frac{\|\hat{\vecw}\|^2}{m}}.
\end{equation}

Our hope is to identify similar types of bounds for SGD-trained neural networks. Such kinds of bounds are possible more generally through different types of {\em uniform convergence} based learning-theoretic tools, like Rademacher complexity and PAC-Bayes, that we will extensively discuss in the next chapter.  \\

In our search for such bounds, there are two substantial challenges we must brace ourselves for. First, we must empirically identify and enumerate many candidate notions of implicit capacity control (or implicit bias). There is no particular notion of what implicit bias could look like. Indeed, many creative notions have been explored, such as the sharpness of the loss landscape at the minima found \citep{keskar17largebatch,neyshabur17exploring}, redundancy of neurons \citep{morcos18importance} and spectral bias \citep{rahaman19spectral}. However, it is not clear which of these is the ``right'' notion that leads to good generalization.

 The second challenge is to translate any empirically identified implicit bias into a concrete theoretical bound. Since neural networks are complicated mathematical objects, this is  always a tricky endeavor. Most analyses end up with spurious dependencies on the parameter count besides the desired norm itself, rendering the bound as questionable as the classical bounds. Nevertheless, many interesting bounds have been proposed in the literature starting from \citet{bartlett1998sample} to the more recent ones like \citet{golowich17size,neyshabur2015norm,bartlett2017spectrally,neyshabur18pacbayes}. These bounds  depend on norms of the weight matrices, such as the Frobeniums norm and spectral norm.\\

The indirect approach to a theory of generalization often leads to simpler, insightful analyses compared to the direct one. However, the price we pay for this simplicity are bounds that are miles away from being precise (with certain exceptions like \citet{dziugaite17nonvacuous}). It is also worth noting that the indirect approach typically does not bother with answering {\em why} SGD leads to a particular form of capacity control --- that question is an independent one that is deferred to the optimization-theorist (although a few works such as \citep{li18learning,zhu18beyond} do handle both the generalization and the optimization aspects).

\section[Why generalization bounds?]{Why do we want a generalization bound?}
\label{sec:generalization-powers}

Before setting out to propose a generalization bound, let us reflect a bit on what might motivate us to do so. There are three possible reasons, and we may care about one or more of these:

\begin{enumerate}
\item \textbf{For its explanatory power}: we may be scientifically curious about {\em explaining} an empirical phenomenon. 
\item \textbf{For its predictive power}: we may want to predict how well our network would perform, perhaps in comparison to other networks.
\item \textbf{For its utilitarian power}: we may want to improve the training algorithm by drawing inspiration from the bound.
\end{enumerate}

It is important to consider these end goals as they would determine two key aspects of the bound:

\begin{itemize}
	\item \textbf{Information}: What information (about the trained model, the data, and the domain) can we utlize while deriving the bound?
	\item \textbf{Evaluation}: How do we measure the quality of the bound? 
\end{itemize}

Let us delve into the implications of each of the powers on the above two aspects.

\subsection{Explanatory power}

\paragraph{Information.} Consider a simple bound that can be derived by estimating the error on held-out data. Via the Hoeffding's inequality (See Lemma~\ref{lem:hoeffding-bound} from Chapter~\ref{chap:uc}), one can say that with high probability of $1-\delta$ over draws of a held-out dataset $T \sim \scrD^{m_{\text{ho}}}$ of $m_{\text{ho}}$ many datapoints,

\begin{equation}
  \scrL_\scrD({\hat{\calW}}) \leq \hat{\scrL}_{T}(\hat{\calW}) + O\left(\sqrt{\frac{1}{m_{ho}}}\right).
  \label{eq:held-out}
\end{equation}

 This bound would be as tight as any bound could get (and can be made arbitrarily tight by simply collecting more held-out data). Yet, this tells us nothing interesting about {\em why} the algorithm generalizes well, as all it tells us is that ``the learner generalizes well to unseen data because it generalizes well on held-out data''. This statement fails to resolve the crux of puzzle: how  did the learner have the foresight to perform well on held-out data that was not revealed during training? 

 Abstracting this a bit, it seems uncontroversial to suggest that \begin{quote}{\em in order to produce a valid explanation for why the learner is able to generalize well with whatever little information it had, the explainer should have access to no more information than the learner.}  \end{quote}
This philosophy has guided the line of work detailed in the indirect approach in Section~\ref{sec:indirect-approach}, where the bounds are computed purely based on the training data. \footnote{However, towards the end of the thesis we will question this philosophy!}

\paragraph{Evaluation.}  An ideal bound would be as {\em numerically} tight as possible, hopefully very close to the exact generalization gap.  Most abstract theoretical tools however are lax in how they deal with constants, and unfortunately, even slight leniency towards small multiplicative factors can render the bound numerically vacuous.  Besides, striving for such severe precision can come at the cost of insight and simplicity. 

Perhaps a more reasonable criterion could be to search for a bound that {\em parallels} the behavior of the generalization gap under varying conditions. For instance, even if the bound may be ridiculously large, does it remain non-increasing with the parameter count like the actual generalization gap? As the labels get noisier, the actual generalization gap widens in practice; does the bound too get larger with label noise? One could generate an endless list of such factors to vary: training set size, width, depth, the learning rate, batch size and so on. Not all of these factors however may be equally important when it comes to evaluating a bound for its insight. For instance, a bound that is unable to reflect the true training-set-size dependence of the generalization gap seems more fundamentally flawed than one that is unable to reflect minute variations in the gap due to variations in other hyperparameters. Nevertheless, the importance one assigns to these factors may have to change depending on whether and how the insight from the bound is eventually repurposed for designing new algorithms (i.e., its utilitarian power). As such the explanatory power of a bound in itself is arguably subjective.

\subsection{Predictive power}

\paragraph{Information.} When we care only about predictive power, we need not be as cautious about using information that was not available to the learner as we were when we desired explanatory power. In fact, the held-out data bound in Equation~\ref{eq:held-out} --- which was hopeless from an explanatory viewpoint --- is not an unreasonable option for prediction, given that it is quite precise. However, we would still want to do much better than this since (a) labeled data is expensive to gather and (b) even if we did gather labeled data, we would rather use that during training to improve the model. 

Other kinds of ``extra'' information may still be useful. For instance, 
 unlabeled held-out data (which is much easier to gather) that may not be easily integrated with the training pipeline, could potentially be incorporated in a predictive bound. Indeed, we will present such bounds in Chapter~\ref{chap:disagreement}.

\paragraph{Evaluation.} A consideration that becomes important when it comes to predictive power is the numerical value of the bound. Especially if we are working in a high-stakes application where we want to be sure about how good our model is, a numerically tight bound would be critical. But perhaps in other situations, where we only care about the \textit{relative} performance of different models (as was advocated in \citet{Jiang2020Fantastic}), the numerical value need not be as sacred.

\subsection{Utilitarian power}

\paragraph{Information.} For a utilitarian bound, there is nothing that stops us from providing extra information that wasn't available to the learner. But recall that we still want to leverage the bound (or any insight within it) to design a new, improved learning algorithm (e.g., this might involve using the bound as a regularizer during training). Therefore, the information that we should grant to the bound should be largely determined by the information that would be available to the ``improved learner''. So for example, if our new learner would have access to unlabeled data that our old learner did not, we should be comfortable providing extra unlabeled data to the bound.

\paragraph{Evaluation.} In this case we have a concrete evaluation criterion: the performance of the new learner. While it is possible that optimizing for this goal would also result in a bound that does well in terms of other criteria (such as being numerically small or paralleling the actual generalization gap), it is perhaps wise not to explicitly optimize for these criteria.

\section{Our contributions}

In \textbf{Part I} of this thesis, we will discuss multiple findings that culminate in a generalization bound for deep networks. As stated before, these results fit within the indirect framework of generalization theory (Section~\ref{sec:indirect-approach}). These chapters rely on the tool of {\em uniform-convergence} which is essentially a learning-theoretic tool that quantifies generalization in terms of complexity --- we discuss this in Chapter~\ref{chap:uc}. In more detail:

\begin{enumerate}
	 \item In Chapter~\ref{chap:dist-from-init}, we will provide empirical and theoretical arguments arguing that measures of complexity and implicit bias must {\em not} be agnostic to the random initialization of the deep network.
	 \item In the next few chapters we will derive a few fundamental results that will help us eventually derive a PAC-Bayesian bound for deep networks in Chapter~\ref{chap:deterministic-pacbayes}:
	 		\begin{enumerate}	
	 			\item PAC-Bayesian techniques typically give us bounds on a stochastic model, where the weights are random variables. Deep networks however are deterministic models. In order to translate the PAC-Bayesian bound to a deterministic network, we would have to formally bound the extent to which deep networks are resilient to parameter perturbations. In Chapter~\ref{chap:noise-resilience}, we provide an extensive analysis of the noise-resilience properties of deep networks.
	 			\item In Chapter~\ref{chap:derandomized}, we will derive a new and general technique to derandomize PAC-Bayesian bounds.
	 			\item Usual norm-based notions of complexity are agnostic to the training data. In Chapter~\ref{chap:datadependent-pacbayes} we will provide a recipe for using the above derandomization technique to derive bounds that involve {\em data-dependent} notions of complexity. 
	 		\end{enumerate}
	 \item In Chapter~\ref{chap:deterministic-pacbayes}, we will combine all the above findings into a data-dependent, deterministic PAC-Bayesian bound for neural networks. Our bound captures two notions of implicit bias --- distance from initialization and flatness of the loss landscape --- and unlike existing bounds, do not suffer from exponential dependence on the depth.\\
\end{enumerate}

In \textbf{Part II} of this thesis, we will take a step back and arrive at a pessimistic view of using uniform convergence to build theories of generalization in deep learning. We will then discuss some initial steps towards moving beyond uniform convergence that result in highly predictive bounds. In particular:

\begin{enumerate}
	\item In Chapter~\ref{chap:inc-norms}, we will report empirical observations showing that many existing norm-based measures of complexity (that arise from uniform convergence based analyses) fail to capture a trivial fact about generalization: that generalization improves with {\em training set size}.
	\item Motivated by the above failings, and by the fact that a tight uniform-convergence-based bound has so far remained elusive, in Chapter~\ref{chap:unif-conv}, we will show that there are settings where {\em any} uniform convergence bound provably fails to explain generalization due to overparameterization.   
	\item In the final part of the thesis, we will end with some optimism. In Chapter~\ref{chap:disagreement}, we will take a radically different approach towards {\em empirically} estimating generalization gap: by using unlabeled data and by using {\em disagreement} between models, rather than complexity. The estimate, as we will see, is remarkably precise in practice.
	\item In Chapter~\ref{chap:calibration}, we will theoretically explain {\em why} the empirical estimate has such remarkable precision, thus making it a theoretically-founded estimate. 
\end{enumerate}

In \textbf{Part III} of the thesis we will conclude the thesis by first providing a summary of our results. We will then discuss some philosophical aspects of what it means to explain generalization, especially in light of the approach in Chapter~\ref{chap:disagreement} which uses unlabeled data to provide an estimate.

\chapter{Preliminaries: Uniform Convergence}

\label{chap:uc}

\label{chap:prelim}

In the introduction, we had informally referred to different notions of complexity, such as ones based on the parameter count or norms. Formally, these notions of complexity stem from the idea of uniform convergence. In this chapter, we will describe this idea technically and discuss how it gives rise to different widely-used learning-theoretic tools like Rademacher complexity and PAC-Bayes.

\section{Notation}

\paragraph{Supervised learning.} Let $\calX$ denote an input space and $\calY$ denote a label space. Typically, we will assume $\calX = \mathbb{R}^N$. In the supervised learning setup, we are given a dataset $S = \{(\vecx_i, y_i) \}_{i=1}^{m} \in (\calX \times \calY)^m$ of $m$ labeled examples drawn independently and identically from an underlying distribution $\scrD$, i.e., $S \sim \scrD^m$. Let $(\vecx, y) \sim S$ denote uniform sampling from $S$. 

We use different notations for binary classification and K-class classification (for $K > 2$). In the case of binary classification, we have $\calY = \{ -1, +1 \}$. Let $\calF$ be a set of hypotheses/functions where each function $f: \calX \to \mathbb{R}$ maps the inputs to a real-value. 
The {\em margin} of $f$ at a datapoint $(\vecx,y)$ is denoted by \begin{equation}\Gamma(f(\vecx), y) \coloneqq f(\vecx) \cdot y.\end{equation}

 In the case of K-class classification ($K$ > 2), $\calY = \{0, 1, \hdots, K-1\}$ and each function $f: \calX \to \mathbb{R}^{K}$ maps to a $K$-dimensional real-valued vector. Note that the values output by $f$ are {\em logits} and not probabilities. We let $f(\vecx)[k]$ denote the output of $f$ for the $k$th class. Then, the {\em margin} of any $f \in \calF$ at a datapoint $(\vecx,y)$ is denoted by 

 \begin{equation}\Gamma(f(\vecx), y) \coloneqq f(\vecx)[y] - \max_{k \neq y} f(\vecx)[k].
 \end{equation}

\paragraph{Loss functions.} Using the above notion of margin, we can define the 0-1 error of $f$ at a datapoint $(\vecx, y)$ as
\begin{equation}
{\scrL}(f(\vecx),y) \coloneqq \mathbb{1}[\Gamma(f(\vecx), y) < 0].
\end{equation}

Note that at times, we will abuse $\scrL$ to denote any generic loss function---this will be clear from context.

We will encounter generalizations of the 0-1 error when we visit different learning-theoretic tools. One generalization is that of the {\em margin-based loss}. Here, we choose a margin threshold $\gamma \geq 0$ and count any point that is not classified by a margin of $\gamma$ or more as a misclassification:

\begin{equation}
{\scrL}^{(\gamma)}(f(\vecx),y) \coloneqq \mathbb{1}[\Gamma(f(\vecx), y) < \gamma].
\end{equation}

The other generalization of the 0-1 error is the {\em ramp loss}. This is less harsh than  the margin loss in that when the margin lies in $[0, \gamma]$, it is penalized ``proportionally'':

\begin{equation}
{\scrL}^{\text{ramp} (\gamma)}(f(\vecx),y) \coloneqq \begin{cases} 
1 & \text{if } \Gamma(f(\vecx), y) < 0 \\
1 - \frac{\Gamma(f(\vecx), y)}{\gamma} & \text{if } \Gamma(f(\vecx), y) \in [0,\gamma] \\
0 & \text{otherwise}
 \end{cases}.
\end{equation}

Note that both the ramp and margin-based losses upper bound the 0-1 error, and when 
 $\gamma$ is zero, they reduce to the 0-1 error. 

For any generic loss $\scrL$, with an abuse of notation, we will denote the empirical/training error/loss as:

\begin{equation}
\hat{\scrL}_S(f) \coloneqq  \frac{1}{m} \sum_{i=1}^{m} \scrL(f(\vecx), y),
\end{equation}

and the test/population error/loss as:

\begin{equation}
{\scrL}_{\scrD}(f) \coloneqq  \Esub{(\vecx,y) \sim \scrD}{\scrL(f(\vecx), y)}.
\end{equation}

\paragraph{Neural network.} We consider a feed-forward network of $D$ layers and $H$ hidden units in each hidden layer, that maps from $\calX = \mathbb{R}^{N}$.   We denote the parameters by $\mathcal{W} = \{   \vecW_1, \vecW_2, \hdots, \vecW_D \}$ and biases $\mathcal{B} = \{ \vec{b}_1, \vec{b}_2, \hdots, \vec{b}_D \}$ so that the function computed by the network can be denoted as 
\begin{equation}
f_{(\mathcal{W},\mathcal{B})}(\vec{x}) \coloneqq \vecW_D \Phi \left( \vecW_{D-1} \phi(\hdots \Phi(\vecW_1\vec{x} + \vec{b}_1) \hdots ) +  \vec{b}_{D-1} \right) +\vec{b}_D.
\end{equation}
Here, $\Phi$ is some non-linearity.  We will use $\relu{}$ to denote the RELU non-linearity i.e., $\relu{t} = t \cdot \mathbb{1}[t \geq 0]$.

While the above network has bias terms, for most of our discussion we will ignore the bias terms to keep the discussion simple. We will denote such a network by replacing the $(\mathcal{W},\mathcal{B})$ subscript with just $\mathcal{W}$:
\begin{equation}
f_{\calW}(\vecX) \coloneqq \vec\vecW_D \Phi(\vec\vecW_{D-1} \hdots \vec\vecW_2 \Phi(\vec\vecW_{1})).
\end{equation}
%  We will denote the output of the $k$th layer before applying the activation function as $f^{(k)}_{(\mathcal{W},\mathcal{B})}(\vec{x}) = \vecW_k \phi(f^{(k-1)}_{(\mathcal{W},\mathcal{B})}(\vec{x})) + \vec{b}_k$, with $f^{(0)}_{(\mathcal{W},\mathcal{B})}(\vec{x})=\vec{x}$.
Sometimes, to avoid clutter in the notation, we will specify $\W$ in the argument rather than the subscript, as $\nn{\W}{}{}{\vec{x}}$. \\

Note that when $D > 1$, $\vecW_1 \in \mathbb{R}^{H \times N}$.  For all $d < D$ and $d > 1$, we will assume that $\vecW_D \in \mathbb{R}^{H \times H}$. That is, every hidden layer has $H$ hidden units. As for the last layer, if the task is binary classification, we assume the network has only a single output and therefore, $\vec\vecW_D \in \mathbb{R}^{1 \times H}$. Else, $\vec\vecW_D \in \mathbb{R}^{K \times H}$. We will use the notation ${\calW}_d$ to denote the first $d$ weight matrices.  We denote the vector of weights input to the $h$th unit on the $d$th layer (which corresponds to the $h$th row in $\vecW_d$) as $\vec{w}^{d}_{h}$. \\

\paragraph{Hidden unit outputs.} We denote the $k$th output of the classifier by $\nn{\W}{}{k}{\vec{x}}$. For any input $\vec{x} \in \mathbb{R}^{N}$, we denote the function computed by the network on that input as  \begin{equation}
\nn{\W}{}{}{\vec{x}} = \vecW_D \relu{\vecW_{D-1} \hdots \relu{\vecW_1 \vec{x}}}.
\end{equation} For any $d = 1, \hdots, D-1$, we denote the output of the $d$th hidden layer after the activation by $\nn{\W}{d}{}{\vec{x}}$. We denote the corresponding pre-activation values for that layer by $\prenn{\W}{d}{}{\vec{x}}$. We denote the value of the $h$th hidden unit on the $d$th layer after and before the activation by $\nn{\W}{d}{h}{\vec{x}}$ and $\prenn{\W}{d}{h}{\vec{x}}$ respectively. Note that for the output layer $d=D$, these two values are equal as we assume only a linear activation. For $d=0$, we define $\nn{\W}{0}{}{\vec{x}}\coloneqq \vec{x}$.  As a result, we have the following recursions:

 \begin{align}
\nn{\W}{d}{}{\vec{x}} & = \Phi{\prenn{\W}{d}{}{\vec{x}}}, d=1,2,\hdots, D-1, \\
\nn{\W}{D}{}{\vec{x}} & = {\prenn{\W}{D}{}{\vec{x}}} = \nn{\W}{}{}{\vec{x}},  \\
\prenn{\W}{d}{}{\vec{x}} & = \vecW_d \nn{\W}{d-1}{}{\vec{x}}, \forall d=1,2,\hdots, D, \\
\prenn{\W}{d}{h}{\vec{x}} & = \vec{w}^d_{h} \cdot \nn{\W}{d-1	}{h}{\vec{x}}, \forall d=1,2,\hdots, D. 
 \end{align}

\paragraph{Layerwise Jacobians.} For layers $d', d$ such that $d' \leq d$, let us define $\jacobian{\W}{d'}{d}{}{\vec{x}}$ to be the Jacobian corresponding to the pre-activation values of layer $d$ with respect to the pre-activation values of layer $d'$ on an input $\vec{x}$. That is,

\begin{equation}
\jacobian{\W}{d'}{d}{}{\vec{x}}  \coloneqq \frac{\partial \prenn{\W}{d}{}{\vec{x}}}{\partial \prenn{\W}{d'}{}{\vec{x}}}.
\end{equation}

In the case of the RELU activation,  this corresponds to the product of the ``activated'' portion of the matrices $\vecW_{d'+1}, \vecW_{d'+2}, \hdots, \vecW_{d}$, where the weights corresponding to inactive inputs are zeroed out. In short, we will call this `Jacobian $\verbaljacobian{d'}{d}$'. Note that each row in this Jacobian corresponds to a unit on the $d$th layer, and each column corresponds to a unit on the $d'$th layer.

% We denote the value of the $h$th hidden unit on the $d$th layer before and after the activation by $\prenn{\W}{d}{h}{\vec{x}}$ and $\nn{\W}{d}{h}{\vec{x}}$ respectively. We define $\jacobian{\W}{d'}{d}{}{\vec{x}}:={\partial \prenn{\W}{d}{}{\vec{x}}}/{ \partial \prenn{\W}{d'}{}{\vec{x}}}$ to be the Jacobian of the pre-activations of layer $d$ with respect to the pre-activations of layer $d'$ for $d' \leq d$ (each row in this Jacobian corresponds to a unit in layer $d$). In short, we will call this, Jacobian $\verbaljacobian{d'}{d}$. Let $\Z$ denote the random initialization of the network.

\sloppy  \paragraph{Perturbation parameters.}
Typically, we will use $\U$ to denote a set of $D$ ``perturbation'' matrices $\vecU_1, \vecU_2, \hdots, \vecU_D$ that are added to the weights $\W$. We use $\W+\U$ to denote the entrywise addition of the perturbation matrices with the original matrices. We will use $\vecU_d$ to denote only the first $d$ of the randomly sampled weight matrices,  and $\W+\vecU_d$ to denote a network where the $d$ random matrices are added to the first $d$ weight matrices in $\W$. Note that $\W + \calU_0 = \W$. Thus, $\nn{\W+\calU_d}{}{}{\vec{x}}$ is the output of a network where the first $d$ weight matrices have been perturbed. In our analyses, we will also need to study a perturbed network where the hidden units are frozen to be at the activation state they were at before the perturbation; we will use the notation  $\W[+\calU_d]$ to denote the weights of such a network. We will denote the parameters of a random initialization of the network by $\Z = (\vecZ_1, \vecZ_2, \hdots, \vecZ_D)$.

\paragraph{Training data dependency.} 
To denote the function learned from the dataset $S$, we will use $\hat{f}$ (a generic function) or $f_{\hat{\calW}}$ (a neural network with weights $\hat{\calW}$). Sometimes, it might be important to emphasize the dependence of the learned network on the random variable (abbreviated as r.v.) $S$, therefore we might also use the notation $\hat{f}_S$ or $f_{\hat{\calW}_S}$.

\paragraph{Other basic quantities.} For any set $A$, we use $\text{Unif}(\{A\})$ to denote the uniform distribution over $A$. We use $\mathcal{N}(\mu,\sigma^2)$ to denote a Gaussian with standard deviation $\sigma$ and mean $\mu$.
 Given two distributions $P$ and $Q$, we let $KL(Q \| P)$ be the KL divergence between $Q$ and $P$. Sometimes, with an abuse of notation, if $X$ is a random variable with the distribution $P$, we will use $KL(X \| P)$ to denote $KL(Q \| P)$.   %We will use $\U \sim \mathcal{N}(0,\sigma^2)$ to denote parameters of a neural network whose entries are all sampled independently from a Gaussian.  Here, $\U$ consists of weight matrices $(\vecU_1, \vecU_2, \hdots, \vecU_D)$.

 For a vector, we will use $\| \cdot \|$ to denote its $\ell_2$ norm and  $\elltwo{\cdot}_{\infty}$  for the $\ell_\infty$  norm.  Let  $\spec{\cdot}, \frob{\cdot}, \matrixnorm{\cdot}{2,\infty}$ denote the spectral norm, Frobenius norm and maximum row $\ell_2$ norm of a matrix, respectively. We will use the notation $\|(\mathcal{W}, \mathcal{B}) - (\mathcal{Z}, \mathcal{C}) \|_F $ to denote $\sqrt{\sum_{k=1}^{d} \|\vecW_k - \vec{Z}_k \|_{F}^2 +  \|\vec{b}_k - \vec{c}_k \|^2 }$.   We use $\frob{\W}^2$ to denote $\sum_{d=1}^{D} \frob{\vecW_d}^2$.

For our statements regarding probability of events, we will use $\land$, $\vee$, and $\lnot$ to denote the intersection, union and complement of events (to disambiguate from the set operators).

In order to make the mathematical derivations easier to read, if we want to emphasize a term, say $x$, we write, $\focus{x}$. We use $\tilde{\mathcal{O}}(\cdot)$ and  $\tilde{\Omega}(\cdot)$ to hide logarithmic factors.

\section{Hold-out bounds}

Let us first revisit the rudimentary hold-out-data based bound from Equation~\ref{eq:held-out} in Chapter~\ref{chap:intro}, and then motivate our way towards more sophisticated uniform convergence bounds. 
Recall that the hold-out bound gives us an estimate of the test error based on the empirical performance of a network on a held-out dataset that was not used during training. We formalize this bound below: 

\begin{theorem}
\label{thm:hold-out}
\textbf{(Held-out-data-based generalization bound)}
Let $T \sim \scrD^{m_{\text{ho}}}$ be a dataset of $m^{\text{ho}}$ many datapoints drawn i.i.d from $\scrD$ and independent of the training dataset $S \sim \scrD^m$. 
Then, with probability at least $1-\delta$ over the draws of $T$:
\begin{equation}
\scrL_{\scrD}(\hat{f}_S) \leq \underbrace{\hat{\scrL}_{\focus{T}}(\hat{f}_S)}_{\text{hold-out error}} + \sqrt{\frac{1}{2m_{\text{ho}}} \ln \frac{2}{\delta}}.
\end{equation}
\end{theorem}

This result follows quite easily from a standard result known as Hoeffding's inequality. The inequality bounds the deviation of an average of bounded i.i.d variables from their expectation:

\begin{lemma}
\label{lem:hoeffding-bound} (\textbf{Hoeffding's inequality for bounded r.v.s})
For $i=1,2, \hdots, n$, let $X_i$ be independent r.v.s bounded in $[0,1]$, with mean $\mu_i$. Then, for all $t \geq 0$, we have,
\begin{equation}
\pr{\sum_{i=1}^{n} (X_i - \mu_i) \geq t} \leq \exp \left( - \frac{2t^2}{n} \right),
\end{equation}
or in other words, for $\delta \in (0,1]$,
\begin{equation}
\pr{\sum_{i=1}^{n} (X_i - \mu_i) \geq \sqrt{ \frac{n}{2} \ln \frac{1}{\delta}}}.
\end{equation}
An identical inequality holds for the symmetric event $\sum_{i=1}^{n} (X_i - \mu_i) \leq -t$.
\end{lemma}

We discuss the proof of the hold-out bound below.

\begin{proof} \textbf{(Proof of Theorem~\ref{thm:hold-out})}
To derive the hold-out bound from Hoeffding's inequality, {\em fix} a particular classifier $f$. Then we can treat the error of $f$ on any set of i.i.d datapoints as also i.i.d. Instantiating this claim for the particular dataset $T$, we can invoke Hoeffding's inequality with $X_i$ equal the r.v. $\scrL(f(\vecx_i),y_i)$ (where $f$ is fixed, but $(\vecx_i,y_i)$ is the $i$'th held-out datapoint that is a r.v. with distribution $\scrD$). By Hoeffding's inequality, we can say  the error of $f$ averaged over the dataset $T$ and the error of $f$ in expectation over $\scrD$ are close with high probability. Furthermore, this statement holds for any arbitrary $f$. Formally:

\begin{equation}
\label{eq:held-out-hoeffding}
\forall f \in \calF, \prsub{T \sim \scrD^{m_{\text{ho}}}}{\scrL_{\scrD}(f) - \hat{\scrL}_T(f) \leq \sqrt{\frac{1}{2m_{\text{ho}}} \ln \frac{2}{\delta}}} \geq 1- \delta.
\end{equation}

Finally, we can instantiate this statement by fixing $f$ to be the function $\hat{f}_S$ that was learned on the dataset $S$ to get the hold-out bound. 
\end{proof}

\subsection{From a hold-out bound to a training data bound: an (incorrect) attempt}

What if we want to say something about how close the test error of $\hat{f}_S$ is to its empirical error on the dataset $S$ that it was trained on? First, we can safely replace $T$ with $S$ in Eq~\ref{eq:held-out-hoeffding} to get:

\begin{equation}
\label{eq:held-out-hoeffding-S}
\forall f \in \calF, \prsub{S \sim \scrD^{m}}{\scrL_{\scrD}(f) - \hat{\scrL}_S(f) \leq \sqrt{\frac{1}{2m} \ln \frac{2}{\delta}}} \geq 1- \delta.
\end{equation}

Next, it might be quite tempting to instantiate the above claim for the specific hypothesis $\hat{f}_S$, and declare that the test and training error are close for $\hat{f}_S$ as,

\begin{equation}
\label{eq:wrong-hoeffding}
\text{Claim: } \text{For } f=\hat{f}_S, \prsub{S \sim \scrD^{m}}{\scrL_{\scrD}(f) - \hat{\scrL}_S(f) \leq \sqrt{\frac{1}{2m} \ln \frac{2}{\delta}}} \geq  1- \delta.  
\end{equation}\\

This claim is flawed, and it is important to internalize why this is flawed as it demonstrates the dangers of not being doubly careful while dealing with dependencies on the training set. Let us try to understand this flaw from two different perspectives:

\begin{enumerate}

 \item \textbf{The ``what's the randomness in the random variable?'' perspective:} Let us pay careful attention to the r.v.s involved and their distributions. Crucially, consider the point in our argument when we {\em fix} $f$ to be a particular $\hat{f}_S$. Is $S$ here a r.v.? No. We are implicitly fixing a particular $S$. More rigorously, Eq~\ref{eq:wrong-hoeffding} begins as ``$\forall$ $S$, for $f = \hat{f}_S, \hdots$'', and when we say ``$\forall$ $S$'', $S$ ceases to be a r.v.---it becomes a constant. When we subsequently invoke Hoeffding's inequality, we however treat $S$ as a r.v. with the distribution $\scrD^m$ within the probability term, thereby contradicting ourselves. Hence, our claim is incorrect.

This however does not mean that Hoeffding's inequality is inapplicable after fixing $S$. If we did invoke it correctly, we'd have that the $X_i$'s are all ``r.v.s'' that take a {\bf \em constant} value of $\scrL(\hat{f}_S(\vec{x}_i),y_i)$ (since neither $\hat{f}_S$ nor the $i$th point in $S$ are random). Then, the expectation of the ``r.v.''  $\frac{1}{m} \sum_i X_i = \hat{\scrL}_S(\hat{f}_S)$  would {\em not} be the test error $\scrL_\scrD(f)$ (as we would want) but would remain as the training error itself, $\hat{\scrL}_S(\hat{f}_S)$.

More formally, here is the ``correct'' version of Eq~\ref{eq:wrong-hoeffding}. Let $T \sim \{ S\}$ denote sampling of a r.v. $T$ that always takes the constant value $S$. Then, on invoking Hoeffding's, the only true statement we can conclude is

\begin{equation}
\label{eq:vacuous-hoeffding}
\forall \; S, \text{ for } f=\hat{f}_S, \prsub{T \sim \{ S\}}{\underbrace{\focus{\hat{\scrL}_T(f)}}_{\text{not test error!}} - \underbrace{\hat{\scrL}_T(f)}_{\text{train error}} \leq \sqrt{\frac{1}{2m} \ln \frac{2}{\delta}}} \geq 1- \delta,
\end{equation}

which is a pointless statement.\\

 Why didn't this issue crop up when we were dealing with the held-out dataset and fixed $f$ to be $\hat{f}_{S}$? There again, by fixing $f$ to be  $\hat{f}_{S}$, $S$ ceased to be a random variable. However, when we invoked Hoeffding's inequality and let $X_i$ be the r.v. $\scrL(\hat{f}_S(\vec{x}_i),y_i)$, $X_i$ was {\em not} a constant. There, $(\vec{x}_i, y_i)$ was still a r.v. with the distribution $\scrD$ since $T$ was independent of $S$ and was {\em not} fixed. Therefore the expectation of $\frac{1}{m} \sum_i X_i$ became the desired test error, $\scrL_{\scrD}(\hat{f}_S)$.\\

\item \textbf{The ``set of all bad datasets'' perspective:} For any $f$, let  $\calS_{\text{bad}}(f)$ denote a set of all ``bad'' datasets $S$ for $f$ in that the empirical error of $f$ on $S$ deviates a lot from the test error of $f$.  When we apply Hoeffding's Inequality to every $f$ in Equation~\ref{eq:held-out-hoeffding-S}, we learn that whatever the set of datasets $\calS_{\text{bad}}(f)$ may be, we can rest assured that it is a small set in that  $\prsub{S \sim \scrD^m}{S \in \calS_{\text{bad}}(f)} \leq \delta$. Unfortunately though, 
Hoeffding's Inequality parts with absolutely no information about {\em what} $\calS_{\text{bad}}(f)$ is, such as what datasets it may or may not contain. In particular, for $f = \hat{f}_S$, does
$S$ belong to $\calS_{\text{bad}}(f)$? We cannot tell. Yet, in Eq~\ref{eq:wrong-hoeffding}, we presumptuously concluded from Hoeffding's inequality that for as many as $1-\delta$ draws of $S$,  $S$ does not belong to $\calS_{\text{bad}}(\hat{f}_S)$ --- this was incorrect.\\

 In fact,  Hoeffding's inequality does not even deny the worst-case possibility that {\em for every $S$}, $S$ is a bad dataset for $\hat{f}_S$! Intuitively, this captures the fact that, without any other assumption, it is possible for $f$ to overfit to $S$ and generalize poorly. 
\end{enumerate}

\section[Uniform convergence]{Uniform convergence: a correct attempt at a training data bound}

The key issue in the above flawed proof was the fact that $\hat{f}_S$ and $S$ are not independent quantities.  To confront that dependency, we need something more mathematically sophisticated than Hoeffding's inequality. This is the precise sophistication that uniform convergence offers us.  In abstract terms, in a u.c. bound,  the ``$\forall f \in \calF$'' --- which was {\em outside} the probability term in Equation~\ref{eq:held-out-hoeffding-S} --- is squeezed {\em into} the probability term. We will shortly see how this rearrangement gives us the power to deal with the training-data-dependency.

We provide the abstract mathematical definition of u.c. below.  Later, we will demonstrate how this abstract definition can be materialized in terms of bounds that are based on the complexity of the hypothesis space $\calF$.  But for now, we'll delibrately avoid thinking about u.c. in terms of the complexity-based intuition, and instead focus on its mathematical structure.

\begin{definition}
\label{def:uc-abstract}
We say that $\epsilon: (\mathbb{N} \times (0,1]) \to \mathbb{R}$ is a uniform convergence bound for the hypothesis class $\calF$, loss function $\scrL$ and distribution $\scrD$, if for every 
$m > 0$ and $\delta > 0$ we have that:

\begin{equation}
\label{eq:uc-abstract}
\prsub{S \sim \scrD^m}{ \focus{\forall f \in \calF}, \; \scrL_{\scrD}(f) - \hat{\scrL}_{S}(f) \leq \epsilon(m, \delta)} \geq 1- \delta.
\end{equation}
\end{definition}

Here think of $\epsilon(m, \delta)$ as an abstract bound that takes the place of the $\sqrt{\frac{1}{m} \ln \frac{1}{\delta}}$ from the Hoeffding-based bound in Eq~\ref{eq:held-out-hoeffding-S}. 

As stated before, the main aspect to pay attention to here is the fact that the ``$\forall f \in \calF$'' which was {\em outside} the
probability term in Equation~\ref{eq:held-out-hoeffding-S} is now part of the probability term.
What this means is that on most draws of the dataset $S$, we can guarantee that the test error and the empirical error are close together simultaneously/{\em uniformly} for all $f \in \calF$. This guarantee is much more powerful than the guarantee provided by Hoeffding's inequality and allows us to straightforwardly conclude what we want: the test error and the {\em train} error are close on most draws of the dataset. Formally,

\begin{theorem}
\label{thm:uc-abstract}
Given a uniform convergence bound $\epsilon$ for a hypothesis class $\calF$ and distribution $\scrD$, and an algorithm that for any $S$ produces a hypothesis $\hat{f}_S \in \calF$, we have that with probability at least $1-\delta$ over the draws of $S \sim \scrD^m$:
\begin{equation}
\scrL_{\scrD}(\hat{f}_S) - \hat{\scrL}_{S}(\hat{f}_S) \leq \epsilon(m, \delta).
\end{equation}
\end{theorem}

\begin{proof}(\textbf{Proof of Theorem~\ref{thm:uc-abstract}})
From the u.c. bound, we have that with probability $1-\delta$ over the draws of $S \sim \scrD^m$, for every $f \in \calF$,
\begin{equation}
\scrL_{\scrD}({f}) - \hat{\scrL}_{S}({f}) \leq \epsilon(m, \delta).
\end{equation} 
Since $\hat{f}_S \in \calF$ for every $S$, we can fix $f = \hat{f}_S$ above to get the desired result.
\end{proof}

To better appreciate the power of u.c., let us revisit how u.c. fixes the shortcomings of Hoeffding's inequality:

\begin{enumerate}
	\item \textbf{The ``what's the randomness in the random variable?'' perspective:} In the incorrect Eq~\ref{eq:wrong-hoeffding}, when we fixed $f$ to be $\hat{f}_S$, we also fixed $S$ to be a particular dataset; and then we contradicted ourselves by treating $S$ as a r.v. with distribution $\scrD^m$. This sort of a contradiction would not arise under u.c. because, at the point where we fix $f$ to be $\hat{f}_S$, we've already drawn a particular dataset $S$ from $\scrD^m$.
	\item \textbf{The ``set of all bad datasets'' perspective:} This perspective perhaps offers a more insightful view. Recall that Hoeffding's inequality does not tell us anything about the set of all bad datasets $\calS_{\text{bad}}(f)$ for each $f$, besides guaranteeing that they are small. However, u.c. tells us something stronger: {\em across all $f$, $\calS_{\text{bad}}(f)$ is identical}. Let us denote the common set of bad datasets as just $\calS_{\text{bad}}$. 

	Now, let's say for some training set $S$, $S \in \calS_{\text{bad}}(\hat{f}_S)$, which means the training error and test error would be far apart for $f$. How often would we draw such ``bad training sets'' $S \in \calS_{\text{bad}}(\hat{f}_S)$? Since $\calS_{\text{bad}}(\hat{f}_S) = \calS_{\text{bad}}$, this is equivalent to asking how often we would draw an $S$ belonging to $\calS_{\text{bad}}$. By the u.c. guarantee, we know that $\prsub{S}{S \in \calS_{\text{bad}}} \leq \delta$. In other words, for most of the datasets $S$, $S \notin \calS_{\text{bad}}(\hat{f}_S)$, implying that the test and training errors will be close on those datasets. \\

	Recall that earlier we couldn't arrive at the above conclusion because Hoeffding's inequality tells us very little about these bad datasets, and we needed some extra ``assumption'' to refute the possiblity $S \in \calS_{\text{bad}}(\hat{f}_S)$. What is the extra assumption that u.c. makes? We will see how  certain notions of complexity of the hypothesis class $\calF$ can act as a u.c. bound; then, intuitively, under the assumption that $\calF$ is a ``simple'' class of functions, u.c. allows us to conclude that the model generalizes well. The condition that $\calF$ must be simple is the key condition that Hoeffding's inequality missed.

\end{enumerate}

\subsection{Other abstract versions of uniform convergence}

Before we wrap up this section, we make a few more remarks about other version of u.c..
\begin{enumerate}
	\item  A more general version of u.c. bounds allows the bound to depend on $f$ i.e., we would define $\epsilon: (\mathcal{F} \times \mathbb{N} \times (0,1]) \to \mathbb{R}$ so that the inequality becomes 
\begin{equation}
	\prsub{S \sim \scrD^m}{ {\forall f \in \calF}, \; \scrL_{\scrD}(f) - \hat{\scrL}_{S}(f) \leq \epsilon(\focus{f}, m, \delta)} \geq 1- \delta.
	\end{equation} 

	The previous discussion would apply to this  version of u.c. as well. We will see these kinds of bounds when we study PAC-Bayesian bounds.
	\item A further general version of u.c. bounds allows the bound to depend on both $f$ and the dataset $S$. These are often termed {\em data-dependent} bounds. The first part of our thesis will deal with these kinds of bounds extensively. Again note that the above discussion would apply to this  version of u.c. as well.
	\item  Note that in Definition~\ref{def:uc-abstract} we specifically defined uniform convergence in terms of a loss function $\scrL$ alongside a hypothesis class $\calF$. This specific definition was given for the sake of convenience. But for the rest of this document, uniform convergence should be more generally understood as any type of bound with the following form:
	\begin{quote}
	 {\em With high probability over the dataset draws, for every hypothesis in a hypothesis class, the difference between some quantity in expectation over a distribution and the same (or similar) quantity averaged over the dataset is bounded.}
	\end{quote}

	\item An orthogonal notion of u.c. deals with convergence over {\em all possible distributions}, simultaneously. These would be bounds of the form \begin{equation}\focus{\forall \scrD}, \prsub{S \sim \scrD^m}{ {\forall f \in \calF}, \; \scrL_{\scrD}(f) - \hat{\scrL}_{S}(f) \leq \epsilon(m, \delta)} \geq 1-\delta.\end{equation} Here, $\epsilon(m, \delta)$ acts as an upper bound not just for a single distribution, but a class of distributions. However, we will not study these kinds of bounds in this thesis, and when we refer to u.c., we will only refer to u.c. with respect to the hypothesis class.
\end{enumerate}

\section{Rademacher complexity}

One of the most popular types of uniform convergence bounds is the Rademacher complexity. The Rademacher complexity of a hypothesis class measures the ability of the class to fit noisy labels on a dataset. Below, we will define this notion for a generic class of functions $\calH$. Our final generalization bound will depend on the complexity of $\calF$ composed with the loss function, $\scrL$.

\begin{definition} (\textbf{Empirical Rademacher Complexity.})
Let $\calH$ be a class of functions $h: \calZ \to \mathbb{R}$. Let $\vecxi = (\xi_1, \xi_2, \hdots, \xi_m)$ be a random vector where each $\xi_i$ (called a {\em Rademacher variable})  is independently sampled from $\text{Unif}(\{ -1, +1\})$. Then, the empirical Rademacher complexity of $\calH$ with respect to a dataset $S \in Z^m$ is defined as\footnote{The Rademacher complexity as such is defined as the empirical Rademacher complexity in expectation over the draws of $S \sim \scrD^m$. But we will not require this definition for any of our forthcoming discussions. We will however use the term Rademacher complexity to loosely refer to empirical Rademacher complexity}:
\begin{equation}
 	\hat{\calR}_{S} \coloneqq \Esub{\vecxi}{ \sup_{h \in \calH} \frac{1}{m} \sum_{i=1}^{m} \xi_i h(z_i)}.
\end{equation}
\end{definition}

Observe that the Rademacher variables can be thought of as ``random labels'' on the dataset $S$ and the $sup_{h \in \calH}$ can be thought of as finding the hypothesis in $\calH$ that maximizes the correlation between the outputs of $h$ and the random labels. The more expressive the hypothesis class $\calH$ is, the larger the correlation that can be achieved, and hence larger the empirical Rademacher complexity. While this definition is abstract, hopefully, for any given class of models, such as say linear classifiers or neural networks, we can analytically simplify its Rademacher complexity in terms of different properties of that class, such as its maximum weight norm or parameter count.\\

Next, we present the central theorem in this section where we will use Rademacher complexity to bound empirical and test averages for $\calH$ uniformly. For proof we refer the reader to \citet{mohri12foundations}.

\begin{theorem}
\label{thm:rademacher-complexity}
Let $\calH$ be a class of functions $h: \calZ \to \mathbb{R}$ and let $\scrD$ be a distribution over $\calZ$. Let $S = \{z_1, z_2, \hdots, z_m \}$ denote a dataset sampled from $\scrD^m$. For any $\delta > 0$, with probability at least $1-\delta$ over the draws of $S \sim \scrD^m$, $\focus{\forall h \in \calH}$:
\begin{equation}
 \Esub{z \sim \scrD}{h(z)} \leq \frac{1}{m}\sum_{i=1}^{m} h(z_i) + 2\hat{\calR}_S(\calH) + \sqrt{\frac{\ln \frac{1}{\delta}}{2m}},
\end{equation}
and also 
\begin{equation}
\frac{1}{m}\sum_{i=1}^{m} h(z_i) \leq \Esub{z \sim \scrD}{h(z)} + 2\hat{\calR}_S(\calH) + 3\sqrt{\frac{\ln \frac{2}{\delta}}{2m}}.
\end{equation}
\end{theorem}

Observe that this is a uniform convergence bound since the bound holds simultaneously for every hypothesis in $\calH$, across most draws of the dataset. The insight this gives us is that, if $\calH$ was a simple class of functions, then it is guaranteed to generalize well (i.e., the test and training averages are going to be close). \\

When we apply Rademacher complexity to analyze classification error, we would have to analyze the complexity of a loss function $\scrL$ composed with a set of classifiers $\calF$. Below, we present {\em Talagrand's lemma}, which tells us how to deal with this situation under a smoothness assumption.

\begin{lemma}\textbf{(Talagrand's lemma.)}
\label{lem:talagrand-lemma}
Let $\Lambda: \mathbb{R} \to \mathbb{R}$ be an $\ell$-Lipschitz function in that for any $t ,t' \in \mathbb{R}$, $|\Lambda(t) - \Lambda(t')| \leq \ell |t-t'|$. Then,
\begin{equation}
\hat{\calR}_S(\Lambda \circ \calH) \leq \ell \cdot \hat{\calR}_S(\calH).
\end{equation}
\end{lemma}

The 0-1 error is unfortunately not a smooth function. However, recall that there are margin-based loss functions that are smooth. Below, we present a standard generalization bound on the 0-1 error that involves the margin-based loss, specifically, the margin the classifier achieves on the training data.

\begin{theorem}
\label{thm:rademacher-binary-class}
Let $\calF$ be a class of functions $f: \calX \to \mathbb{R}$. Let $\gamma > 0$ be a fixed margin threshold.  Let $\scrL$ be the 0-1 error and $\hat{\scrL}^{\focus{\text{ramp}(\gamma)}}$ be the ramp loss. Let $\scrD$ be a distribution over $(\calX \times \{-1, +1 \})^m$. For any $\delta > 0$, with probability at least $1-\delta$ over the draws of $S \sim \scrD^m$, for every $f \in \calF$,
\begin{equation}
 \scrL_{\scrD}(f) \leq \hat{\scrL}^{\focus{\text{ramp}(\gamma)}}_S(f) + \frac{2}{\focus{\gamma}} \hat{\calR}_S(\calF) + \sqrt{\frac{\ln \frac{1}{\delta}}{2m}}.
\end{equation}
\end{theorem}

Let us parse this (uniform convergence) upper bound. The first term in the right is a ramp-loss based training error term. This term is an upper bound on the 0-1 training error. Larger our initially chosen value of $\gamma$, the larger will the training error term be. The next quantity is the Rademacher complexity, but this term grows inversely with $\gamma$. Hence, for larger values of $\gamma$, this quantity gets smaller. 
 To get a smaller Rademacher term, we would want to substitute a large value of $\gamma$, but doing so might potentially hurt the training error term if the training data was not classified by a large margin. Thus, overall the bound conveys that if our classifier fits the training data by a large margin, then it should generalize well.

We provide only a quick proof sketch for this result and direct the reader to \citet{mohri12foundations} for the full proof.
\begin{proof}(\textbf{Proof sketch for Theorem~\ref{thm:rademacher-binary-class}})
The result follows by first applying Theorem~\ref{thm:rademacher-complexity} on the hypothesis class corresponding to $\calF$ over which the ramp loss $\scrL^{\text{ramp}(\gamma)}$ is composed. This would result in a Rademacher complexity term involving the composed function class. We can then invoke Talagrand's lemma and use the fact that the ramp loss is $1/\gamma$-Lipschitz with respect to the margin of the classifier. This would result in the desired upper bound but on the test ramp loss $\scrL_{\scrD}^{\text{ramp}(\gamma)}$. However, by noting that the test ramp loss is an upper bound on the test 0-1 error, we get the final statement.
\end{proof}

\section{PAC-Bayes}
\label{sec:prelim-pac-bayes}

The PAC-Bayesian framework \citep{allester99pacbayes,mcallester03simplified,langford02pacbayes,mcallester99model} allows us to derive generalization bounds for a {\em stochastic} classifier. Specifically, let $\tilde{\W}$ be a random variable in the parameter space. The loss of a stochastic classifier $f_{\tilde{\W}}$ at an datapoint $(\vecx,y)$ is given by $\Esub{\W \sim \tilde{\calW}}{\scrL(f_{\W}(\vecx), y)}$. With an abuse of notation, we will denote the training loss of a stochastic classifier as:

\begin{equation}
\hat{\scrL}_S(f_{\tilde{\W}}) \coloneqq \frac{1}{m} \sum_{i=1}^{m} \Esub{\W \sim \tilde{\W}}{\scrL(f_{\W}(\vecx_i), y_i)},
\end{equation}
and its test loss as,
\begin{equation}
\hat{\scrL}_S(f_{\tilde{\W}}) \coloneqq \Esub{(\vecx,y)\sim\scrD}{\Esub{\W \sim \tilde{\W}}{\scrL(f_{\W}(\vecx), y)}}.
\end{equation}

In order to derive a PAC-Bayesian bound, we must first fix a ``prior'' distribution $P$ in the parameter space, {\em chosen independent of the training data}. The PAC-Bayesian framework then yields a generalization bound for every stochastic classifiers $\tilde{\W}$ in terms of the KL-divergence between $\W$ and $P$. This captures the fact that if  we learn a stochastic classifier that is close to the data-independent $P$, it means the classifier is only ``weakly dependent'' on the dataset $S$. Such a classifier would not rely on overly specific details in $S$ and should therefore
generalize well. \\

We state the formal PAC-Bayesian theorem below.

\begin{theorem}
\label{thm:basic-pac-bayes}
Let $\calF$ be a class of binary or multiclass classifiers parameterized by $\W$. Let $\scrL$ be {\em any} loss function that is  bounded in that it maps to $[0,1]$. Let $\scrD$ be a distribution over $(\calX \times \calY)^m$. Let $P$ be any fixed distribution in the parameter space. For any $\delta > 0$, with probability at least $1-\delta$ over the draws of $S \sim \scrD^m$, \hl{for every possible stochastic classifier} $\tilde{\W}$,

\begin{equation}
KL( \hat{\scrL}_S(f_{\tilde{\W}})\| {\scrL}_{\scrD}(f_{\tilde{\W}})) \leq {\frac{2 \left(KL(\tilde{\W} \| P) + \ln \frac{2m}{\delta} \right)}{m-1}}, 
\end{equation}
as a corollary of which,
\begin{equation}
{\scrL}_{\scrD}(f_{\tilde{\W}}) \leq \hat{\scrL}_S(f_{\tilde{\W}}) + 2\sqrt{\frac{2 \left(KL(\tilde{\W} \| P) + \ln \frac{2m}{\delta} \right)}{m-1}}.
\end{equation}
\end{theorem}

Again, note that this is a uniform convergence bound in that, the bound simultaneously holds for every possible stochastic classifier, across the $1-\delta$ draws of the training data.\\

How do we use a PAC-Bayesian bound to say something about deterministic classifiers that we typically use in practice? 
The following {\em derandomized} bound on the deterministic classifier derives from the above bound (although, we will derive a new derandomized bound in this thesis). Similar to the margin-based Rademacher complexity bound, this bound too relies on the margin of the classifier. This was proven in \citet{neyshabur18pacbayes} (and we direct the reader to \citet{neyshabur18pacbayes} for the proof). Less general versions originally proven in \citet{langford02pacbayes,mcallester03simplified}. 

\begin{theorem}
\label{thm:basic-derandomized}
Let $\calF$ be a class of binary or multiclass classifiers parameterized by $\W$. Let $\gamma > 0$ be any margin threshold. Let $\scrL$ be the 0-1 error and $\scrL^{(\gamma)}$ be the margin loss. Let $\scrD$ be a distribution over $(\calX \times \calY)^m$.  Let $P$ be any fixed distribution in the parameter space. For any $\delta > 0$, with probability at least $1-\delta$ over the draws of $S \sim \scrD^m$, for every parameter $f_\W \in \calF$, and for any random {\em perturbation} parameters $\U$ such that 
\begin{equation}
\prsub{\U}{ {\max_{\vecx \in \calX} |f_{\W}(\vecx) - f_{\W \focus{+\U}}(\vecx)|_{\infty}} \leq \frac{\gamma}{4}},
\end{equation}
we have:
\begin{equation}
 \scrL(f_{\W}) \leq \hat{L}_{S}^{(\gamma)}(f_{\W}) + 4 \sqrt{\frac{KL(\W + \U \| P) + \ln \frac{6m}{\delta}}{m-1}}.
\end{equation}
\end{theorem}

In words, this bound requires us to take the deterministic classifier given by $\W$ and create a stochastic classifier from it by adding a random perturbation $\U$ to the weights. As long as the random perturbation is minute enough not to perturb the margin of the original classifier, then one can derive a generalization bound on the deterministic classifier in terms of the KL-divergence between the stochastic parameters $\W + \U$ (where $\W$ is fixed) and the prior $P$. 

Typically, when we can afford a $\U$ with a large variance, the KL-divergence term becomes smaller. For example, this is the case when $\U$ is Gaussian noise (which is also often the typical choice for the perturbation). Therefore, intuitively, the theorem suggests that if the classifier achieves a large margin on the training set and if the classifier is highly {\em noise-resilient}, then it generalizes well.

\section{Uniform convergence and the generalization puzzle}

Let us now reframe the generalization puzzle in deep learning in terms of the formal notions of uniform convergence that we have seen in this chapter.
For the sake of this discussion, let us focus on binary classification. Let $\calF$ denote the set of all functions that can be represented by an overparameterized deep network of a particular architecture. By overparameterization, recall that we mean the number of parameters in the model is greater than the number of training datapoints, $m$.
Now, the standard sense in which u.c. bounds are computed in deep learning would be to apply u.c. on $\calF$. By ``apply u.c. on $\calF$'' we mean, the bound would hold simultaneously for every $f \in \calF$ for most draws of the dataset $S$. The resulting bound would depend on some notion of complexity of the whole hypothesis class $\calF$. The claim that was either directly or indirectly made in \citet{zhang17generalization,neyshabur2014search,bartlett1998sample,breiman2018reflections} is that such an application of u.c. would only yield vacuous bounds: 

\begin{proposition}
Any uniform convergence bound that applies to the whole hypothesis class $\calF$ representable by an overparameterized model would be vacuous.
\end{proposition}

\begin{proof}(\textbf{Sketch})
In the case of VC-dimension (which is also based on uniform convergence), this follows from the fact that the VC-dimension of $\calF$ is as large as the parameter count. Since VC-dimension bounds the gap between test and train error via $\sqrt{\frac{\text{VC-dim}(\calF)}{m}}$, this quantity becomes larger than $1$, which is a vacuous statement.

Let us illustrate this for Rademacher complexity bounds. From Theorem~\ref{thm:rademacher-binary-class}, we roughly have
that w.h.p. over $S$, for all $f \in \calF$,

\begin{equation}
	 \scrL_{\scrD}(f) \leq \hat{\scrL}^{{\text{ramp}(\gamma)}}_S(f) + \frac{2}{{\gamma}} \underbrace{\Esub{\vecxi}{ \sup_{f \in \calF} \frac{1}{m} \sum_{i=1}^{m} \xi_i f(z_i)}}_{\hat{\calR}_S(\calF)}.
\end{equation} 

How large can $\hat{\calR}_S(\calF)$  get for a massively overparameterized neural network? Recall that the Rademacher complexity measures the ability of the network to fit random labels on the dataset $S$. As \citet{zhang17generalization} showed, overparameterized networks can be trained simply via gradient descent to fit a dataset with any set of labels, however noisy they are. In fact, by sufficiently overparameterizing the network, one can express any function to arbitrary precision \citep{hornik89multilayer}. More specifically, we can claim that for any given $\gamma$, and for a sufficiently overparameterized model, for every possible $\vecxi$, there exists a hypothesis $f \in \calF$ such that for every $i$, $f(\vecx_i) = 0.5 \gamma \cdot \xi_i$. Subsequently, the Rademacher complexity boils down to a value of $1$ thereby rendering the bound vacuous. 
\end{proof}

As we discuss in Section~\ref{sec:indirect-approach}, the reason for the failure of this kind of uniform convergence is the fact that it does not incorporate any information about how the algorithm works on the particular data distribution. A more effective idea would be {\em \bf algorithm-dependent} uniform convergence.  Formally, for a given learning algorithm $\calA$, assume we know a small class $\calF_\calA \subseteq \calF$ such that across most draws of $S$, $\hat{f}_S$ is picked only from $\calF_\calA$. Then, we could get away with applying uniform convergence on only $\calF_\calA$ and get a valid generalization bound (this directly follows from Theorem~\ref{thm:uc-abstract}). Such a generalization bound would depend on the complexity of $\calF_\calA$, potentially via norms and other properties that are controlled by $\calA$.

For instance, for a given distribution $\scrD$ (say CIFAR-10), suppose we empircally find that SGD is implicitly biased towards networks where the spectral norms of the weight matrices satisfies $\| \vec\vecW_d\|_2 \leq 10$. Hypothetically, this could be a form of empirically observed implicit bias of SGD. We can then develop u.c. bounds that are applied to networks with bounded spectral norms e.g., for any constant $\beta$, let $\calF_{\|\cdot\| \leq \beta}$ be the set of all possible functions that can be represented by the network with weight matrices of spectral norm at most $\beta$. By applying u.c. on $\calF_{\|\cdot\| \leq \beta}$, we could hope to get a generalization bound that depends on the spectral norm quantity $\beta$. Hopefully, the empirically observed upper bound on the spectral norm is small enough that when we substitute that into the bound, the bound is also sufficiently small.

\subsection{Spectrally normalized margin bounds}
\label{sec:prelim-spec}
Indeed, some of the most popular generalization bounds for deep networks are based on spectral norms. Let $D$ denote the 
depth of the deep network and $H$ the width  (i.e., number hidden units in each layer). Ignoring log factors, the bound in \citet{bartlett17spectral} can be written as follows. For some margin threshold $\gamma > 0$, 
\begin{align}
\scrL_{\scrD}(f_{\W}) & \leq  \hat{\scrL}^{\text{ramp}(\gamma)} + \tilde{\mathcal{O}}\left( \frac{BD\sqrt{H}}{\gamma\sqrt{m}}\prod_{d=1}^{D} \| \vecW_d\|_2 \times  \frac{1}{D\sqrt{H}} \left( {\sum_{d=1}^{D}\left(\frac{\|\vecW_d \|_{2,1}}{\|\vecW_d\|_2}\right)^{2/3}}\right)^{3/2} \right), \\
\intertext{and the bound in \citet{neyshabur18pacbayes} can be written as:}
\scrL_{\scrD}(f_{\W}) &\leq \hat{\scrL}^{(\gamma)} + \tilde{\mathcal{O}}\left( \frac{BD\sqrt{H}}{\gamma\sqrt{m}}\prod_{d=1}^{D} \| \vecW_d\|_2 \times \sqrt{\sum_{d=1}^{D}\frac{\|\vecW_d \|_F^2}{\|\vecW_d\|^2_2}} \right),
\end{align}

In these margin-based bounds, the margin is said to be {\em normalized}. An unnormalized margin bound would look something like this:
\begin{equation}
\scrL_{\scrD}(f_{\W}) \leq \hat{\scrL}^{(\gamma)} + \tilde{\mathcal{O}}\left( \frac{BD\sqrt{H}}{\gamma\sqrt{m}} \right),
\end{equation}

The normalization however is critical to get a valid bound. To see why, consider a set of weights $\calW$ such that on all training data, the margin of the network satisfies $\Gamma(f_{\calW}(\vec{x_i}), y_i) \geq \gamma^\star_i$. Now imagine scaling the top layer weights of $\calW$ by some constant $\alpha > 0$, resulting in new weights $\calW'$. This network would have the same classification error but its margins would satisfy $\Gamma(f_{\calW'}(\vec{x_i}), y_i) \geq \alpha \cdot \gamma^\star_i$. By plugging in $\gamma = \alpha \gamma^\star$, the unnormalized bound for the new network would be:
\begin{equation}
\scrL_{\scrD}(f_{\W'}) \leq \tilde{\mathcal{O}}\left( \frac{BD\sqrt{H}}{\alpha \gamma^\star \sqrt{m}} \right),
\end{equation}

since $\hat{\scrL}^{(\alpha \gamma^\star)} = 0$.  Now, observe that we can arbitrarily increase $\alpha$ to a large value to completely kill the right hand side. This is problematic: rescaling the weights does not affect the classification error of the network, and yet we are able to claim perfect generalization simply by rescaling. Indeed, the unnormalized bound is invalid. The normalized bound however is valid, and does not run into pathological rescaling issues.

\section{Useful Lemmas}

We wrap this chapter up with a few useful standard lemmas. In this section, we state some standard results we will use in our proofs. We first define some constants: $c_1\coloneqq1/2048$, $c_2 \coloneqq\sqrt{15/16}$ and $c_3 \coloneqq \sqrt{17/16}$ and $c_4 \coloneqq\sqrt{2}$.\\

The first few results deal with the concentration of Gaussian random variables. 
We begin with a statement of the Hoeffding bound for sub-Gaussian random variable. 

\begin{lemma}
\label{lem:uc-hoeffding}
\label{lem:hoeffding-gaussian} 
Let $X_1, X_2, \hdots, X_n$ be independently drawn sub-Gaussian variables with mean $0$ and sub-gaussian parameter $\sigma_i$. Then,
\begin{equation}
\pr{\sum_{i=1}^{n} X_i \geq t } \leq \exp\left(-\frac{t^2}{2\sum_{i=1}^{n} \sigma_i^2}\right).
\end{equation}
An identical inequality holds good symmetrically for the event $\sum_{i=1}^{n} X_i \leq -t $.
\end{lemma}

As a result of this we have the following inequality on the dot product of a Gaussian vector with another vector. 

\begin{corollary}
\label{cor:gaussian}
For any $\vec{u} = (u_1, u_2, \hdots, u_n) \in \mathbb{R}^n$,
for $X_1, X_2, \hdots, X_n \sim \mathcal{N}(0,1)$,
\begin{equation}
\pr{ \left|\sum_{i=1}^{n} u_i X_i \right| \geq \| \vec{u} \|_2 \cdot c_4  \sqrt{\ln \frac{2}{\delta}} } \leq \delta.
\end{equation}

\end{corollary}

% Next, we present a version of Hoeffding's inequality that holds for Gaussian variables.

% \begin{lemma}
% \label{lem:hoeffding-gaussian} \textbf{(Hoeffding's Inequality for Gaussians)}
% For $i=1,2, \hdots, n$, let $X_i$ be independent random variables sampled from a Gaussian with mean $\mu_i$ and variance $\sigma_i^2$. Then for all $t \geq 0$, we have:
% \[
% \pr{\sum_{i=1}^{n} (X_i - \mu_i) \geq t} \leq \exp \left( - \frac{t^2}{2 \sum_{i=1}^{n} \sigma_i^2} \right).
% \]

% Or alternatively, for $\delta \in (0,1]$
% \[
% \pr{\sum_{i=1}^{n} (X_i - \mu_i) \geq  \sqrt{2 \sum_{i=1}^{n} \sigma_i^2\ln \frac{1}{\delta}}} \leq \delta
% \]
% An identical inequality holds good symmetrically for the event $\sum_{i=1}^{n} X_i - \mu_i \leq -t$.
% \end{lemma}

% \begin{lemma}
% %\label{lem:normsq}
% (\textbf{$\ell_2$ norm of a Gaussian vector})
% If $z_{1}, z_2, \hdots, z_k \sim \mathcal{N}(0,1)$ be $k$ standard normal random variables drawn independently, then for all $t \in (0,1)$:
% \[
% \pr{\left|\frac{1}{k}\sum_{i=1}^{k} z_{i}^2 - 1 \right| \geq t} \leq 2e^{-kt^2/8}. 
% \]
% \end{lemma} 
% %
%Then, as a corollary we have that:
%
%\begin{corollary}
%\label{lem:frob-bound}
%If $\vec{W}$ is a $n_1 \times n_2$ matrix with entries all drawn independently at random from  $\mathcal{N}(0,\sigma)$, then with probability $1-\delta$, $\| \vec{W}\|_F = \mathcal{O}\left(\sigma\sqrt{n_1n_2 \log\frac{1}{\delta}}\right)$.
%\end{corollary}

Next, we state a tail bound for sub-exponential random variables \citep{wainwright19high}.

%  chi-squared random variables i.e., sum of squared normal variables.
% \begin{lemma}
% Let $z_1, z_2, \hdots, z_D \sim \mathcal{N}(0,1)$, then  for all $t \in (0,1)$
% \[
% Pr\left[ \left| \frac{1}{N}\sum_{d=1}^{N} z_d^2 - 1 \right| \geq t\right] \leq 2\exp(-Dt^2/8)
% \]
% \end{lemma}
\begin{lemma}\textbf{(Tail bound on sub-exponential random variables)}
For a sub-exponential random variable $X$ with parameters $(\nu, b)$ and mean $\mu$, for all $t > 0$,
\begin{equation}
\pr{|X - \mu| \geq t}\leq 2 \exp\left(-\frac{1}{2} \min \left(\frac{t}{b}, \frac{t^2}{\nu^2}\right) \right). \\
\end{equation}

\end{lemma}

As a corollary, we have the following bound on the sum of squared normal variables:
% We can then restate the above as follows, for use in our proof:

\begin{corollary}
\label{cor:chi}
\label{lem:normsq}
For $z_1, z_2, \hdots, z_D \sim \mathcal{N}(0,1)$, we have that
\begin{equation}
Pr\left[  \frac{1}{N}\sum_{j=1}^{N} z_j^2 \in [c_2^2, c_3^2]\right] \leq 2\exp(-c_1 D).
\end{equation}
\end{corollary}

\vspace{5pt}

Next, we use the following theorem based on Theorem 2.1.1 in \cite{wainwright15nachdiplom}, to bound the spectral norm of a matrix with random gaussian entries.
\begin{lemma}
\label{lem:spectralnorm}\textbf{(Spectral norm of an entrywise Gaussian matrix)}
Let $\vec{W}$ be a $n_1 \times n_2$ matrix with entries all drawn independently at random from $\mathcal{N}(0,1)$. Then, given $n_1 \geq n_2$, 
\begin{equation}
\pr{ \left| \sup_{\vecu: \| \vecu\| = 1}   \frac{\| \vec{W} \vecu  \|^2}{n_1} - 1 \right| \geq \mathcal{O}\left(\sqrt{\frac{n_2}{n_1} } + t\right)} \leq \mathcal{O} \left( e^{-n_1t^2/2}\right).
\end{equation}
\end{lemma}
%Note that the above concentration inequality is essentially a bound on the spectral norm of the matrix $\vec{W}$.

The following result \citep{tropp12tailbounds} provides a similar, more precise bound for the case where the dimensions of the matrix are identical.

\begin{lemma}
\label{lem:spec}\textbf{(Spectral norm of an entrywise Gaussian square matrix)}
Let $\vecU$ be a $H \times H$ matrix. Then,
\begin{equation}\prsub{\vecU \sim \N(0,\sigma^2I)}{\spec{\vecU} > t} \leq 2H \exp(-t^2/2H\sigma^2)\end{equation}
or alternatively, for any $\delta > 0$,
\begin{equation}\prsub{\vecU \sim \N(0,\sigma^2I)}{\spec{\vecU} > \sigma \sqrt{2H \ln \frac{2H}{\delta}} } \leq \delta. \end{equation}
\end{lemma}

As a result of Corollary~\ref{lem:normsq} and Lemma~\ref{lem:spectralnorm}, we can bound the norms of a Xavier-initialized networks as follows:
\begin{corollary}
\label{cor:weight-matrix-bounds} \textbf{(Norms of a randomly initialized network)}
For a network of more than $1$ hidden layer, and $H$ hidden units per layer, when its initialization $\mathcal{Z}$ is according to Xavier i.e.,  when all entries are drawn from $\mathcal{N}(0,\mathcal{O}(1/\sqrt{H}))$, with high probability we have: 
\begin{itemize}
\item $\|\vec{Z}_1\|_F = \tilde{{\Theta}}(\sqrt{n})$, $\|\vec{Z}_d\|_F = \tilde{{\Theta}}(\sqrt{H})$ for $D > d > 1$, $\|\vec{Z}_D\|_F = \tilde{{\Theta}}(1)$.
\item  $\|\vec{Z}_d\|_2 = \tilde{{\Theta}}(1)$ for all $d$.
\end{itemize}
\end{corollary}

The following result bounds the norm of the product of an entrywise Gaussian matrix and an arbitrary vector.
\begin{lemma}
\label{lem:gaussian-operator-norm}
Let $\vecU$ be a $H_1 \times H_2$ matrix where each entry is sampled from $\mathcal{N}(0,\sigma^2)$. Let 
 $\vec{x}$ be an arbitrary vector in $\mathbb{R}^{H_2}$. Then, $\vecU\vec{x} \sim \N(0,  \|\vec{x} \|_2^2 \sigma^2 \vecI)$. 
%  Furthermore,

% \[
% \pr{ \left| \frac{\|  U \vec{x}\|_2^2}{ H_1 \sigma^2 \| \vec{x}\|_2^2} - 1 \right| \geq \sqrt{\frac{8}{H_1}\ln \frac{2}{\delta}}} \leq \delta
% \]
\end{lemma}

\begin{proof}
$\vecU \vec{x}$ is a random vector sampled from a multivariate Gaussian with mean $\E[\vecU\vec{x}] = 0$ and co-variance $\E[\vecU\vec{x}\vec{x}^T \vecU^T]$. The $(i,j)$th entry in this covariance matrix is $\E[(\vec{u}_i^T\vec{x}) (\vec{u}_j^T\vec{x})]$ where $\vec{u}_i$ and $\vec{u}_j$ are the $i$th and $j$th row in $\vecU$. When $i=j$, $\E[(\vec{u}_i^T\vec{x}) (\vec{u}_j^T\vec{x})] = \E[\|\vec{u}_i^T \vec{x}\|^2] = \sum_{h=1}^{H_2} \E[u_{ih}^2] x_h^2 = \sigma^2 \|\vec{x} \|_2^2$. When $i\neq j$, since $\vec{u}_i$ and $\vec{u}_j$ are independent random variables, we will have $\E[(\vec{u}_i^T\vec{x}) (\vec{u}_j^T\vec{x})] = \sum_{h=1}^{H_2} \E[u_{ih} x_h]  \sum_{h=1}^{H_2} \E[u_{jh} x_h] = 0$. 

%Thus, we can apply of Lemma~\ref{lem:chi-square} by considering the $H_1$ random variables $\frac{\vec{u}_i \vec{x}}{\sigma^2\|\vec{x}\|_2} \sim \N(0, 1)$, and setting $t = \sqrt{\frac{8}{H_1}\ln \frac{2}{\delta}}$.

 \end{proof}

Next, we present the Khintchine-Kahane inequality which is used to bound a Rademacher complexity-like term ignoring the supremum within the expectation.

\begin{theorem}\textbf{(Khintchine-Kahane inequality)}
\label{thm:kk}
For any $0 < p < \infty$ and set of scalar values $ \{ x_1, x_2, \hdots, x_m\}$, when $\vec{\xi}$ is a Rademacher vector sampled uniformly from $\{ -1, 1\}^m$:
\begin{equation}
\left( \mathbb{E}_{\vec{\xi}} \left[\sum_{i=1}^{m} \left| \xi_i x_i \right|^p\right]\right)^{1/p} \leq C_{p} \left( \sum_{i=1}^{m} {x}_i^2\right)^{1/2}.
\end{equation}

where $C_p$ is a constant dependent only on $p$.
\end{theorem}

It is simple to extend this to vector-valued variables for $p=2$, which is what we will need specifically for our discussion:

\begin{corollary}
\label{cor:kk}
For a set of vectors $ \{ \vec{x}_1, \vec{x}_2, \hdots, \vec{x}_m\}$, when $\vec{\xi}$ is a Rademacher vector sampled uniformly from $\{ -1, 1\}^m$:
\begin{equation}
\mathbb{E}_{\vec{\xi}} \left[\left\| \sum_{i=1}^{m}  \xi_i \vec{x}_i \right\|\right] \leq \left( \mathbb{E}_{\vec{\xi}} \left[ \left\|\sum_{i=1}^{m}  \xi_i\vec{x}_i \right\|^2\right]\right)^{1/2} \leq C_{p} \left( \sum_{i=1}^{m} \|\vec{x}_i\|^2\right)^{1/2}.
\end{equation}
\end{corollary}

Here, the first inequality follows from Jensen's inequality. To derive the next, we first apply the Khintchine-Kahane inequality for each dimension of these vectors. We then square the inequalities and sum them up, after which we take the square root of both sides.\\

We will use the following KL divergence equality to bound the generalization error in our PAC-Bayesian analyses.

\begin{lemma}\textbf{(KL Diverge of Gaussians)}
\label{lem:kldivergence}
Let $P$ be the spherical Gaussian $\N(\vec{\mu}_1, \sigma^2 \vecI)$ and $Q$ be the spherical Gaussian $\N(\vec{\mu}_2, \sigma^2 \vecI)$. Then, the KL-divergence between $Q$ and $P$ is:
\begin{equation}
\text{KL}(Q \| P) = \frac{\elltwo{\vec{\mu}_2 - \vec{\mu}_1}^2}{2\sigma^2}.
\end{equation}

\end{lemma}

\part{Tighter, Data-dependent \\ Uniform Convergence Bounds}
\graphicspath{{dist-from-init/}}

\chapter{The Role of Distance from Initialization}
\label{chap:dist-from-init}

%Why does training deep neural networks using stochastic gradient descent (SGD)  result in a generalization error that does not worsen with the number of parameters in the network? 

%To answer this question, we advocate 
% a notion of effective model capacity that is dependent  on {\em a given random initialization of the network} and not just the training algorithm and the data distribution. We provide empirical evidences that demonstrate that the model capacity of SGD-trained deep networks is in fact restricted through implicit regularization of {\em the $\ell_2$ distance from the initialization}.  We also provide  theoretical arguments that further highlight the need for initialization-dependent notions of model capacity. 

\section{Introduction}
\label{sec:init-intro}
%  \cite{neyshabur15inductive} and \cite{zhang17generalization} proposed an important open question in deep learning that has been actively studied over the last year. The question stems from the observation that
%  %While the first question has been a long-standing problem in deep learning, in this paper we focus on the second, which was posed only recently \citep{zhang17generalization} and has received much attention since. 
% an overparametrized deep model that can be trained to fit completely random labels, can also be trained to fit real-world data while achieving low test error. This contradicts the conventional understanding in machine learning about the tradeoff between the expressivity of a model and its ability to generalize. 

When the generalization puzzle was first posed, it was immediately clear that we must move away from algorithm-independent notions of model capacity and search for notions of capacity/complexity that take into account the training algorithm, namely GD or SGD or other variants of it used in practice. For example, \citep{neyshabur17exploring} explored many notions of algorithm-dependent model capacity such as the $\ell_2$ norms of the learned weights, path norms, spectral norms and so on.

In this chapter, we will extend this line of work by specifically advocating that the notion of capacity must not only be algorithm-dependent, but also {\em initialization-dependent}. For example, we argue that to truly understand generalization and get tighter bounds, we must look at the $\ell_2$ {\em distance from initialization} rather than the total $\ell_2$ norm of the weights which is agnostic to the initialization of the network. The idea of incorporating the initialization was first done in the PAC-Bayesian bounds of \cite{dziugaite17nonvacuous}. Their rationale was that this helps account for symmetries in the network. We
provide complementary arguments supporting the need for doing this. In particular, we argue that the distance of the learned network from its initialization is implicitly regularized by SGD to a {\em width-independent} value (see Section~\ref{sec:init-experiments}). Then, in Sections~\ref{sec:init-linear-network} and ~\ref{sec:init-bad-norms}, we provide theoretical arguments highlighting how initialization-dependent model capacity is much tighter than initialization-independent notions. We leave as open questions how and why distance from initialization is regularized, and whether it is sufficient to explain generalization. The results in this chapter have been previously published in \cite{nagarajan17role}.

\section{Initialization-dependent model capacity}
%\label{sec:init-problem}
\label{sec:init-dep-capacity}

We first formally define a notion of effective model capacity based on which we will study implicit regularization and generalization. 

%Let $\scrD$ be a distribution over data in $\mathbb{R}^{n} \times [0,1]$. Let $\mathcal{L}: \mathbb{R}^n \times \mathbb{R} \to [0,1]$ be a loss function.  Given $m$ datapoints, our goal is to explain why training a network of $H$ hidden units per layer using SGD to zero training loss, 
%(i.e., finding $(\mathcal{W}^\star, \mathcal{B}^\star)$ such that  $\frac{1}{m}\sum_{(\vec{x},y) \in \mathcal{S}} \mathcal{L}(f_{(\mathcal{W}^\star, \mathcal{B}^\star)}(\vec{x}), y) = 0$)
 %results in a generalization error dependent only on $m$ and independent of $H$.  In order to do this, we argue (through empirical and theoretical evidences) that the effective capacity of the model must be take into account not just the training algorithm and the underlying distribution but also {\em the given random initialization}:\\

\begin{definition}  \label{def:effective-capacity} For a particular model (i.e., network architecture), we define the   {\em \bf effective capacity}  $\calF_{m,\delta}[\scrD, (\mathcal{Z}, \mathcal{C}), \mathcal{A}]$ of a (distribution, initialization, algorithm)-tuple to be a set of parameter configurations such that with high probability $1-\delta$ over the draws of a dataset of $m$ samples from the distribution $\scrD$, training the network initialized with $ (\mathcal{Z}, \mathcal{C})$,  to zero loss on these samples using algorithm \footnote{If $\mathcal{A}$ is stochastic, we could either incorporate it in the ``high probability'' component of the definition, or we could ``freeze'' it by including it as an argument to $\calF$ like the random initialization.} $\mathcal{A}$, obtains a parameter configuration that lies in the set  $\calF_{m,\delta}[\scrD, (\mathcal{Z}, \mathcal{C}), \mathcal{A}]$. \\
\end{definition}

This is notion is more abstract and/or more refined than existing notions of effective model capacity. 
First, this is an abstraction of the idea of \cite{dziugaite17nonvacuous} who incorporated the initialization into their PAC-Bayesian analysis by arguing that it will take into account the symmetries of the network. This notion is however more refined than the one in \cite{arpit17closer}  which is independent of both the data distribution and the (random) initializations. Similarly, \cite{neyshabur17exploring} consider an indirect notion of effective capacity by evaluating a norm (such as the $\ell_2$ norm) for each parameter configuration and investigating whether the algorithm restricts itself to configurations of low norm; however, these norms are calculated independent of the random initialization. \\

Our goal now is to identify as precise a characterization of $\calF_{m,\delta}[\scrD, (\mathcal{Z}, \mathcal{C}), \mathcal{A}]$ as possible.  Effectively, instead of seeking a `global' quantity regularized across all initializations (such as the $\ell_2$ norm in \cite{neyshabur17exploring}), we seek one that is specific to the initialization.
 %Hopefully this has a learning-theoretic complexity independent of the parameter count.
 %width $H$. 
 % This would then hopefully allow us to derive parameter-count-independent generalization guarantees. 

% show that the generalization error 
% must be independenent of $H$ too
% (see Lemma~\ref{lem:rademacher}).

The focus of this chapter is arguably the simplest such quantity, one that was originally considered  in \cite{dziugaite17nonvacuous}: the distance of the weights $(\mathcal{W}, \mathcal{B})$ from the initialization, $\| (\mathcal{W}, \mathcal{B}) - (\mathcal{Z}, \mathcal{C}) \|_F$. Specifically, \cite{dziugaite17nonvacuous} presented a PAC-Bayesian bound involving the distance from initialization (rather than one involving the distance from origin) and showed that SGD can be made to {\em explicitly} regularize such a bound in a way that a non-vacuous PAC-Bayesian bound holds on the resulting network. We show that distance from initialization is in fact {\em implicitly} regularized by SGD and we investigate it in much greater detail in terms of its dependence on the parameter count and its ability to explain generalization.

\section{Experiments}
\label{sec:init-experiments}
We first report empirical results demonstrating the effect of SGD-training on distance from initialization.
We conduct experiments on the CIFAR-10 \& MNIST datasets, where we train networks of 4 hidden layers with varying width $H$ to minimize cross entropy loss and squared error loss. We study how distance from initialization (which we will denote in short as $r$) varies with width $H$ and training set size $m$ both for real data and partially/fully corrupted labels like in \cite{zhang17generalization} and \cite{arpit17closer}.   We summarize our observations below.% and present more experiments and details in Appendix~\ref{app:experiments}.

\begin{figure}[t!]
        \begin{minipage}{.5\textwidth}
        \centering
        \includegraphics[scale=0.25]{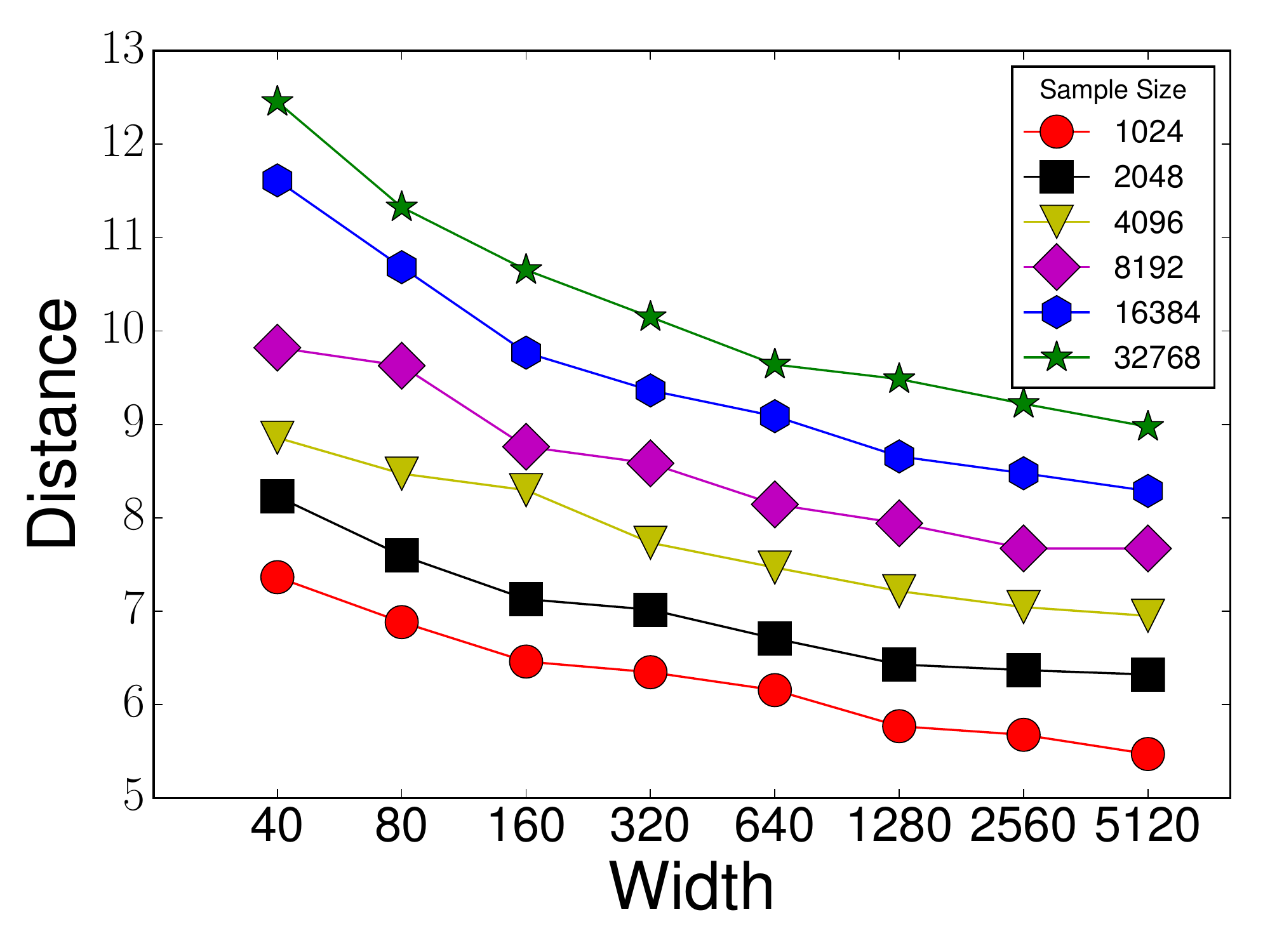}\\
    \end{minipage}%
        \begin{minipage}{.5\textwidth}
        \centering
        \includegraphics[scale=0.25]{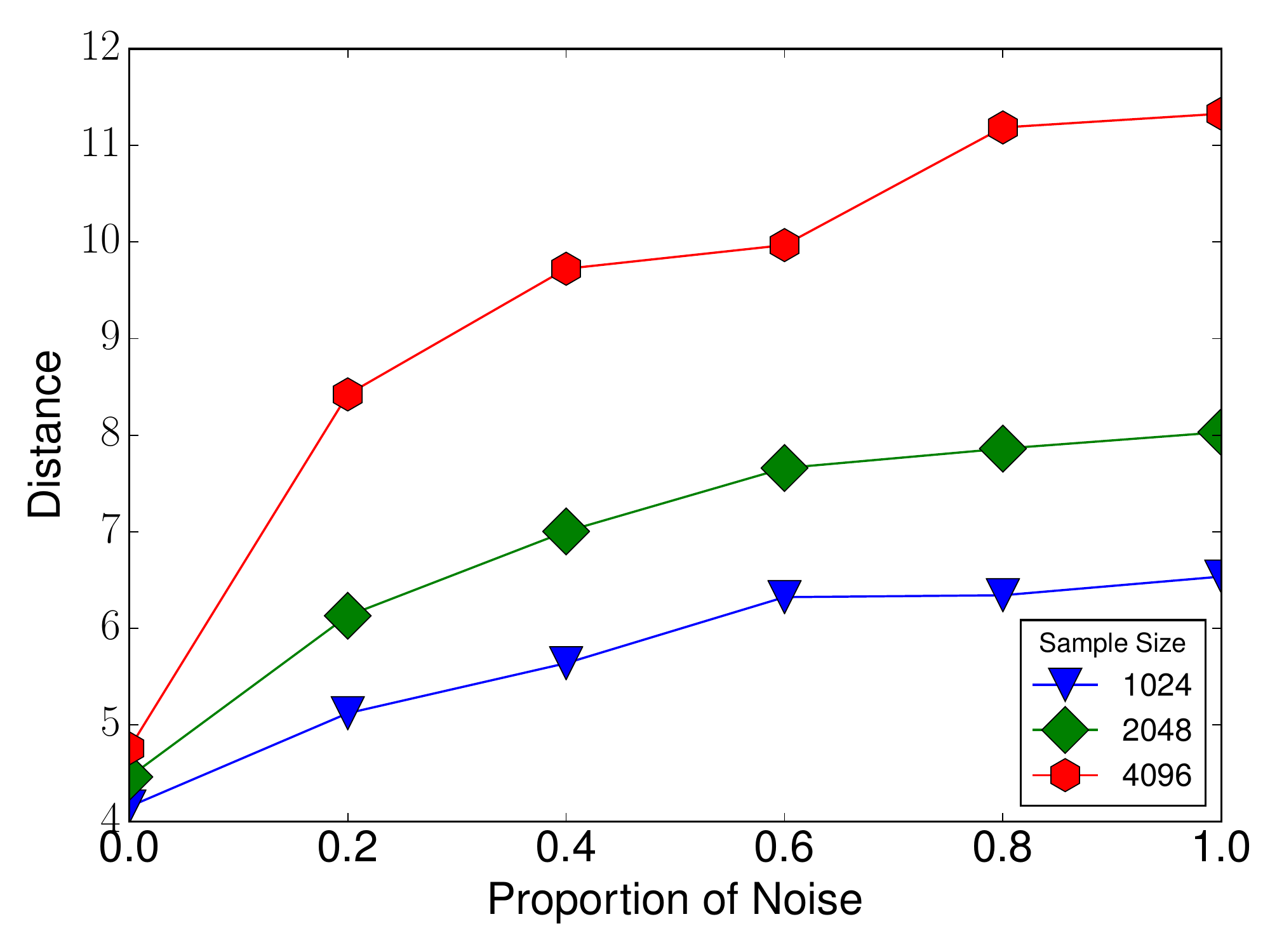}\\
    \end{minipage}
	\caption{{Distance from initialization for MNIST classification}}
    \label{fig:init-cross-entropy}
\end{figure}

\begin{figure}[t!]
        \begin{minipage}{.25\textwidth}
        \centering
        \includegraphics[scale=0.25]{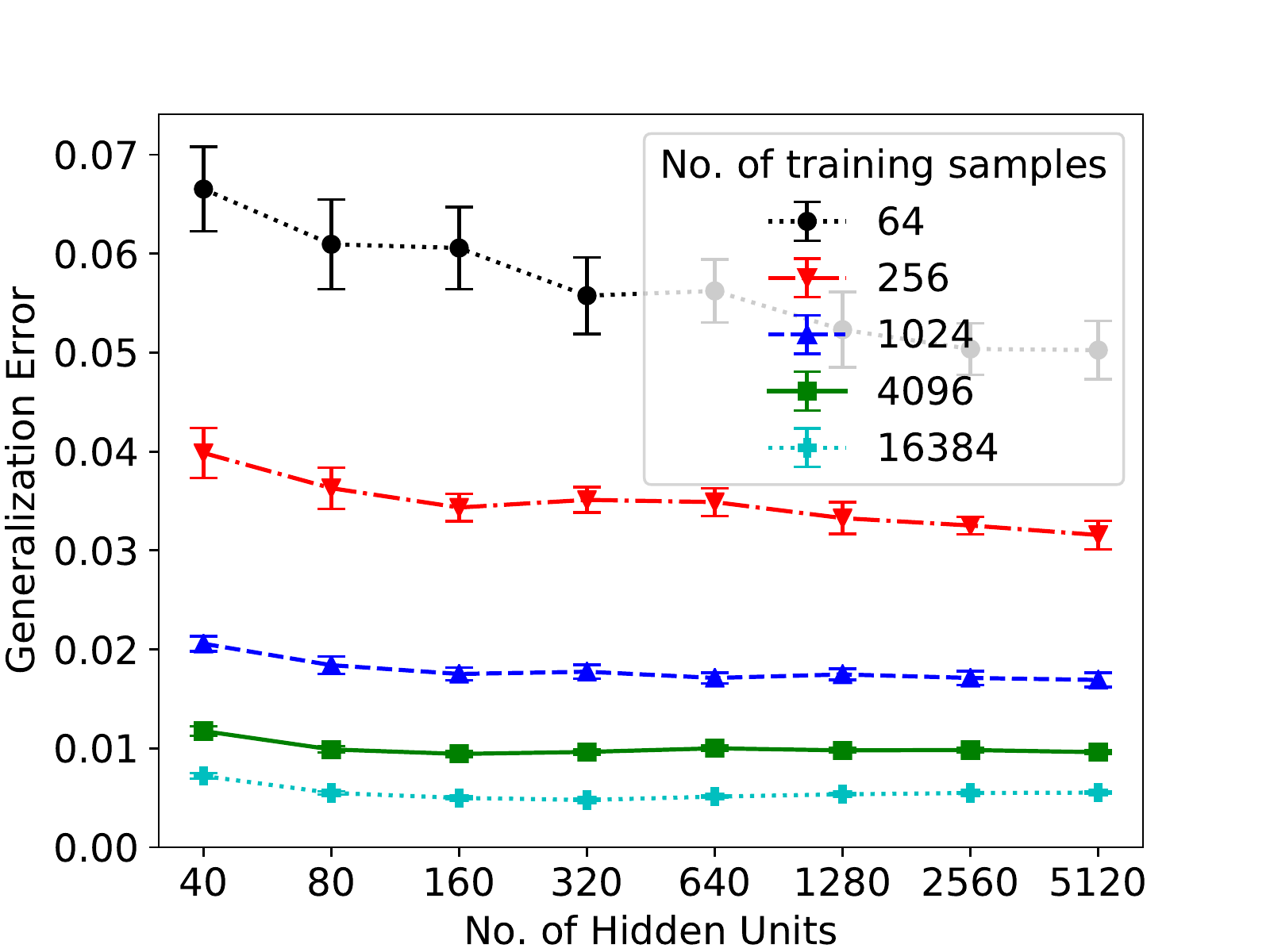}
    \end{minipage}%
               \begin{minipage}{.25\textwidth}
        \centering
                \includegraphics[scale=0.25]{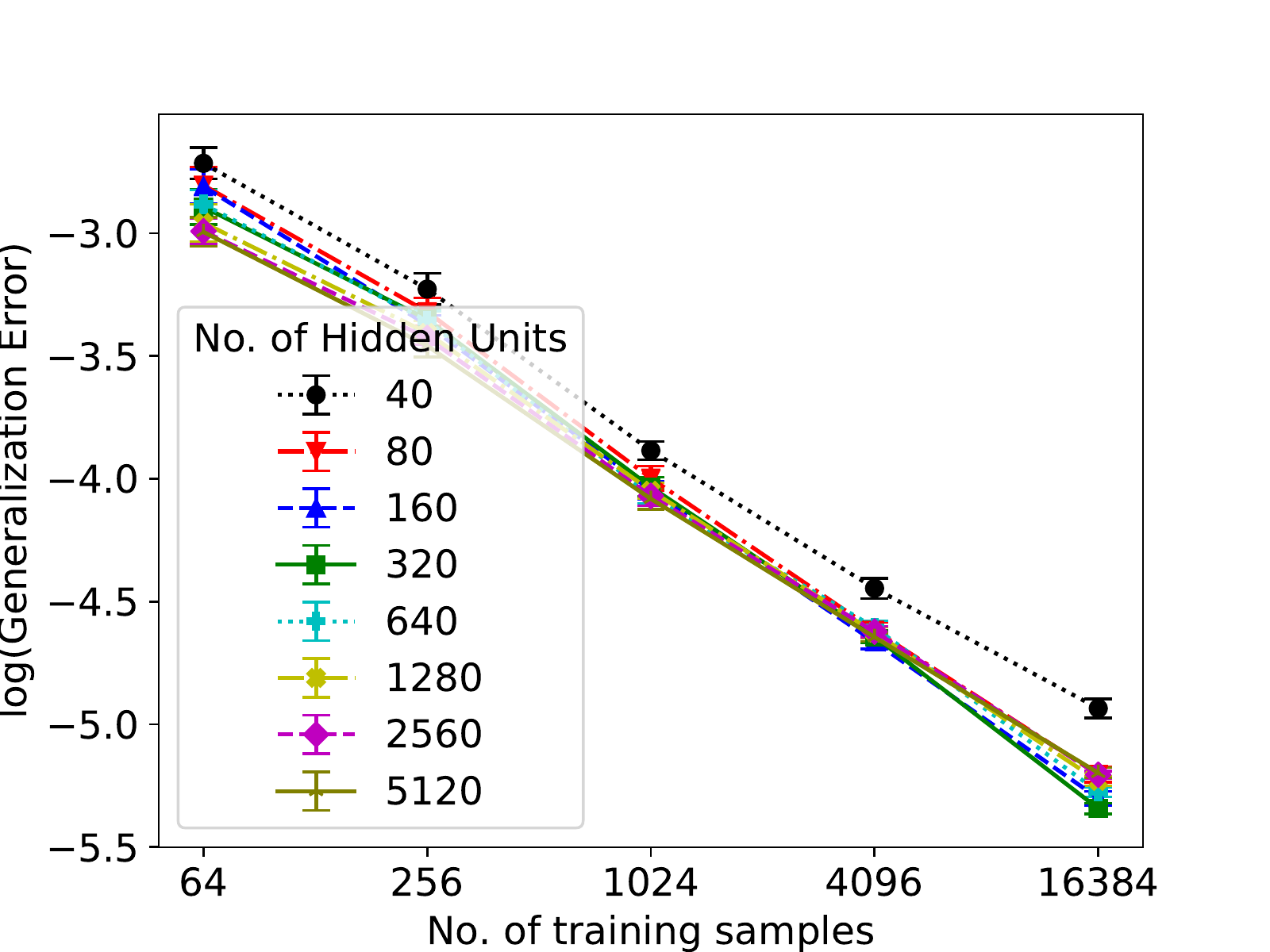}
        \end{minipage}%
                \begin{minipage}{.25\textwidth}
        \centering
                \includegraphics[scale=0.25]{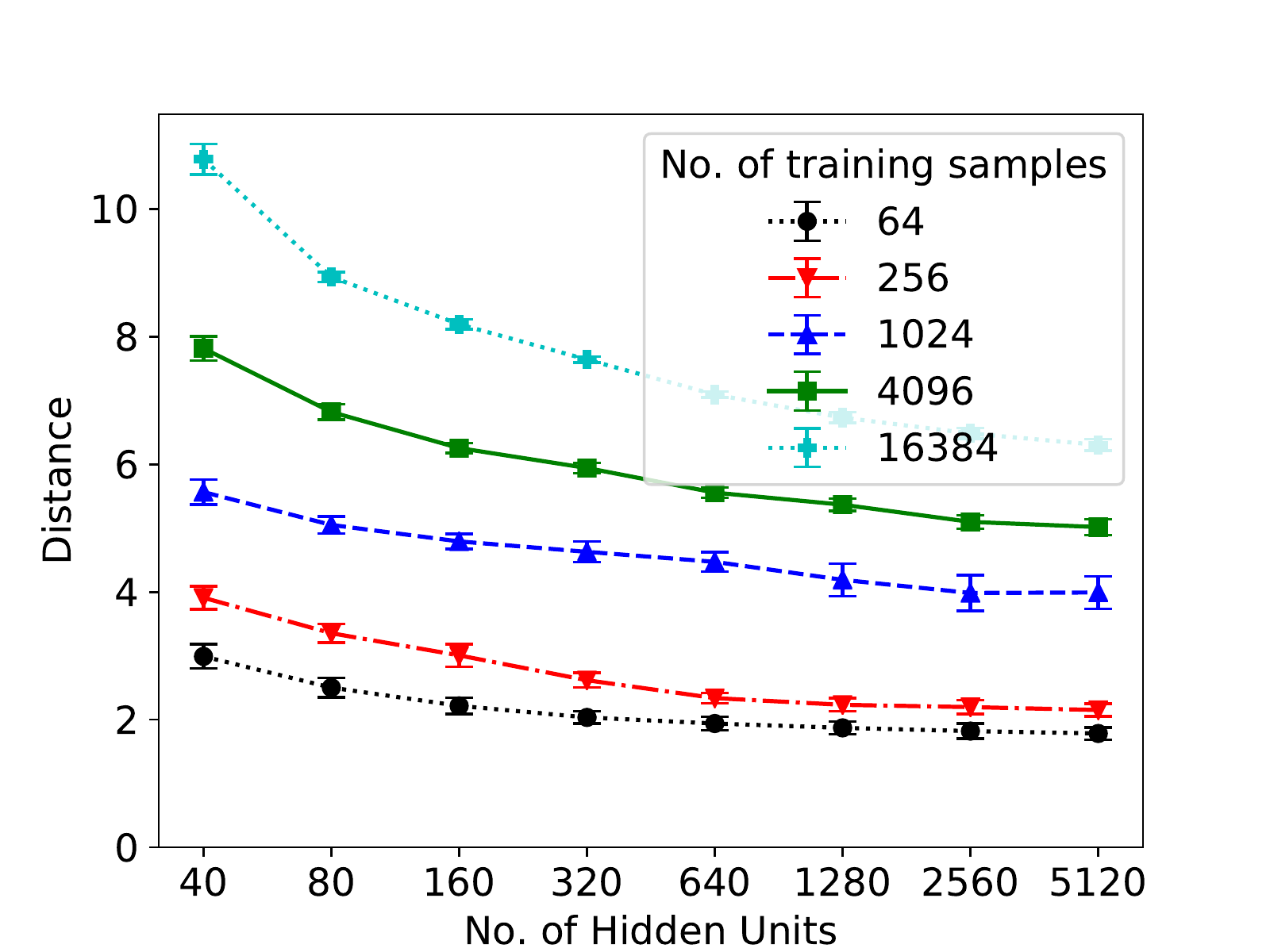}
    \end{minipage}%
        \begin{minipage}{.25\textwidth}
        \centering
        \includegraphics[scale=0.25]{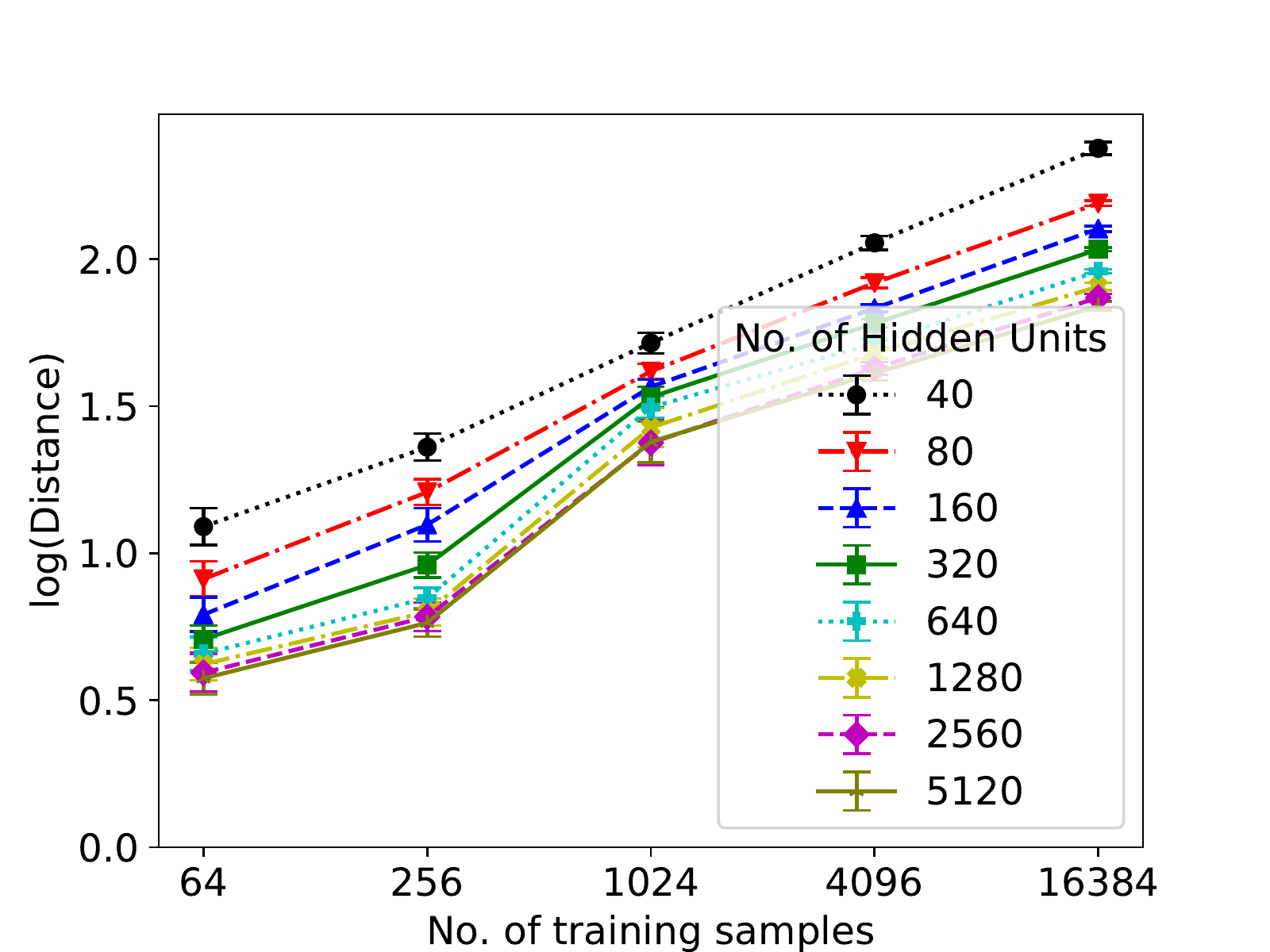}
        \end{minipage}
    \caption{{Distance from initialization for regression with Momentum SGD on MNIST}}
    \label{fig:init-mnist-momentum}
\end{figure}

\begin{figure}[t!]
        \begin{minipage}{.25\textwidth}
        \centering
        \includegraphics[scale=0.25]{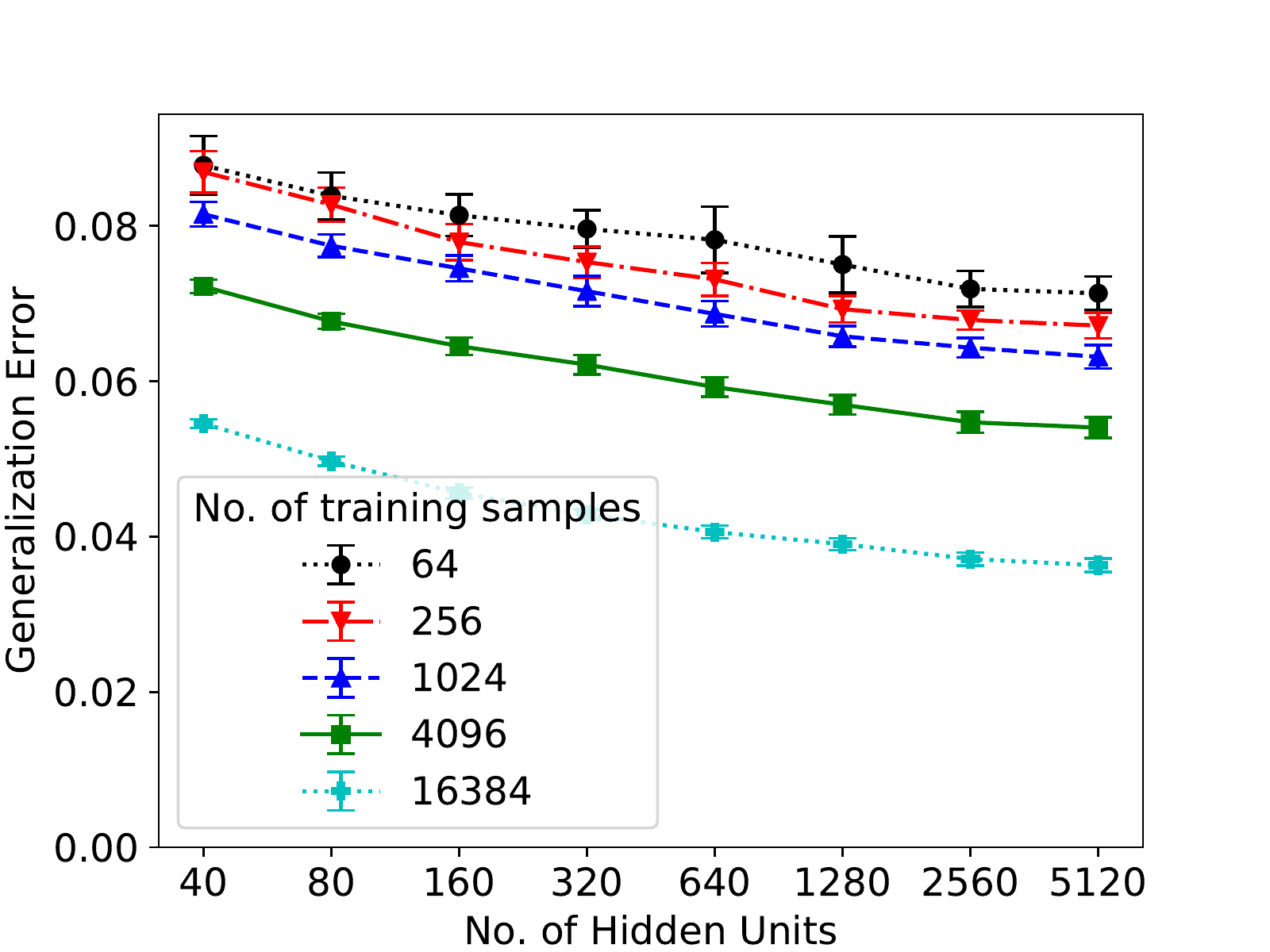}
    \end{minipage}%
        \begin{minipage}{.25\textwidth}
        \centering
        \includegraphics[scale=0.25]{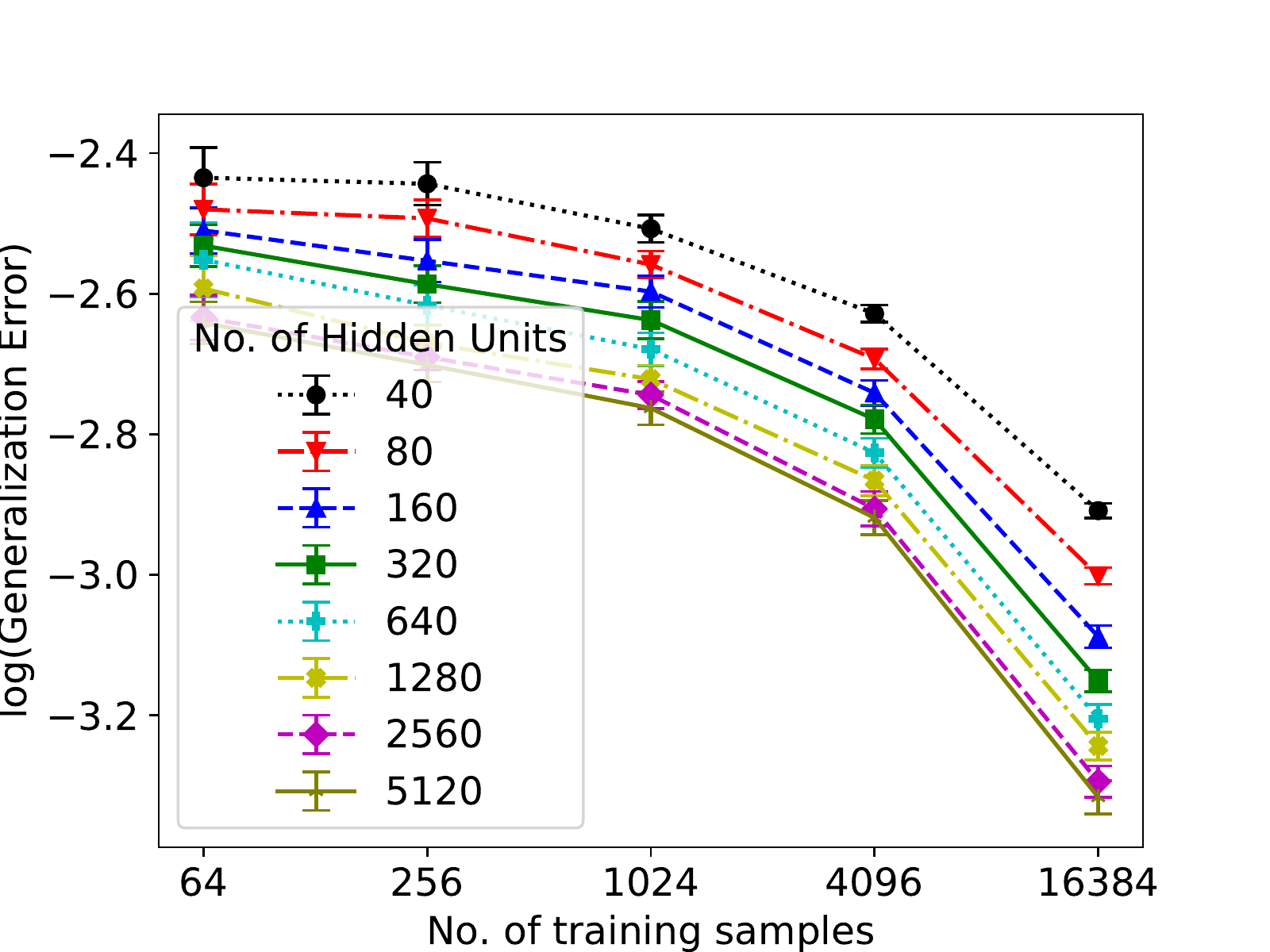}
    \end{minipage}\hfill
               \begin{minipage}{.25\textwidth}
        \centering
        \includegraphics[scale=0.25]{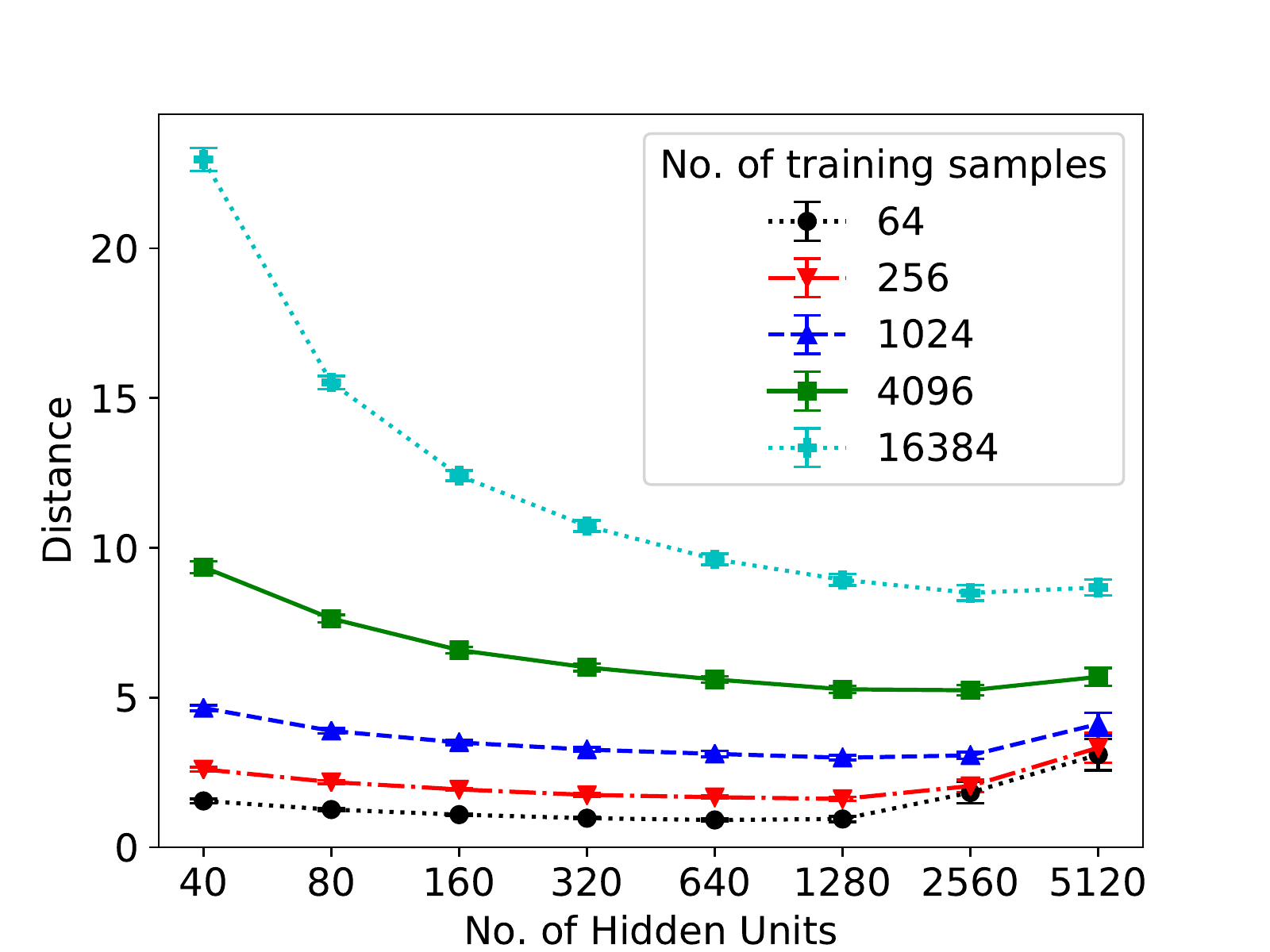} 
        \end{minipage}%
        \begin{minipage}{.25\textwidth}
        \centering
        \includegraphics[scale=0.25]{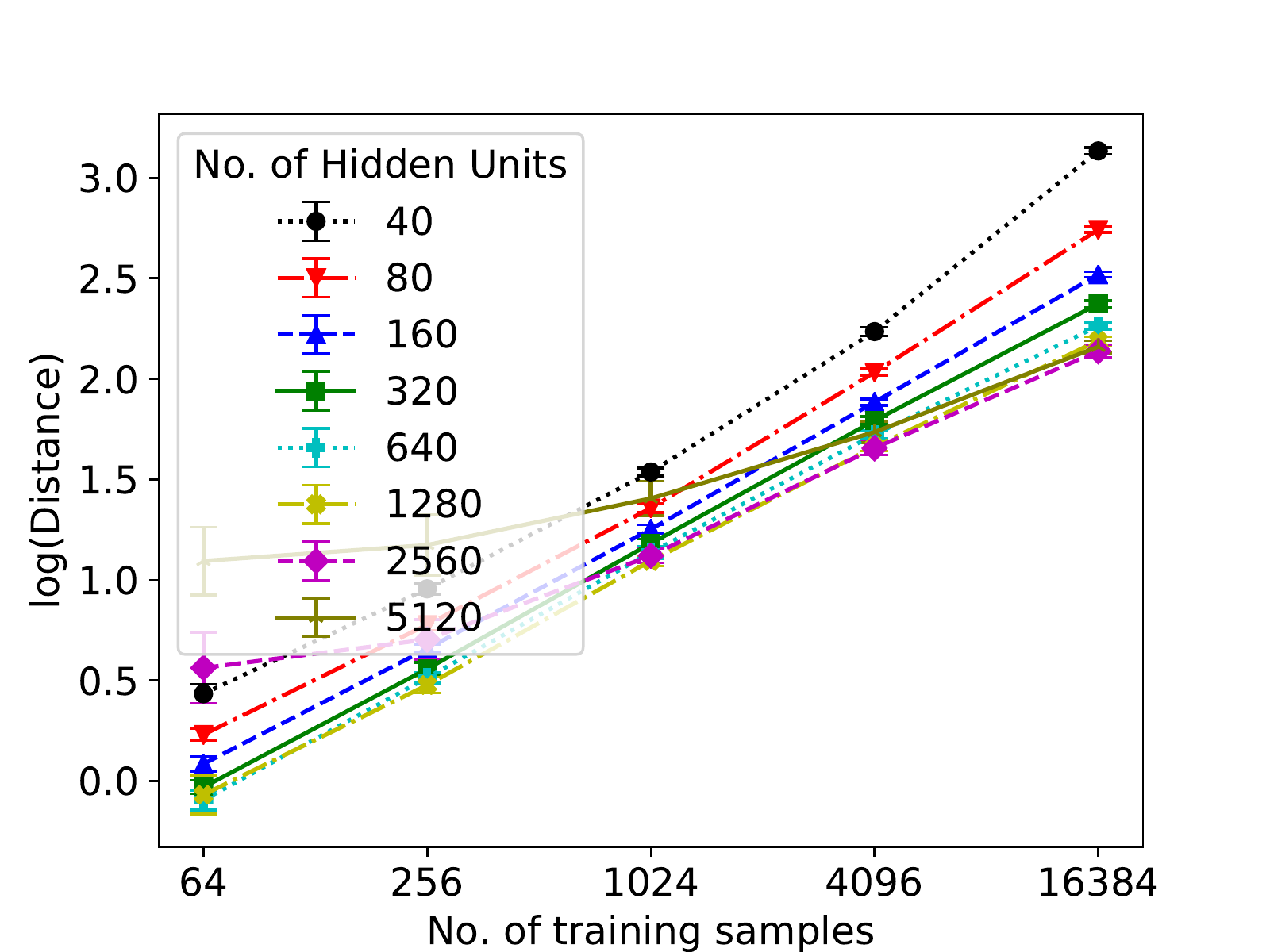}
        \end{minipage}\hfill
    \caption{{Distance from initialization for regression with SGD on CIFAR-10}}
    \label{fig:init-cifar}
\end{figure}

\begin{figure}[t!]
        \begin{minipage}{.25\textwidth}
        \centering
        \includegraphics[scale=0.25]{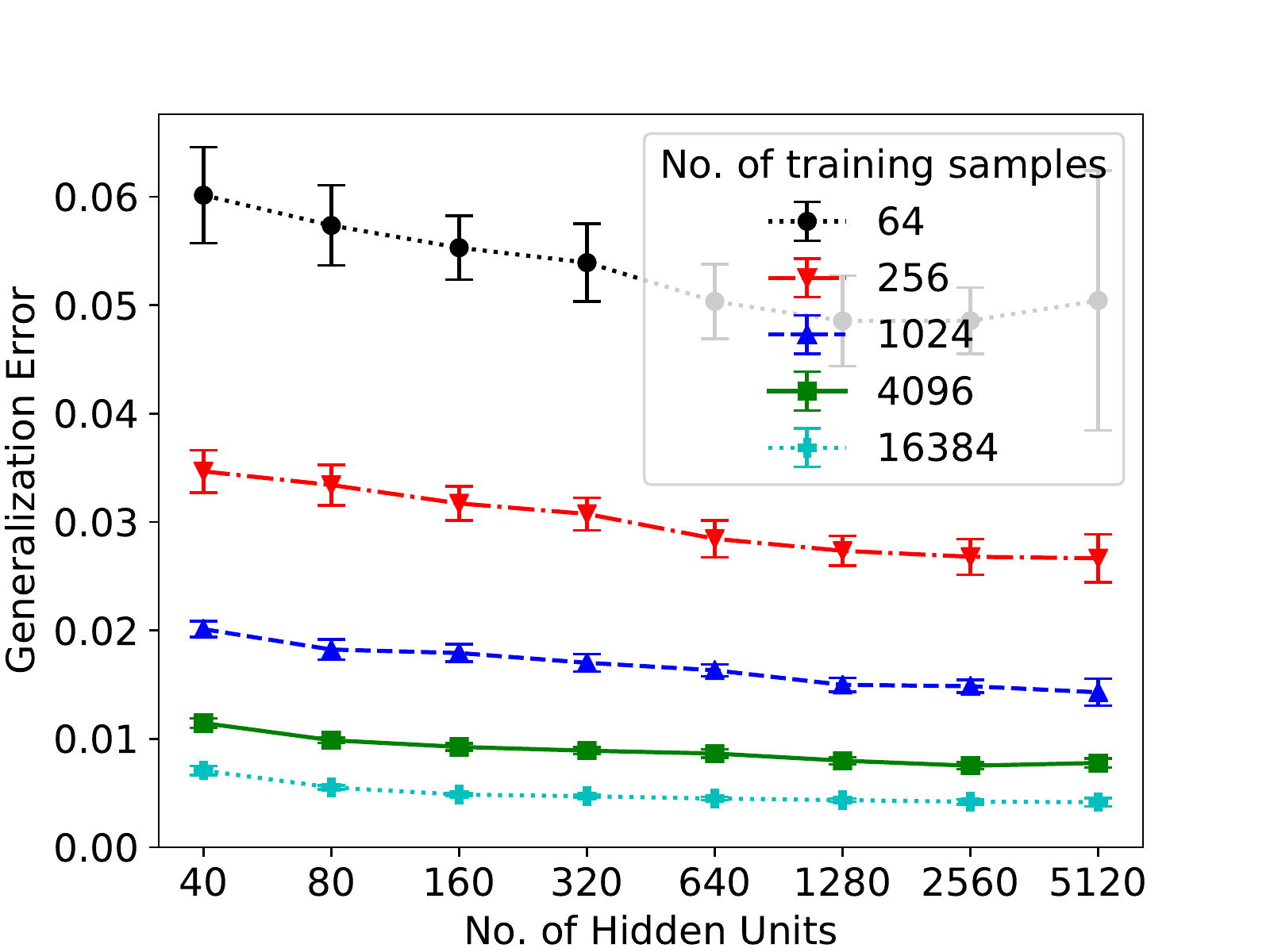}
    \end{minipage}%
        \begin{minipage}{.25\textwidth}
        \centering
        \includegraphics[scale=0.25]{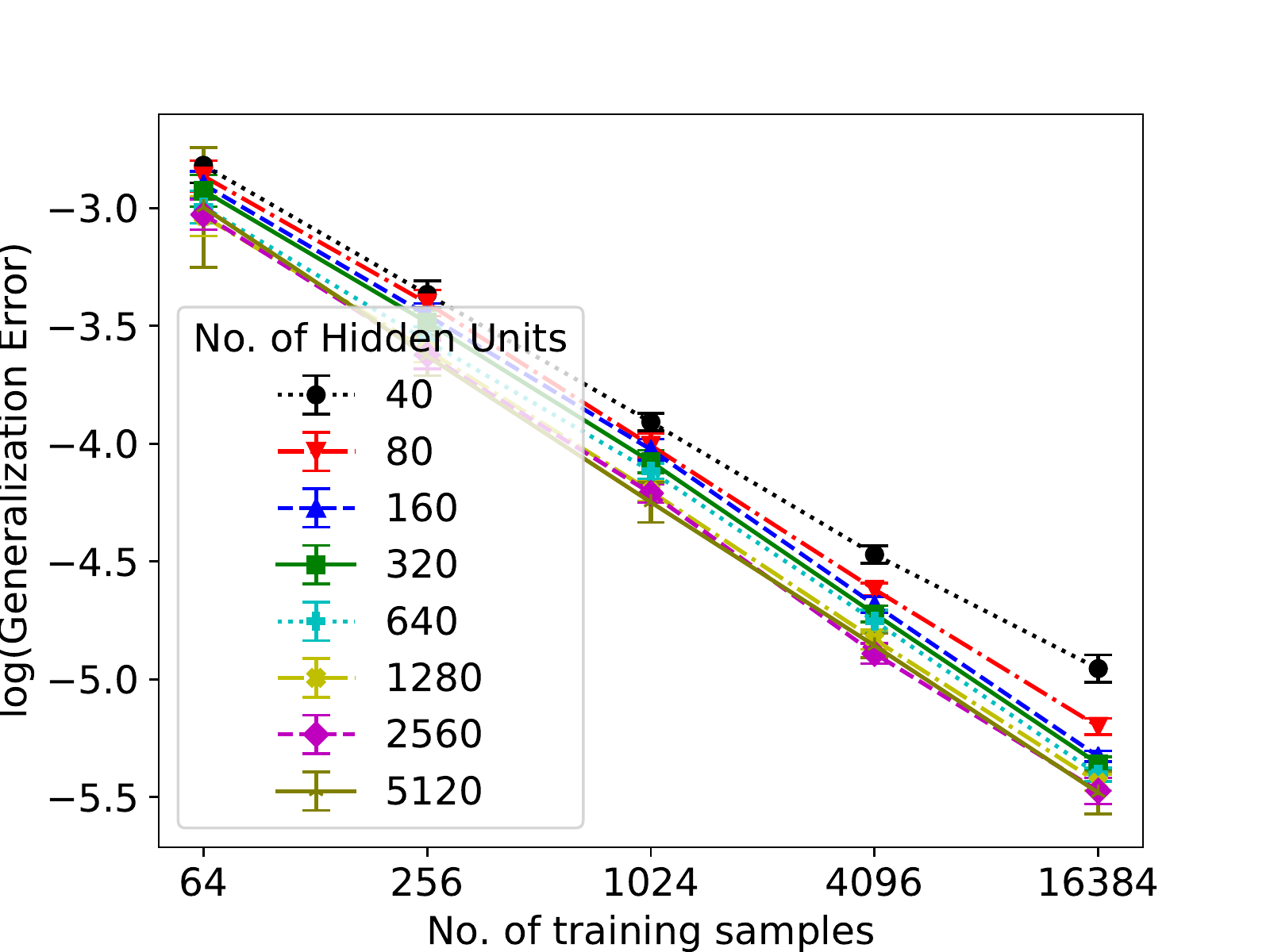}
    \end{minipage}\hfill
               \begin{minipage}{.25\textwidth}
        \centering
        \includegraphics[scale=0.25]{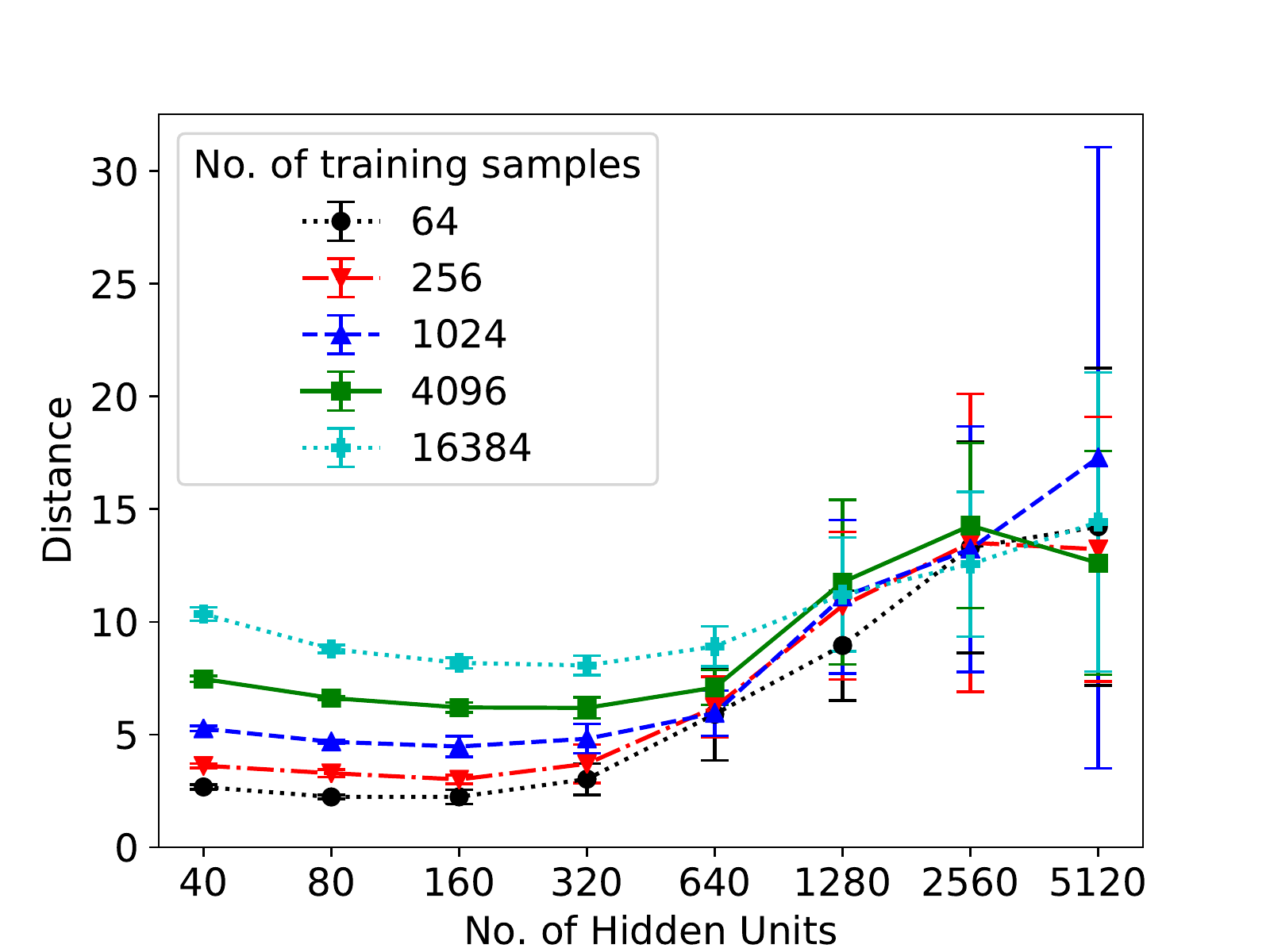} 
        \end{minipage}%
        \begin{minipage}{.25\textwidth}
        \centering
        \includegraphics[scale=0.25]{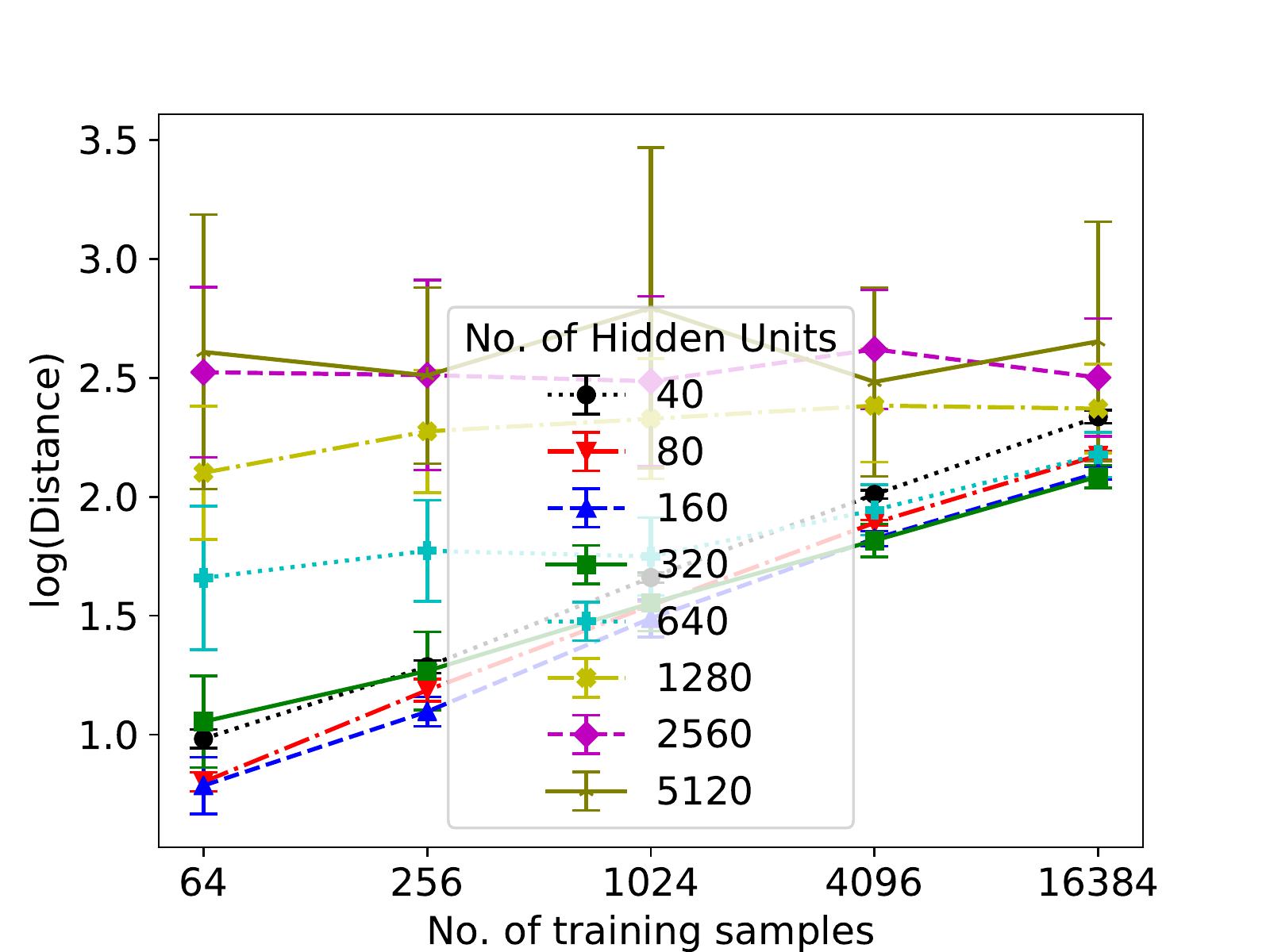}
        \end{minipage}\hfill
    \caption{{Distance from initialization for regression with SGD on MNIST}}
    \label{fig:init-mnist}
\end{figure}

\paragraph{Experimental details.} For the cross-entropy loss experiments in Figure~\ref{fig:init-cross-entropy}, we minimize the loss until at least $99\%$ of the data is classified by a margin of at least $10$, i.e., for $99\%$ of $(\vecx,y) \in S$, $\Gamma(f(\vecx),y) \geq 10$. 
We use SGD with a batch size of $64$. For the squared error loss, we minimize the squared error difference between a one-hot ground truth encoding of the $K=10$ classes and the output of the network,
formulated as $\scrL_{\text{sq}}(f(\vecx), \vec{y}) = \frac{1}{K}\sum_{d} (f(\vecx)[k] - y_d)^2$. For these experiments, 
in the case of MNIST data set, we use SGD with learning rate $0.01$ and momentum $0.9$ until the loss is less than $0.001$ (see Figure~\ref{fig:init-mnist-momentum}). For CIFAR-10 data set, we use SGD with learning rate $0.5$ until the loss is less than $0.02$ (see Figure~\ref{fig:init-cifar}). We also consider another set of experiments on the MNIST dataset using 
 SGD with learning rate $1$ (see Figure~\ref{fig:init-mnist}).

\begin{figure}[t!]
        \begin{minipage}{.25\textwidth}
        \centering
        \includegraphics[scale=0.25]{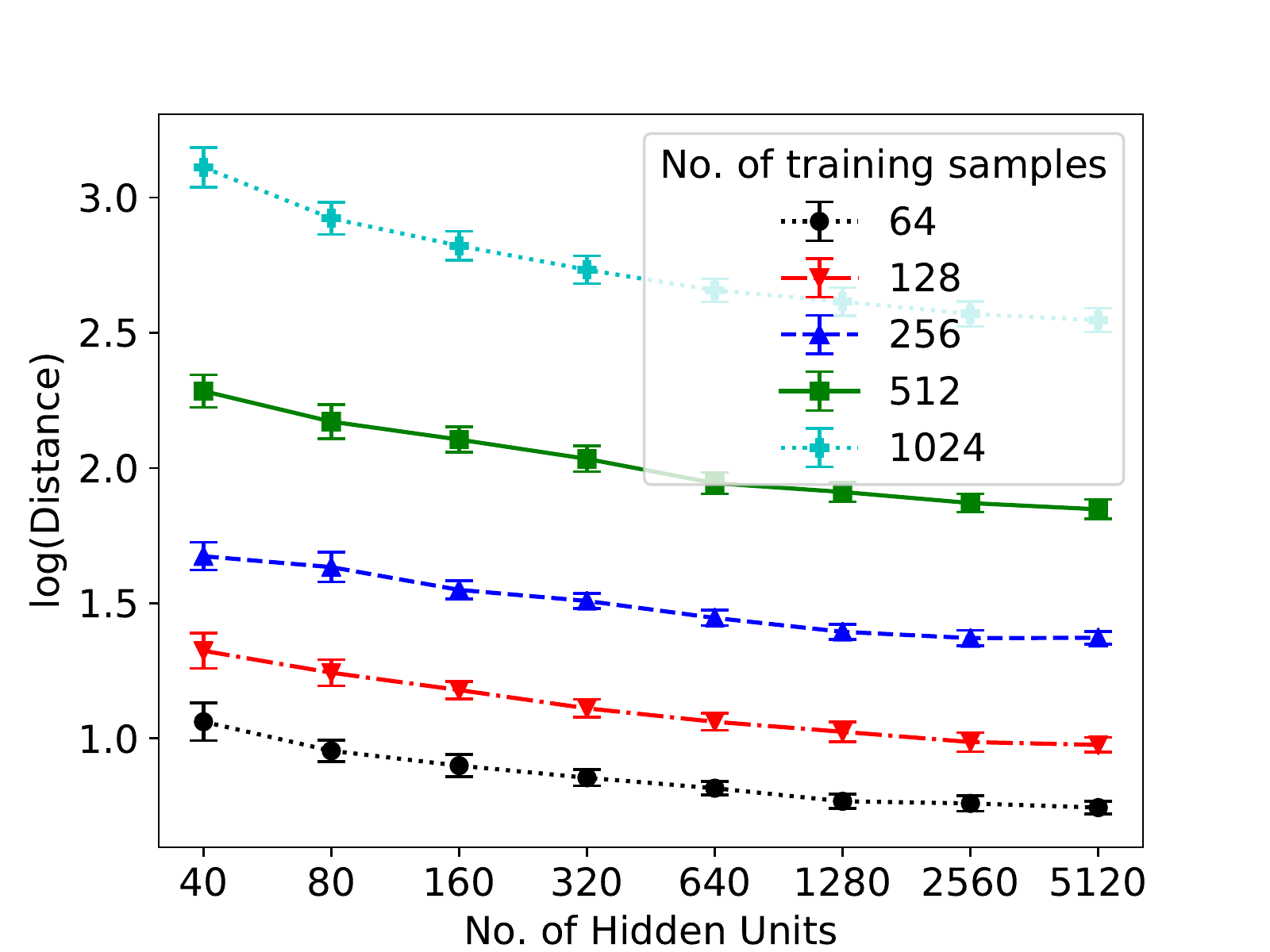}\\
        (a) MNIST
    \end{minipage}%
        \begin{minipage}{.25\textwidth}
        \centering
        \includegraphics[scale=0.25]{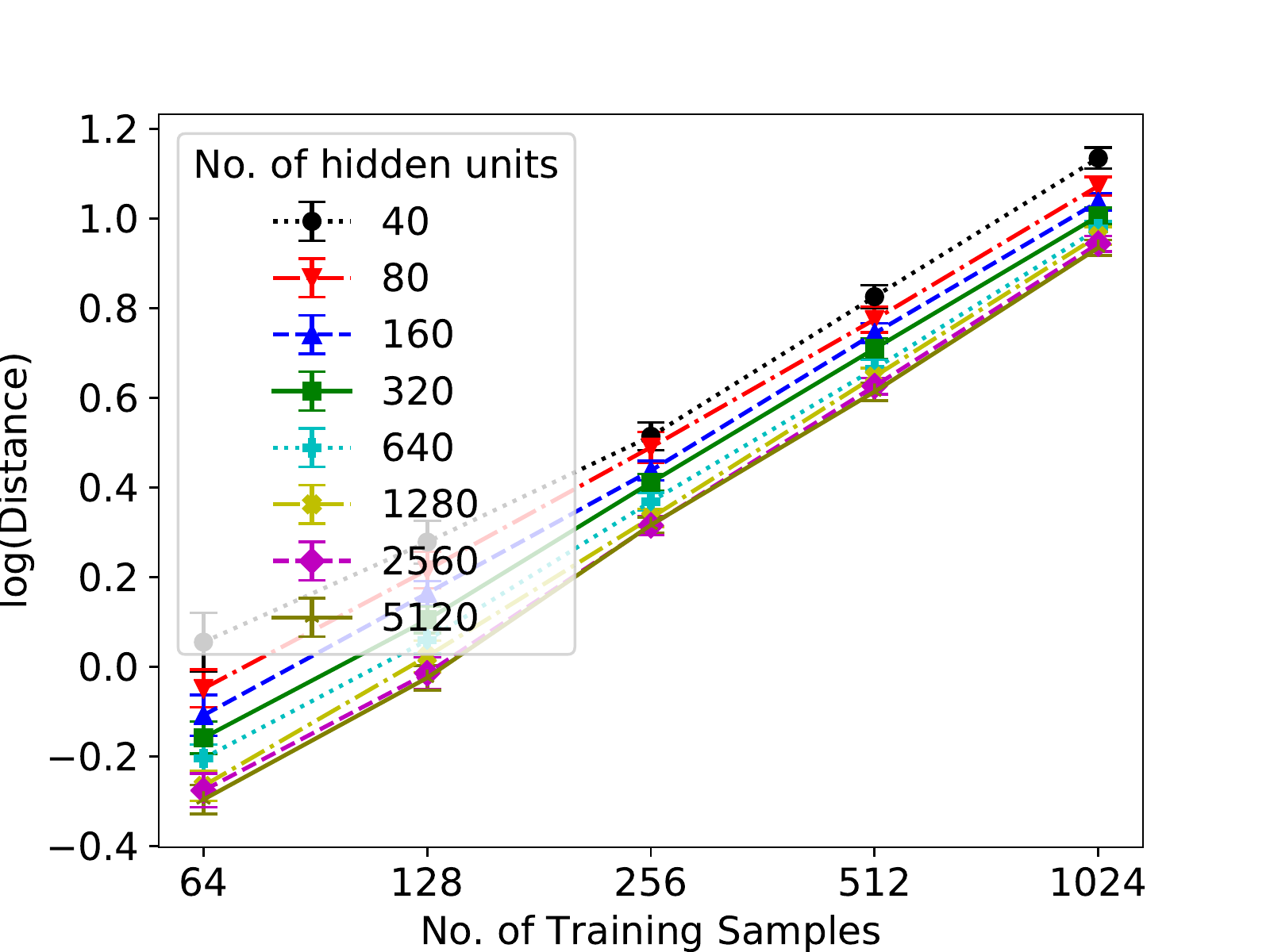}\\
        (b) MNIST
    \end{minipage}
               \begin{minipage}{.25\textwidth}
        \centering
        \includegraphics[scale=0.25]{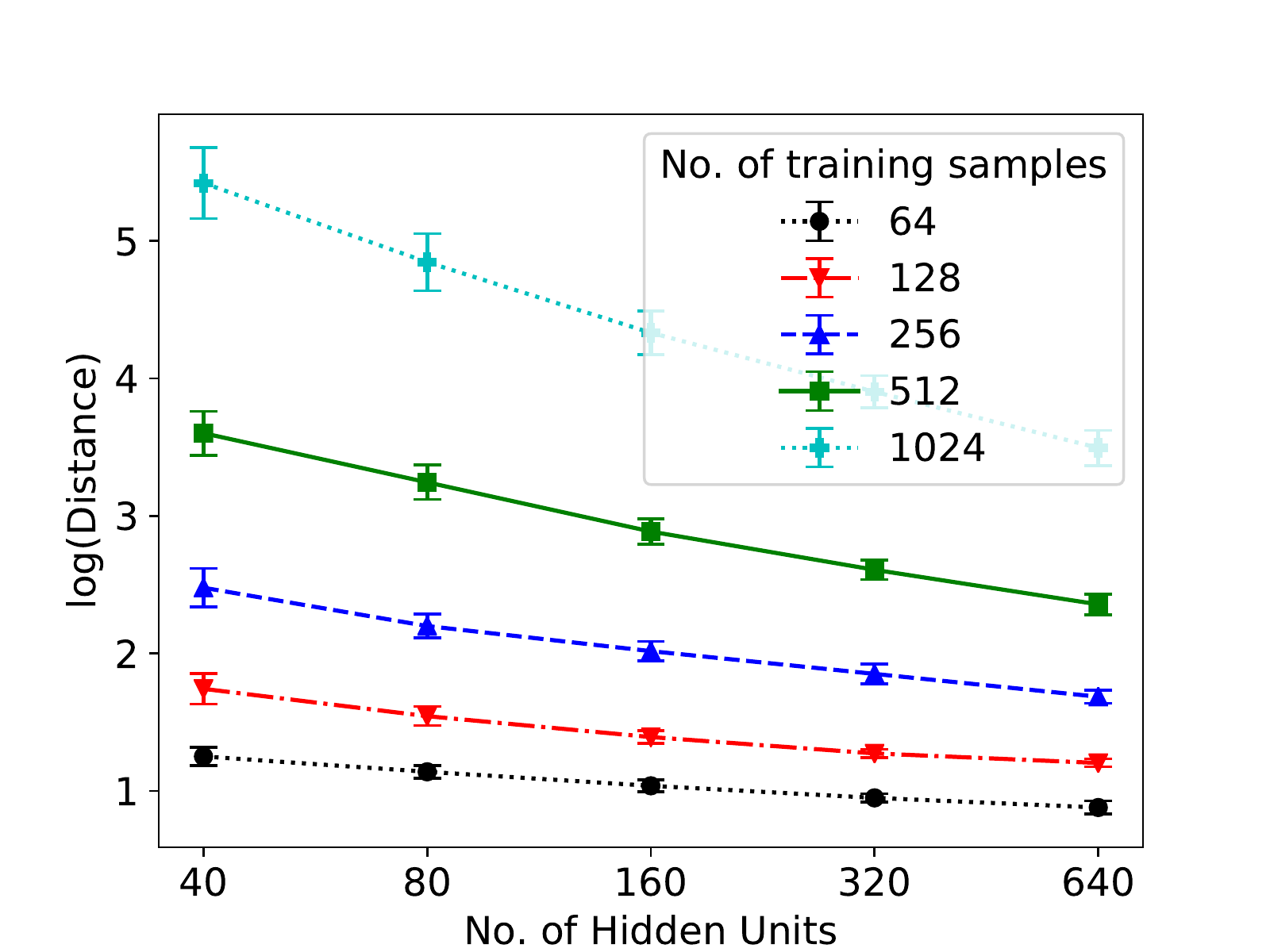} \\
        (c) CIFAR
        \end{minipage}%
        \begin{minipage}{.25\textwidth}
        \centering
        \includegraphics[scale=0.25]{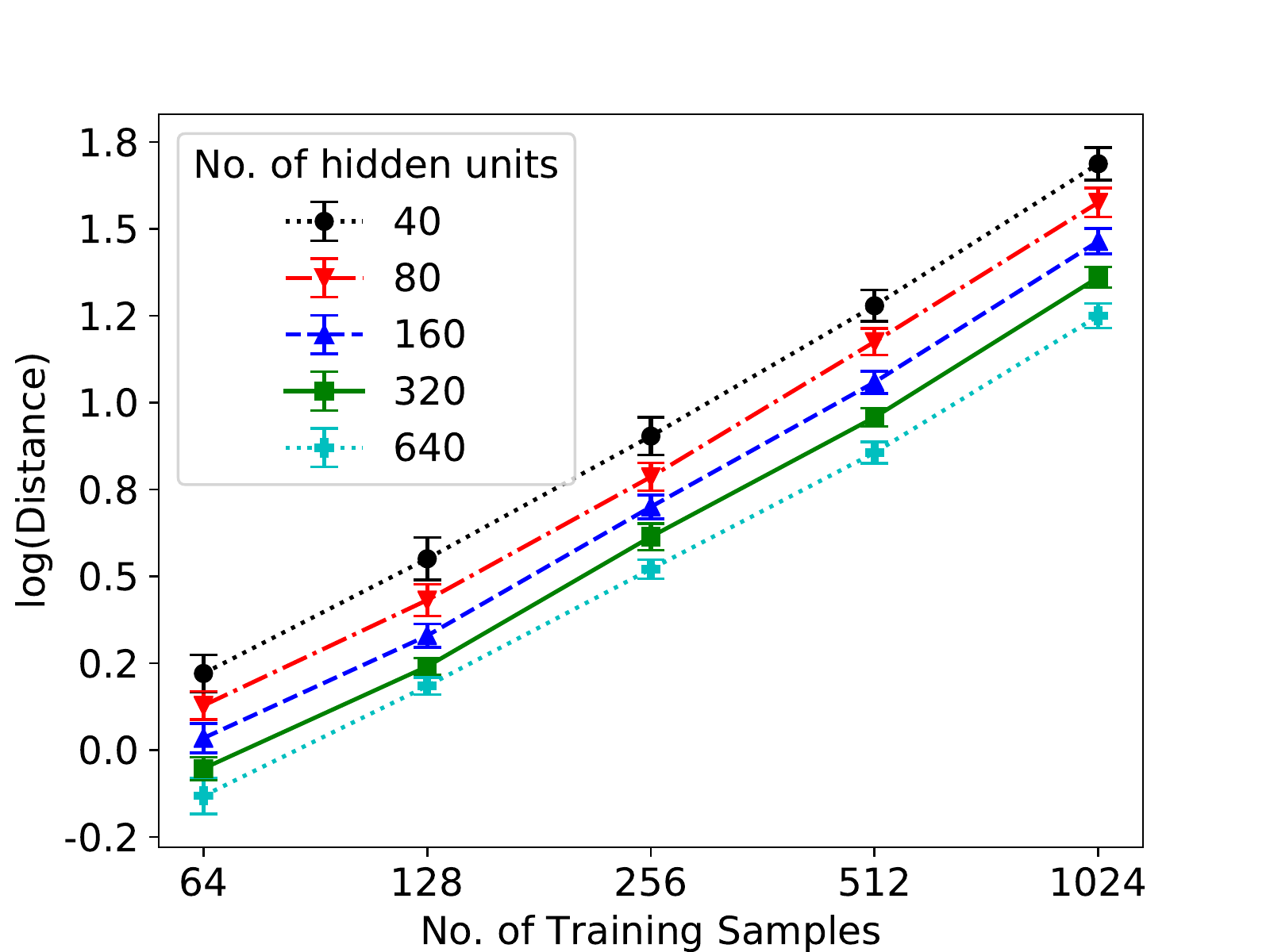}\\
    (d) CIFAR
        \end{minipage}
    \caption{{Distance from initialization for completely noisy labels}}
    \label{fig:init-noise}
\end{figure}

\begin{figure}[t!]
        \begin{minipage}{.25\textwidth}
        \centering
        \includegraphics[scale=0.25]{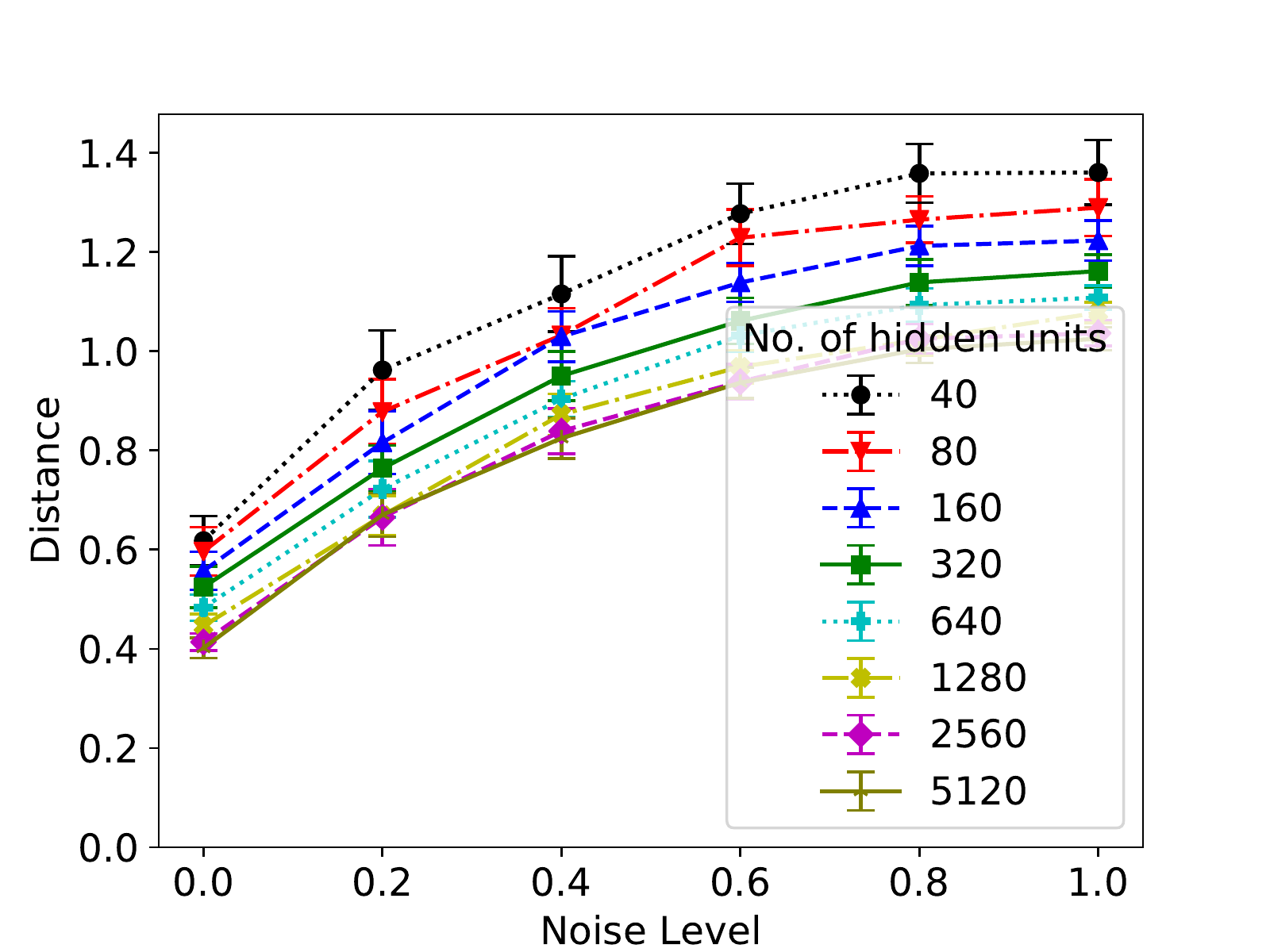}\\
        (a) $m=128$
    \end{minipage}%
        \begin{minipage}{.25\textwidth}
        \centering
        \includegraphics[scale=0.25]{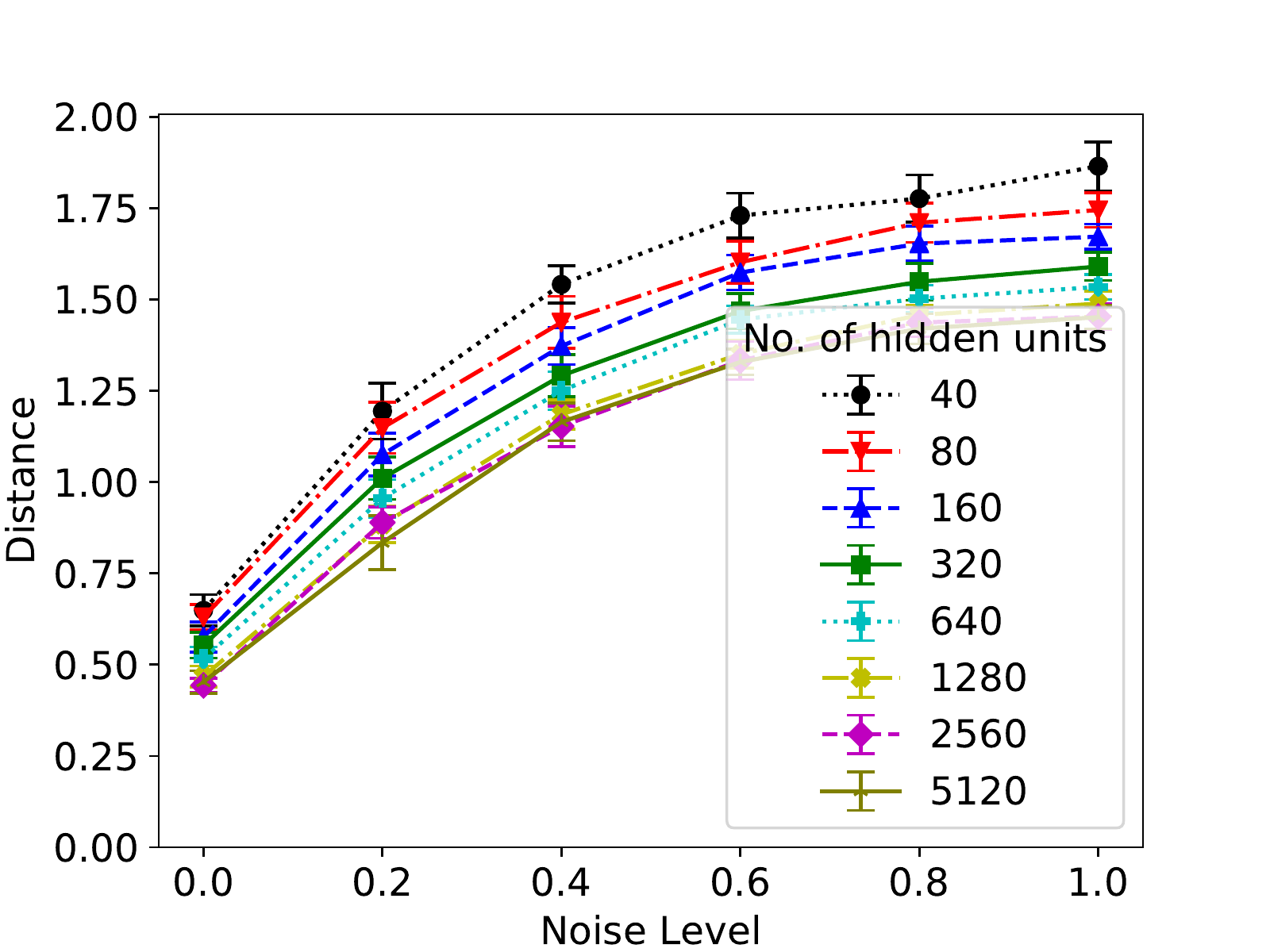}\\
        (b) $m=256$
    \end{minipage}
               \begin{minipage}{.25\textwidth}
        \centering
        \includegraphics[scale=0.25]{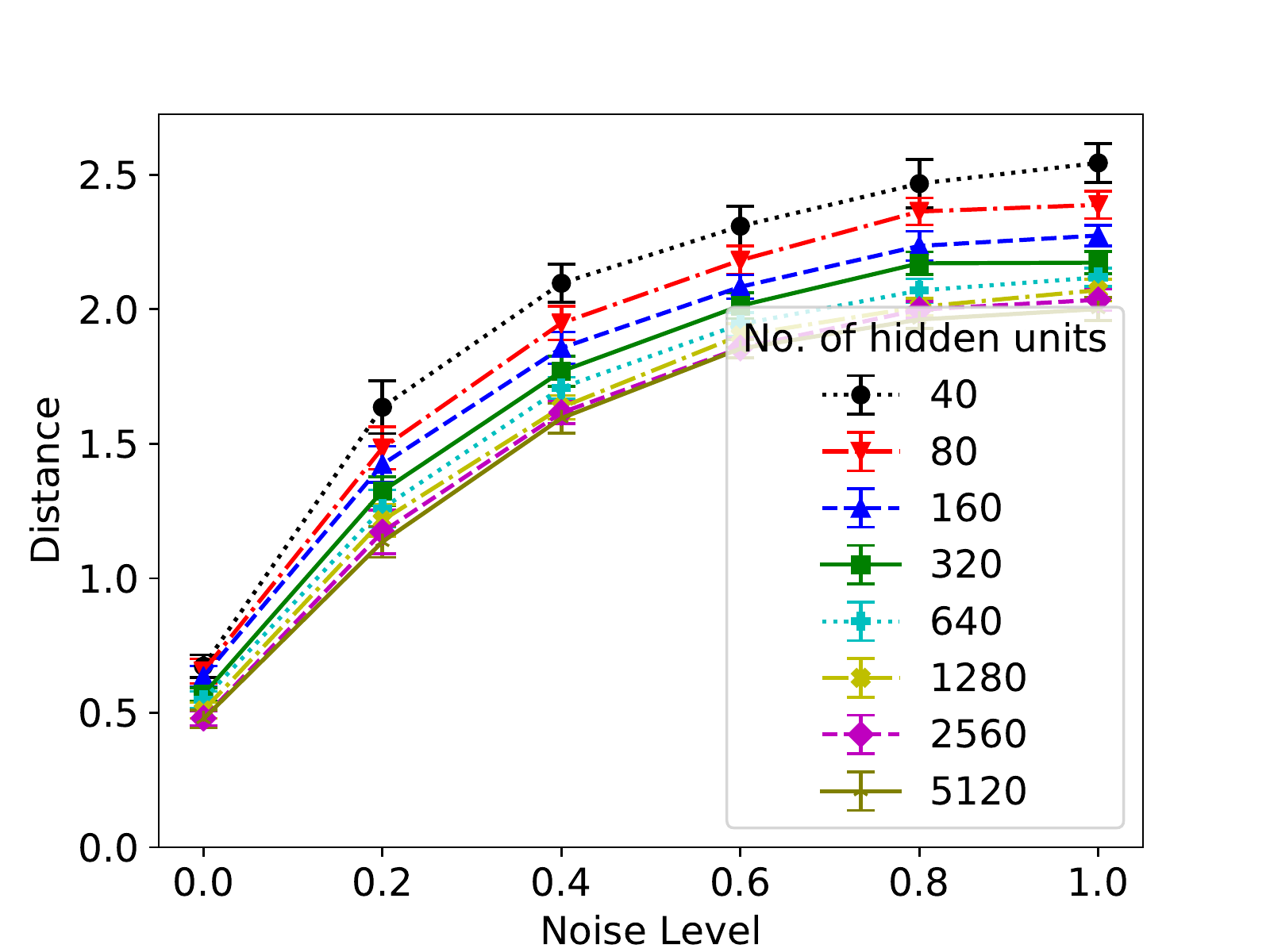} \\
        (c) $m=512$
        \end{minipage}%
        \begin{minipage}{.25\textwidth}
        \centering
        \includegraphics[scale=0.25]{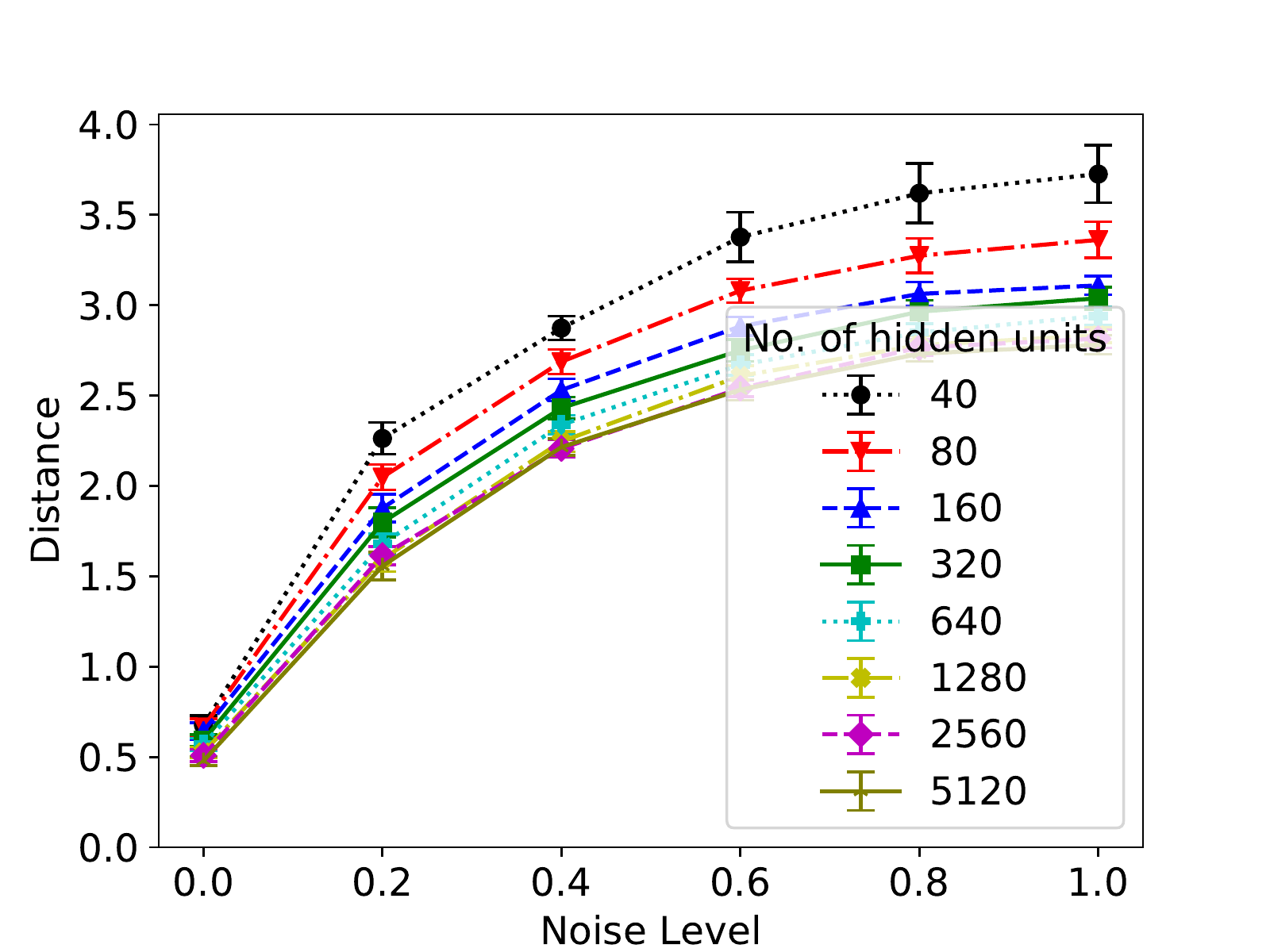}\\
    (d) $m=1024$
        \end{minipage}
    \caption{Distance from initialization for varying levels of noise in the labels}
    \label{fig:init-varied-noise}
\end{figure}

We also investigate how the distances vary when the labels of the datapoints are randomly chosen to be $1$ or $-1$ with equal probability independently (see Figure~\ref{fig:init-noise}).  For MNIST, we train using SGD with learning rate $0.01$ until the loss goes to $0.1$ and for CIFAR, we train using SGD with learning rate $0.001$ until the loss goes to $0.1$. 
In another set of experiments (see Figure~\ref{fig:init-varied-noise}), we select datapoints of just two classes in the MNIST dataset, and corrupt a part of its labels randomly. We train using SGD with learning rate $0.01$ until the loss diminishes to $0.1$. We examine how the distance moved from the random initialization varies with the level of noise.  Note that here the X axis, which is basically the proportion of points that have been corrupted.

%Here again, we notice that the distance from random initialization only decreases with the number of hidden units,  and increase with the number of samples typically between $\mathcal{O}(d^{0.4})$ or $\mathcal{O}(d^{0.5})$.

Note that while we do train the network to around $1\%$ of the initial loss for some of the experiments, for other experiments (including the one involving noisy labels) we only train the network until around $10\%$ of the original loss. While experiments in past work have studied the generalization phenomenon by training to zero or near-zero loss,  we note that it is still interesting to explore generalization without doing so because, even in this setting we still observe the unexplained phenomenon of ``non-increasing (or even, decreasing) generalization errors with increasing width''.

Finally, the $X$ axis in of our plots --- which is typically either the width $H$ or the number of samples $m$ --- is not linear but logarithmic. In some of the plots, we also use a logarithmic $Y$ axis to understand what is the power of $X$ which determines $Y$ i.e., what is $k$ if $Y \propto X^k$. Note that the generalization error only decreases or remains constant as we increase the number of hidden units.

\paragraph{Observations.} First, we observe acros the board, we observe that the distance $r$ mostly remains constant {\em or surprisingly, even decreases} with width $H$ \footnote{Although,sometimes for very large $H$, $r$ shows only a slight increase that scales logarithmically with $H$.} This is true for both MNIST and CIFAR-10, and for both cross-entropy loss and squared error loss, and for both the original labels and for noisy labels.
%  For reference, see Figure~\ref{fig:init-cross-entropy} (left)  and sub-figure (c) in Figures ~\ref{fig:init-mnist-momentum}, ~\ref{fig:init-cifar}, and ~\ref{fig:init-mnist}.

Second,  as we can see in Figures~\ref{fig:init-cross-entropy} (right) and also in Figure~\ref{fig:init-noise} and Figure~\ref{fig:init-varied-noise}, $r$ increases with more noise in the labels, and this increase is more pronounced when sample size $m$ is larger. This demonstrates that larger distances need to be traveled  in order to achieve stronger levels of memorization, implying that distance from initialization is indeed an informative measure of complexity.

  Finally, we note that even though $r$ is regularized to a width-independent value, it does
{\em grow with the training set size $m$}, typically at between the rates of  $m^{0.25}$ to $m^{0.4}$ (when there is no noise in the training data). The growth rate is more prominent for smaller $H$ or when there is more noise in the labels
as is evident from Figure~\ref{fig:init-noise} and Figure~\ref{fig:init-noise}.

\section[Complexity of linear networks]{Complexity of linear networks within fixed distance from initialization}
\label{sec:init-linear-network}
Is the distance regularization observed above sufficient to explain generalization?  While many norm-based generalization bounds have already been derived for ReLU networks \citep{neyshabur15norm,neyshabur18pacbayes} which can be potentially improved with this observation, it seems non-trivial to prove an $H$-independent generalization bound with this observation alone. For example, it is easy to incorporate this quantity (in place of distance from the origin) in PAC-Bayesian analysis such as \cite{neyshabur18pacbayes}, as was already done in \cite{dziugaite17nonvacuous}. While this would result in bounds that are tighter by a factor of $\sqrt{H}$ (because distance from origin grows as $\sqrt{H}$, as we will discuss shortly in Section~\ref{sec:init-bad-norms}), the resulting bound still has dependence on the network width.

As a first step to test the usefulness of the observed distance regularization, we consider a network with all hidden units as simple linear units. If the Rademacher complexity of this space of networks was {\em not} independent of width $H$, then there would be no hope in expecting the same networks but with non-linearities to have a width-independent complexity. Fortunately, we can show that this is not the case for linear networks.

We will consider deep linear networks with biases.  Furthermore, we will assume that the weights are initialized as $(\mathcal{Z}, \mathcal{C})$ where $\mathcal{Z}$ is initialized according to Xavier initialization and $\mathcal{C} = 0$.  We will focus on networks of depth greater than $2$, which would mean that each parameter in $\mathcal{Z}$ is drawn independently from $\mathcal{N}(0,{\Theta}(1/\sqrt{H}))$.  Our proof is based on how a Xavier-initialized network has weight matrices with width-independent spectral norms with high probability. 

%We defer the proof for this to Appendix~\ref{app:proofs} (and present some additional results in Appendix~\ref{sec:init-cool-properties}). \\

\begin{restatable}{theorem}{linearnetwork}
\label{thm:linear-network}
%Assume that the input data is normalized so that with high probability, any $\vec{x}$ drawn at random satisfies $\| \vec{x} \| = \tilde{\mathcal{O}}(1)$.
Consider a network where $\Phi(\cdot)$ is the identity function. Define the distance-from-initialization-bounded class of functions realized by this neural network as:

 \begin{equation}
 \calF_{\|(\cdot,\cdot)-(\calZ,\calC)\|_F \leq r} \coloneqq \{ f_{(\calW, \calB)} | \,  \exists (\calW, \calB) \text{ s.t. } \|(\calW, \calB)-(\calZ,\calC)\|_F \leq r  \}.
 \end{equation}
%when the loss function is such that $\|\mathcal{L}(f_1(\vec{x}, y)) - \mathcal{L}(f_2(\vec{x}, y)) \| = \mathcal{O}(|f_1(\vec{x}) - f_2(\vec{x}))|) $
 The empirical Rademacher complexity of this distance bounded class of functions is independent of the width $H$ and more precisely satisfies:
% corresponding to the hypotheses that are at an $\ell_2$ distance of at most $r$ satisfies:
\begin{equation}
\hat{\calR}_S\left(\calF_{\|(\cdot,\cdot)-(\calZ,\calC)\|_F \leq r} \right) = \tilde{\mathcal{O}} \left( \frac{D c^D (r+1)^{D}\max_i \| \vec{x}_i\|}{\sqrt{m}} \right),
\end{equation}
where $c=\tilde{\Theta}(1)$. \\
\end{restatable}

%TODO : max x must be fixed in proof

\begin{proof}
%TODO : Need to talk about contraction lemma from loss function to $f$
Crucial to our proof is the fact that for the random initialization $\mathcal{Z}$, with high probability, we can bound the spectral norms of all the $H\times H$ matrices in $\mathcal{Z}$ as $\| \vec{Z}_d \|_2 = \tilde{\Theta}(1)$ (for $1 < d \leq D$) and $\| \vec{Z}_1\|_2 =  \tilde{\Theta}(\sqrt{N})$ (where $N$ is the input dimensionality). We present these and a few other relevant bounds in Corollary~\ref{cor:weight-matrix-bounds}.

Now our approach is to remove the network parameters in the expression for the Rademacher complexity layer by layer while applying this bound. For shorthand, we will simply write $\sup$ to denote the supremum over the space ${(\mathcal{W}, \mathcal{B}): \| (\mathcal{Z}, \mathcal{C}) -  (\mathcal{W}, \mathcal{B})\| \leq r }$. Then, we get the following recursive bound for the layer $d > 1$:

\begin{align*}
 \mathbb{E}_{\vec{\xi}} \left[ \sup  \left \| \sum_{i=1}^{m}  \xi_i   \nn{\mathcal{W}, \mathcal{B}}{d}{}{\vec{x}_i}\right\| \right]  & =  \mathbb{E}_{\vec{\xi}} \left[ \sup  \left \| \sum_{i=1}^{m}  \xi_i \left( \vec{W}_{d} \nn{\mathcal{W}, \mathcal{B}}{d-1}{}{\vecx_i} + \vec{b}_{d} \right)\right\| \right] \numberthis \\
 & \leq \mathbb{E}_{\vec{\xi}} \left[ \sup  \left \| \sum_{i=1}^{m}  \xi_i \vec{W}_{d} \nn{\mathcal{W}, \mathcal{B}}{d-1}{}{\vecx_i}  \right\| \right] + \mathbb{E}_{\vec{\xi}} \left[ \sup  \left \| \sum_{i=1}^{m}  \xi_i  \vec{b}_{d} \right\| \right] \numberthis \\   
 & \leq \mathbb{E}_{\vec{\xi}} \left[ \sup \|\vec{W}_{d} \|_2  \left \| \sum_{i=1}^{m}  \xi_i  \nn{\mathcal{W}, \mathcal{B}}{d-1}{}{\vecx_i}  \right\| \right] + \mathbb{E}_{\vec{\xi}} \left[ \sup  \left \| \sum_{i=1}^{m}  \xi_i  \vec{b}_{d} \right\| \right] \numberthis \\
 & \leq \tilde{\mathcal{O}}(\|\vec{W}_d - \vec{Z}_d \|_F + \| \vec{Z}_d \|_2)   \mathbb{E}_{\vec{\xi}} \left[ \sup \left \| \sum_{i=1}^{m}  \xi_i  \nn{\mathcal{W}, \mathcal{B}}{d-1}{}{\vecx_i}  \right\| \right] \\
 & +  \tilde{\mathcal{O}}(\|\vec{b}_d \|_2 )  \mathbb{E}_{\vec{\xi}} \left[ \sup  \left \| \sum_{i=1}^{m}  \xi_i  \right\| \right] \numberthis \\
 & \leq   \tilde{\mathcal{O}}(r + \|\vec{Z}_d \|_2)   \mathbb{E}_{\vec{\xi}} \left[ \left \| \sum_{i=1}^{m}  \xi_i  \nn{\mathcal{W}, \mathcal{B}}{d-1}{}{\vecx_i}  \right\| \right] +  \tilde{\mathcal{O}}(r ) \sqrt{m}. \numberthis \\
\end{align*}

Above, we have used the Khintchine-Kahane inequality (see Theorem~\ref{thm:kk} and Corollary~\ref{cor:kk})  to bound $\mathbb{E}_{\vec{\xi}} \left[ \sup  \left \| \sum_{i=1}^{m}  \xi_i  \right \|  \right] $. 

Finally, for the base case $d=0$, we get:
\begin{align}
\mathbb{E}_{\vec{\xi}} \left[ \sup  \left \| \sum_{i=1}^{m}  \xi_i   \nn{\mathcal{W}, \mathcal{B}}{d}{}{\vec{x}_i}\right\| \right]   = \mathbb{E}_{\vec{\xi}} \left[ \sup  \left \| \sum_{i=1}^{m}  \xi_i   \vec{x}_i \right\| \right]  \leq \mathcal{O}(\sqrt{\sum \|\vec{x}_i \|^2}) \leq \tilde{\mathcal{O}}(\sqrt{m} \max_i \| \vec{x}\|_i).
\end{align}

Here again we have used the Khintchine-Kahane inequality to bound $\mathbb{E}_{\vec{\xi}} \left[ \sup  \left \| \sum_{i=1}^{m}  \xi_i   \vec{x}_i \right\| \right] $. Finally, our claim then follows from repeated application of these recursive bounds. We have included the linear factor of $D$ to account for the term $ \tilde{\mathcal{O}}(r  \sqrt{m} )$ that is added due to the biases in each layer. Similarly the constant $c$ corresponds to the constant within the asymptotic bounds obtained in each recursive application of the above bound. 
\end{proof}

\section[Ineffectiveness of initialization-independent norms]{The ineffectiveness of initialization-independent norms}
\label{sec:init-bad-norms}

We now go back and look at some norms studied in \cite{neyshabur17exploring} and evaluate why they were unable to explain generalization.
First, consider the product of $\ell_2$ norms proportional
%\footnote{In their paper, there are other constant terms independent of $H$ including an extra term that corresponds to the margin of the data which we ignore.} 
to $\prod_{i=1}^{D}\| \vec{W}_d\|_{F}^2$,
% for the possibility of implicit regularization;
which they observe increases with the width $H$ for large $H$. Unfortunately, this could not explain why generalization error is width-independent because the best known bound on the Rademacher complexity of the class of $\ell_2$-norm-bounded networks grows with the norm bound and hence also grows with $H$ \citep{neyshabur15norm}. 
Through the proposition below, we present a more theoretically grounded perspective as to why this norm may not explain generalization: \\

\begin{proposition}
\label{prop:badnorms}
%For a network of $H$ hidden units per layer, with probability at least $1-1/\delta$, 
With high probability over the draws of the random initialization, even though the untrained network provably has a $H$-independent generalization error $\tilde{\mathcal{O}}\left( 1/\sqrt{m}\right)$, its $\ell_2$ norm $\prod_{d=1}^{D}\| \vec{Z}_d \|_{F}^2$ grows as $\tilde{\Omega}(H^{D-2})$. \\
\end{proposition}

The main takeaway from the above proposition is that a norm-based capacity measure for neural networks may not be useful for explaining generalization if it is blind to the random initialization and instead measures any kind of distance of the weights from the origin. This is because, {\em for larger and larger $H$, most random initializations and the origin, all lie farther and farther away from each other}. Therefore, it may not be reasonable to expect that for these initializations, SGD goes all the way close to the origin to find solutions.\\

\begin{proof} \textbf(Proof of Proposition~\ref{prop:badnorms})
In the terminology of Definition~\ref{def:effective-capacity},   the effective capacity when the algorithm is simply one which outputs the initialization itself, is the singleton set consisting of that initialization i.e., $\calF_{m,\delta}[\scrD,(\mathcal{Z},\mathcal{C}),\mathcal{A}] = \{f_(\mathcal{Z}, \mathcal{C}) \}$. The generalization error of this algorithm then follows from applying Hoeffding's inequality (Lemma~\ref{lem:hoeffding-bound}) for bounded i.i.d random variables, with the random variables here being the loss of this network on a $m$ random i.i.d `training' input from the underlying distribution. For standard 0-1 error, this random variable is by default bounded. We can also show that the squared error loss is bounded to a width-independent value, since the output of this randomly initialized network is bounded independent of $H$ (Theorem~\ref{thm:output-bound}). Thus the generalization error of this network is width-independent.

The second part of our claim follows directly from the Frobenius norm bounds in Corollary~\ref{cor:weight-matrix-bounds}.
\end{proof}

%finds solutions that are close to the origin. 

\textbf{On the spectral norm}:  \cite{neyshabur17exploring} also study a spectral norm proportional to $H^{d-1} \prod_{d=1}^{D} \| \vec{W}_d \|_2$. First we note that, like the $\ell_2$ norm above, even this grows with $H$ as $\tilde{\Omega}(H^{d-1})$ for the random initialization, because the spectral norm of the random matrices are $\tilde{\Theta}(1)$.  \cite{neyshabur17exploring} then ask whether the factor of $H^{d-1}$ is ``necessary'' in this measure or not (in Section 2.2 of their paper). Formally, we frame this question as: is it sufficient if $\prod_{d=1}^{D} \| \vec{W}_d\|_2$ is regularized to an $H$-independent value (as against ensuring the same for $H^{d-1} \prod_{d=1}^{D} \| \vec{W}_d\|_2$) for guaranteeing $H$-independent generalization?  

Through our proposition below,  we argue that a more useful question can be asked. In particular, we show that given the observation that distance from initialization is regularized to a $H$-independent value, then one can already conclude that $\prod_{d=1}^{D} \| \vec{W}_d \|_2$. Therefore, the bounds on the distance from initialization is a stronger form of implicit bias. Hence, we would rather want to answer whether this stronger form of bias is sufficient to derive -independent generalization. Effectively, this would 
boil down to extending Theorem~\ref{thm:linear-network} to non-linear networks. \\

\begin{restatable}{proposition}{spectral}
$\prod_{d=1}^{D} \| \vec{W}_d \|_2 \leq \tilde{\mathcal{O}}\left(c^D\left(1+\|(\mathcal{W}, \mathcal{B}) - (\mathcal{Z},\mathcal{C})\|_F\right)^{D} \right)$ for some $c = \tilde{\mathcal{O}}(1)$.\\
\end{restatable}

\begin{proof}

For any $d$, we have that $\| \vec{W}_{d} \|_2 \leq \| \vec{Z}_{d}\|_2 + \|\vec{Z}_d - \vec{W}_d \|_F \leq  \| \vec{Z}_{d}\|_2+ \|(\mathcal{W}, \mathcal{B}) - (\mathcal{Z},\mathcal{C})\|_F$.  The result then immediately follows from the spectral norm bounds in Corollary~\ref{cor:weight-matrix-bounds}. 

\end{proof}

\section[Width-independent properties of ReLU networks]{Some width-independent properties of distance-bounded ReLU networks}
\label{sec:init-cool-properties}
In this section we lay out two useful properties of neural networks in terms of how far away their weight are from their random initialization. In particular, we show that both the output and the gradient of a network with respect to its parameters is bounded purely by the distance from its random initialization and not on the number of hidden units. As always, we assume that the initialization $\mathcal{Z}$ is according to Xavier initialization (i.e., in this case the weights are drawn from a zero-mean gaussian with standard deviation $\mathcal{O}({1}/{\sqrt{H}})$) and $\mathcal{C}$ is zero.  Note that in the proofs in this section, we will drop the symbols $(\calW, \calB)$ from the expression for the network $f$ to avoid clutter. 

\begin{theorem}
\label{thm:output-bound}
$|f_{(\mathcal{W}, \mathcal{B})}(\vec{x})| \leq  \tilde{\mathcal{O}}(c^D(r+1)^{D}(\|\vec{x} \|+1)) $, where $r = \| (\mathcal{Z}, \mathcal{C}) - (\mathcal{W},\mathcal{B}) \|_F$ and $c=\tilde{\mathcal{O}}(1)$.
\end{theorem}

\begin{proof}
We will bound the magnitude of the output $\nn{}{d}{}{\vecx}$ as follows: 

\begin{align}
\| \nn{}{d}{}{\vecx} \| & =  \| \vec{W}_{d}\relu{\nn{}{d-1}{}{\vecx}} + \vec{b}_{d} \|  \leq \|\vec{W}_{d}\relu{\nn{}{d-1}{}{\vecx}} \| + \|\vec{b}_d \|\\ 
& \leq \|\vec{W}_d \|_2 \| \relu{\nn{}{d-1}{}{\vecx}} \|_2 + r \\
& \leq (\|\vec{W}_d  - \vec{Z}_d \|_F + \| \vec{Z}_d \|_2) \| \nn{}{d-1}{}{\vecx} \|_2 + r \\
& \leq \mathcal{O}(r + 1 ) \| \nn{}{d-1}{}{\vecx} \|_2 + r. 
\end{align}

In the third line above, we use the fact that for any scalar value, $|\relu{x}| \leq |x|$ as $\relu{}$ is the ReLU activation. Following that, we use the spectral norm bounds from Corollary~\ref{cor:weight-matrix-bounds}. 
Our bound then follows from repeated applications of these bounds recursively. Note that the value $c$ corresponds to the constant present in the asymptotic bound applied in each recursion.
\end{proof}

As a corrollary, we can bound the initial squared error loss of the network on a set of datapoints, independent of $H$:
\begin{corollary}
\label{cor:loss-bound}
Let ${(\vec{x}_1, {y}_1), \hdots, (\vec{x}_m, {y}_m)}$ be a set of training datapoints. For a randomly initialized network of any size, with high probability, the initial loss can be bounded independent of $H$ as
\begin{equation}
\frac{1}{m}\sum_{i=1}^{m} (f_{(\mathcal{Z}, \mathcal{C})}(\vec{x}) - y)^2 \leq \left(\tilde{\mathcal{O}}(c^D (\max_{i}\|\vec{x}_i \|+1)) + \max_i |y_i| \right)^2.
\end{equation}
\end{corollary}

Next, we bound the gradient of the function with respect to the parameters $\mathcal{W}$, independent of $H$. 
\begin{theorem}
\begin{equation}
\left\| \frac{ \partial f_{(\mathcal{W}, \mathcal{B})}(\vec{x})}{\partial \mathcal{W}}\right\|   \leq  \tilde{\mathcal{O}}( D c^D(r+1)^{D}(\|\vec{x} \|+1)) 
\end{equation}
where $r = \| (\mathcal{Z}, \mathcal{C}) - (\mathcal{W},\mathcal{B}) \|_F$ and $c=\tilde{\mathcal{O}}(1)$. 
\end{theorem}

\begin{proof}
The derivative with respect to $\vec{W}_D$ is easy to bound:
\begin{align}
\left\|\frac{ \partial f_{(\mathcal{W}, \mathcal{B})}(\vec{x})}{\partial \vec{W}_D} \right\| =  \|\relu{\nn{\calW, \calB}{D-1}{}{\vecx}}\| \leq \|f^{(D-1)}_{(\mathcal{W}, \mathcal{B})} (\vec{x})\|.
\end{align}

Above, we make use of the fact that for any scalar value $u$, $|\relu{u}| \leq |u|$. After applying the above inequality, $\|\nn{\calW, \calB}{d-1}{}{\vecx}\|$ can be bounded by the recursive bounds presented  in the proof of Theorem~\ref{thm:output-bound}. 

Next, for $\vec{W}_{d}$, we have that: 

\begin{align}
\left\|\frac{ \partial f_{(\mathcal{W}, \mathcal{B})}(\vec{x})}{\partial \vec{W}_d} \right\| &\leq  \left\| \vec{W}_{d} \frac{ \partial \relu{\nn{\calW, \calB}{d-1}{}{\vecx}}}{\partial \vec{W}_d}\right\| \\
& \leq \| \vec{W}_d\|_2 \left\| \frac{ \partial \relu{\nn{\calW, \calB}{d-1}{}{\vecx}}}{\partial \vec{W}_d}\right\|  \\
&= \tilde{\mathcal{O}}\left((1+r)\left\| \frac{ \partial \relu{\nn{\calW, \calB}{d-1}{}{\vecx}}}{\partial \vec{W}_d}\right\|  \right).
\end{align}
We have used the bound $\| \vec{W}_d\|_2 = \|\vec{Z}_d\|_2 + \| \vec{W}_d - \vec{Z}_d\|_F \leq \tilde{\mathcal{O}}(1+r)$.
Note that the last term above contains the derivative of a vector with respect to a matrix, the norm of which is essentially the norm of the gradient corresponding to every pair of term from the vector and the matrix. Now, to bound this term, we need to consider the case where $d=D-1$ and the case where $d < D-1$. However, instead of deriving the derivative for these particular cases, we will consider two more general cases, the first of which is below:
\begin{align}
\left\| \frac{ \partial \relu{\nn{}{d}{}{\vecx}} }{\partial \vec{W}_{d}}\right\| 
 &= 
 \left\| \frac{\partial \relu{\vec{W}_{d} \relu{\nn{}{d-1}{}{\vecx}}}}{\partial \vec{W}_{d}} \right\|  \\
 &=  \left\|   \relu{}'\left( \vec{W}_{d} \relu{\nn{}{d-1}{}{\vecx}}\right) \circ \relu{\nn{}{d-1}{}{\vecx} }\right\| \\
 & \leq \|  \relu{\nn{}{d-1}{}{\vecx}}  \leq \|\nn{}{d-1}{}{\vecx} \| 
\end{align}

Here, $\circ$ denotes the element-wise product of two vectors. The last inequality follows from the fact that $\relu{}'$ is either $0$ or $1$ when $\relu{}$ is a ReLU activation. We can bound $\|  \nn{}{d-1}{}{\vecx}\| $ with the recursive bounds presented in Theorem~\ref{thm:output-bound}.\\

Next, we consider the following case that remains, where $l < d$ and $d > 1$:
 \begin{align}
 \left\| \frac{ \partial \relu{ \nn{}{d}{}{\vecx}} }{\partial \vec{W}_{l}}\right\| 
 & = \left\| \frac{ \partial \relu{ \vec{W}_{d} \relu{\nn{}{d-1}{}{\vecx}}}}{\partial \vec{W}_{l}} \right\| \\
 &=  \left\|\relu{}'( \vec{W}_{d} \relu{\nn{}{d-1}{}{\vecx}})  \circ  \vec{W}_{d}  \frac{ \partial  \relu{\nn{}{d-1}{}{\vecx} }}{\partial \vec{W}_{l}}\right\| \\
 & \leq   \| \vec{W}_{d}\|_2\left\|   \frac{\partial  \relu{\nn{}{d-1}{}{\vecx}}}{\partial \vec{W}_{l}}\right\| \\
 & \leq \tilde{\mathcal{O}} \left( (1+r)\left\| \frac{\partial \relu{\nn{}{d-1}{}{\vecx} }}{\partial \vec{W}_{l}} \right\|\right).
 \end{align}

For the sake of simplicity, we have abused notation here: in particular, in the second equality we have used $\circ$ to denote that each term in the first vector $\relu{}'$ is multiplied with a corresponding row in $\vec{W}_d$. Since the first vector is 0-1 vector, this results in a matrix with some rows zeroed out. The next inequality follows from the fact that the spectral norm of such a partially-zeroed-out matrix is at most the spectral norm of the original matrix.

Through these recursive bounds, we arrive at our claim.

\end{proof}

\section{Conclusion}
\label{sec:init-conclusions}
To explain generalization in deep networks, we highlight the need to understand the effective capacity of a model for a given random initialization of the network. Furthermore, our experiments suggest that distance moved by the training algorithm from its random initialization is a key form of implicit bias. This leads to multiple concrete open questions.
First, why is distance from the initialization regularized by the training algorithm? Can we precisely bound this distance independent of the number of hidden units, $H$? 
Next, is this observation alone sufficient to explain generalization? More concretely, can we prove an $H$-independent bound on the empirical Rademacher complexity (or any other learning-theoretic complexity) for distance-regularized networks, like we could for linear networks in Theorem~\ref{thm:linear-network}?
If that is not possible, can we identify a more precise characterization of the effective capacity as defined in Definition~\ref{def:effective-capacity}? That is, {for a fixed random initialization, do the solutions obtained by the training algorithm on most training sets lie within a smaller subspace inside a ball of $H$-independent radius around the random initialization?}

\chapter{Noise-Resilience of Deep Networks}
\label{chap:noise-resilience}

\section{Introduction}
\label{sec:nr-intro}
In the previous chapter, we looked at distance from initialization as a way of quantifying the implicit bias of the training algorithm. Another interesting notion of inductive bias that has been empirically linked to generalization is that of the width of the minimum. Specifically, it has been observed that stochastic gradient descent (SGD) tends to find solutions that lie in ``flat, wide minima'' in the training loss \citep{hochreiter97flat,hinton93mdl,keskar17largebatch}. Over the course of the next few chapters, we will work towards deriving a generalization bound that take into account both these notions of implicit biases. 

The notion of flatness (or the sharpness) of the minimum in particular can be incorporated neatly via PAC-Bayesian techniques. Recall from Section~\ref{sec:prelim-pac-bayes} that PAC-Bayesian bounds hold only on a stochastic classifier. However, we {\em can} derandomize these bounds to say something about a deterministic classifier. In particular, if the deterministic classifier is resilient to perturbations in its parameter, the derandomized bound is tighter. Noise-resilience is a consequence of flatness: if the minimum that is found is flat, then random perturbations in parameter are less likely to affect the behavior of the network. \\

In this chapter, we specifically provide theoretical bounds on the noise-resilience of the deep network in different aspects. These bounds are not generalization bounds. Rather in each of these bounds, we will fix a particular input point $(\vec{x},y)$. Then, we will look at some quantity related to the network evaluated at that point, such as the output of the network, or the pre-activation value of a particular unit $h$ at a particular layer $d$, or the Frobenius norm sof its active weight matrices. Then, we will consider independent and identical Gaussian perturbations on the network parameters, and examine how much perturbation these quantities suffer, with high probability over the random parameter perturbations. From here on, we will refer to these quantities as {\em properties} of the network (sometimes, {\em input-dependent} properties).

A key feature of all our noise resilience bounds is that, unlike other naive analyses, {\em they do not involve the product of the spectral norm of the weight matrices}.  This will be of relevance in a few chapters from now (Chapter~\ref{chap:deterministic-pacbayes}) since it will save us an exponential factor in the final generalization bound (when compared to other existing generalization bounds).  

 Instead of spectral-norm-products, our bounds
%quantity $T(\W, \vec{x}, y)$ 
will be in terms of i) the magnitude of the some other ``preceding'' properties (typically, these are properties of the lower layers) of the network, and ii) how those preceding properties themselves respond to perturbations. For example, an upper bound in the perturbation of the $d$th layer's output would involve the $\ell_2$ norm of the lower layers $d' < d$, and how much they would blow up under these perturbations.

The results in this chapter have previously been published in \cite{nagarajan18deterministic}.

\section{Some notations.}
To formulate our results statement succinctly, we design a notation wherein we define a set of ``tolerance parameters'' which we will use to denote the extent of perturbation suffered by a particular property of the network.

Let $\tolset$  denote a ``set'' (more on what exactly we mean by a set below) of positive tolerance values, consisting of the following elements:

\begin{enumerate}
\item $\toloutput{d}$, for each layer $d=1,\hdots, D-1$ (a tolerance value for the $\ell_2$ norm of the output of layer $d$)
\item $\tolpreact{d}$ for each layer $d=1,\hdots, D$ (a tolerance value for the magnitude of the pre-activations of layer $d$)
\item  $\toljacob{d'}{d}$ for each layer $d= 1, 2, \hdots, D$, and $d' = 1, \hdots, d$ (a tolerance value for the $\ell_2$ norm of each row of the Jacobians at layer $d$)
\item  $\tolspec{d'}{d}$ for each layer $d= 1, 2, \hdots, D$, and $d' = 1, \hdots, d$ (a tolerance value for the spectral norm of the Jacobians at layer $d$)
\end{enumerate}

{\bf Notes about (abuse of) notation:} 
\begin{itemize}
\item We call $\tolset$  a `set' to denote a group of related constants into a single symbol. Each element in this set has a particular semantic associated with it, unlike the standard notation of a set, and so when we refer to, say $\toljacob{d'}{d} \in \tolset$, we are indexing into the set to pick a particular element. 
\item We will use the subscripted $\tolset_d$ to index into a subset of only those tolerance values corresponding to layers from $1$ until $d$. \\
\end{itemize}

Next we define two events. The first event formulates the scenario that for a given input, a particular perturbation of the weights until layer $d$ brings about very little change in the properties of these layers (within some tolerance levels). The second event formulates the scenario that the perturbation did not flip the {\em activation states} of the network.

\begin{definition}
\label{def:tolerance-parameters}
Given an input $\vec{x}$, and an arbitrary set of constants $\tolset'$, for any perturbation $\U$ of $\W$, 
we denote by $\boundedperturbationevent(\W\focus{+\U},\tolset', \vec{x})$ the event that:
\begin{itemize}
	\item for each $\toloutput{d}' \in \tolset'$, the perturbation in the $\ell_2$ norm of layer $d$ activations is bounded as $\ellone{\elltwo{\nn{\W}{d}{}{\vec{x}}} - \elltwo{\nn{\W\focus{+\U}}{d}{}{\vec{x}}}} \leq \toloutput{d}'$.
	\item for each $\tolpreact{d}' \in \tolset'$, the maximum perturbation in the preactivation of hidden units on layer $d$ is bounded as $\max_h \ellone{{\nn{\W}{d}{h}{\vec{x}}} - {\nn{\W\focus{+\U}}{d}{h}{\vec{x}}}}\leq \tolpreact{d}'$.
	\item for each $\toljacob{d'}{d}' \in \tolset'$, the  maximum perturbation in the $\ell_2$ norm of a  row of the Jacobian  $\verbaljacobian{d'}{d}$ is bounded as $\max_h \ellone{\elltwo{\jacobian{\W}{d'}{d}{h}{\vec{x}}}-\elltwo{\jacobian{\W\focus{+\U}}{d'}{d}{h}{\vec{x}}}} \leq \toljacob{d'}{d}'$.
	\item for each $\tolspec{d'}{d}' \in \tolset'$, the  perturbation in the spectral norm of  the Jacobian  $\verbaljacobian{d'}{d}$ is bounded as $\ellone{\spec{\jacobian{\W}{d'}{d}{}{\vec{x}}}-\spec{\jacobian{\W\focus{+\U}}{d'}{d}{}{\vec{x}}}} \leq \tolspec{d'}{d}'$.
\end{itemize} 
\end{definition}

\paragraph{Note:} If we supply only a subset of $\tolset$ (say $\tolset_d$ instead of the whole of $\tolset$) to the above event, $\boundedperturbationevent(\W+\U,\cdot, \vec{x})$, then it would denote the event that the perturbations suffered by only that subset of properties is within the respective tolerance values.\\

Next, we define the event that the perturbations do not affect the activation states of the network. %(and so the perturbation is within a `linear' neighborhood around the network's parameters.)

\begin{definition}
For any perturbation $\U$ of the matrices $\W$, let  $\unchangedacts_{d}(\W+\U, \vec{x})$ denote the event that none of the activation states of the first $d$ layers change on perturbation. 
\end{definition}

\section{Noise-resilience lemma.}

%Due to the above dependencies, later in our generalization analysis, we will be able apply the noise-resilience bound on a particular property only after having guaranteed that the `preceding properties' participating in its noise-resilience bound are guaranteed to be i) norm-bounded and also ii) noise-resilient under perturbations. To facilitate this, we state our bounds in a recursive fashion. To formulate our recursion statement succinctly,

%Our results here are styled similar to the equations required by Equation~\ref{eq:generic-noise-resilience} presented in the main paper.

In the following lemma, we provide noise-resilience bounds for every property listed in the previous section. For a given input point and for a particular property of the network, roughly, we bound the
the probability that a perturbation affects that property while none of the ``preceding''
 properties (e.g., the previous layer properties) themselves are perturbed beyond a certain tolerance level -- we bound this particular combination of events since we will require such bounds later when we develop our PAC-Bayesian framework in Chapter~\ref{chap:datadependent-pacbayes}. Also note that while we won't explicitly write down which property precedes which, there is a clear ordering that can be inferred from the noise-resilience bounds in the following lemma.

\begin{lemma}
\label{lem:noise-resilience-induction}
Fix a set of constants $\tolset$ that denote the amount of perturbation in the properties preceding a considered property.  For any $\toldelta > 0$, below define a set of constants $\tolset'$ which will act a bound on the perturbation of a considered property. These constants are written in terms of $\tolset$ and the variance in the Gaussian parameter perturbation $\sigma$ as follows.  For all $d=1,2,\hdots, D$ and for all  $d'= d-1, \hdots, 1$ 
\begin{align*}
\toloutput{d}' & := \sigma   \sum_{d'=1}^{d} \frob{\jacobian{\W}{d'}{d}{}{\vec{x}}} \left(\|{\nn{\W}{d'-1}{}{\vec{x}}}\| + \toloutput{d'-1}\right) \sqrt{2 \ln \frac{2DH}{\toldelta}}  \numberthis\\
\tolpreact{d}' & := \sigma \sum_{d'=1}^{d}  \matrixnorm{\jacobian{\W}{d'}{d}{}{\vec{x}}}{2,\infty} \left(\|{\nn{\W}{d'-1}{}{\vec{x}}}\| + \toloutput{d'-1}\right) \sqrt{2 \ln \frac{2DH}{\toldelta}}  \numberthis \\
\toljacob{d'}{d}'&:= \sigma\left( \frob{ \jacobian{\W}{d'}{d-1}{}{\vec{x}}} + \toljacob{d'}{d-1} \sqrt{H}\right)\sqrt{4\ln \frac{DH}{\toldelta}} \\
& + \sigma \sum_{d''=d'+1}^{d-1} \matrixnorm{W_d}{2,\infty} \spec{\jacobian{\W}{d''}{d-1}{}{\vec{x}}}
  \left( \frob{ \jacobian{\W}{d'}{d''-1}{}{\vec{x}}} + \toljacob{d'}{d''-1} \sqrt{H}\right)\sqrt{4\ln \frac{DH}{\toldelta}} \numberthis \\
\tolspec{d'}{d}'&:=  \sigma \sqrt{H}  \left( \spec{ \jacobian{\W}{d'}{d-1}{}{\vec{x}}} + \tolspec{d'}{d-1}\right)\sqrt{2\ln \frac{2DH}{\toldelta}}\\
& + \sigma \sqrt{H} \sum_{d''=d'+1}^{d-1} \spec{\W_d}\spec{\jacobian{\W}{d''}{d-1}{}{\vec{x}}} \left( \spec{ \jacobian{\W}{d'}{d''-1}{}{\vec{x}}} + \tolspec{d'}{d''-1}\right)\sqrt{2\ln \frac{2DH}{\toldelta}} \numberthis\\
\toloutput{0}' &= 0 \numberthis\\
\toljacob{d}{d}' & := 0 \numberthis\\
\tolspec{d}{d}' & := 0 \numberthis.
\end{align*}
Let $\U_d$ be sampled entrywise from $\N(0,\sigma^2)$ for any $d$. 
Then, the following statements hold good:
\textbf{1. Bound on perturbation of  of $\ell_2$ norm of the output of layer $d$.} For all $d = 1, 2, \hdots, D$,
\begin{align*}
  & \mathbb{P}_{\U} \Big[ 
 \lnot  \boundedperturbationevent(\W+\U, {\{ \toloutput{d}' \}},\vec{x}) \; \wedge \\
  & {\boundedperturbationevent(\W+\U,\tolset_{d-1}\bigcup \{ \toljacob{d'}{d} \}^{d}_{d'= 1},\vec{x}) \; \wedge \; \unchangedacts_{d-1}(\W+\U,\vec{x})}  \Big] \leq  \toldelta. \numberthis
\end{align*}
\textbf{2. Bound on perturbation of pre-activations at layer $d$.} For all $d = 1, 2, \hdots, D$,
\begin{align*}
& \mathbb{P}_{\U} \Big[ 
 \lnot \boundedperturbationevent(\W+\U,{ \{ \tolpreact{d}' \}}
 ,\vec{x}) \; \wedge \\
 & {\boundedperturbationevent(\W+\U,\tolset_{d-1}\bigcup \{ \toljacob{d'}{d} \}^{d}_{d'=1} \bigcup {\{ \toloutput{d} \}} ,\vec{x})  \; \wedge \; \unchangedacts_{d-1}(\W+\U,\vec{x})}  \Big] \leq  \toldelta. \numberthis
\end{align*}
\textbf{3. Bound on perturbation of $\ell_2$ norm  on the rows of the Jacobians $\verbaljacobian{d'}{d}$.} 
\begin{align*}
 & \mathbb{P}_{\U} \Big[ 
 \lnot \boundedperturbationevent(\W+\U, {\{ \toljacob{d'}{d}'\}_{d'=1}^{d}}   ,\vec{x}) \; \wedge \\
  & 
 \boundedperturbationevent(\W+\U,\tolset_{d-1},\vec{x}) \; \wedge \; \unchangedacts_{d-1}(\W+\U,\vec{x}) \Big] \leq  \toldelta. \numberthis
\end{align*}
\textbf{4. Bound on perturbation of spectral norm of the Jacobians $\verbaljacobian{d'}{d}$.}
\begin{align*}
 & \mathbb{P}_{\U} \Big[ 
 \lnot \boundedperturbationevent(\W+\U, {\{ \tolspec{d'}{d}'\}_{d'=1}^{d}}   ,\vec{x}) \; \wedge \\
  & 
 \boundedperturbationevent(\W+\U,\tolset_{d-1},\vec{x}) \; \wedge \; \unchangedacts_{d-1}(\W+\U,\vec{x}) \Big] \leq  \toldelta. \numberthis
\end{align*}

\end{lemma}

\section[Proof]{Proof of Lemma~\ref{lem:noise-resilience-induction}}
\begin{proof}
For the most part of this discussion, we will consider a perturbed network where all the hidden units  are frozen to be at the same activation state as they were at, before the perturbation. We will denote the weights of such a network by $\W [+\U]$ and its output at the $d$th layer by $\nn{\W[+\U]}{d}{}{\vec{x}}$. % At the end of the analysis, we will extend our results  to the perturbed network $\W+\U_{d}$ without such `frozen activation states' by accounting for the probability mass of those values of $\U_{d}$ for which the activation states do get flipped. 
By having the activations states frozen, the Gaussian perturbations propagate linearly through the activations, effectively remaining as Gaussian perturbations; then, we can enjoy the well-established properties of the Gaussian even after they propagate.

%Note that will prove the parts of the above lemma in an order different from how it is stated; specifically, we will study the perturbation of the Jacobian's row-wise $\ell_2$ norm the last, as it is slightly more involved. % However, the reason for stating the lemma in that order is to keep it consistent with the order in which our generalization analysis applies parts of it later.

\paragraph{Perturbation bound on the $\ell_2$ norm of layer $d$.} We bound the change in the $\ell_2$ norm of the $d$th layer's output by  applying a triangle inequality\footnote{Specifically, for two vectors $\vec{a}, \vec{b}$, we have from triangle inequality that $\elltwo{\vec{b}} \leq  \elltwo{\vec{a}} + \elltwo{\vec{b}-\vec{a}}$ and $\elltwo{\vec{a}} \leq \elltwo{\vec{b}} + \elltwo{\vec{a}-\vec{b}}$. As a result of this, we have: $-\elltwo{\vec{a}-\vec{b}} \leq \elltwo{\vec{a}} - \elltwo{\vec{b}} \leq \elltwo{\vec{a}-\vec{b}}$. We use this inequality in our proof.} after splitting it into a sum of vectors. Each summand here (which we define as $\vec{v}_{d'}$ for each $d' \leq d$) is the difference in the $d$th layer output on introducing noise in weight matrix $d'$ after having introduced noise into all the first $d'-1$ weight matrices.

\begin{align}
\ellone{\elltwo{\nn{\W[+\U_{d}]}{d}{}{\vec{x}}} - \elltwo{\nn{\W}{d}{}{\vec{x}} }  } & \leq \elltwo{\nn{\W[+\U_{d}]}{d}{}{\vec{x}} - \nn{\W}{d}{}{\vec{x}}}.  \\
\intertext{Since the activations are ReLU, we can replace this with the perturbation of the pre-activation as} 
\ellone{\elltwo{\nn{\W[+\U_{d}]}{d}{}{\vec{x}}} - \elltwo{\nn{\W}{d}{}{\vec{x}} }  }   & \leq \elltwo{\prenn{\W[+\U_{d}]}{d}{}{\vec{x}} - \prenn{\W}{d}{}{\vec{x}}} \\
& \leq \elltwo{\sum_{d'=1}^{d} \left( \underbrace{\prenn{\W[+\U_{d'}]}{d}{}{\vec{x}} - \prenn{\W[+\U_{d'\focus{-1}}]}{d}{}{\vec{x}}}_{:= \vec{v}_{d'}} \right)} \\
& \leq \sum_{d'=1}^{d} \elltwo{\vec{v}_{d'}} =  {\sum_{d'=1}^{d} \sqrt{\sum_{h} { v^2_{d',h}}} }.       \label{align:l2normeqn} 
% & \leq  \sqrt{\sum_{d'=1}^{d} \elltwo{\vec{v}_{d'}}^2   + 2\sum_{d'=1}^{d}\sum_{d''=1}^{d'} \vec{v}_{d'-1} \vec{v}_{d''}} \numberthis      \label{align:l2normeqn} \\
\end{align}

Here, $v_{d',h}$ is the perturbation in the preactivation of hidden unit $h$ on layer $d'$, brought about by perturbation of the $d'$th weight matrix in a network where only the first $d'-1$ weight matrices have already been perturbed.\\

Now, for each $h$, we bound $v_{d',h}$ in Equation~\ref{align:l2normeqn}. Since the activations have been frozen we can rewrite each $v_{d',h}$ as the product of the $h$th row of the unperturbed network's Jacobian $\verbaljacobian{d'}{d}$ , followed by only the perturbation matrix $\vecU_{d'}$, and then the output of the layer $d'-1$. Concretely, we have\footnote{Below, we have used $H_d$ to denote the number of units on the $d$th layer (and this equals $H$ for the hidden units and $K$ for the output layer).}  \footnote{Note that the succinct formula below holds good even for the corner case $d'=d$, where the first Jacobian-row term becomes a vector with zeros on all but the $h$th entry and therefore only the $h$th row of the perturbation matrix $\vecU_{d'}$ will participate in the expression of $v_{d',h}$.
}:
\begin{align}
{{v}}_{d',h} = 
\overbrace{\jacobian{\W}{d'}{d}{h}{\vec{x}}}^{1 \times H_{d'}} \underbrace{ \overbrace{\vecU_{d'}}^{H_{d'} \times H_{d'-1}} \overbrace{\nn{\W[+\U_{d'-1}]}{d'-1}{}{\vec{x}}}^{H_{d'-1} \times 1} }_{\text{spherical Gaussian}}.
\end{align}

% we can rewrite each $v_{d',h}$ as follows: i) when $d' < d$, as the product of the $h$th row of the unperturbed network's Jacobian of the top layers from $d$ until $d'$ , followed by only the perturbation matrix $U_{d'}$, and then the output of the layer $d'-1$ and ii) when $d' = d$, as the product of the perturbation vector $\vec{u}^{d}_{h}$  and the output of the layer $d-1$.
% \begin{align*}
% {{v}}_{d',h} = 
% \begin{cases}
% \jacobian{\W}{d'}{d}{h}{\vec{x}} U_{d'} \nn{\W[+\U_{d'-1}]}{d'-1}{}{\vec{x}}  & d' < d \\
% \vec{u}^{d}_{h} \nn{\W[+\U_{d-1}]}{d-1}{}{\vec{x}}  & \text{otherwise} \\
% \end{cases}
% \end{align*}

%\paragraph{Distribution of ${{v}}_{d',h}$ when $d' < d$.} 
What do these random variables $v_{d',h}$ look like? 
Conditioned on $\U_{d'-1}$, the second part of our expansion of ${{v}}_{d',h}$, namely, $\vecU_{d'} \nn{\W[+\U_{d'-1}]}{d'-1}{}{\vec{x}}$ is a multivariate spherical Gaussian (see Lemma~\ref{lem:gaussian-operator-norm}) of the form $\N(0, \sigma^2 \| \nn{\W[+\U_{d'-1}]}{d'-1}{}{\vec{x}}\|^2 \vecI)$. As a result, conditioned on $\U_{d'-1}$, ${{v}}_{d',h}$ is a univariate Gaussian $\N(0, \sigma^2 \|{\jacobian{\W}{d'}{d}{h}{\vec{x}}}\|^2 \|{\nn{\W[+\U_{d'-1}]}{d'-1}{}{\vec{x}}}\|^2)$. \\

Then, we can apply a standard Gaussian tail bound (see Lemma~\ref{lem:hoeffding-gaussian}) to conclude that with probability $1-\toldelta/DH$ over the draws of $U_{d'}$ (conditioned on any $\U_{d'-1}$), $v_{d',h}$ is bounded as:

\begin{align}
|v_{d',h}| & \leq \sigma \elltwo{\jacobian{\W}{d'}{d}{h}{\vec{x}}} \|{\nn{\W[+\U_{d'-1}]}{d'-1}{}{\vec{x}}}\| \sqrt{2 \ln \frac{2DH}{\toldelta}}.   \label{align:l2normeqn2}
\end{align}

Then, by a union bound over all the hidden units on layer $d$, and for each $d'$, we have that with probability $1-\toldelta$, Equation~\ref{align:l2normeqn} is upper bounded as:

\begin{align*}
\sum_{d'} \sqrt{\sum_{h} v_{d',h}^2} \leq \sum_{d'=1}^{d} \sigma \frob{\jacobian{\W}{d'}{d}{}{\vec{x}}} \|{\nn{\W[+\U_{d'-1}]}{d'-1}{}{\vec{x}}}\| \sqrt{2 \ln \frac{2DH}{\toldelta}}. \numberthis \label{align:l2normeqn3}
\end{align*}\\

Using this we prove the probability bound in the lemma statement. To simplify notations, let us denote $\tolset_{d-1} \bigcup \{ \toljacob{d'}{d} \}^{d}_{ d'= 1}$ by $\prev{\tolset}$. Furthermore, we will drop redundant symbols in the arguments of the events we have defined. Then, recall that we want to upper bound the following probability (we ignore the arguments $\W+\U$ and $\vec{x}$ for brevity):

%by the sum of the failure probability for the bound in Equation~\ref{align:l2normeqn3}, and the failure probability for the event $\boundedperturbationevent(\W+\U,\tolset_{d-1} \bigcup \{ \toljacob{d''}{d} \}^{d}_{ d''= d}, \vec{x})  \bigcup  \unchangedacts_{d-1}(\W+\U,\vec{x})$. As a result we get the claim in the lemma:
%We show the calculations here for the sake of completeness
\begin{align*}
& \pr{
 \left(\lnot \boundedperturbationevent\left(  { \{ \toloutput{d}'\} }\right)  \right) \wedge \boundedperturbationevent(\prev{\tolset}) \wedge \unchangedacts_{d-1}}
 \end{align*}

Recall that Equation~\ref{align:l2normeqn3} is a bound on the perturbation of the $\ell_2$ norm of the  $d$th layer's output when the activation states are explicitly frozen. If the perturbation we randomly draw happens to satisfy 
$\unchangedacts_{d-1}$ then the bound in Equation~\ref{align:l2normeqn3} holds good even in the case where the activation states are not explicitly frozen. Furthermore, when $\boundedperturbationevent(\prev{\tolset})$ holds, the bound in Equation~\ref{align:l2normeqn3} can be upper-bounded by $\toloutput{d}'$ as defined in the lemma statement, because under $\boundedperturbationevent(\prev{\tolset})$, the middle term in Equation~\ref{align:l2normeqn3} can be upper bounded using triangle inequality as $\|{\nn{\W[+\U_{d'-1}]}{d'-1}{}{\vec{x}}}\| \leq \|{\nn{\W}{d'-1}{}{\vec{x}}}\| + \toloutput{d'-1}$. Hence, the event above happens only for the perturbations for which Equation~\ref{align:l2normeqn3} fails and hence we have that the above probability term is upper bounded by $\toldelta$.

\paragraph{Perturbation bound on the preactivation values of layer $d$.} Following the same analysis as above, the bound we are seeking here is essentially $\max_h \sum_{d'=1}^{d} |v_{d',h}|$. The bound follows similarly from Equation~\ref{align:l2normeqn2}.

\paragraph{Perturbation bound on the $\ell_2$ norm of the rows of the Jacobian $\verbaljacobian{d'}{d}$.} 
We split this term like we did in the previous subsection, and apply triangle equality as follows:
\begin{align*}
& \max_h \ellone{{\elltwo{\jacobian{\W}{d'}{d}{h}{\vec{x}}}-\elltwo{\jacobian{\W[+\U_{d}]}{d'}{d}{h}{\vec{x}}}}} \\
& \leq \max_h \elltwo{\jacobian{\W}{d'}{d}{h}{\vec{x}}-{\jacobian{\W[+\U_{d}]}{d'}{d}{h}{\vec{x}}}} \numberthis \\
& \leq \elltwo{\sum_{d''=1}^{d} \left( \underbrace{\jacobian{\W[+\U_{d''}]}{d'}{d}{h}{\vec{x}}-{\jacobian{\W[+\U_{d''\focus{-1}}]}{d'}{d}{h}{\vec{x}}}}_{:= \vec{y}^{d''}_{h}} \right)} \numberthis \\
& \leq \max_h {\sum_{d''=1}^{d} \elltwo{\vec{y}^{d''}_{h}}} = \max_h \sum_{d''=1}^{d} \sqrt{\sum_{h'} \sq{ y_{d'',h,h'}} }. \numberthis      \label{align:jacobnormeqn} \\
% & \leq  \sqrt{\sum_{d'=1}^{d} \elltwo{\vec{v}_{d'}}^2   + 2\sum_{d'=1}^{d}\sum_{d''=1}^{d'} \vec{v}_{d'-1} \vec{v}_{d''}} \numberthis      \label{align:l2normeqn1} \\
\end{align*}

Here, we have defined $\vec{y}^{d''}_h$ to be the vector that corresponds to the difference in the $h$th row of the Jacobian $\verbaljacobian{d'}{d}$ brought about by perturbing the $d''$th weight matrix, given that the first $d''-1$ matrices have already been perturbed.  We use $h$ to iterate over the units in the $d$th layer and $h'$ to iterate over the units in the $d'$th layer. \\

Now, under the frozen activation states, when we perturb the weight matrices from $1$ uptil $d'$, since these matrices are not involved in the Jacobian $\verbaljacobian{d'}{d}$, fortunately, the Jacobian  $\verbaljacobian{d'}{d}$ is not perturbed (as the set of active weights in $\verbaljacobian{d'}{d}$ are the same when we perturb $\W$ as $\W[+\U_{d'}]$).  So, we will only need to bound $y_{d'',h,h'}$ for $d'' > d'$. 

What does the distribution of $y_{d'',h,h'}$ look like for $d'' > d'$? We can expand\footnote{Again, note that the below succinct formula works even for corner cases like $d'' = d'$ or $d'' = d$.} $y_{d'',h,h'}$ as the product of i) the $h$th row of the Jacobian $\verbaljacobian{d''}{d}$ ii) the perturbation matrix $\U_{d''}$ and iii) the $h'$th column of the Jacobian $\verbaljacobian{d''-1}{d'}$ for the perturbed network:

\begin{align*}
y_{d'',h,h'} = \overbrace{\jacobian{\W}{d''}{d}{h}{\vec{x}}}^{1 \times H_{d''}} \; \underbrace{ \overbrace{\vecU_{d''}}^{H_{d''} \times H_{d''-1}} \; \overbrace{\jacobian{\W[+\U_{d''-1}]}{d'}{d''-1}{:,h'}{\vec{x}}}^{H_{d''-1} \times 1}  }_{\text{spherical Gaussian}}
\end{align*}

Conditioned on $\U_{d''-1}$, the second part of this expansion, namely, $\vecU_{d''} \jacobian{\W[+\U_{d''-1}]}{d'}{d''-1}{:,h'}{\vec{x}}$ is a multivariate spherical Gaussian (see Lemma~\ref{lem:gaussian-operator-norm}) of the form $\N(0, \sigma^2 \| \jacobian{\W[+\U_{d''-1}]}{d'}{d''-1}{:,h'}{\vec{x}}\|^2 \vecI)$. As a result, conditioned on $\U_{d''-1}$, $y_{d'',h,h'}$ is a univariate Gaussian $\N(0, \sigma^2\|\jacobian{\W}{d''}{d}{h}{\vec{x}} \|^2\| \jacobian{\W[+\U_{d''-1}]}{d'}{d''-1}{:,h'}{\vec{x}}\|^2 )$. \\

Then, by applying a standard Gaussian tail bound we have that with probability $1-\frac{\toldelta}{D^2H^2}$ over the draws of $\vecU_{d''}$ conditioned on $\U_{d''-1}$, each of these quantities is bounded as:

\begin{align*}
|y_{d'',h,h'}| \leq \sigma \|\jacobian{\W}{d''}{d}{h}{\vec{x}} \| \cdot \| \jacobian{\W[+\U_{d''-1}]}{d'}{d''-1}{:,h'}{\vec{x}}\| \sqrt{2\ln \frac{D^2H^2}{\toldelta}}. \numberthis\label{eq:vanilla-jacobian-bound}
\end{align*}\\

We simplify the bound on the right hand side a bit further so that it does not involve any Jacobian of layer $d$. Specifically, when $d'' < d$, $\elltwo{\jacobian{\W}{d''}{d}{h}{\vec{x}}}$ can be written as the product of the spectral norm of the Jacobian $\verbaljacobian{d''}{d'-1}$ and the $\ell_2$ norm of the $h$th row of Jacobian $\verbaljacobian{d}{d-1}$. Here, the latter can be upper bounded by the $\ell_2$ norm of the $h$th row of $W_{d}$ since the Jacobian (for a ReLU network) is essentially $W_d$ but with some columns zerod out. When $d=d''$, $\elltwo{\jacobian{\W}{d'}{d}{h}{\vec{x}}}$ is essentially $1$ as the Jacobian is merely the identity matrix. Thus, we have:
\begin{align}
|y_{d'',h,h'}|  \leq \begin{cases}  \sigma \elltwo{\vec{w}^{d}_{h}} \spec{\jacobian{\W}{d''}{d-1}{}{\vec{x}}} \| \jacobian{\W[+\U_{d''-1}]}{d'}{d''-1}{:,h'}{\vec{x}}\| \sqrt{4\ln \frac{DH}{\toldelta}} & d'' < d, \\
  \sigma \| \jacobian{\W[+\U_{d''-1}]}{d'}{d''-1}{:,h'}{\vec{x}}\| \sqrt{4\ln \frac{DH}{\toldelta}} & d'' = d.  \\ 
\end{cases}
\end{align}\\

By a union bound on all $d''$, we then get that with probability $1-\frac{\toldelta}{D}$ over the draws of $\U_{d}$, we can upper bound Equation~\ref{align:jacobnormeqn} as:

\begin{align*}
&\max_h \sum_{d''=1}^{d}\sqrt{\sum_{h'}\sq{y_{d'',h,h'}}} \leq  \sigma  \frob{ \jacobian{\W[+\U_{d''-1}]}{d'}{d-1}{}{\vec{x}}} \sqrt{4\ln \frac{DH}{\toldelta}} +  \\
&   \sum_{d''=d'+1}^{d-1} \sigma \max_h \elltwo{\vec{w}^{d}_{h}} \spec{\jacobian{\W}{d''}{d-1}{}{\vec{x}}}  \frob{ \jacobian{\W[+\U_{d''-1}]}{d'}{d''-1}{}{\vec{x}}} \sqrt{4\ln \frac{DH}{\toldelta}}. \numberthis
\end{align*}

By again applying a union bound for all $d'$, we get the above bound to hold simultaneously for all $d'$ with probability at least $1-\toldelta$. Then, by a similar argument as in the case of the perturbation bound on the output of each layer,  we get the result of the lemma.

\paragraph{Perturbation bound on the spectral norm of the Jacobian $\verbaljacobian{d'}{d}$.} 

Again,  we split this term and apply triangle equality as follows:
\begin{align*}
& \ellone{{\spec{\jacobian{\W}{d'}{d}{}{\vec{x}}}-\spec{\jacobian{\W[+\U_{d}]}{d'}{d}{}{\vec{x}}}}} \\
& \leq \spec{\jacobian{\W}{d'}{d}{}{\vec{x}}-{\jacobian{\W[+\U_{d}]}{d'}{d}{}{\vec{x}}}} \numberthis \\
& \leq \spec{\sum_{d''=1}^{d} \left( \underbrace{\jacobian{\W[+\U_{d''}]}{d'}{d}{}{\vec{x}}-{\jacobian{\W[+\U_{d''\focus{-1}}]}{d'}{d}{}{\vec{x}}}}_{:= \vecY_{d''}} \right)}\numberthis  \\
& \leq \sum_{d''=1}^{d} \spec{\vecY_{d''}}. \numberthis      \label{align:jacobspeceqn}
\end{align*}

Here, we have defined $\vecY_{d''}$ to be the matrix that corresponds to the difference in the Jacobian $\verbaljacobian{d'}{d}$ brought about by perturbaing the the $d''$th weight matrix, given that the first $d''-1$ matrices have already been perturbed. \\

As argued before, under the frozen activation states, when we perturb the weight matrices from $1$ uptil $d'$, since these matrices are not involved in the Jacobian $\verbaljacobian{d'}{d}$, fortunately, the Jacobian  $\verbaljacobian{d'}{d}$ is not perturbed (as the set of active weights in $\verbaljacobian{d'}{d}$ are the same when we perturb $\W$ as $\W[+\U_{d'}]$).  So, we will only need to bound $\vecY_{d''}$ for $d'' > d'$.

Recall that we can expand $\vecY_{d''}$ for $d'' > d'$, $y_{d'',h,h'}$ as the product of i) Jacobian $\verbaljacobian{d''}{d}$ ii) the perturbation matrix $\U_{d''}$ and iii)  the Jacobian $\verbaljacobian{d''-1}{d'}$ for the perturbed network\footnote{Again, note that the below succinct formula works even for corner cases like $d'' = d'$ or $d'' = d$.}:

\begin{align*}
Y_{d''} = \overbrace{\jacobian{\W}{d''}{d}{}{\vec{x}}}^{H_{d} \times H_{d''}} \; \underbrace{ \overbrace{\vecU_{d''}}^{H_{d''} \times H_{d''-1}} \; \overbrace{\jacobian{\W[+\U_{d''-1}]}{d'}{d''-1}{}{\vec{x}}}^{H_{d''-1} \times H_{d'}}  }_{\text{spherical Gaussian}}
\end{align*}\\

Now, the spectral norm of $\vecY_{d''}$ is at most the products of the spectral norms of each of these three matrices. 
Using Lemma~\ref{lem:spec}, the spectral norm of the middle term $\vecU_{d''}$ can be bounded by $\sigma \sqrt{2H \ln \frac{2DH}{\toldelta}}$ with high probability $1-\frac{\toldelta}{D}$ over the draws of $\vecU_{d''}$. \footnote{Although Lemma~\ref{lem:spec} applies only to the case where $U_{d''}$ is a $H \times H$ matrix, it can be easily extended to the corner cases when $d''=1$ or $d'' =D$. When $d''=1$, $U_{d''}$ would be a $H \times N$ matrix, where $H > N$; one could imagine adding more random columns to this matrix, and applying Lemma~\ref{lem:spec}. Since adding columns does not reduce the spectral norm, the bound on the larger matrix would apply on the original matrix too. A similar argument would apply to $d''=D$, where the matrix would be $K \times H$. } 

We will also decompose the spectral norm of the first term so that our final bound does not involve any Jacobian of the $d$th layer. When $d'' = d$, this term has spectral norm $1$ because the Jacobian $\verbaljacobian{d}{d}$ is essentially the identity matrix. When $d'' < d$,  we have that $\spec{\jacobian{\W}{d''}{d}{}{\vec{x}}} \leq \spec{\jacobian{\W}{d-1}{d}{}{\vec{x}}} \spec{\jacobian{\W}{d''}{d-1}{}{\vec{x}}}$. Furthermore, since, for a ReLU network, $\jacobian{\W}{d-1}{d}{}{\vec{x}}$ is effectively $W_{d}$ with some columns zerod out, the spectral norm of the Jacobian is upper bounded by the spectral norm of $\vecW_d$.\\

Putting all these together, we have that with probability $1-\frac{\toldelta}{D}$ over the draws of $\vecU_{d''}$, the following holds good:

\begin{align}
\spec{\vecY_{d''}} \leq \begin{cases} \sigma \spec{\vecW_d} \spec{\jacobian{\W}{d''}{d-1}{}{\vec{x}}}  \spec{\jacobian{\W[+\U_{d''-1}]}{d'}{d''-1}{}{\vec{x}}} \sqrt{2H \ln \frac{2DH}{\toldelta}} & d'' < d\\
\sigma \spec{\jacobian{\W[+\U_{d''-1}]}{d'}{d''-1}{}{\vec{x}}} \sqrt{2H \ln \frac{2DH}{\toldelta}} & d'' = d.\\
\end{cases}
\end{align}

By a union bound, we then get that with probability $1-{\toldelta}$ over the draws of $\U_{d}$, we can upper bound Equation~\ref{align:jacobspeceqn} as:

\begin{align}
\sum_{d''=1}^{d}\spec{\vecY_{d'}} \leq & \sigma \spec{\jacobian{\W[+\U_{d''-1}]}{d'}{d''-1}{}{\vec{x}}} \sqrt{2H \ln \frac{2DH}{\toldelta}} \\
& + \sigma \sum_{d''=d'+1}^{d-1}  \spec{\vecW_d} \spec{\jacobian{\W}{d''}{d-1}{}{\vec{x}}} \spec{\jacobian{\W[+\U_{d''-1}]}{d'}{d''-1}{}{\vec{x}}} \sqrt{2H \ln \frac{2DH}{\toldelta}}
\end{align}

Note that the above bound simultaneously holds over all $d'$ (without the application of a union bound). Finally we get the result of the lemma by a similar argument as in the case of the perturbation bound on the output of each layer.

\end{proof}

\graphicspath{{deterministic-pacbayes/}}

\chapter{A Derandomized PAC-Bayesian Bound}
\label{chap:derandomized}

\section{Introduction}

In this chapter, we will develop a fundamental technique to derandomize PAC-Bayesian bounds for an {\em arbitrary} classifier. Crucially, our technique exploits the noise-resilience of a classifier more powerfully than existing derandomization techniques. Empowered with this general technique, and with the noise-resilience bounds for neural networks from the previous chapter, we will later tackle the specific case of neural networks.

Concretely, our result extends the generalization bound provided by conventional PAC-Bayesian analysis \citep{mcallester03simplified} -- which is a generalization bound on the expected loss of a {\em distribution} of classifiers i.e., a stochastic classifier -- to a generalization bound on a deterministic classifier. The way we reduce the PAC-Bayesian bound to a standard generalization bound, is different from existing techniques pursued in previous works like \citep{neyshabur18pacbayes,langford02pacbayes}.

Furthermore, rather than providing a generalization bound for the specific case of the 0-1 error, we will provide a bound for a more generic function. This will give us some flexibility in how we apply these bounds in the case of deep networks. For example, we later want to use this theorem to be able to say statements like ``if on most training data, the $\ell_2$ norm of the first layer's activations is bounded by $10$, then on most test data as well, the $\ell_2$ norm of the first layer's preactivations is bounded by $10$''.

The results in this chapter have previously been published in \cite{nagarajan18deterministic}.

\section[Our technique]{Our derandomization technique}
%The theorem that we will state below is a more general version than the one we presented in the main paper.  In particular, the generalization bound in the main paper was a bound on the classification error (as are typical generalization bounds). However, recall from our discussion in Section~\ref{sec:derandomized-noise-resilience}, that we will need to study simulatenous generalization of many other quantities related to the network (e.g., we want to bound the proportion of test points for which the $\ell_2$ norm of a particular layer is small, given that it is small on most training datapoints). 
So to state a bound that is general enough, consider a set of functions $\rho_r(\W, \vec{x}, \vec{y})$ for $r=1,2,\hdots R'$ (we will reserve $R$ for a more important notation in the future section). Each of these functions computes a scalar value.  As an example, this could simply be the margin, $\Gamma(\nn{\W}{}{}{\vec{x}}, y)$. But it could also be the quantity $\mathbb{1}[\|\nn{\W}{d}{}{\vec{x}}\|_2 \leq 10]$, which corresponds to whether or not the $\ell_2$ norm of the $d$th layer activations is bounded by $10$. Recall from the previous chapter that, since these functions essentially compute some property of the network that is also dependent on the input, we will refer to these functions as {\em \bf input-dependent properties}. However, for the sake of simplicity, the reader can think of these functions as (fancy kinds) of loss functions.\\

Next, we define a notion of noise-resilience with respect to these properties. Intuitively, at any given input point and parameter configuration, the model is noise-resilient with respect to the $\rho$ functions, if the outputs of these functions do not suffer much perturbation when the parameters themselves are randomly perturbed (by a Gaussian). In order to measure the noise-resilience, let us fix some ``margin threshold'' $\Delta_1, \Delta_2, \hdots, \Delta_{R'}$. If any of the properties suffer a perturbation much larger than the chosen margin threshold, we would consider it to be a lack of noise-resilience. 

For convenience, we will denote these pairs of thresholds and input-dependent properties as:
\begin{equation}
(\vecrho, \vecDelta) \coloneqq \{(\rho_r,\Delta_r)\}_{r=1}^{R'}
\end{equation}\\

Based on these pairs, we define noise-resilience below.

\begin{definition}\textbf{(Noise-resilience.)}
%Given a set of input-dependent properties $\rho_1, \rho_2, \hdots, \rho_{R'}$ and a corresponding set of positive margin thresholds $\Delta_1, \Delta_2, \hdots, \Delta_{R'}$, a model with weights $\W$ is said to be \textbf{$(\sigma, \nu)$-noise-resilient} with respect to the (input-dependent property, threshold) pairs $\{(\rho_r,\Delta_r)\}_{r=1}^{R'}$ at an input $(\vecx,y)$ if:
%Define $T(\cdot, \cdot, \cdot)$ to be  noise resilient  (for this choice of $\U$) at $(\W, \vec{x}, y)$ if 
We say that a model with weights $\W$ is \textbf{$(\sigma, \nu)$-noise-resilient} with respect to the (input-dependent property, threshold) pairs $(\vecrho, \vecDelta) $  at an input $(\vecx, y)$ if:
\begin{equation}
\label{eq:noise-resilience}
\prsub{\U \sim \N(0,\focus{\sigma^2})}{  \exists r \;  : \; \ellone{\rho_r(\W, \vec{x}, y) - \rho_r(\W + \U, \vec{x}, y)} > \frac{\Delta_r}{2}  }  \leq \focus{\nu}.
\end{equation}
Additionally, we define $\nr^{(\vecrho, \vecDelta)}_{\sigma, \nu}(\W, \vecx, y)$ to be the event that the Eq~\ref{eq:noise-resilience} holds.

% \begin{equation}
% \nr^{(\vecrho, \vecDelta)}_{\sigma, \nu}(\W, \vecx, y) = \begin{cases}
%  1 & \text{if Eq~\ref{eq:noise-resilience} holds} \\
%  0 & \text{otherwise}.
% \end{cases}
% \end{equation}
\end{definition}

Now we are ready to state our main result. The outline of our result is that, the proportion of test points where the these $\rho$ functions are small ($< 0$) can be bounded by (a) the proportion of training points where these $\rho$ functions are small ($< \Delta$) and (b) the proportion of training and test points where these $\rho$ functions are not noise-resilient and (c) a standard PAC-Bayesian KL-divergence term between a posterior centered at $\calW$ and the prior.

\begin{restatable}{theorem}{pacbayesian}
\label{thm:pac-bayesian}
Let $P$ be a prior distribution over the parameter space that is chosen independent of the training dataset. Let $\U$ be a random variable sampled entrywise from $\N(0,\sigma^2)$. 
%Let $\rho_{r}(\cdot, \cdot, \cdot)$ and $\Delta_r > 0$ for $r=1, 2, \hdots R'$, be a set of input-dependent properties and their corresponding margins.  %For a particular value of $\Delta$ and a particular data point $(\vec{x}, y)$, let us 
Let us denote the proportion of test and train points where $\W$ is {\bf not} $(\sigma, \focus{\frac{1}{\sqrt{m}}})$-noise-resilient as:
\begin{align}
\mu_{\D}((\vecrho, \vecDelta) ,\W) & \coloneqq \prsub{(\vecx,y) \sim \scrD}{\lnot \nr_{(\sigma, {{1}/{\sqrt{m}}})}^{(\vecrho, \vecDelta)}(\W, \vecx, y)} \\
\hat{\mu}_{S}((\vecrho, \vecDelta) ,\W) & \coloneqq \prsub{(\vecx,y) \sim S}{\lnot \nr_{(\sigma, {{1}/{\sqrt{m}}})}^{(\vecrho, \vecDelta)}(\W, \vecx, y)} \\
\end{align}

%  probability over the random draw of a point $(\vec{x},y)$ drawn from $\D$, that the network with weights $\W$ is {\bf not} noise-resilient 
% % $T(\W, \cdot, \cdot)$ is not noise resilient at 
%  at $(\vec{x},y)$ according to Equation~\ref{eq:noise-resilience}. That is, let $\mu_{\D}(\{(\rho_r,\Delta_r)\}_{r=1}^{R'},\W) := $
%  \begin{equation}
% \prsub{(\vec{x},y)\sim \D}{  \prsub{\U \sim \N(0,\sigma^2)}{  \exists r \;  : \; \ellone{\rho_r(\W, \vec{x}, y) - \rho_r(\W + \U, \vec{x}, y)} > \frac{\Delta_r}{2}  }  > \frac{1}{\sqrt{m}}  }
%  \end{equation}

%  Similarly, let $\hat{\mu}_{S}(\{(\rho_r,\Delta_r)\}_{r=1}^{R'},\W)$ denote the fraction of data points $(\vec{x},y)$ in a dataset $S$ for which the network is {\bf not} noise-resilient according to Equation~\ref{eq:noise-resilience}.
 % that $T(\W, \cdot, \cdot)$ is not noise resilient at.
Then for any $\delta$, with probability $1-\delta$ over the draws of a sample set $S = \{ (\vec{x}_i, y_i) \sim \D \; | i=1,2,\hdots, m  \}$, for any $\W$ we have: 

\begin{align}
 \prsub{(\vec{x},y) \sim \D}{\exists r \; : \; \rho_r(\W, \vec{x}, y) < 0}  \leq & \; \; \prsub{(\vec{x},y) \sim S}{ \exists r \; : \; \rho_r(\W, \vec{x}, y) <  \Delta_r}   \\
 & +\hat{\mu}_{S}((\vecrho, \vecDelta),\W)  + \mu_{\D}((\vecrho, \vecDelta), \W) \\ 
& + 2\sqrt{\frac{2 KL(\N(\W, \sigma^2 I) \| P) + \ln \frac{2m}{\delta}}{m-1}} + \frac{2}{\sqrt{m}-1}.
\end{align}
\end{restatable}

The reader maybe curious about how one would bound the term $\mu_{\D}$ in the above bound, as this term corresponds to noise-resilience with respect to {\em test} data. This is precisely what we will address in the next chapter.

\section[Advantage of our technique]{Key advantage of our derandomization technique.}
\label{sec:derandomized-advantage}
%Fix the mu_D reference back!
 %The high level argument is that, to convert the PAC-Bayesian bound, these latter works relied on a looser output perturbation bound, one that holds on all possible inputs, with high probability over all perturbations i.e., a bound on $\max_{\vec{x}} \ellinfty{\nn{\W}{}{}{\vec{x}}- \nn{\W+\U}{}{}{\vec{x}}}$ w.h.p over draws of $\U$. In contrast, our technique relies on a subtly different but significantly tighter bound: a bound on the output perturbation that holds with high probability {\em given an input} i.e., a bound on $\ellinfty{\nn{\W}{}{}{\vec{x}}- \nn{\W+\U}{}{}{\vec{x}}}$ w.h.p over draws of $\U$ for each $\vec{x}$. When we do instantiate our framework as in the next section, this subtle difference is critical in being able to bound the output perturbation without suffering from a factor proportional to the product of the spectral norms of the weight matrices (which is the case in \cite{neyshabur18pacbayes}).

The above approach differs from previous derandomization approaches used by \citet{neyshabur18pacbayes,langford02pacbayes} in how strong a noise-resilience we require of the classifier to provide the generalization guarantee. The stronger the noise-resilience requirement, the more price we have to pay when we jump from the PAC-Bayesian guarantee on the stochastic classifier to a guarantee on the deterministic classifier. We argue that {\em our noise-resilience requirement is a much milder condition and therefore promises tighter guarantees}. Our requirement is philosophically similar to \cite{london16stability,mcallester03simplified}, although technically different. \\

More concretely, to arrive at a reasonable generalization guarantee in our setup, observe that it is sufficient if we can show that $\mu_{\scrD}$ and $\hat{\mu}_{S}$ are as only as large as $\mathcal{O}(1/\sqrt{m})$ (we cannot get a better convergence rate with respect to $m$ anyway). In other words, we would want the following for $(\vec{x},y) \sim \D$ and for $(\vec{x},y) \sim S$:
\begin{equation}
\prsub{(\vec{x},y)} { \prsub{\U \sim \N(0,\sigma^2)}{ \exists r \;  : \; \ellone{\rho_r(\W, \vec{x}, y) - \rho_r(\W + \U, \vec{x}, y)} > \frac{\Delta_r}{2}  }  > \frac{1}{\sqrt{m}} } = \mathcal{O}(1/\sqrt{m}) .
\end{equation}

Let us contrast this sort of a noise-resilience requirement with the noise-resilience requirement from previous derandomization techniques. Previous works require a noise resilience condition of the form that with high probability a particular perturbation does not perturb the classifier output on {\em any input}. For example, the noise-resilience condition used in \citet{neyshabur18pacbayes} (discussed in Theorem~\ref{thm:basic-derandomized}) written in terms of our notations, would be:
\begin{equation}
\prsub{\U \sim \N(0,\sigma^2)}{ \focus{\exists {\vec{x}}} \; : \; \exists r \;  : \; \ellone{\rho_r(\W, \vec{x}, y) - \rho_r(\W + \U, \vec{x}, y)} > \frac{\Delta_r}{2}  } \leq \frac{1}{2}.
\end{equation}\\

The main difference between the above two formulations is in what makes a particular perturbation (un)favorable for the classifier. In our case, we deem a perturbation unfavorable only after fixing the datapoint (given by the fact that $\mathbb{P}_{\vecx}$ {\em precedes} $\mathbb{P}_\calU$ in the former equation). However, in the earlier works, a perturbation is deemed unfavorable if it perturbs the classifier output sufficiently on {\em some} datapoint from the domain of the distribution (given by the fact that $\exists x$ follows {\em after} $\mathbb{P}_\calU$). While this difference is subtle, the earlier approach would lead to a much more pessimistic analysis of these perturbations.  In our analysis, this weakened noise resilience condition will be critical in analyzing the Gaussian perturbations more carefully than in \citet{neyshabur18pacbayes} i.e., we can bound the perturbation in the classifier output more tightly by analyzing the Gaussian perturbation for a fixed input point.\\

Note that one way our noise resilience condition would seem stronger is that on a given datapoint we want less than $1/\sqrt{m}$ mass of the perturbations to be unfavorable for us, while in previous bounds, there can be as much as $1/2$ probability mass of perturbations that are unfavorable. In our analysis, this will only weaken our generalization bound by a $\ln \sqrt{m}$ factor in comparison to previous bounds (while we save other significant factors).
% In Appendix~\ref{sec:derandomized-pac-bayesian}, we also outline how our approach significantly differs from past work in the technicalities of how noise-resilience is used to modify the PAC-Bayesian bound.

%The key advantage of our bound, in comparison to  previous works, lies in the fact that our bound demands a weaker notion of noise-resilience from $\W$ (which as we saw, is required to extend the PAC-Bayesian bound on the stochastic classifier $\tilde{\W}$ to the deterministic classifier $\W$). We discuss the precise technical difference between these bounds in Appendix~\ref{sec:derandomized-advantage}. 

%Specifically, previous formulations required that there must be only a small proportion of random perturbations that affect the output of $\W$ on {\em any} input.  Our bound however requires that for most inputs, there must be only a small proportion of random perturbations that affect the output of $\W$ {\em for that specific input}. We describe this more formally in Appendix~\ref{sec:derandomized-pac-bayesian}.

%This is precisely what we will address extensively in the next section.

%\pacbayesian*
\section[Proof]{Proof of Theorem~\ref{thm:pac-bayesian}}

\begin{proof}

The starting point of our proof is the standard PAC-Bayesian theorem  \cite{mcallester03simplified} which bounds the generalization error of a stochastic classifier (stated in Theorem~\ref{thm:basic-pac-bayes}). Recall that the result applied to any bounded loss function. Let us restate that result but with slightly different notion for the generic loss function so that it is easier to adapt it to our $\rho$ functions here. 

In particular, let $\mathcal{L}(\W, \vec{x}, y)$ be any loss function that takes as input the network parameter, and a datapoint $\vec{x}$ and its  true label $y$ and outputs a value in $[0,1]$. Let $P$ be a data-independent prior over the parameter space. Then, we have that, with probability $1-\delta$ over the draw of $S \sim \D^m$, for every distribution $Q$ over the parameter space, the following holds:

\begin{equation}
\label{eq:main-pac-bayes}
\Esub{\tilde{\W} \sim Q}{\Esub{(\vec{x},y) \sim \D}{\mathcal{L}(\tilde{\W} , \vec{x}, y)}} \leq \Esub{\tilde{\W} \sim Q}{\frac{1}{m}\sum_{(\vec{x}, y) \in S}\mathcal{L}(\tilde{\W} , \vec{x}, y)} + 2\sqrt{\frac{2 KL(Q \| P) + \ln \frac{2m}{\delta}}{m-1}}
\end{equation}

%In other words, the statement tells us that except for a $\delta$ proportion of bad draws of $m$ samples, the test loss of the stochastic classifier $\tilde{\W} \sim Q$ would be close to its train loss. This holds for every possible distribution $Q$, which allows us to cleverly choose $Q$ based on $S$. 

 We choose $Q$ to be the distribution of the stochastic classifier picked from $\N(\W,\sigma^2 I)$ i.e., a Gaussian perturbation of the deterministic classifier $\W$. \\

% Since it holds for every $Q$, and since we would want a generalization bound for the deterministic classifier learned by SGD by training on $S$, a good choice of $Q$ would be one which is centered around the classifier learned. Hence, for the rest of the discussion we will choose $Q$ to be

Now our task is to bound a loss for the deterministic classifier $\W$, where the loss is defined in terms of $\vecrho$ as $\prsub{(\vec{x},y) \sim \D}{\exists r \; | \; \rho_r(\W , \vec{x}, y) < 0}$. To this end, let us define the following margin-based variation of this loss for some $c \geq 0$:

\begin{equation}
\mathcal{L}^{(c)}(\W, \vec{x}, y) = \begin{cases}
1 & \exists r \; : \; \rho_r(\W, \vec{x}, y) < c \Delta_r \\
0 & \text{otherwise},
\end{cases}
\end{equation}
and so we have $\prsub{(\vec{x},y) \sim \D}{ \exists r \; | \; \rho_r(\W , \vec{x}, y) < 0} = \mathbb{E}_{(\vec{x},y) \sim \D} \left[ \mathcal{L}^{(0)}(\W, \vec{x}, y)\right]$. 

 The proof from here follows two main stages. In the first stage, we will upper bound the test loss of a deterministic classifier with that of a stochastic classifier, and then apply a PAC-Bayesian bound on it. In the next stage, we will upper bound the training loss of a stochastic classifier (that would arise in the PAC-Bayesian bound), with the training loss of the deterministic classifier.

\paragraph{Relating test loss of stochastic classifier to deterministic classifier.}
First, we will bound the expected test $\mathcal{L}^{(0)}$ loss of a deterministic classifier by the expected $\mathcal{L}^{(1/2)}$ of the stochastic classifier; then we will bound the test $\mathcal{L}^{(1/2)}$ of the stochastic classifier using the PAC-Bayesian bound.

We will split the expected loss of the deterministic classifier into an expectation over datapoints for which it is noise-resilient with respect to Gaussian noise and an expectation over the rest. To simplify notations, we will write 
$\nr_{(\sigma, {{1}/{\sqrt{m}}})}$ as just $\nr$. 

\begin{align*}
\mathbb{E}_{(\vec{x},y) \sim \D} \left[ \mathcal{L}^{(0)}(\W, \vec{x}, y)\right]  & = \Esub{(\vec{x},y) \sim \D}{\left. {\mathcal{L}^{(0)}({\W}, \vec{x}, y)}  \right| \; \mathfrak{N}(\W,\vec{x},y) } \prsub{(\vec{x},y) \sim \D}{\mathfrak{N}(\W,\vec{x},y)} \\
& +\underbrace{ \Esub{(\vec{x},y) \sim \D}{ \left. \mathcal{L}^{(0)}({\W}, \vec{x}, y)  \right| \; \lnot \mathfrak{N}(\W,\vec{x},y) }}_{\leq 1} \underbrace{ \prsub{(\vec{x},y) \sim \D}{\lnot \mathfrak{N}(\W,\vec{x},y)}}_{\mu_{\D}((\vecrho, \vecDelta),\W)} \numberthis \\
& \leq \Esub{(\vec{x},y) \sim \D}{\left. {\mathcal{L}^{(0)}({\W}, \vec{x}, y)}  \right|\; \mathfrak{N}(\W,\vec{x},y) } \prsub{(\vec{x},y) \sim \D}{\mathfrak{N}(\W,\vec{x},y)} \\
 &+ \mu_{\D}((\vecrho, \vecDelta),\W). \numberthis\label{eq:deterministic-loss-upper-bound} 
\end{align*}\\

To further continue the upper bound on the right hand side, we will try to bound $\Esub{(\vec{x},y) \sim \D}{\left. {\mathcal{L}^{(0)}({\W}, \vec{x}, y)}  \right|\; \mathfrak{N}(\W,\vec{x},y) }$, which corresponds to the deterministic classifier's loss on the noise-resilient part of the distribution. In particular, we will bound this in terms of the stochastic classifier's loss on the noise-resilient part of the distribution.
%recall that if for a point $(\vec{x},y)$, $\mathfrak{N}(\W,\vec{x},y)$ holds it means that $\prsub{\tilde{\W}\sim Q}{ \exists r \; | \; |\rho_r(\W,\vec{x},y) - \rho_r(\tilde{\W},\vec{x},y)| \geq \Delta_r/2} \leq 1/4m^2$. %Thus, we can derive an upper bound on ${\mathcal{L}^{(0)}({\W}, \vec{x}, y)}$ as follows, starting from
%We can then 
% next, we upper bound the first term on the right hand side,

For simplicity of notations, we will write $\D'$ to denote the distribution $\D$ conditioned on $\mathfrak{N}(\W,\vec{x},y)$. Also, let $\mathfrak{U}(\tilde{\W},\vec{x},y)$ be the favorable event that for a given data point $(\vec{x}, y)$ and a fixed draw of the stochastic classifier, $\tilde{\W}$, it is the case that for every $r$, $|\rho_r(\W,\vec{x},y) - \rho_r(\tilde{\W},\vec{x},y)| \leq \Delta_r/2$.
Then, the stochastic classifier's loss $\mathcal{L}^{(1/2)}$ on $\D'$ can be written in terms of $\mathfrak{U}$ as:
\begin{align*}
&\Esub{\tilde{\W}\sim Q}{\Esub{(\vec{x},y) \sim \D'}{ \mathcal{L}^{(1/2)}(\tilde{\W}, \vec{x}, y) }} \\
 &= \Esub{(\vec{x},y) \sim \D'}{\Esub{\tilde{\W}\sim Q}{\mathcal{L}^{(1/2)}(\tilde{\W}, \vec{x}, y)}} \numberthis \\
& = \Esub{(\vec{x},y) \sim \D'}{\Esub{\tilde{\W}\sim Q}{ \left. \mathcal{L}^{(1/2)}(\tilde{\W}, \vec{x}, y)  \right| \; \mathfrak{U}(\tilde{\W},\vec{x},y)}\prsub{\tilde{\W}\sim Q}{\mathfrak{U}(\tilde{\W},\vec{x},y)}} \\
& + \underbrace{\Esub{(\vec{x},y) \sim \D'}{{ \Esub{\tilde{\W}\sim Q}{ \left. \mathcal{L}^{(1/2)}(\tilde{\W}, \vec{x}, y)  \right|\; \lnot \mathfrak{U}(\tilde{\W},\vec{x},y) }}\prsub{\tilde{\W}\sim Q}{\lnot \mathfrak{U}(\tilde{\W},\vec{x},y)}}}_{\geq 0} \numberthis \\
& \geq \Esub{(\vec{x},y) \sim \D'}{\Esub{\tilde{\W}\sim Q}{ \left. \mathcal{L}^{(1/2)}(\tilde{\W}, \vec{x}, y)  \right| \; \mathfrak{U}(\tilde{\W},\vec{x},y)}\prsub{\tilde{\W}\sim Q}{\mathfrak{U}(\tilde{\W},\vec{x},y)}}. \numberthis 
\end{align*}\\

Next, we use the following fact: if $\mathcal{L}^{(1/2)}(\tilde{\W},\vec{x},y) = 0$, then for all $r$, $\rho_r(\tilde{\W},\vecx,y) \geq \Delta_r/2$ and if $\tilde{\W}$ is a favorable perturbation of $\W$, then for all $r$, $\rho_r(\W,\vecx,y) \geq \rho_r(\tilde{\W},\vec{x},y) - \Delta_r/2 > 0$ i.e., $\mathcal{L}^{(1/2)}(\tilde{\W},\vec{x},y) = 0$ implies $\mathcal{L}^{(0)}({\W}, \vec{x}, y)  = 0$. Hence if $\tilde{\W}$ is a favorable perturbation then, $\mathcal{L}^{(1/2)}(\tilde{\W}, \vec{x}, y)  \geq \mathcal{L}^{(0)}({\W}, \vec{x}, y) $. Therefore, we can lower bound  the above series of inequalities by replacing the stochastic classifier with the deterministic classifier (and thus ridding ourselves of the expectation over $Q$):

\begin{align}
\Esub{\tilde{\W}\sim Q}{\Esub{(\vec{x},y) \sim \D'}{ \mathcal{L}^{(1/2)}(\tilde{\W}, \vec{x}, y) }} & \geq \Esub{(\vec{x},y) \sim \D'}{{ \mathcal{L}^{(0)}({\W}, \vec{x}, y) } \prsub{\tilde{\W}\sim Q}{\mathfrak{U}(\tilde{\W},\vec{x},y)} }.  
\end{align}\\

Since the favorable perturbations for a fixed datapoint drawn from $\D'$ have sufficiently high probability (that is, $\prsub{\tilde{\W}\sim Q}{\mathfrak{U}(\tilde{\W},\vec{x},y)} \geq 1-1/\sqrt{m}$), we have:

\begin{align}
\Esub{\tilde{\W}\sim Q}{\Esub{(\vec{x},y) \sim \D'}{ \mathcal{L}^{(1/2)}(\tilde{\W}, \vec{x}, y) }}  \geq  \left( 1-\frac{1}{\sqrt{m}} \right) \Esub{(\vec{x},y) \sim \D'}{  \mathcal{L}^{(0)}({\W}, \vec{x}, y)  }. 
\end{align}

Thus, we have a lower bound on the stochastic classifier's loss that is in terms of the deterministic classifier's loss on the noise-resilient datapoints. Rearranging it, we get an upper bound on the latter:

\begin{align*}
\Esub{(\vec{x},y) \sim \D'}{  \mathcal{L}^{(0)}({\W}, \vec{x}, y)  } & \leq \frac{1}{\left(1-\frac{1}{\sqrt{m}} \right)} \Esub{\tilde{\W}\sim Q}{\Esub{(\vec{x},y) \sim \D'}{ \mathcal{L}^{(1/2)}(\tilde{\W}, \vec{x}, y) }} \numberthis  \\
& \leq \left(1 + \frac{1}{\sqrt{m}-1} \right) \Esub{\tilde{\W}\sim Q}{\Esub{(\vec{x},y) \sim \D'}{ \mathcal{L}^{(1/2)}(\tilde{\W}, \vec{x}, y) }} \numberthis \\
&  \leq \Esub{\tilde{\W}\sim Q}{\Esub{(\vec{x},y) \sim \D'}{ \mathcal{L}^{(1/2)}(\tilde{\W}, \vec{x}, y) }}\\
& +  \frac{1}{\sqrt{m}-1}  \underbrace{\Esub{\tilde{\W}\sim Q}{\Esub{(\vec{x},y) \sim \D'}{ \mathcal{L}^{(1/2)}(\tilde{\W}, \vec{x}, y) }}}_{\leq 1} \numberthis\\
& \leq \Esub{\tilde{\W}\sim Q}{\Esub{(\vec{x},y) \sim \D'}{ \mathcal{L}^{(1/2)}(\tilde{\W}, \vec{x}, y) }} + \frac{1}{\sqrt{m}-1}. \numberthis
\end{align*}

Thus, we have an upper bound on the expected loss of the deterministic classifier $\W$ on the noise-resilient part of the distribution. Plugging this back in the first term of the upper bound on the deterministic classifier's loss on the whole distribution $\D$ in Equation~\ref{eq:deterministic-loss-upper-bound} we get :

\begin{align*}
&\mathbb{E}_{(\vec{x},y) \sim \D} \left[ \mathcal{L}^{(0)}(\W, \vec{x}, y)\right] \leq  \\
 & \left( \Esub{\tilde{\W}\sim Q}{\Esub{(\vec{x},y) \sim \D'}{ \mathcal{L}^{(1/2)}(\tilde{\W}, \vec{x}, y) }} + \frac{1}{\sqrt{m}-1} \right) \prsub{(\vec{x},y) \sim \D}{\mathfrak{N}(\W,\vec{x},y)} + \mu_{\D}((\vecrho, \vecDelta),\W). \numberthis
\end{align*}

Rearranging, we get:
\begin{align*}
\mathbb{E}_{(\vec{x},y) \sim \D} \left[ \mathcal{L}^{(0)}(\W, \vec{x}, y)\right] & \leq   \left( \Esub{\tilde{\W}\sim Q}{\Esub{(\vec{x},y) \sim \D'}{ \mathcal{L}^{(1/2)}(\tilde{\W}, \vec{x}, y) }} \right) \prsub{(\vec{x},y) \sim \D}{\mathfrak{N}(\W,\vec{x},y)} + \\
&  \mu_{\D}((\vecrho, \vecDelta),\W) +  \frac{1}{\sqrt{m}-1} \underbrace{\prsub{(\vec{x},y) \sim \D}{\mathfrak{N}(\W,\vec{x},y)}}_{\leq 1}. \numberthis
\end{align*}

Rewriting the expectation over $\D'$  explicitly as an expectation over $\D$ conditioned on $\mathfrak{N}(\W,\vec{x},y)$, we get: 
\begin{align*}
&\mathbb{E}_{(\vec{x},y) \sim \D} \left[ \mathcal{L}^{(0)}(\W, \vec{x}, y)\right] \leq  \\
 & \left( \Esub{\tilde{\W}\sim Q}{\Esub{(\vec{x},y) \sim \D}{ \left. \mathcal{L}^{(1/2)}(\tilde{\W}, \vec{x}, y) \right| \; \mathfrak{N}(\W,\vec{x},y)} \prsub{(\vec{x},y) \sim \D}{\mathfrak{N}(\W,\vec{x},y)}  }   \right)  +\\
& \mu_{\D}((\vecrho, \vecDelta),\W)  + \frac{1}{\sqrt{m}-1}. \numberthis
\end{align*}\\

The first term in the right hand side is essentially an expectation of a loss over the distribution $\D$ with the loss set to be zero over the non-noise-resilient datapoints and set to be $\mathcal{L}^{(1/2)}$ over the noise-resilient datapoints; thus we can upper bound it with the expectation of the $\mathcal{L}^{(1/2)}$ loss over the whole distribution $\D$:
\begin{align*}
\mathbb{E}_{(\vec{x},y) \sim \D} \left[ \mathcal{L}^{(0)}(\W, \vec{x}, y)\right] \leq    \Esub{\tilde{\W}\sim Q}{\Esub{(\vec{x},y) \sim \D}{ \mathcal{L}^{(1/2)}(\tilde{\W}, \vec{x}, y) }} +\cdot \mu_{\D}((\vecrho, \vecDelta),\W)  + \frac{1}{\sqrt{m}-1}. \numberthis \label{eq:deterministic-loss-final-upper-bound}
\end{align*}

Now observe that we can upper bound the first term in the R.H.S. using the PAC-Bayesian bound by plugging in $\mathcal{L}^{(1/2)}$ for the generic $\mathcal{L}$ in Equation~\ref{eq:main-pac-bayes}; however, the bound would still be in terms of the stochastic classifier's train error. To get the generalization bound we seek, which involves the deterministic classifier's train error, we need to take another step mirroring these tricks on the train loss. 

\paragraph{Relating the stochastic classifier's train loss to deterministic classifier's train loss.}

Our analysis here is almost identical to the above analysis. Instead of working with the distribution $\D$ and $\D'$ we will work with the training data set $S$ and a subset of it $S'$ for which noise resilience property is satisfied by $\W$. %Below, to make the presentation neater, we use $(\vec{x},y) \sim S$ to denote uniform sampling from $S$.% and $\Esub{(\vec{x},y) \sim S}{\cdot}$ to denote the average over $S$.

First, we upper bound the stochastic classifier's train loss ($\mathcal{L}^{(1/2)}$) by splitting it over the noise-resilient points $S'$ ($(\vec{x},y) \in S$ for which $\mathfrak{N}(\W,\vec{x},y)$ holds) like in Equation~\ref{eq:deterministic-loss-upper-bound}:

\begin{align*}
\Esub{\tilde{\W}\sim Q}{\Esub{(\vec{x},y) \sim S}{ \mathcal{L}^{(1/2)}(\tilde{\W}, \vec{x}, y) }}  &= \Esub{(\vec{x},y) \sim S}{\Esub{\tilde{\W}\sim Q}{\mathcal{L}^{(1/2)}(\tilde{\W}, \vec{x}, y)}} \numberthis \\
& \leq \Esub{(\vec{x},y) \sim S'}{\Esub{\tilde{\W}\sim Q}{\mathcal{L}^{(1/2)}(\tilde{\W}, \vec{x}, y)}} \prsub{(\vec{x},y) \sim S}{(\vec{x},y) \in S'} \\ & + \hat{\mu}_{S}((\vecrho, \vecDelta),\W). \numberthis \label{eq:stochastic-loss-upper-bound} 
\end{align*} \\

We can upper bound the first term by first splitting it over the favorable and unfavorable perturbations like we did before:

\begin{align*}
& \Esub{(\vec{x},y) \sim S'}{\Esub{\tilde{\W}\sim Q}{\mathcal{L}^{(1/2)}(\tilde{\W}, \vec{x}, y)}} \\
& = \Esub{(\vec{x},y) \sim S'}{\Esub{\tilde{\W}\sim Q}{ \left. \mathcal{L}^{(1/2)}(\tilde{\W}, \vec{x}, y)  \right| \; \mathfrak{U}(\tilde{\W},\vec{x},y)}\prsub{\tilde{\W}\sim Q}{\mathfrak{U}(\tilde{\W},\vec{x},y)}} \\
& + \Esub{(\vec{x},y) \sim S'}{{ \Esub{\tilde{\W}\sim Q}{ \left. \mathcal{L}^{(1/2)}(\tilde{\W}, \vec{x}, y)  \right|\; \lnot \mathfrak{U}(\tilde{\W},\vec{x},y) }}\prsub{\tilde{\W}\sim Q}{\lnot \mathfrak{U}(\tilde{\W},\vec{x},y)}}. \numberthis
\end{align*}\\

To upper bound this, we apply a similar argument. First, if $\mathcal{L}^{(1/2)}(\tilde{\W},\vec{x},y) = 1$, then $\exists r$ such that $\rho_r(\tilde{\W},\vecx,y) < \Delta_r/2$ and if $\tilde{\W}$ is a favorable perturbation then for that value of $r$, $\rho_r(\W,\vecx,y) < \rho_r(\tilde{\W},\vec{x},y) + \Delta_r/2 < \Delta_r$. Thus if $\tilde{\W}$ is a favorable perturbation then, $\mathcal{L}^{(1)}({\W}, \vec{x}, y)  = 1$ whenever $\mathcal{L}^{(1/2)}(\tilde{\W}, \vec{x}, y) = 1$ i.e., $\mathcal{L}^{(1/2)}(\tilde{\W}, \vec{x}, y)  \leq \mathcal{L}^{(1)}({\W}, \vec{x}, y) $. Next, we use the fact that the unfavorable perturbations for a fixed datapoint drawn from $S'$ have sufficiently low probability i.e., $\prsub{\tilde{\W}\sim Q}{\lnot \mathfrak{U}(\tilde{\W},\vec{x},y)} \leq 1/\sqrt{m}$. Then, we get the following upper bound on the above equations, by replacing the stochastic classifier with the deterministic classifier (and thus ignoring the expectation over $Q$):

\begin{align*}
 &\Esub{(\vec{x},y) \sim S'}{\Esub{\tilde{\W}\sim Q}{\mathcal{L}^{(1/2)}(\tilde{\W}, \vec{x}, y)}}  \\
 & \leq \Esub{(\vec{x},y) \sim S'}{\Esub{\tilde{\W}\sim Q}{ \left. \mathcal{L}^{(1)}({\W}, \vec{x}, y)  \right| \; \mathfrak{U}(\tilde{\W},\vec{x},y)}  \underbrace{\prsub{\tilde{\W}\sim Q}{\mathfrak{U}(\tilde{\W},\vec{x},y)}}_{\leq 1}   } \\ 
&+ \Esub{(\vec{x},y) \sim S'}{{ \underbrace{\Esub{\tilde{\W}\sim Q}{ \left. \mathcal{L}^{(1/2)}(\tilde{\W}, \vec{x}, y)  \right|\; \lnot \mathfrak{U}(\tilde{\W},\vec{x},y) }}_{\leq 1}} \frac{1}{\sqrt{m}}} \numberthis \\
& \leq \Esub{(\vec{x},y) \sim S'}{ \mathcal{L}^{(1)}({\W}, \vec{x}, y) } + \frac{1}{\sqrt{m}}.  \numberthis
\end{align*}\\

Plugging this back in the first term of Equation~\ref{eq:stochastic-loss-upper-bound}, we get:

\begin{align*}
\Esub{\tilde{\W}\sim Q}{\Esub{(\vec{x},y) \sim S}{ \mathcal{L}^{(1/2)}(\tilde{\W}, \vec{x}, y) }} 
  \leq & \left( \Esub{(\vec{x},y) \sim S'}{ \mathcal{L}^{(1)}({\W}, \vec{x}, y) } + \frac{1}{\sqrt{m}}  \right) \prsub{(\vec{x},y) \sim S}{(\vec{x},y) \in S'}\\
 &  + \hat{\mu}_{S}((\vecrho, \vecDelta),\W)  \numberthis \\
 \leq &   \Esub{(\vec{x},y) \sim S'}{ \mathcal{L}^{(1)}({\W}, \vec{x}, y) }   \prsub{(\vec{x},y) \sim S}{(\vec{x},y) \in S'}  \\
  &  + \hat{\mu}_{S}((\vecrho, \vecDelta),\W)  + \frac{1}{\sqrt{m}}  \underbrace{\prsub{(\vec{x},y) \sim S}{(\vec{x},y) \in S'}}_{\leq 1}  \numberthis \\
 \leq  &  \Esub{(\vec{x},y) \sim S'}{ \mathcal{L}^{(1)}({\W}, \vec{x}, y) }  \prsub{(\vec{x},y) \sim S}{(\vec{x},y) \in S'} +\frac{1}{\sqrt{m}} \\
 &  + \hat{\mu}_{S}((\vecrho, \vecDelta),\W) \numberthis
\end{align*}

Since the first term is effectively the expectation of a loss over the whole distribution  with the loss set to be zero on the non-noise-resilient points and set to $\mathcal{L}^{(1)}$ over the rest, we can upper bound it by setting the loss to be $\mathcal{L}^{(1)}$ over the whole distribution:
\begin{align*}
  \Esub{\tilde{\W}\sim Q}{\Esub{(\vec{x},y) \sim S}{ \mathcal{L}^{(1/2)}(\tilde{\W}, \vec{x}, y) }}  & \leq  \Esub{(\vec{x},y) \sim S}{ \mathcal{L}^{(1)}({\W}, \vec{x}, y) } + \hat{\mu}_{S}((\vecrho, \vecDelta),\W) +\frac{1}{\sqrt{m}}  \numberthis \\ 
\end{align*}

Applying the above upper bound and the bound in Equation~\ref{eq:deterministic-loss-final-upper-bound} into the PAC-Bayesian result of Equation~\ref{eq:main-pac-bayes} yields our result (Note that combining these equations would produce the term $\frac{1}{\sqrt{m}} + \frac{1}{\sqrt{m}-1}$ which is at most $\frac{2}{\sqrt{m}-1}$, which we reflect in the final bound. % For $m \geq 2$, this is at most $\frac{5}{\sqrt{m-1}}$; we use this bound in our final result.
).

\end{proof}

\graphicspath{{deterministic-pacbayes/}}

\chapter{Data-dependent (Derandomized) PAC-Bayesian Bounds}
\label{chap:datadependent-pacbayes}

% This is not proceeding the way we want to
\section{Introduction}

 If we know that the output of classifier is highly noise-resilient on most training data and most test data, we can use the tools from Chapter~\ref{chap:derandomized} to derive a strong PAC-Bayesian bound for the error of the deterministic classifier. But how do we {\em know} how noise-resilient a classifier is at a particular input? For neural networks, it must be clear from Chapter~\ref{chap:noise-resilience}, that in order to tightly characterize noise-resilience, it is not enough to merely look at the weights that were learned. We also need to look at lower layer properties like the $\ell_2$ norm of the hidden layer representations and the inter-layer Jacobians, and also how noise-resilient those lower layer properties themselves are. All these properties crucially depend on the given input and in particular how the input activates the weights of the network. Thus, to {\em know} whether the classifier is highly noise-resilient, we must know something about the input.

In the context of generalization theory, we ``know'' what the training data is, and what the trained weights are. Hence, we can easily derive tight noise-resilience guarantees on the training data. This is akin to training a neural network and empirically observing an implicit bias and incorporating that in the generalization analysis. However, since the generalization bound must not ``know'' the test data \footnote{Recall the discussion from Section~\ref{sec:generalization-powers} on what it means for a generalization bound to have explanatory power}, there is no immediate way to derive tight noise-resilience guarantees for the classifier on test data.\\

The way we tackle this issue is by {\em generalizing these input-dependent properties themselves}. That is, we ``observe'' the fact that the classifier has small hidden layer norms on most training data, and ``generalize the property'' to say that the classifier has small hidden layer norms even on test data. We can {\em successively} do this in a particular order, from the lower most layer to the top most layer, and then eventually generalize the output property, namely the error itself. It is important to generalize these properties in a particular order: since the 5th layer $\ell_2$ norm is small only if the 2nd layer $\ell_2$ norm is small, we first need a test-time guarantee on the 2nd layer $\ell_2$ norm, before seeking a guarantee for the 5th layer. 

The bound that we eventually arrive at will be purely based upon input-dependent properties that are computed on the (training) data. In other words, the bound would correspond to a certain notion of {\em data-dependent} complexity (such as Jacobian norms, hidden layer norms etc.,). Contrast this with {\em data-independent} notions of complexity like the Frobenius norms of the weights\footnote{Although, admittedly, one could technically argue that even Frobenius norms are data-dependent since the weights themselves are data-dependent}.

The results in this chapter have previously been published in \cite{nagarajan18deterministic}.

\section{Our abstract framework}
\label{sec:framework}

We now discuss how noise-resilience can be formalized in an abstract framework 
 through certain conditions on the weight matrices. Much of our discussion below is dedicated to how these conditions must be designed, as these details carry the key ideas behind how  noise-resilience can be generalized from training to test data. We then present our main generalization bound and some intuition about our proof technique. Note that all of this discussion is abstract and can apply to any classifier, not necessarily neural networks.

% lay out the key components of our framework, and constraints on how these should be designed; following which we present our main result and some intuition about our proof technique.

%-- conditions that are meant to imply that the output of the network is noise-resilient at a particular datapoint. For example, one such condition could be that the pre-activation values of the $d$th layer are all sufficiently large, i.e., $\min_h |\prenn{\W}{d}{h}{\vec{x}}| \geq 0.01$, or that the $\ell_2$ norm of the $d$th layer is sufficiently small i.e., $\| \nn{\W}{d}{}{\vec{x}} \| \leq 3$. 

\subsection{Input-dependent properties of weights} 

Recall that, at a high level, the noise-resilience of a network corresponds to how little the network reacts to random parameter perturbations. Naturally, this would vary depending on the input. Hence, in our framework, we will analyze the noise-resilience of the network as a function of a given input. Specifically, we will characterize noise-resilience through conditions on input-dependent properties of the weights. For example, one condition could be``the preactivation values of the hidden units in layer $d$ have magnitude larger than some small positive constant''.  The idea is that when these conditions involving the weights and the input  are satisfied, if we add noise to the weights, the output of the classifier for that input will provably suffer only little perturbation.  \\
%We will more generally refer to each scalar quantity involved in these conditions, such as each of the pre-activation values, as an {\em input-dependent property of the weights}. 

We will now formulate these input-dependent properties and the conditions on them, for a generic classifier, and in the next chapter, we will see how they can be instantiated in the case of deep networks. Consider a classifier for which we can {\em hypothetically} define $R$ different conditions, which when satisfied on a given input, will help us guarantee the classifier's noise-resilience at that input i.e., 
bound the output perturbation under random parameter perturbations. In the case of deep networks, we will have a condition for each layer, and so $R$ will scale with depth.

In more detail, let the $r$th condition be a bound involving a particular set of input-dependent properties of the weights denoted by $\{ \rho_{r,1}(\W, \vec{x},y), \rho_{r,2}(\W, \vec{x},y), \hdots, \}$. Here, each element $\rho_{r,l}(\W, \vec{x},y)$ is a scalar value that depends on the weights and the input, just like pre-activation values\footnote{As we will see in the next chapter, most of these properties depend on only the unlabeled input $\vec{x}$ and not on $y$. But for the sake of convenience, we include $y$ in the formulation of the input-dependent property, and use the word input to refer to $\vec{x}$ or $(\vec{x},y)$ depending on the context}.  Note that here the first subscript $l$ is the index of the element in the set, and the second subscript $r$ is the index of the set itself.  Now for each of these properties, we will define a corresponding set of \textit{positive} constants (that are independent of $\W, \vec{x}$ and $y$), denoted by $\{ \trainmargin_{r,1}, \trainmargin_{r,2}, \hdots \}$, which we will use to specify our conditions. In particular,

\begin{definition}
 We say that the weights $\W$ satisfy the $r$th \textit{condition} on the input $(\vec{x},y)$ if\footnote{When we say $\forall l$ below, we refer to the set of all possible indices $l$ in the $r$th set, noting that different sets may have different cardinality.}:
\begin{equation}
\forall l,  \; \rho_{r,l}(\W,\vec{x},y) >  \trainmargin_{r,l}.
\label{eq:condition}
\end{equation}
\end{definition}
 % In our framework, we characterize noise-resilience through a set of say, $R$ (abstract) conditions on how the activated weight matrices interact with each other for a given input. To define these noise-resilience-related conditions, we will consider $R$ sets of the form  for $r=1,2,3, \hdots, R$; each $\rho_r^{(l)}(\W, \vec{x},y)$ computes a scalar value which corresponds to an {\em input-dependent property} of the weights (and $l$ is an index for the elements within a set). The reader can think of each set here as the set of properties of a particular layer in the network (the reason for grouping them in this manner will be apparent soon). For example, in the next section, for ReLU networks, we will consider a set for each layer, each consisting of the pre-activations of each layer $\{ \prenn{\W}{d}{1}{\vec{x}}, \hdots, \prenn{\W}{d}{H}{\vec{x}} \}$ for $d=1,2,\hdots D$. (Although, as we will see, we will actually consider $R=\mathcal{O}(D^2)$ sets in total.)

% Given this $r$th set of properties, we assume that the $r$th condition can be expressed as $T_r(\W,\vec{x},y) > 0$, where $T_r$ is a function that aggregates these properties, e.g., $T_r(\W,\vec{x},y)  = \min_r \rho_{r,l}(\W, \vec{x}, y)$.  

For example, the $r$th condition could be ``every pre-activation unit in the $r$th layer must be at least as large as $10$''. This sort of a condition can be realized if we let $l$ iterate over the hidden units, and $\rho$ correspond to the pre-activations and $\Delta$ vaues equal $10$.

For convenience, we also define an additional $R+1$th set to be the singleton set containing the \textit{margin} of the classifier on the input: $\nn{\W}{}{}{\vec{x}}[y]- \max_{j\neq y} \nn{\W}{}{}{\vec{x}}[j]$. Note that if this term is positive (negative) then the classification is (in)correct. We will also denote the corresponding constant $\trainmargin_{R+1,1}$ as $\classmargin$.

 % Next, we define conditions that bound the properties. To do this, for each set, we define a function $T_r(\W,\vec{x},y)$ that aggregates the values of the properties in that set; then, a {\em condition} of the form $T_r(\W,\vec{x},y) > 0$ would imply a bound on the properties in the set. For example, on ReLU networks, we will define $T_d(\W,\vec{x},y) :=  \min_{h=1}^{H}|\prenn{\W}{d}{h}{\vec{x}}| - \Delta$ (for some $\Delta > 0$) which aggregates the pre-activation values of the $d$th layer; the condition $T_d(\W,\vec{x},y) > 0$ imposed on all the training data, would imply that  these pre-activation values on the training set have magnitude at least $\Delta$. 

%\paragraph{Constraint 1: Aggregation of the properties}
%
%In this framework, we restrict the 
%aggregator functions $T_r$ to be linear in terms of the corresponding properties, in the sense that, $\forall \;  r, \W, \W'$: 
%%  the extreme values (or extreme absolute value)  (effectively limiting us to ) from the corresponding set of properties, like the example function $T_d(\W,\vec{x},y) =  \min_{h=1}^{H}|\prenn{\W}{d}{h}{\vec{x}}| - \Delta$. We formulate this as the following constraint:
% \vspace*{-0.1cm}
%
% \begin{equation} \label{eq:constraint-1} |T_r(\W,\vec{x},y) - T_r(\W',\vec{x},y)| \leq \max_l |\rho_{r,l}(\W,\vec{x},y) - \rho_{r,l}(\W',\vec{x},y)| 
% \end{equation}

\subsection{Conditional noise-resilience of the properties}

Recall that in the case of neural networks, there is a natural ordering between these properties, and furthermore, properties of a layer are noise-resilient if we know that the previous layer properties are well-behaved. E.g.,  for any given input, the perturbation in the pre-activation values of the $d$th layer  is small
%i.e., $\max_h |\prenn{\W}{d}{h}{\vec{x}} - \prenn{\W+\U}{d}{h}{\vec{x}}|$, is small (i.e., doesn't scale with the product of the spectral norms of the weights) w.h.p over $\U \sim \N(0,\sigma^2 I)$
 if the absolute pre-activation values in the layers below $d-1$ are large, and a few other norm-bounds on the lower layer weights are satisfied. \\

 Let us capture this sort of a ``conditional noise-resilience'' abstractly.
%As we will see later, this constraint will help us  prove how these properties themselves are noise-resilient on test data i.e., with random parameter perturbations, even properties like the pre-activations do not suffer much perturbation.
 Roughly speaking, we want to formulate the fact that for a given input, {\em if the first $r-1$ sets of properties approximately satisfy the condition in Equation~\ref{eq:condition}, then the properties in the $r$th set are noise-resilient} i.e., under random parameter perturbations, these properties do not suffer much perturbation.

We formalize the above requirement by defining quantities $\Delta_{r,l}(\sigma)$ that bound the perturbation in the properties $\rho_{r,l}$, in terms of the variance $\sigma^2$ of the parameter perturbations. 

\begin{definition}
For every $r \leq R+1$ and $l$, we define $\Delta_{r,l}(\sigma)$ to be a quantity such that 
for any  $(\vec{x},y)$: %be such that $T_q(\W,{\vec{x},y}) > 0$ for all the preceding $q < r$. Then
\begin{align*}
& \focus{\text{if }} \forall q < r, \forall l, \rho_{q,l}(\W,\vec{x},y) > 0 \text{ then} \\
& Pr_{\U \sim \N(0,\sigma^2 I)} \Big[ 
{
\exists l  \; |\rho_{r,l}(\W+\U,{\vec{x},y}) - \rho_{r,l}(\W,{\vec{x},y}) | > \frac{\Delta_{r,l}(\sigma)}{2}
 }
 \; \; \text{ and } \\
& \; \; \; \; \; \; \; \; \; \; \; \; \; \; \; { \forall q {<} r,  \forall l \; \;  |\rho_{q,l}(\W+\U,{\vec{x},y}) - \rho_{q,l}(\W,{\vec{x},y}) |{<} \frac{\Delta_{q,l}(\sigma)}{2} } \Big]  \leq \frac{1}{(R+1)\sqrt{m}}. \numberthis \label{eq:generic-noise-resilience}
\end{align*}

\end{definition}

%\begin{align*}
%& \text{if } \forall q < r, T_q(\W,{\vec{x},y}) > 0 \text{ then} \\
%& Pr_{\U \sim \N(0,\sigma^2 I)} \Big[ 
%\overbrace{
% \max_l  \; |\rho_{r,l}(\W+\U,{\vec{x},y}) - \rho_{r,l}(\W,{\vec{x},y}) | > \frac{\Delta_r(\sigma)}{2}
% }^{\text{some property in the $r$th set suffers large perturbation}}  
% \; \; \text{ and } \\
%& \; \; \; \; \; \; \; \; \; \; \; \; \; \; \;  \underbrace{ \forall q {<} r, \max_l \; \;  |\rho_{q,l}(\W+\U,{\vec{x},y}) - \rho_{q,l}(\W,{\vec{x},y}) |{<} \frac{\Delta_q(\sigma)}{2} }_{
%\substack{\text{the first } r-1 \text{ sets of properties do not get perturbed much}}
%} \Big]  \leq \frac{1}{R\sqrt{m}}. \numberthis \label{eq:generic-noise-resilience}
%\end{align*}
Let us unpack the above constraint. First, although the above constraint must hold for all inputs $(\vec{x},y)$, it effectively applies only to those inputs that satisfy the pre-condition of the if-then statement: namely, it applies
 only to inputs $(\vec{x},y)$ that {\em approximately} satisfy the first $r-1$ conditions in Equation~\ref{eq:condition} in that $\rho_{q,l}(\W,\vec{x},y) > 0$ (approximatley, in that this is not $\rho_{q,l}(\W,\vec{x},y) > \trainmargin_{q,l}$).  

 Next, we discuss the second part of the above if-then statement which specifies a probability term that is required to be small for all such inputs. In words, the first event within the probability term above is the event that for a given random perturbation $\U$, the properties involved in the $r$th condition 
suffer a large perturbation. The second is the event that the properties involved in the first $r-1$ conditions do \textit{not} suffer much perturbation; but, given that these $r-1$ conditions already hold approximately, this second event implies that these conditions are still preserved approximately under perturbation.

 In summary, our constraint requires the following:  for any input on which the first $r-1$ conditions hold, there should be very few parameter perturbations that significantly perturb the $r$th set of properties while preserving the first $r-1$ conditions. When we instantiate the framework, we have to derive closed form expressions for the perturbation bounds $\Delta_{r,l}(\sigma)$ (in terms of only  $\sigma$ and the constants $\trainmargin_{r,l}$). This is indeed what we have already done in
our noise-resilience analysis of deep networks in Chapter~\ref{chap:noise-resilience}.
%As we will see, for ReLU networks, we will choose the properties in a way that this constraint naturally falls into place in a way that the perturbation bounds $\Delta_{r,l}(\sigma)$ do not grow with the product of spectral norms (Lemma~\ref{lem:noise-resilience-induction}).

\subsection{Theorem Statement}

 In this setup, we have the following margin-based generalization guarantee on the original network.  Our generalization guarantee, which scales linearly with the number of conditions $R$,  holds under the setting that the training algorithm always finds weights such that on the training data, the conditions in Equation~\ref{eq:condition}  is  satisfied for all $r = 1, \hdots, R$. \\

\begin{theorem}
\label{thm:generic-generalization} 
%Let $\trainmargin_1, \trainmargin_2, \hdots, \trainmargin_R$ be a set of positive constants that will act as bounds on the input-dependent properties. Let $\trainmargin_R = \classmargin/2$. Let  $T_1, T_2, \hdots, T_R$ be a set of aggregator functions that satisfy Constraint 1.  
Let $\sigma^*$ be the standard deviation of a Gaussian parameter perturbation\footnote{Ideally, we must choose the largest possible such perturbation to get the smallest bound.} such that the constraint in Equation~\ref{eq:generic-noise-resilience} holds with $\Delta_{r,l}(\sigma^\star) \leq \Delta^{\star}_{r,l}$ $\forall r \leq R+1$ and $\forall l$. 
%$\sigma$ such that for all $r=1, \hdots, R$, Constraint 2 holds with the perturbation bounds $\Delta_r(\sigma)$ satisfying $\Delta_r(\sigma) \leq \trainmargin_r$. 
Then, for any $\delta > 0$, with probability $1-\delta$ over the draw of samples $S$ from $\D^m$, for any $\W$ we have that, if $\W$ satisfies the conditions in Equation~\ref{eq:condition}  for all $r \leq R$ and for all training examples $(\vec{x},y) \in S$, then
\begin{align} 
%\prsub{(\vec{x},y) \sim \D}{ \mathcal{L}_{0}(\nn{\W}{}{}{\vec{x}},y)}  \leq &  \prsub{(\vec{x},y) \sim S}{ \mathcal{L}_{\classmargin}(\nn{\W}{}{}{\vec{x}},y)} \\& 
\scrL_{\scrD}(f_{\calW}) \leq \scrL^{(\classmargin)}_{f_{\calW}}
+ \tilde{\mathcal{O}} \left({R} \sqrt{\frac{2 KL(\N(\W, (\sigma^\star)^2 \vecI) \| P)+ \ln \frac{2mR}{\delta}}{m-1} }  \right)
\end{align}
\end{theorem}

The crux of our proof lies in generalizing the conditions of Equation~\ref{eq:condition} satisfied on the training data to test data {\em one after the other}, by proving that they are noise-resilient on both training and test data. Crucially, after we generalize the first $r-1$ conditions from training data to test data (i.e., on most test and training data, the $r-1$ conditions are satisfied), we will have from Equation~\ref{eq:generic-noise-resilience} that the $r$th set of properties are noise-resilient on both training and test data. Using the noise-resilience of the $r$th set of properties on test/train data, we can generalize even the $r$th condition to test data. Our result crucially relies on the fundamental derandomization technique introduced in Theorem~\ref{thm:pac-bayesian} in the previous chapter.

\section[Proof]{Proof of Theorem~\ref{thm:generic-generalization}}
\begin{proof}
 Our proof is based on the following recursive inequality that we demonstrate for all $r \leq R$ (we will prove a similar, but slightly different inequality for $r=R+1$):

\begin{align*} 
\prsub{(\vec{x},y) \sim \D}{\exists q \focus{\leq} r, \exists l \; \rho_{q,l}(\W, \vec{x},y) < 0  }  \leq & \; \prsub{(\vec{x},y) \sim \D}{\exists q \focus{ < } r, \exists l \; \rho_{q,l}(\W, \vec{x},y) < 0  } \\  
& + \underbrace{\tilde{\mathcal{O}} \left( \sqrt{\frac{2 KL(\N(\W, \sigma^2 I) \| P) }{m-1}}  \right)}_{\text{generalization error for condition }r} \numberthis \label{eq:recursion}
\end{align*}

To interpret this inequality,  recall that the $r$th condition in Equation~\ref{eq:condition} is that $\forall l$, $\rho_{r,l} > \Delta^{\star}_{r,l}$. If this was approximately satisfied, we would expect $\rho_{r,l} > 0$.
Above, we bound the probability mass of test points such that any one of the first $r$ conditions in Equation~\ref{eq:condition} is not even approximately satisfied, in terms of the probability mass of points where one of the first $r-1$ conditions is not even approximately satisfied, and a term that corresponds to how much error there can be in generalizing the $r$th condition from the training data. \\

Our proof crucially relies on Theorem~\ref{thm:pac-bayesian}. This theorem provides an upper bound on the proportion of test data that fail to satisfy a set of conditions, in terms of four quantities. The first quantity is the proportion of training data that do not satisfy the conditions; the second and third quantities, which we will in short refer to as $\hat{\mu}_S$ and $\hat{\mu}_{\D}$, correspond to the proportion of training and test data on which the properties involved in the conditions are not noise-resilient. The fourth quantity is the PAC-Bayesian KL divergence term. \\

First, we consider the base case when $r=1$, and apply the PAC-Bayes-based guarantee from Theorem~\ref{thm:pac-bayesian} on the first set of properties $\{\rho_{r,1},\rho_{r,2},\hdots \}$ and their corresponding constants $\{\Delta^\star_{r,1},\Delta^\star_{r,2},\hdots \}$. First we have from our assumption (in the main theorem statement) that on all the training data, the condition $\rho_{1,l}(\W,\vec{x},y) > \Delta_{1,l}^\star$ is satisfied for all possible $l$. Thus, the first term in the upper bound in Theorem~\ref{thm:pac-bayesian} is zero. Next, we can show that the terms $\hat{\mu}_S$ and $\hat{\mu}_{\D}$ would be zero too. This follows from the fact that the constraint in Equation~\ref{eq:generic-noise-resilience} holds in this framework. Specifically, applying this equation for $r=1$, for $\sigma = \sigma^\star$, we get that for all possible $(\vec{x},y)$ the following inequality holds:
\begin{equation}
\prsub{\U \sim \N(0,(\sigma^\star)^2 I)}{\exists l\; \;|\rho_{1,l}(\W+\U,{\vec{x},y}) - \rho_{l,1}(\W,{\vec{x},y}) | > \frac{\Delta_{1,l}(\sigma^\star)}{2}} \leq \frac{1}{R\sqrt{m}}.
\end{equation}
Since, $\sigma^\star$ was chosen such that $\Delta_1(\sigma^\star) \leq \Delta_1^\star$, we have:
\begin{equation}
\prsub{\U}{\exists l\; \;|\rho_{1,l}(\W+\U,{\vec{x},y}) - \rho_{l,1}(\W,{\vec{x},y}) | > \frac{\Delta_1^\star}{2}} \leq \frac{1}{R\sqrt{m}}.
\end{equation}

Effectively this establishes that the noise-resilience requirement of Equation~\ref{eq:noise-resilience} in Theorem~\ref{thm:pac-bayesian} holds on all possible inputs, thus proving our claim that the terms $\hat{\mu}_S$ and $\hat{\mu}_{\D}$ would be zero.
Thus, we will get that 

\begin{align*} 
\prsub{(\vec{x},y) \sim \D}{\exists l \; \rho_{1,l}(\W, \vec{x},y) < 0  } \leq & \;  {\tilde{\mathcal{O}} \left( \sqrt{\frac{2 KL(\N(\W, \sigma^2 I) \| P) }{m-1}}  \right)}
\end{align*}

which proves the recursion statement for the base case.\\

To prove the recursion for some arbitrary $r \leq R$, we again apply the PAC-Bayes-based guarantee from Theorem~\ref{thm:pac-bayesian}, but on the union of the first $r$ sets of properties. Again, we will have that the first term in the guarantee would be zero, since the corresponding conditions are satisfied on the training data. Now, to bound the proportion of bad points $\hat{\mu}_S$ and $\hat{\mu}_{\D}$, we make the following claim:

\begin{quote}
the network is noise-resilient as per Equation~\ref{eq:noise-resilience} in Theorem~\ref{thm:pac-bayesian} for any input that satisfies the $r-1$ conditions approximately i.e., $\forall q \leq r-1$ and $\forall l$, $\rho_{q,l}(\W, \vec{x}, y) > 0$. 
\end{quote}

The above claim can be used to prove Equation~\ref{eq:recursion} as follows. Since all the conditions are assumed to be satisfied by a margin on the training data, this claim immediately implies that $\hat{\mu}_{S}$ is zero. Similarly, this claim implies that for the test data, we can bound $\mu_{\D}$ in terms of $\prsub{(\vec{x},y) \sim \D}{\exists q \focus{ < } r \; \exists l \; \rho_{q,l}(\W, \vec{x},y) < 0  }$, thus giving rise to the recursion in Equation~\ref{eq:recursion}.\\

Now, to prove our claim, consider an input $(\vec{x},y)$ such that $\rho_{q,l}(\W, \vec{x}, y) > 0$ for $q=1,2,\hdots, r-1$ and for all possible $l$. First from the assumption in our theorem statement that $\Delta_{q,l}(\sigma^\star) \leq \Delta_{q,l}^\star$, we  have the following upper bound on the proportion of parameter perturbations under which any of the properties in the first $r$ sets suffer a large perturbation:

\begin{align*}
& \prsub{\U}{  \exists q \leq r \; \exists l \; : \; \ellone{\rho_{q,l}(\W, \vec{x}, y) - \rho_{q,l}(\W + \U, \vec{x}, y)} > \frac{\Delta_{q,l}^*}{2}  }  \\ 
& \leq \prsub{\U}{  \exists q \leq r \; \exists l \; : \; \ellone{\rho_{q,l}(\W, \vec{x}, y) - \rho_{q,l}(\W + \U, \vec{x}, y)}  > \frac{\Delta_{q,l}(\sigma^\star)}{2}  } \numberthis \\
& \leq \sum_{q=1}^{r} Pr\Big[ \exists l  \; |\rho_{q,l}(\W+\U,{\vec{x},y}) - \rho_{q,l}(\W,{\vec{x},y}) | > \frac{\Delta_q(\sigma^\star)}{2} \wedge\\
& \; \; \;\; \; \; \; \; \; \forall q' < q, \; \exists l \ellone{\rho_{q',l}(\W, \vec{x}, y) - \rho_{q',l}(\W + \U, \vec{x}, y)} {<} \frac{\Delta_{q'}(\sigma^\star)}{2} \Big]. \numberthis
 \intertext{Now, we are considering an input $(\vec{x},y)$ that satisfies $\rho_{q,l}(\W, \vec{x}, y) > 0$ for $q=1,2,\hdots, r-1$ and for all possible $l$, by the constraint assumed in Equation~\ref{eq:generic-noise-resilience}, each term in the RHS is bounded above by $\frac{1}{(R+1)\sqrt{m}}$. So: } \\
& \prsub{\U}{  \exists q \leq r \; \exists l \; : \; \ellone{\rho_{q,l}(\W, \vec{x}, y) - \rho_{q,l}(\W + \U, \vec{x}, y)} > \frac{\Delta_{q,l}^*}{2}  } \leq \sum_{q=1}^{r} \frac{1}{(R+1)\sqrt{m}} \leq \frac{1}{\sqrt{m}}. \numberthis \\
\end{align*}

Thus,we have proven above that $(\vec{x},y)$ satisfies the noise-resilience condition from Equation~\ref{eq:noise-resilience} in Theorem~\ref{thm:pac-bayesian} if it also satisfies $\rho_{q,l}(\W, \vec{x}, y) > 0$ for $q=1,2,\hdots, r-1$ and for all possible $l$. This proves our claim, and hence in turn proves the recursion in Equation~\ref{eq:recursion}.\\

Finally, we can apply a similar argument for the $R+1$th set of input-dependent properties (which is a singleton set consisting of the margin of the network) with a small change since the first term in the guarantee from Theorem~\ref{thm:pac-bayesian} is not explicitly assumed to be zero; we will get an inequality in terms of the number of training points that are not classified correctly by a margin, giving rise to the margin-based bound:

\begin{align*} 
\prsub{(\vec{x},y) \sim \D}{\exists q {\leq} R\focus{+1} \; \exists l, \rho_{q,l}(\W, \vec{x},y) < 0  }  \leq & \; \frac{1}{m}\sum_{(\vec{x},y) \in S} \mathbf{1}[\rho_{R+1,1}(\W,\vec{x},y) < \Delta_{R,1}] \\
& +\prsub{(\vec{x},y) \sim \D}{\exists q \leq R \; \exists l \; \rho_{q,l}(\W, \vec{x},y) < 0  } \\  
& + {\tilde{\mathcal{O}} \left( \sqrt{\frac{2 KL(\N(\W, \sigma^2 I) \| P) }{m-1}}  \right)} 
\end{align*}

Note that in the first term on the right hand side, $\rho_{R+1,1}(\W,\vec{x},y)$ corresponds to the margin of the classifier on $(\vec{x},y)$.  Now, by using the fact that the test error is upper bounded by the left hand side in the above equation, applying the recursion on the right hand side $R+1$ times, we get our final result.
\end{proof}

\graphicspath{{deterministic-pacbayes/}}

\chapter[Exponentially Tighter Bounds for Deep Networks]{Exponentially Tighter Bounds for Deep Networks via Generalizing Noise-Resilience}
\label{chap:deterministic-pacbayes}

\section{Introduction}

% Modern deep neural networks contain millions of parameters and are trained on
% relatively few samples. Conventional wisdom in machine learning suggests that such models should massively overfit on the training data, as these models have the capacity to memorize even a randomly labeled dataset of similar
% size \citep{zhang17generalization,neyshabur15inductive}. Yet these models have achieved state-of-the-art generalization error on many real-world tasks. This observation has spurred an active line of research \citep{soudry18sgd,brutzkus18sgd,li18learning} that has tried to understand what properties are possessed by stochastic gradient descent (SGD) training of deep networks that allows these networks to generalize well. 

%In the previous chapter, we looked at distance from initialization as a way of quantifying the implicit bias of the training algorithm. Another interesting notion of inductive bias that has been empirically linked to generalization is that of the width of the minimum. Specifically, it has been observed that stochastic gradient descent (SGD) tends to find solutions that lie in ``flat, wide minima'' in the training loss \citep{hochreiter97flat,hinton93mdl,keskar17largebatch}. Since this is a form of inductive bias, we can hope to derive generalization bounds that exploit this bias. 

One particularly promising line of work on flatness of loss landscapes \citep{neyshabur17exploring,arora18compression} has been bounds that utilize the {\em noise-resilience} of deep networks on training data i.e., how much the  training loss of the network changes with noise injected into the parameters, or roughly, how wide is the training loss minimum. While these have yielded generalization bounds that do not have a severe exponential dependence on depth (unlike other bounds that grow with the product of spectral norms of the weight matrices), these bounds are quite limited: they either apply to a \emph{stochastic} version of the classifier (where the parameters are drawn from a distribution) or a \emph{compressed} version of the classifier (where the parameters are modified and represented using fewer bits).\\ 

%  based upon PAC-Bayesian analysis \cite{}, which is able to provide parameter-indepdnent bounds on the generalization performance, but which does so (in a naive application) only for a \emph{stochastic} version of the classifier, not for the actual deterministic network.  Futhermore, those works which \emph{do} extend the PAC-Bayes analysis to the deterministic setting pay a high cost, and result in generalization bounds that scale exponentially with the depth of the network.

In this chapter, we revisit the PAC-Bayesian analysis of deep networks in \cite{neyshabur17exploring,neyshabur18pacbayes}. In particular, we build on the general PAC-Bayesian tools we have derived in the last two chapters and use noise-resilience of the deep network on training data to provide a bound on the original deterministic and uncompressed network. We achieve this by arguing that if on the training data, the interaction between the ``activated weight matrices'' (weight matrices where the weights incoming from/outgoing to inactive units are zeroed out)  satisfy certain conditions which results in a wide training loss minimum, these conditions 
themselves generalize to the weight matrix interactions on the test data. 

Our generalization bound accomplishes two goals simultaneously: i) it applies to the original network and ii)
it does not scale exponentially with depth in terms of the products of the spectral norms of the weight matrices; instead our bound scales with more meaningful terms that capture the interactions between the weight matrices and do not have such a severe dependence on depth in practice. Besides this, our bound also incorporates the distance from initialization as discussed in Chapter~\ref{chap:dist-from-init}.\\

 We note that all but one of these terms are indeed quite small on networks in practice.  The one particularly (empirically) large term that we use is the reciprocal of the magnitude of the network pre-activations on the training data (and so our bound would be small only in the scenario where the pre-activations are not too small).  We emphasize that this drawback is more of a limitation in how we characterize noise-resilience through the specific conditions we chose for the ReLU network, rather than a drawback in our PAC-Bayesian framework itself.  Our hope is that, since our technique is quite general and flexible, by carefully identifying the right set of conditions, in the future, one might be able to derive a similar generalization guarantee that is smaller in practice.

The results in this chapter have previously been published in \cite{nagarajan18deterministic}.

\section[Background]{Background and related work}

One of the most important aspects of the generalization puzzle that has been
studied is that of the flatness/width of the training loss at the minimum found by SGD. The general understanding is that flatter minima are correlated with better generalization
behavior, and this should somehow help explain the generalization behavior
\citep{hochreiter97flat,hinton93mdl,keskar17largebatch}.  Flatness of the training loss minimum is also correlated with the observation that on training data, adding noise to the parameters of the network results only in little change in the output of the network -- or in other words, the network is noise-resilient.  Deep networks are known to be similarly resilient to noise injected into the inputs \citep{novak2018sensitivity}; but note that our theoretical analysis relies on resilience to parameter perturbations.

While some progress has been made in understanding the convergence and generalization behavior of SGD training of simple models like two-layered hidden neural networks under simple data distributions \citep{neyshabur15inductive,soudry18sgd,brutzkus18sgd,li18learning}, all known generalization guarantees for SGD on deeper networks --- through analyses that do not use noise-resilience properties of the networks --- have strong exponential dependence on depth. In particular, these bounds scale either with the product of the spectral norms  of the weight matrices \citep{neyshabur18pacbayes,bartlett17spectral}  or their Frobenius norms \citep{golowich17size}. In practice, the weight matrices have a spectral norm that is as large as $2$ or $3$, and an even larger Frobenius norm that scales with $\sqrt{H}$ where $H$ is the width of the network i.e., maximum number of hidden units per layer. \footnote{To understand why these values are of this order in magnitude, consider the initial matrix that is randomly initialized with independent entries with variance $1/\sqrt{H}$. It can be shown that the spectral norm of this matrix, with high probability, lies near its expected value, near $2$ and the Frobenius norm near its expected value which is $\sqrt{H}$. Since SGD is observed not to move too far away from the initialization regardless of $H$ \citep{nagarajan17role}, these values are more or less preserved for the final weight matrices.}  Thus, the generalization bound scales as say, $2^D$ or $H^{D/2}$, where $D$ is the depth of the network. \\

At a high level, the reason these bounds suffer from such an exponential dependence on depth is that they effectively perform a worst case approximation of how the weight matrices interact with each other. For example, the product of the spectral norms arises from a naive approximation of the Lipschitz constant of the neural network, which would hold only when the singular values of the weight matrices all align with each other. However, in practice, for most inputs to the network, the interactions between the activated weight matrices are not as adverse.

By using noise-resilience of the networks, prior approaches \citep{arora18compression,neyshabur17exploring} have been able to derive bounds that replace the above worst-case approximation with smaller terms that realistically capture these interactions. However, these works are limited in critical ways. \citet{arora18compression} use noise-resilience of the network to \textit{modify} and ``compress'' the parameter representation of the network, and derive a generalization bound on the compressed network. While this bound enjoys a better dependence on depth because its applies to a compressed network, the main drawback of this bound is that it does not apply on the original network. On the other hand, \citet{neyshabur17exploring} take advantage of noise-resilience on training data by incorporating it within a
 PAC-Bayesian generalization bound \citep{allester99pacbayes}. However, their final guarantee is only a bound on the expected test loss of a stochastic network.\\

In this chapter, we revisit the idea in \citet{neyshabur17exploring}, by pursuing the PAC-Bayesian framework \citep{allester99pacbayes} to answer this question. 
The standard PAC-Bayesian framework provides generalization bounds for the expected loss of a stochastic classifier, where the stochasticity typically corresponds to Gaussian noise injected into the parameters output by the learning algorithm. However, if the classifier is noise-resilient on both training and test data, one could extend the PAC-Bayesian bound to a standard generalization guarantee on the deterministic classifier.

Other works have used PAC-Bayesian bounds in different ways in the context of neural networks. \citet{langford01not,dziugaite17nonvacuous} optimize the stochasticity and/or the weights of the network in order to {\em numerically} compute good (i.e., non-vacuous) generalization bounds on the stochastic network. \citet{neyshabur18pacbayes}  derive generalization bounds on the original, deterministic network by working from the PAC-Bayesian bound on the stochastic network. However, as stated earlier, their work does not make use of noise resilience in the networks learned by SGD. 

% The key contribution in our work is a novel, general PAC-Bayes based technique that leverages noise-resilience of deep networks to derive a generalization guarantee on the test loss of the network output by an algorithm like SGD -- an uncompressed, deterministic network.  More concretely, we make the following contributions:

\paragraph{Our Contributions}  

The key contribution in this chapter is to apply the abstract data-dependent PAC-Bayesian bound we have developed in the previous chapter for the case of fully-connected ReLU networks. Recall that the abstract framework requires enumerating a list of input-dependent properties of the classifier, and how they perturb under Gaussian noise. We determine what thesw properties are based on our noise-resilience analysis in Chapter~\ref{chap:noise-resilience}.
 While very similar conditions have already been identified in prior work \citep{arora18compression,neyshabur17exploring}(see Section~\ref{app:comparison} for an extensive discussion of this), our contribution here is in showing how these conditions generalize from training to test data. Crucially, like these works, our bound does not have severe exponential dependence on depth in terms of products of spectral norms. \\

We note that in reality, all but one of our conditions on the network do hold on training data as necessitated by the framework. The strong, non-realistic condition we make is that the pre-activation values of the network are sufficiently large, although only on training data; however, in practice a small proportion of the pre-activation values can be arbitrarily small.  Our generalization bound scales inversely with the smallest absolute value of the pre-activations on the training data, and hence in practice, our bound would be large.

Intuitively, we make this assumption to ensure that under sufficiently small parameter perturbations, the activation states of the units are guaranteed not to flip.  It is worth noting that 
\citet{arora18compression,neyshabur17exploring} too
require similar, but more realistic assumptions about pre-activation values that effectively assume only a small proportion of units flip under noise. However, even under our stronger condition that no such units exist, it is not apparent how these approaches would yield a similar bound on the deterministic, uncompressed network without generalizing their conditions to test data. We hope that in the future our work could be developed further to accommodate the more realistic conditions from \citet{arora18compression,neyshabur17exploring}.

% Additionally, we emphasize that this drawback is more of a limitation in how we characterize noise-resilience through our conditions rather than a drawback in our PAC-Bayesian framework itself.  Our hope is that in the future, by considering a better set of conditions -- perhaps drawing inspiration from other works like \citet{arora18compression,neyshabur17exploring} -- and applying them within our general framework, one might be able to derive a similar but tighter generalization bound in practice.

\section[Bound for ReLU networks]{Application of our framework to ReLU Networks}
\label{sec:application}

% \paragraph{Notation.}
%In this section, we apply our framework to feedforward fully connected ReLU networks of depth $D$ (we care about $D > 2$) and width $H$ (which we will assume is larger than the input dimensionality $N$, to simplify our proofs) and derive a generalization bound on the original network that does not scale with the product of spectral norms of the weight matrices.  Let  $\relu{\cdot}$  denote the ReLU activation. We consider a network parameterized by $\mathcal{W} = (W_1, W_2, \hdots, W_D)$ such that the output of the network is computed as $\nn{\W}{}{}{\vec{x}} = W_D \relu{W_{D-1} \hdots \relu{W_1 \vec{x}}}$.  We denote the value of the $h$th hidden unit on the $d$th layer before and after the activation by $\prenn{\W}{d}{h}{\vec{x}}$ and $\nn{\W}{d}{h}{\vec{x}}$ respectively. We define $\jacobian{\W}{d'}{d}{}{\vec{x}}\coloneqq{\partial \prenn{\W}{d}{}{\vec{x}}}/{ \partial \prenn{\W}{d'}{}{\vec{x}}}$ to be the Jacobian of the pre-activations of layer $d$ with respect to the pre-activations of layer $d'$ for $d' \leq d$ (each row in this Jacobian corresponds to a unit in layer $d$). In short, we will call this, Jacobian $\verbaljacobian{d'}{d}$. Let $\Z$ denote the random initialization of the network.

Informally, we consider a setting where the learning algorithm satisfies the following conditions on the training data that make it noise-resilient on training data:  a) the $\ell_2$ norm of the hidden layers are all  small, b) the pre-activation values are all sufficiently large in magnitude, c) the Jacobian of any layer with respect to a lower layer, has rows with a small $\ell_2$ norm, and has a small spectral norm.

%The abstract framework in Chapter~\ref{chap:datadependent-pacbayes} requires us to formally cast these conditions  In Chapter~\ref{chap:noise-resilience}, we have already cast these conditions in the form of Equation~\ref{eq:condition} by appropriately defining the properties $\rho$'s and the margins $\Delta^\star$'s in the general framework. We note that these properties are quite similar to those already explored in \cite{arora18compression,neyshabur17exploring}; we provide more intuition about these properties, and how we cast them in our framework in Appendix~\ref{app:noise-resilience-overview}.
%Having defined these properties, we first prove in Lemma~\ref{lem:noise-resilience-induction} in Appendix~\ref{app:noise-resilience} a guarantee equivalent to the abstract inequality in Equation~\ref{eq:generic-noise-resilience}. Essentially, we show that under random perturbations of the parameters, the perturbation in the output of the network and the perturbation in the input-dependent properties involved in (a), (b), (c) themselves can all be bounded in terms of each other. Crucially, these perturbation bounds do not grow with the spectral norms of the network. % and  hold in a way that induces an ordering amongst these properties as required by our framework. 

%Having instantiated the framework as above, we then instantiate the bound provided by the framework. 

Our generalization bound scales with the bounds on the properties in (a) and (c) above as satisfied on the training data, and with the reciprocal of the property in (b) i.e., the smallest absolute value of the pre-activations on the training data. Additionally, our bound has an explicit dependence on the depth of the network, which arises from the fact that we generalize $R=\mathcal{O}(D)$ conditions. Most importantly,
our bound does not have a dependence on the product of the spectral norms of the weight matrices. \\

 \begin{theorem}
\label{thm:relu-generalization}
% Consider a set of positive training-set related constants $\trainset$.  
% \begin{enumerate}
% \item  $\trainjacob{d'}{d}$ for each layer $d= 1, 2, \hdots, D$, and $d' = 1, \hdots, d$
% \item $\trainoutput{d}$, for each hidden layer $d=1,\hdots, D-1$,
% \item $\trainpreact{d}$ for each layer $d = 1, \hdots, D$.
% \end{enumerate}
For any $\delta > 0$, with probability $1-\delta$ over the draw of samples $S\sim\D^m$, for any $\W$,

\begin{align}
\scrL_{\scrD}(f_{\calW})  \leq & \hat{\scrL}^{(\classmargin)}_{S}(f)  +  \bigoh{D \cdot \sqrt{\frac{1}{m-1} \cdot \left( 2 \sum_{d=1}^{D} \elltwo{W_d-Z_d}^2 \frac{1}{(\sigma^\star)^2} + \ln \frac{Dm}{\delta} \right) }}, 
\end{align}
where 
\begin{align}
\frac{1}{\sigma^\star} \coloneqq \bigoh{ \sqrt{H} \cdot \sqrt{\ln \left(DH\sqrt{m}\right)} \times \max \{\boundoutput,\boundhiddenpreact, \boundjacob, \boundspec,  \boundoutputpreact \}},
\end{align}

where,
\begin{align}
\boundoutput& \coloneqq \bigoh{  \max_{1 \leq d < D} \frac{\sum_{d'=1}^{d} {\trainjacob{d'}{d}} \trainoutput{d'-1} }{ \trainoutput{d} } },  \\
 \boundhiddenpreact & \coloneqq \bigoh{ \max_{1 \leq d < D}
 \frac{\sum_{d'=1}^{d}  \trainjacob{d'}{d} \trainoutput{d'-1} }{    \sqrt{H} \trainpreact{d} }},   \\
 \boundjacob & \coloneqq  \bigoh{ \max_{1 \leq d < D}  \max_{1 \leq d' < d \leq D}  \frac{{\trainjacob{d'}{d-1} } + \matrixnorm{W_d}{2,\infty} \sum_{d''=d'+1}^{d-1} \trainspec{d''}{d-1}  \trainjacob{d'}{d''-1}}{  \trainjacob{d'}{d} } },   \\
  \boundspec & \coloneqq  \bigoh{ \max_{1 \leq d < D}  \max_{1 \leq d' < d \leq D}  \frac{ \trainspec{d'}{d-1} + \spec{W_d} \sum_{d''=d'+1}^{d-1} \trainspec{d''}{d-1} \trainspec{d'}{d''-1}  }{  \trainspec{d'}{d} } }, \\
  \boundoutputpreact & \coloneqq \bigoh{ 
 \frac{\sum_{d=1}^{D} \trainjacob{d}{D} \trainoutput{d-1} }{   \sqrt{H} \classmargin }},  
\end{align}
where,
\begin{enumerate}
\item  $\trainoutput{d} \coloneqq \max \left(\max_{(\vec{x},y) \in S} \elltwo{\nn{\W}{d}{}{\vec{x}}}, 1 \right)$ is an upper bound on the $\ell_2$ norm of the output of each hidden layer $d=0,1,\hdots, D-1$ on the training data. Note that for layer $0$, this would correspond to the $\ell_2$ norm of the input.
\item $\trainpreact{d} \coloneqq \min_{(\vec{x},y) \in S} \min_h \ellone{\nn{\W}{d}{h}{\vec{x}}}$ is a lower bound on the absolute values of the pre-activations for each layer $d = 1, \hdots, D$ on the training data.
\item $\trainjacob{d'}{d} \coloneqq \max \left(\max_{(\vec{x},y) \in S} \matrixnorm{\jacobian{\W}{d'}{d}{}{\vec{x}}}{2,\infty}, 1 \right)$ is an upper bound on the row $\ell_2$ norms of the Jacobian for each layer $d= 1, 2, \hdots, D$, and $d' = 1, \hdots, d$ on the training data. 
\item $\trainspec{d'}{d} \coloneqq \max \left(\max_{(\vec{x},y) \in S} \spec{\jacobian{\W}{d'}{d}{}{\vec{x}}}, 1 \right)$ is an upper bound on the spectral norm of the Jacobian for each layer $d= 1, 2, \hdots, D$, and $d' = 1, \hdots, d$ on the training data.
	\end{enumerate}
\end{theorem}

\section[Proof]{Proof of Theorem~\ref{thm:relu-generalization}}

\begin{proof}
Overall, our idea is to use the noise-resilience results for deep networks given in Chapter~\ref{chap:noise-resilience}, and apply them within the abstract PAC-Bayesian framework of Chapter~\ref{chap:datadependent-pacbayes}.

\subsection{Notations.}
To make the presentation of our proof cleaner, we will set up some notations. First, we use $\trainset$  to denote the ``set'' of constants related to the norm bounds on training set defined in the Theorem above. (Here we use the term set loosely, like we noted in Chapter~\ref{chap:noise-resilience}.) Based on these training set related constants, we also define $\testset$ to be the following constants corresponding to weaker norm-bounds related to the {\em test data}:

%   the following positive constants which will correspond to norm bounds on different properties of the network {\em on the training data}:
% \begin{enumerate}
% \item $\trainoutput{d}$, for each hidden layer $d=0,1,\hdots, D-1$, (we will use this to bound $\ell_2$ norms of the outputs of the layers of the network on a training input)\footnote{Note that $\trainoutput{0}$ is essentially the $\ell_2$-norm bound on the input $\vec{x}$.}
% \item $\trainpreact{d}$ for each layer $d = 1, \hdots, D$, (we will use this to bound magnitudes of the preactivations values of the network on a training input).
% \item  $\trainjacob{d'}{d}$ for each layer $d= 1, 2, \hdots, D$, and $d' = 1, \hdots, d$ (we will use this to bound $\ell_2$ norms of rows in the Jacobians of the network for a training input)
% \item  $\trainspec{d'}{d}$ for each layer $d= 1, 2, \hdots, D$, and $d' = 1, \hdots, d$ (we will use this to bound spectral norms of the Jacobians of the network for a training input)
% \end{enumerate}
%As we did in the previous section, to simplify our discussion, we will group these constants into a `set' $\trainset$.

\begin{enumerate}
\item $\testoutput{d} \coloneqq \focus{2}\trainoutput{d}$, for each hidden layer $d=0,1,\hdots, D-1$, (we will use this to bound $\ell_2$ norms of the outputs of the layers of the network on a test input)
\item $\testpreact{d} \coloneqq \trainpreact{d}/\focus{2}$ for each layer $d = 1, \hdots, D$, (we will use this to bound magnitudes of the preactivations values of the network on a test input).
\item  $\testjacob{d'}{d}\coloneqq \focus{2}\trainjacob{d'}{d}$ for each layer $d= 1, 2, \hdots, D$, and $d' = 1, \hdots, d$ (we will use this to bound $\ell_2$ norms of rows in the Jacobians of the network for a test input)
\item  $\testspec{d'}{d}\coloneqq \focus{2}\trainspec{d'}{d}$ for each layer $d= 1, 2, \hdots, D$, and $d' = 1, \hdots, d$ (we will use this to bound spectral norms of the Jacobians of the network for a test input)\\

\end{enumerate}
%{\bf Note:} For any $c \in [0,1]$, we will use $c C_D$ to denote the constants in $C_D$ multiplied by a factor of $c$.

% {\bf Note about (abuse of) notation:} We reiterate a point about our notation which we also made in Appendix~\ref{app:noise-resilience}.
% %\begin{itemize}
% %\item 
% We call $\trainset$ and $\testset$ a `set' to denote a group of related constants by a single symbol. Each element in this set has a particular semantic associated with it, unlike the standard notation of a set.
% %\item For an arbitrary set of constants $\tempset$, we will use the subscript $\tempset_d$ to `index' into the set of only those constants corresponding to layers from $1$ until $d$. $\tempset_D$ denotes the entire set of constants.
% %\end{itemize}

Now, for any given set of constants $\tempset$, for a particular weight configuration $\W$, and for a given input $\vec{x}$, we define the following event which holds when the network satisfies certain norm-bounds defined by the constants $\tempset$ (that are favorable for noise-resilience). %(Throughout our discussion, it is assumed that the network has weights $\W$, so we will not take the trouble to include this in our notation.)

\begin{definition}
\label{def:good-point}
For a set of constants $\tempset$, for network parameters $\W$ and for any input $\vec{x}$, we define $\boundednormevent(\W,\tempset,  \vec{x})$ to be the event that all the following hold good: 
\begin{enumerate}
	\item for all $\tempoutput{d} \in \tempset$,
	 % for every $d \in \{1,2, \hdots, D \}$, 
	 $\elltwo{\nn{\W}{d}{}{\vec{x}}} <\tempoutput{d}$ (Output of the layer does not have too large an $\ell_2$ norm). 
	\item for all $\temppreact{d} \in \tempset$,
	 %$d \in \{1,2, \hdots D \}$, 
	 $\min_h \ellone{\nn{\W}{d}{h}{\vec{x}}} > \temppreact{d}$. (Pre-activation values are not too small). 
	\item for all $\tempjacob{d'}{d} \in \tempset$, 
	%for every $d \in \{1, 2, \hdots, D \}$, and every $d' \in \{1, 2, \hdots,d \}$ 
	$\max_h  \elltwo{\jacobian{\W}{d'}{d}{h}{\vec{x}}} < \tempjacob{d'}{d}$ (Rows of Jacobian do not have too large an $\ell_2$ norm). 
	\item for all $\tempspec{d'}{d} \in \tempset$, 
	%for every $d \in \{1, 2, \hdots, D \}$, and every $d' \in \{1, 2, \hdots,d \}$ 
	$\spec{\jacobian{\W}{d'}{d}{}{\vec{x}}} < \tempspec{d'}{d}$ (Jacobian does not have too large a spectral norm). 
\end{enumerate}
\end{definition}

\paragraph{Note:} (Similar to a note under Definition~\ref{def:tolerance-parameters}) A subtle point in the above definition (which we will make use of, to state our theorems) is that if we supply only a subset of $\tempset$  to the above event, $\boundednormevent(\W,\cdot, \vec{x})$ then it would denote the event that  only those subset of properties satsify the respective norm bounds. \\

% We will first show how the bounds on these three kinds of properties generalize to the test dataset by a layerwise induction.That is, for each layer $d$, we show that on most test data,
  % the $\ell_2$ norm of each row in its Jacobians are upper bounded, the $\ell_2$ norm of that layer's activations is upper bounded, and its magnitude of the preactivations are sufficiently large. Formally we will bound the proportion of test datapoints for which the event $\boundednormevent_D(\W,\testset,\vec{x})$ does not hold (where recall that we have defined $\testset$ to a weaker norm bound than $\trainset$). Having generalized the bounds on these properties, we will be able to quantify the noise-resilience of the output of the network on test data points, which will allow us to derive the main generalization bound.

To apply the framework from Chapter~\ref{chap:datadependent-pacbayes}, we will have to first define and {\em order} the input-dependent properties $\rho$ and the corresponding margins $\Delta^{\star}$ used in Theorem~\ref{thm:generic-generalization}. 
%\footnote{Note that in the statement of Theorem~\ref{thm:generic-generalization}, $T$ had a third argument which corresponded to the true label of the input, $y$. In the below choices for $T$, we have dropped this argument as these properties are independent of $y$. However, we will need that argument when we choose a $T$ for our final bound on the classification error.}
We will define these properties in terms of the following functions: $\elltwo{\nn{\W}{d}{}{\vec{x}}}$, $\nn{\W}{d}{h}{\vec{x}}$,  $\elltwo{\jacobian{\W}{d'}{d}{h}{\vec{x}}}$ and $\spec{\jacobian{\W}{d'}{d}{}{\vec{x}}}$. Following this definition, we will create an ordered grouping of these properties. 
%We define the following set of functions $\propertyset$, and the corresponding set of margin values  $\marginset$:
\begin{definition}
\label{def:properties-and-margins}
For ReLU networks, we enumerate the input-dependent properties (on the left below) and their corresponding margins (on the right below) denoted with a superscript $\Delta$:
\[
\begin{array}{ccl}
 2\trainoutput{d}-\elltwo{\nn{\W}{d}{}{\vec{x}}} &  \marginoutput{d} \coloneqq \trainoutput{d} & \text{for } d= 0, 1, 2, \hdots, D-1, \\
\ellone{\nn{\W}{d}{h}{\vec{x}}} - \frac{\trainpreact{d}}{2} & \marginpreact{d} \coloneqq \frac{\trainpreact{d}}{2} & \text{for } d= 1, 2, \hdots, D-1 \; \; \text{for all possible } h, \\ 
  2\trainjacob{d'}{d} -  \elltwo{\jacobian{\W}{d'}{d}{h}{\vec{x}}} & \marginjacob{d'}{d} \coloneqq  \trainjacob{d'}{d}  &  \text{for } \; d= 1,\hdots, D \;  \; d' = 1, \hdots, d-1 \; \; \text{for all possible } h,\\ 
  2\trainspec{d'}{d} - \spec{\jacobian{\W}{d'}{d}{}{\vec{x}}} & \marginspec{d'}{d} \coloneqq  \trainspec{d'}{d}  &  \text{for } \; d= 1,\hdots, D \; \; d' = 1, \hdots, d-1,
\end{array}
\]
and for the output layer $D$:
\begin{align*}
 {\nn{\W}{}{}{\vec{x}}}[y] -  \max_{j \neq y}{\nn{\W}{}{}{\vec{x}}}[j] & \; \; & \marginpreact{D}  \coloneqq {\classmargin}. \\
\end{align*}
We will use the notation $\marginset$ to denote the sets of all margin terms defined on the right side above, and $\marginset/2$ to denote the values in that set divided by $2$.
\end{definition}

\paragraph{On the choice of the above functions and margin values.}
 Recall that for a specific input-dependent property $\rho$ and its margin $\Delta^\star$, the condition in Equation~\ref{eq:condition} requires that $\rho(\W,\vec{x},y) > \Delta^\star$. When we generalize these conditions in Theorem~\ref{thm:pac-bayesian}, we will assume that these are satisfied on the training data, and we show that on the test data the approximate version of these conditions, namely $\rho(\W,\vec{x},y) > 0$ hold. Below, we show what these conditions and their approximate versions translate to, in terms of norm-bounds on  $\elltwo{\nn{\W}{d}{}{\vec{x}}}$, $\nn{\W}{d}{h}{\vec{x}}$,  $\elltwo{\jacobian{\W}{d'}{d}{h}{\vec{x}}}$ and $\spec{\jacobian{\W}{d'}{d}{}{\vec{x}}}$; we encapsulate our statements in the following fact for easy reference later in our proof.

\begin{fact}
\label{fact:norm-bounds}
 When $\rho, \Delta^\star$ correspond to  
 $2\trainoutput{d}-\elltwo{\nn{\W}{d}{}{\vec{x}}}$ and  $\marginoutput{d}$, the conditions in Equation~\ref{eq:condition}  translate to upper bounds on the $\ell_2$ norm of the layer as:
 \begin{align}
 \rho(\W,\vec{x},y) > \Delta^\star  \equiv &  {\elltwo{\nn{\W}{d}{}{\vec{x}}} < \trainoutput{d}},  \\
 \rho(\W,\vec{x},y) > 0  \equiv & \elltwo{\nn{\W}{d}{}{\vec{x}}} < 2\trainoutput{d} = \testoutput{d}.
 \end{align}

 When $\rho, \Delta^\star$ correspond to  
$\ellone{\nn{\W}{d}{h}{\vec{x}}} - \frac{\trainpreact{d}}{2}$ and $\marginpreact{d} \coloneqq \frac{\trainpreact{d}}{2}$, 
then the conditions translate to lower bounds on the pre-activation values as:
 \begin{align}
 \rho(\W,\vec{x},y) > \Delta^\star  \equiv &  { \ellone{\nn{\W}{d}{h}{\vec{x}}} > \trainpreact{d}},  \\
 \rho(\W,\vec{x},y) > 0  \equiv & \ellone{\nn{\W}{d}{h}{\vec{x}}} > \trainpreact{d}/2 = \testpreact{d}.
 \end{align}

 When $\rho, \Delta^\star$ correspond to  
  $2\trainjacob{d'}{d} -  \elltwo{\jacobian{\W}{d'}{d}{h}{\vec{x}}}$ and $\marginjacob{d'}{d}$, the conditions translate to upper bounds on the row $\ell_2$ norm of the Jacobian as:

   \begin{align}
 \rho(\W,\vec{x},y) > \Delta^\star  \equiv & {\elltwo{\jacobian{\W}{d'}{d}{h}{\vec{x}}} < \trainjacob{d'}{d}}, \\
 \rho(\W,\vec{x},y) > 0  \equiv & {\elltwo{\jacobian{\W}{d'}{d}{h}{\vec{x}}} < 2 \trainjacob{d'}{d}} = \testjacob{d'}{d}.
 \end{align}

 When $\rho, \Delta^\star$ correspond to  
$2\trainspec{d'}{d} - \spec{\jacobian{\W}{d'}{d}{}{\vec{x}}}$ and  $\marginspec{d'}{d}$, the conditions translate to upper bounds on the spectral norms of the Jacobian as:

   \begin{align}
 \rho(\W,\vec{x},y) > \Delta^\star  \equiv & \spec{\jacobian{\W}{d'}{d}{}{\vec{x}}} < \trainspec{d'}{d},\\
 \rho(\W,\vec{x},y) > 0  \equiv & \spec{\jacobian{\W}{d'}{d}{}{\vec{x}}} > 2 \trainspec{d'}{d} = \testspec{d'}{d}.
 \end{align}

 When $\rho, \Delta^\star$ correspond to  
$ {\nn{\W}{}{}{\vec{x}}}[y] -  \max_{j \neq y}{\nn{\W}{}{}{\vec{x}}}[j]$ and $ \marginpreact{D} $, the conditions translate to lower bounds on the margin:
   \begin{align}
 \rho(\W,\vec{x},y) > \Delta^\star  \equiv & {\nn{\W}{}{}{\vec{x}}}[y] -  \max_{j \neq y}{\nn{\W}{}{}{\vec{x}}}[j] > 0, \\
 \rho(\W,\vec{x},y) > 0  \equiv & {\nn{\W}{}{}{\vec{x}}}[y] -  \max_{j \neq y}{\nn{\W}{}{}{\vec{x}}}[j] > \classmargin.
 \end{align}
\end{fact}

\paragraph{Grouping and ordering the properties.}
Now to apply the abstract generalization bound in Theorem~\ref{thm:generic-generalization}, recall that we need to come up with an ordered grouping of the functions above such that we can realize the constraint given in Equation~\ref{eq:generic-noise-resilience}. Specifically, this constraint effectively required that, for a given input, the perturbation in the properties grouped in a particular set be small, given that all the properties in the preceding sets satisfy the corresponding conditions on them. To this end, we make use of Lemma~\ref{lem:noise-resilience-induction} where we have proven perturbation bounds relevant to the properties we have defined above. Our lemma also naturally induces dependencies between these properties in a way that they can be ordered as required by our framework.

The order in which we traverse the properties is as follows, as dictated by Lemma~\ref{lem:noise-resilience-induction}. We will go from layer $0$ uptil $D$. For a particular layer  $d$, we will first group the properties corresponding to the spectral norms of the Jacobians of that layer whose corresponding margins are $\{\marginspec{d'}{d}\}_{d'=1}^{d}$. Next, we will group the row $\ell_2$ norms of the Jacobians of layer $d$, whose corresponding margins are $\{\marginjacob{d'}{d}\}_{d'=1}^{d}$. Followed by this, we will have a singleton set of the layer output's $\ell_2$ norm whose corresponding margin is $\marginoutput{d}$. We then will group the pre-activations of layer $d$, each of which has the corresponding margin $\marginpreact{d}$. For the output layer, instead of the pre-activations or the output $\ell_2$ norm, we will consider the margin-based property we have defined above.  \footnote{For layer $0$, the only property that we have defined is the $\ell_2$ norm of the input.}
\footnote{Note that the Jacobian for $\verbaljacobian{d}{d}$ is nothing but an identity matrix regardless of the input datapoint; thus we do not need any generalization analysis to bound its value on a test datapoint. Hence, we ignore it in our analysis, as can be seen from the list of properties that we have defined.} Observe that the number of sets $R$ that we have created in this manner, is at most $4D$ since there are at most $4$ sets of properties in each layer.

\paragraph{Proving Constraint in Equation~\ref{eq:generic-noise-resilience}.}
Recall the constraint in Equation~\ref{eq:generic-noise-resilience}  that is required by our framework. For any $r$, the $r$th set of properties need to satisfy the following statement:
\begin{align*}
& \text{if } \forall q < r, \forall l, \rho_{q,l}(\W,\vec{x},y) > 0 \text{ then} \\
& Pr_{\U \sim \N(0,\sigma^2 I)} \Big[ 
{
\exists l  \; |\rho_{r,l}(\W+\U,{\vec{x},y}) - \rho_{r,l}(\W,{\vec{x},y}) | > \frac{\Delta_{r,l}(\sigma)}{2}
 }
 \; \; \text{ and } \\
& \; \; \; \; \; \; \; \; \; \; \; \; \; \; \; { \forall q {<} r,  \forall l \; \;  |\rho_{q,l}(\W+\U,{\vec{x},y}) - \rho_{q,l}(\W,{\vec{x},y}) |{<} \frac{\Delta_{q,l}(\sigma)}{2} } \Big]  \leq \frac{1}{(R+1)\sqrt{m}}. \;\;\; (\ref{eq:generic-noise-resilience})
\end{align*}

Furthermore, we want the perturbation bounds $\Delta_{r,l}(\sigma)$ to satisfy $\Delta_{r,l}(\sigma^\star) \leq \Delta_{r,l}^{\star} $, where $\sigma^\star$ is the standard deviation of the parameter perturbation chosen in the PAC-Bayesian analysis.\\

The next step in our proof is to show that our choice of $\sigma^\star$, and the input-dependent properties, all satisfy the above requirements. To do this, we instantiate Lemma~\ref{lem:noise-resilience-induction} with $\sigma=\sigma^\star$  as in Theorem~\ref{thm:relu-generalization} (choosing appropriate constants), $\toldelta = \frac{1}{4D\sqrt{m}}$ and $\tolset = \marginset/2$. Then, it can be verified that the values of the perturbation bounds in $\tolset'$  in Lemma~\ref{lem:noise-resilience-induction} can be upper bounded by the corresponding value in $\marginset/2$. In other words, we have that for our chosen value of $\sigma$, the perturbations in all the properties and the output of the network can be bounded by the constants specified in $\marginset/2$.
%\footnote{The precise upper bounds that $\sigma^\star$ would provide, for the Jacobian row $\ell_2$ and spectral norms and the layer output $\ell_2$ norms, is slightly different. In particular, the upper bound would correspond to the margin from $\marginset/2$ only if this margin value is at least as large as a constant; if not, the way we have defined $\sigma^\star$ would force the upper bound to be a constant. This is because, even if these norms on the training set are very small, we can afford to have a large $\sigma$ that can perturb these norms by a constant values, to ensure that the network output varies within a margin proportional $\classmargin$. Hence, for this discussion one should imagine $\marginset/2$ to be a set where any small margin values are replaced by a constant. However, for the sake of simplicity, we don't explicitly define it this way.} 
Succinctly, let us say:
\begin{equation}
\tolset' \leq \marginset/2 \label{eq:tolerance-bound}
 \end{equation}\\

% for all but the last function $\proppreact{D}$ and ii) the value $\classmargin/4$ for $\proppreact{D}$.

Given that these perturbation bounds hold for our chosen value of $\sigma$, we will focus on showing that a constraint of the form Equation~\ref{eq:generic-noise-resilience} holds for the row $\ell_2$ norms of the Jacobians $\verbaljacobian{d'}{d}$ for all $d' < d$. A similar approach would apply for the other properties. %For brevity, we will define $\prev{\testset} =   \testset_{d-1}\bigcup \{ \testjacob{d''}{d} \}^{d}_{{d''=d'+1}}$ and $\prev{\tolset} = \tolset_{d-1}\bigcup \{ \toljacob{d''}{d} \}^{d}_{{d''=d'+1}}$

First, we note that the sets of properties preceding the ones corresponding to the row $\ell_2$ norms of Jacobian $\verbaljacobian{d'}{d}$, consists of all the properties upto layer $d-1$. Therefore, 
the precondition for Equation~\ref{eq:generic-noise-resilience} which is of the form $\rho(\W,\vec{x},y) > 0$ for all the previous properties $\rho$, translates to norm bound on these properties involving the constants $\testset_{d-1}$ as discussed in Fact~\ref{fact:norm-bounds}. Succinctly, these norm bounds can be expressed as $\boundednormevent(\W+\U,\testset_{d-1},\vec{x})$. \\

Given that these norm bounds hold for a particular $\vec{x}$, our goal is to argue that the rest of the constraint in Equation~\ref{eq:generic-noise-resilience} holds.
To do this, we first argue that given these norm bounds,  if $\boundedperturbationevent(\W+\U,\marginset_{d-1}/2,\vec{x})$ holds, then so does 
$\unchangedacts_{d-1}(\W+\U,\vec{x})$. This is because, the event $\boundedperturbationevent(\W+\U,\marginset_{d-1}/2   ,\vec{x})$ implies that the pre-activation values of layer $d-1$ suffer a perturbation of at most  $\marginpreact{d-1}/2 = \trainpreact{d-1}/4$ i.e., $\max_h |\nn{\W+\U}{d-1}{h}{\vec{x}} - \nn{\W}{d-1}{h}{\vec{x}}| \leq \trainpreact{d-1}/4$. However, since $\boundednormevent(\W,\testset_{d-1},\vec{x})$ holds, we have that the preactivation values of this layer have a magnitude of at least $\testpreact{d-1} = \trainpreact{d-1}/2$ before perturbation i.e., $\min_h |\nn{\W}{d-1}{h}{\vec{x}}| \geq \trainpreact{d-1}/2$. From these two equations, we have that the hidden units even at layer $d-1$ of the network do not change their activation state (i.e., the sign of the pre-activation does not change) under this perturbation. We can similarly argue for the layers below $d-1$, thus proving that $\unchangedacts_{d-1}(\W+\U,\vec{x})$ holds under $\boundedperturbationevent(\W+\U,\marginset_{d-1}/2,\vec{x})$.\\

Then, from the above discussion on the activation states, and from Equation~\ref{eq:tolerance-bound}, we have that Lemma~\ref{lem:noise-resilience-induction} boils down to the following inequality, when we plug $\sigma = \sigma^\star$:

\begin{align*}
 Pr_{\U} \Big[ 
 \lnot \boundedperturbationevent(\W+\U, {\{ \marginjacob{d'}{d}/2\}}_{d'=1}^{d-1},\vec{x}) \wedge 
 \boundedperturbationevent(\W+\U,\marginset_{d-1}/2,\vec{x}) \Big] \leq  \frac{1}{4D \sqrt{m}}
\end{align*}

 First note that this inequality has the same form as the constraint required by Equation~\ref{eq:generic-noise-resilience} in our framework. Specifically, in place of the generic perturbation bound $\Delta(\sigma^\star)$, we have $\marginjacob{d'}{d}$. Furthermore, recall that our abstract generalization theorem in Theorem~\ref{thm:generic-generalization} required that the perturbation bound $\Delta_{r,l}(\sigma^\star)$ be smaller than the corresponding margin $\Delta_{r,l}^\star$. Since the margin here is $\marginjacob{d'}{d}$, this is indeed the case. 
Through identical arguments for the other sets of input-dependent properties that we have defined, we can similarly show how the constraint in Equation~\ref{eq:generic-noise-resilience} holds. 

Thus, the input-dependent properties we have devised satisfy all the requirements of our framework, allowing us to apply Theorem~\ref{thm:generic-generalization}, with $R \leq 4D$.  Here, we use a prior centered at the random initialization $\Z$; Lemma~\ref{lem:kldivergence} helps simplify the KL-divergence term between the posterior centered at $\W$ and the prior at the random initialization $\Z$.

\paragraph{Covering argument.} To complete our proof, we need to take one more final step. First note that the guarantee in Theorem~\ref{thm:generic-generalization} requires that both $\sigma^\star$ and the margin constants $\Delta_{r,l}^{\star}$ in Equation~\ref{eq:condition} are all chosen before drawing the training dataset. Thus, to apply this bound in practice, one would have to train the network on multiple independent draws of the training dataset (roughly $\mathcal{O}(1/\delta)$ many draws), and then compute norm-bounds on the input-dependent properties across all these runs, and then choose the largest  $\sigma^\star$ based on all these norm-bounds. We emphasize that theoretically speaking, this sort of a bound is still a valid generalization bound that essentially applies to a restricted, norm-bounded class of neural networks. Indeed, the hope is that the implicit bias of stochastic gradient descent ensures that the networks it learns do satisfy these norm-bounds on the input-dependent properties across most draws of the training dataset. 

But for practical purposes, one may not be able to empirically determine norm-bounds that hold on $1-\delta$ of the training set draws, and one might want to get a generalization bound based on norm-bounds that hold on just a single draw. We take this final step in our proof in order to derive such a generalization bound.
%The goal of our final step is only to make sure that we also have a practical bound that can be computed from a single draw of the training dataset. 
We do this via the standard theoretical trick of `covering' the space of all possible norm-bounds. That is, consider the set of $\leq 4D^2$ different constants in $\trainset$ (that bound the different norms), based on which we choose $\sigma^\star$. We will create a `grid' of constants (independent of the training data) such that for any particular run of the algorithm, we can find a point on this grid (that corresponds to a configuration of the constants) for which the norm-bounds still hold for that run. These bounds will be looser, but only by a constant multiplicative factor. This will ensure that the bound resulting from choosing $\sigma^\star$ based on this point on the grid, is only a constant factor looser than choosing $\sigma^\star$ based on the actual norm-bounds for that training set. Then, we will instantiate Theorem~\ref{thm:generic-generalization} for all the points on the grid, and apply a union bound over all of these to get our final bound. \\

We create the grid based as follows. Observe that the bound we get from $\boundoutputpreact$ is at least as large as $\trainoutput{d-1}/(\sqrt{H}\classmargin)$. Then, for any value of $\trainoutput{d-1} = \omega{\sqrt{H}\classmargin \sqrt{m}}$, we will choose a value of $1/\sigma^\star$ that is $\omega{\sqrt{m}}$, rendering the final bound vacuous. Also note that $\trainoutput{d-1} \geq 1$. Thus, we will focus on the interval $[1,\bigoh{\sqrt{H}\classmargin \sqrt{m}} ]$, and grid it based on the points $1,2,4,8, \hdots, \bigoh{\sqrt{H}\classmargin \sqrt{m}}$. Observe that any value of $\trainoutput{d-1}$ can be approximated by one of these points within a multiplicative factor of $2$. Furthermore, this gives rise to at most $\bigoh{\log_2 \sqrt{H}\classmargin \sqrt{m}}$ many points on this grid.  Next, for a given point on this grid, by examining $\boundoutput$ and $\boundoutputpreact$, we can similarly argue how the range of values of $\trainjacob{d'}{d}$ is limited between $1$ and a polynomial in terms of $H$ and $\classmargin$; this range of values can similarly be split into a grid. Then, by examining $\boundhiddenpreact$, we can arrive at a similar grid for the quantity $1/\trainpreact{d}$; by examining $\boundjacob$, we can get a grid for $\trainspec{d'}{d}$ too. In this manner, we can grid the space of all possible configurations of the constants into at most $(poly(H,D,m,\classmargin))^{4D^2}$ many points (since there are not more than $4D^2$ different constants). 

For any given run, we can pick a point from this grid such that the norm-bounds are loose only by a constant multiplicative factor. Finally, we apply Theorem~\ref{thm:generic-generalization} for each of these grids by setting the failure probability to be $\delta/(poly(H,D,m,\classmargin))^{4D^2}$, and then combine them via a union bound. Note, that the resulting bound would have a $\sqrt{\log \frac{(poly(H,D,m,\classmargin))^{4D^2}}{\delta}}/\sqrt{m}$ term, that would only result in a $\sqrt{\frac{D^2 \log poly(H,D,m,\classmargin)}{m}}$ term that does not affect our bound in an asymptotic sense.

\end{proof}

\section[Empirical values]{Empirical study of our bound}
Our bound involves many different terms. To get a sense of what dependencies these terms may or may not have, let us empirically study these quantities.

\paragraph{Experimental details.} In all the experiments (except the one in Figure~\ref{fig:overall-bounds-depth} (b)) we use SGD with learning rate $0.1$ and mini-batch size $64$. We train the network on a subset of $4096$ random training examples from the MNIST dataset to minimize cross entropy loss. We stop training when we classify at least $0.99$ of the data perfectly, with a margin of $\classmargin=10$. In  Figure~\ref{fig:overall-bounds-depth} (b) where we train networks of depth $D=28$, the above training algorithm is quite unstable. Instead, we use Adam with a learning rate of $10^{-5}$ until the network achieves an accuracy of $0.95$ on the training dataset. Finally, we note that all logarithmic transformations in our plots are to the base $10$.

\begin{figure}[t!]
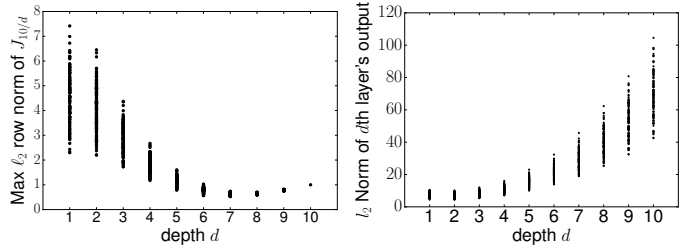

    \centering
    \begin{minipage}[t]{\textwidth}
    \centering
            \begin{minipage}{.28\textwidth}
        \centering
        \adjincludegraphics[width=\textwidth,trim={0 {0} 0 0},clip,valign=t]{Images/jacobian} %
    \end{minipage}%  
        \begin{minipage}{.28\textwidth}
        \centering
        \adjincludegraphics[width=\textwidth,trim={0 {0} 0 0},clip,,valign=t]{Images/alpha} %
    \end{minipage}%        
    \end{minipage}
          \caption{Magnitudes of the Jacobian and hidden layer norms across different datapoints.}
 \label{fig:alpha-jacobian}
\end{figure}

\paragraph{Depth dependence of norm-bounds.} In Figure~\ref{fig:alpha-jacobian} we show how the norm-bounds on the input-dependent properties of the network do not scale as large as the product of spectral norms. In both these plots, we train a network with $D=11$, $H=1280$. In the left plot, each point corresponds to the maximum row $\ell_2$ norm of the Jacobian $\verbaljacobian{d}{10}$ for a particular input. Observe that for any $d$, these quantities are nowhere near as large as a naive upper bound that would roughly scale as $\prod_{d'=d}^{10} \|W_d\|_2 = 2^{10-d}$. On the right, each points corresponds  corresponds to the $\ell_2$ norm of the output of layer $d$ for a particular datapoint.  A naive upper bound on this value would be $\| \vec{x}\| \prod_{d'=1}^{d} \| W_d\|_2 \approx 10 \cdot 2^d$, which would be at least  $100$ times larger than the observed value for $d=10$.

\subsection{Depth dependence of the terms in our bound.}
\begin{figure}[t!]
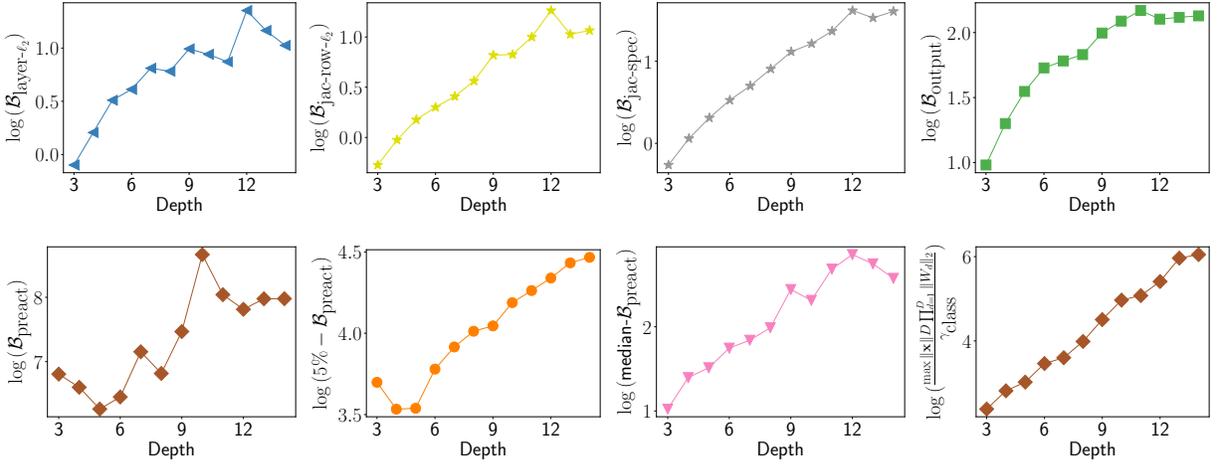

    \centering
    \begin{minipage}[t]{\textwidth}
    \centering
            \begin{minipage}{.25\textwidth}
        \centering
        \adjincludegraphics[width=1\textwidth,trim={0 {0} 0 0},clip,valign=t]{Images/b_out_vs_depth_log_width=40} %
    \end{minipage}%
            \begin{minipage}{.25\textwidth}
        \centering
        \adjincludegraphics[width=1\textwidth,trim={0 {0} 0 0},clip,,valign=t]{Images/b_jac_vs_depth_log_width=40} %
    \end{minipage}%
    \begin{minipage}{.25\textwidth}
        \centering
        \adjincludegraphics[width=1\textwidth,trim={0 {0} 0 0},clip,,valign=t]{Images/b_spectral_vs_depth_log_width=40} %
    \end{minipage}%
     \begin{minipage}{.25\textwidth}
        \centering
        \adjincludegraphics[width=1\textwidth,trim={0 {0} 0 0},clip,valign=t]{Images/b_output_preact_vs_depth_log_width=40} %
    \end{minipage}% 
    \\
    \vspace{5pt}
                    \begin{minipage}{.25\textwidth}
        \centering
        \adjincludegraphics[width=1\textwidth,trim={0 {0} 0 0},clip,valign=t]{Images/b_hidden_preact_vs_depth_log_width=40} %
    \end{minipage}%
                \begin{minipage}{.25\textwidth}
        \centering
        \adjincludegraphics[width=1\textwidth,trim={0 {0} 0 0},clip,valign=t]{Images/five_pc_b_hidden_preact_vs_depth_log_width=40} %
    \end{minipage}%
                    \begin{minipage}{.25\textwidth}
        \centering
        \adjincludegraphics[width=1\textwidth,trim={0 {0} 0 0},clip,valign=t]{Images/median_b_hidden_preact_vs_depth_log_width=40} %
    \end{minipage}%
               \begin{minipage}{.25\textwidth}
        \centering
        \adjincludegraphics[width=1\textwidth,trim={0 {0} 0 0},clip,valign=t]{Images/spectral_norm_term_vs_depth_log_width=40} %
    \end{minipage}%
    \end{minipage}
       \caption{ {Depth-dependence of the terms in our PAC-Bayesian bound and the product of spectral norms}}
       %In the above figure, we plot the logarithm (to the base 10) of values of the terms occurring in the upper bound of $1/\sigma^\star$ for networks of varying depth, with $H=40$. Additionally, we  plot variations of $\boundhiddenpreact$, namely $5\%$-$\boundhiddenpreact$ and median-$\boundhiddenpreact$ as discussed in the text. We also plot the equivalent term from \cite{neyshabur18pacbayes} corresponding to $\max_{\vec{x}} \|\vec{x} \|D\prod_{d=1}^{D} \| W_d\|_2 /\classmargin$. Note that if the slope of the $\log y$ vs $D$ graph is $c$, then the $y \propto (10^c)^D$. 
       \label{fig:quantities-vs-depth}
\end{figure}

 In Figure~\ref{fig:quantities-vs-depth}, we show how the terms in our bound vary for networks of varying depth with a small width of $H=40$ on the MNIST dataset. We observe that $ \boundoutput, \boundoutputpreact,\boundjacob,\boundspec$ typically lie in the range of $[10^0, 10^2]$ and scale with depth as $\propto 1.57^D$. In contrast, the equivalent term from \cite{neyshabur18pacbayes} consisting of the product of spectral norms can be as large as $10^3$ or $10^5$ and scale with $D$ more severely as $2.15^D$.

The bottleneck in our bound is $\boundhiddenpreact$, which scales inversely with the magnitude of the smallest absolute pre-activation value of the network. In practice, this term can be arbitrarily large, even though it does not depend on the product of spectral norms/depth.  This is because some hidden units can have arbitrarily small absolute pre-activation values -- although this is true only for a small proportion of these units.  
   
  To give an idea of the typical, non-pathological magnitude of the pre-activation values, we plot two other  variations of $\boundhiddenpreact$: a) $5\%$-$\boundhiddenpreact$ which is calculated by ignoring $5\%$ of the training datapoints with the smallest absolute pre-activation values and  b) median-$\boundhiddenpreact$ which is calculated by ignoring half the hidden units in each layer with the smallest absolute pre-activation values for each input. We observe that median-$\boundhiddenpreact$  is quite small (of the order of $10^2$), while 
  $5\%$-$\boundhiddenpreact$, while large (of the order of $10^4$), is still orders of magnitude smaller than $\boundhiddenpreact$.

\begin{figure}[t!]
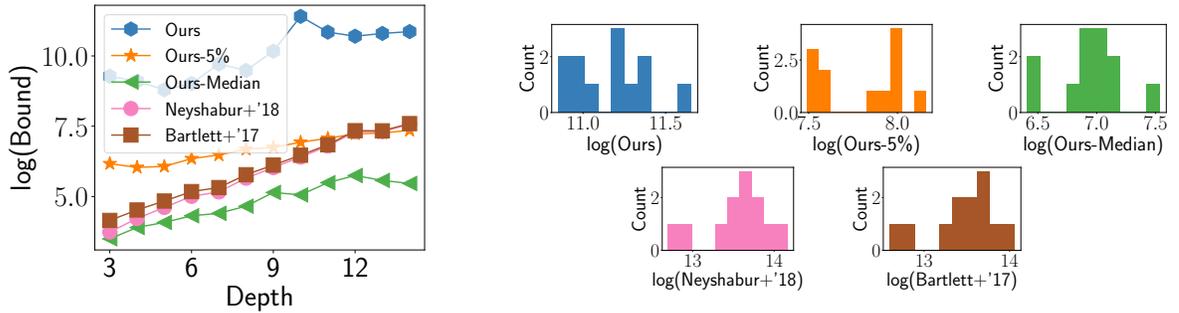

    \centering
    \begin{minipage}[t]{\textwidth}
    \centering
         \begin{minipage}{.45\textwidth}
        \centering
        \adjincludegraphics[width=1\textwidth,trim={0 {0} 0 0},clip,valign=t,scale=0.8]{Images/overall_bound_vs_depth_log_width=40}
	%\caption*{a) Bound vs. Depth}        
         %
    		\end{minipage}%
                \begin{minipage}{.55\textwidth}
			      \begin{minipage}{\textwidth}
			      \begin{minipage}{0.35\textwidth}
        \centering
        \adjincludegraphics[width=1\textwidth,trim={0 {0} 0 0},clip,valign=t,scale=0.8]{Images/overall_Ours_bound_vs_large_depth_3}
    		\end{minipage}%
    					      \begin{minipage}{0.35\textwidth}
        \centering
        \adjincludegraphics[width=1\textwidth,trim={0 {0} 0 0},clip,valign=t,scale=0.8]{Images/overall_Ours-5_bound_vs_large_depth_3}
    		\end{minipage}%
    					      \begin{minipage}{0.35\textwidth}
        \centering
        \adjincludegraphics[width=1\textwidth,trim={0 {0} 0 0},clip,valign=t,scale=0.8]{Images/overall_Ours-Median_bound_vs_large_depth_3}
    		\end{minipage}
	\end{minipage}    		
    		 \\
    					     \begin{minipage}{\textwidth}
    					     \centering
    					      \begin{minipage}{0.33\textwidth}
        \centering
        \adjincludegraphics[width=1\textwidth,trim={0 {0} 0 0},clip,valign=t,scale=0.8]{Images/overall_Neyshabur+18_bound_vs_large_depth_3}
    		\end{minipage}%
    					      \begin{minipage}{0.33\textwidth}
        \centering
        \adjincludegraphics[width=1\textwidth,trim={0 {0} 0 0},clip,valign=t,scale=0.8]{Images/overall_Bartlett+17_bound_vs_large_depth_3}
    		\end{minipage}
	\end{minipage}

%        \begin{tabular}{|c|c|}
%        \hline
%         & $\substack{\textrm{Numerical value of bound} \\ \textrm{for } H=40,D=26}$ \\ \hline
%        Ours & $3.4 \times 10^{13}$ \\ \hline
%        Ours-5\% & $2.3 \times 10^9$ \\ \hline
%        Ours-Median & $2.5 \times 10^8$ \\ \hline
%        Neyshabur et al '18 & $1.1 \times 10^{14}$ \\ \hline
%         Bartlett et al '17 & $9.9 \times 10^{13}$ \\ \hline
%        \end{tabular}
        %\caption{b) Distribution of bounds for $H=40, D=28$}
    \end{minipage}% 
    \end{minipage}
        \caption{\textbf{Behavior of our overall bound.} In the \textbf{left}, we show how our bound has exponentially better depth dependence than other bounds. On the \textbf{right}, we report smaller numerical values compared to other bounds for large depth. } \label{fig:overall-bounds-depth}
\end{figure}

  In Figure~\ref{fig:overall-bounds-depth} we show how our overall bound and existing product-of-spectral-norm-based bounds \citep{bartlett17spectral,neyshabur18pacbayes} vary with depth. We vary the depth of the network (fixing $H=40$) and plot the logarithm of various generalization bounds ignoring the dependence on the training dataset size and a $\log (DH)$ factor in all of the considered bounds. Specifically, we consider our bound, the hypothetical versions of our bound involving $5\%$-$\boundhiddenpreact$ and median-$\boundhiddenpreact$ respectively, and the bounds from \cite{neyshabur18pacbayes} $\frac{\max_{\vec{x}} \| \vec{x}\|_2 D\sqrt{H} \prod_{d=1}^{D} \|W_d \|_2}{\classmargin} \cdot \sqrt{\sum_{d=1}^{D} \frac{\frob{W_d-Z_d}^2}{\| W_d\|_2^2}}$ and \cite{bartlett17spectral} $\frac{\max_{\vec{x}} \| \vec{x}\|_2 \prod_{d=1}^{D} \|W_d \|_2}{\classmargin} \cdot \left({\sum_{d=1}^{D} \left(\frac{\|{W_d-Z_d}\|_{2,1}}{\| W_d\|}\right)^{2/3}}\right)^{3/2}$ both of which have been modified to include distance from initialization instead of distance from origin for a fair comparison.\\ % Observe the last two bounds have a plot with a larger slope than the other bounds indicating that they might potentially do worse for a sufficiently large $D$. Indeed, this can be observed from the plots on the right where we report the distribution of the logarithm of these bounds for $D=28$ (although under training settings different from the experiments on the left; see Appendix~\ref{app:interpret-generalization} for the exact details).

  While our bound is orders of magnitude larger than prior bounds, the key point here is that  our bound grows with depth as $1.57^D$ while prior bounds grow with depth as $2.15^D$ indicating that our bound should perform  asymptotically better with respect to depth. Indeed, we verify that our bound obtains better values than the other existing bounds when $D=28$ (see Figure~\ref{fig:overall-bounds-depth} b). For this figure, we report values for 12 different runs.

Both of our hypothetical bounds where we replace $\boundhiddenpreact$ with   $5\%$-$\boundhiddenpreact$ (see ``Ours-5\%'') and median-$\boundhiddenpreact$  (see ``Ours-Median'') 
  perform orders of magnitude better than our actual bound (note that these two hypothetical bounds do not actually hold good). In fact for larger depth, the bound with $5\%$-$\boundhiddenpreact$ performs better than all other bounds (including existing bounds). This indicates that the only bottleneck in our bound comes from the dependence on the smallest pre-activation magnitudes, and if this particular dependence is addressed, our bound has the potential to achieve tighter guarantees for even smaller $D$ such as $D=8$.

In the next subsection, we show how our bound varies with depth for $H=1280$ (Figure~\ref{fig:alpha-jacobian}, \ref{fig:quantities-vs-depth_1280})  and with width for  $D=8,14$ (Figures~\ref{fig:quantities-vs-width_7} and ~\ref{fig:quantities-vs-width_13}). We dedicate a separate section to these plots because these plots were generated for a computationally cheaper version of our bound (and hence are slightly looser).

\subsection{A computationally cheaper bound}

We  will present a slightly looser bound than the one presented in our main result, motivated by the fact that computing our actual bound is expensive as it involves computing spectral norms of $\Theta(D^2)$ Jacobians on $m$ training datapoints. We note that even this looser bound does not have a dependence on the product of spectral norms, and has similar overall dependence on the depth.

 Specifically, we will consider a bound that is based on a slightly modified noise-resilience analysis. Recall that in Lemma~\ref{lem:noise-resilience-induction}, when we considered the perturbation in the row $\ell_2$ norm Jacobian $\verbaljacobian{d'}{d}$, we bounded Equation~\ref{eq:vanilla-jacobian-bound} in terms of the spectral norms of the Jacobians. Instead of taking this route, if we retained the bound in Equation~\ref{eq:vanilla-jacobian-bound}, we will get a slightly different upper bound on the perturbation of the Jacobian row  $\ell_2$ norm as:
\begin{align*}
\toljacob{d'}{d}' \coloneqq \sigma \sum_{d''=d'+1}^{d} \frob{\jacobian{\W}{d''}{d}{}{\vec{x}}} (\frob{\jacobian{\W}{d'}{d''-1}{}{\vec{x}}} + \toljacob{d'}{d''-1}\sqrt{H})\sqrt{2\ln \frac{D^2H^2}{\toldelta}}
\end{align*}

By using this bound in our analysis, we can ignore the spectral norm terms $\trainspec{d'}{d}$ and derive a generalization bound that does not involve these terms. However, we would now have $\mathcal{O}(D^2)$ conditions instead of $\mathcal{O}(D)$. This is because, the perturbation bound for the row norms of Jacobian $\verbaljacobian{d'}{d}$ now depends on the row norms of Jacobian $\verbaljacobian{d''}{d}$, for all $d'' > d'$. Thus, the row $\ell_2$ norms of these Jacobians must be split into separate sets of properties, and the bound on them generalized one after the other (instead of grouped into one set and generalized all at one go as before). This would give us a similar generalization bound that is looser by a factor of $D$, does not involve $\boundspec$, and where $\boundjacob$ is redefined as:

\[\boundjacob \coloneqq  \bigoh{ \max_{1 \leq d < D}  \max_{1 \leq d' < d \leq D}  \frac{\sum_{d''=d'+1}^{d} \trainjacob{d''}{d-1}  \trainjacob{d'}{d''-1}}{  \trainjacob{d'}{d} } }   \]

All other terms remain the same. In the rest of the discussion, we plot this generalization bound that is looser by a $D$ factor, but still does not depend on the product of the spectral norms. 

\begin{figure}[t!]
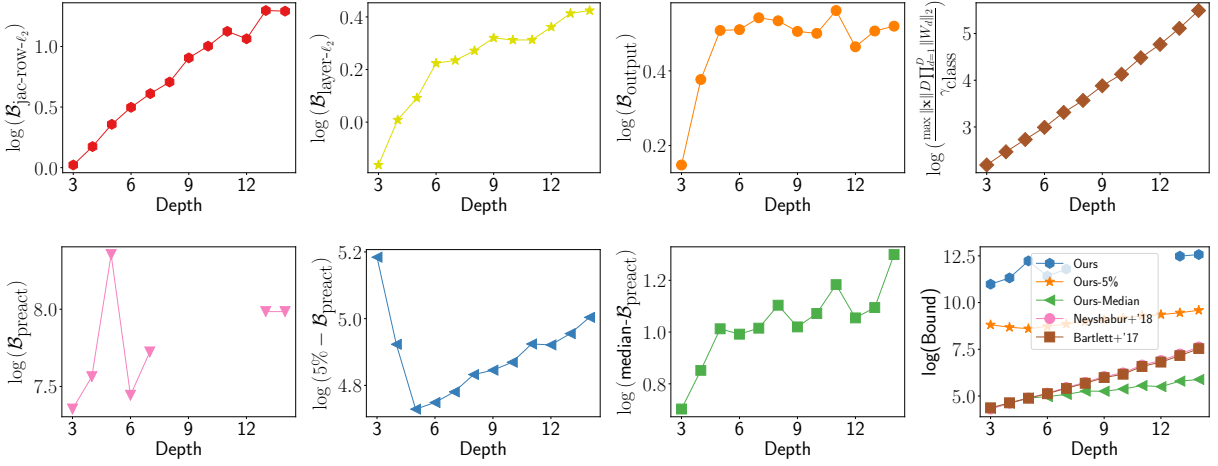

    \centering
    \begin{minipage}[t]{\textwidth}
    \centering
        \begin{minipage}{.25\textwidth}
        \centering
        \adjincludegraphics[width=1\textwidth,trim={0 {0} 0 0},clip,,valign=t]{Images/b_jac_vs_depth_log_width=1280} %
    \end{minipage}%
            \begin{minipage}{.25\textwidth}
        \centering
        \adjincludegraphics[width=1\textwidth,trim={0 {0} 0 0},clip,valign=t]{Images/b_out_vs_depth_log_width=1280} %
    \end{minipage}%
                \begin{minipage}{.25\textwidth}
        \centering
        \adjincludegraphics[width=1\textwidth,trim={0 {0} 0 0},clip,valign=t]{Images/b_output_preact_vs_depth_log_width=1280} %
    \end{minipage}% 
               \begin{minipage}{.25\textwidth}
        \centering
        \adjincludegraphics[width=1\textwidth,trim={0 {0} 0 0},clip,valign=t]{Images/spectral_norm_term_vs_depth_log_width=1280} %
    \end{minipage}%
    \\
    \vspace{5pt}
                    \begin{minipage}{.25\textwidth}
        \centering
        \adjincludegraphics[width=1\textwidth,trim={0 {0} 0 0},clip,valign=t]{Images/b_hidden_preact_vs_depth_log_width=1280} %
    \end{minipage}%
                \begin{minipage}{.25\textwidth}
        \centering
        \adjincludegraphics[width=1\textwidth,trim={0 {0} 0 0},clip,valign=t]{Images/five_pc_b_hidden_preact_vs_depth_log_width=1280} %
    \end{minipage}%
                    \begin{minipage}{.25\textwidth}
        \centering
        \adjincludegraphics[width=1\textwidth,trim={0 {0} 0 0},clip,valign=t]{Images/median_b_hidden_preact_vs_depth_log_width=1280} %
    \end{minipage}%
                   \begin{minipage}{.25\textwidth}
        \centering
        \adjincludegraphics[width=1\textwidth,trim={0 {0} 0 0},clip,valign=t]{Images/overall_bound_vs_depth_log_width=1280} %
    \end{minipage}%
    \end{minipage}
        \caption{Behavior of the terms in our bound and the spectral norm product for a very wide network ($H=1280$).
         } \label{fig:quantities-vs-depth_1280}
\end{figure}

\textbf{Observations.} In Figure~\ref{fig:quantities-vs-depth_1280} we show how the quantities in this bound and the bound itself varies with depth, for a  network of $H=1280$, wider than what we considered in Figure~\ref{fig:quantities-vs-depth}. We observe that  $\boundjacob, \boundoutput, \boundoutputpreact$ typically lie in the range of $[10^0, 10^2]$. In contrast, the equivalent term from \cite{neyshabur18pacbayes} consisting of the product of spectral norms can be as large as $10^5$ for $D=10$. Unfortunately, for large $H$, due to numerical precision issues, the smallest pre-activation value is rounded off to zero and hence $\boundhiddenpreact$ becomes undefined in such situations. However, as noted before, the hypothetical variations  $5\%$-$\boundhiddenpreact$  and median-$\boundhiddenpreact$ are bounded better and achieve significantly smaller values. 
        Finally, observe that our overall bound and all its hypothetical variations have a smaller slope than previous bounds.\\

\begin{figure}[!h]
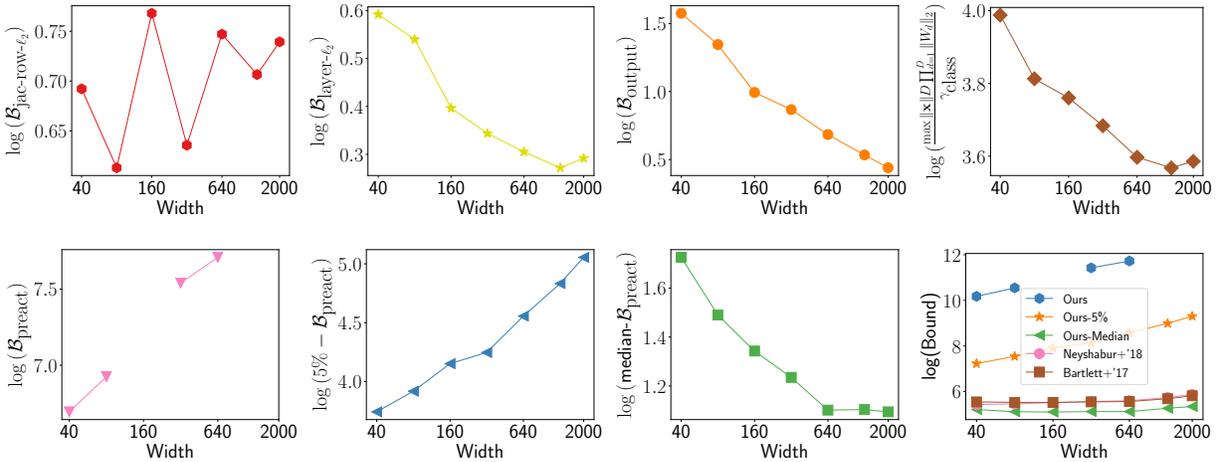

    \centering
    \begin{minipage}[t]{\textwidth}
    \centering
        \begin{minipage}{.25\textwidth}
        \centering
        \adjincludegraphics[width=1\textwidth,trim={0 {0} 0 0},clip,,valign=t]{Images/b_jac_vs_width_log_log_depth=7} %
    \end{minipage}%
            \begin{minipage}{.25\textwidth}
        \centering
        \adjincludegraphics[width=1\textwidth,trim={0 {0} 0 0},clip,valign=t]{Images/b_out_vs_width_log_log_depth=7} %
    \end{minipage}%
                \begin{minipage}{.25\textwidth}
        \centering
        \adjincludegraphics[width=1\textwidth,trim={0 {0} 0 0},clip,valign=t]{Images/b_output_preact_vs_width_log_log_depth=7} %
    \end{minipage}% 
               \begin{minipage}{.25\textwidth}
        \centering
        \adjincludegraphics[width=1\textwidth,trim={0 {0} 0 0},clip,valign=t]{Images/spectral_norm_term_vs_width_log_log_depth=7} %
    \end{minipage}%
    \\
    \vspace{5pt}
                    \begin{minipage}{.25\textwidth}
        \centering
        \adjincludegraphics[width=1\textwidth,trim={0 {0} 0 0},clip,valign=t]{Images/b_hidden_preact_vs_width_log_log_depth=7} %
    \end{minipage}%
                \begin{minipage}{.25\textwidth}
        \centering
        \adjincludegraphics[width=1\textwidth,trim={0 {0} 0 0},clip,valign=t]{Images/five_pc_b_hidden_preact_vs_width_log_log_depth=7} %
    \end{minipage}%
                    \begin{minipage}{.25\textwidth}
        \centering
        \adjincludegraphics[width=1\textwidth,trim={0 {0} 0 0},clip,valign=t]{Images/median_b_hidden_preact_vs_width_log_log_depth=7} %
    \end{minipage}%
                   \begin{minipage}{.25\textwidth}
        \centering
        \adjincludegraphics[width=1\textwidth,trim={0 {0} 0 0},clip,valign=t]{Images/overall_bound_vs_width_log_log_depth=7} %
    \end{minipage}%
    \end{minipage}
        \caption{Log-log plots of various terms in our bound for $D=8$ and varying width $H$.  
         } \label{fig:quantities-vs-width_7}
\end{figure}

\begin{figure}[!h]
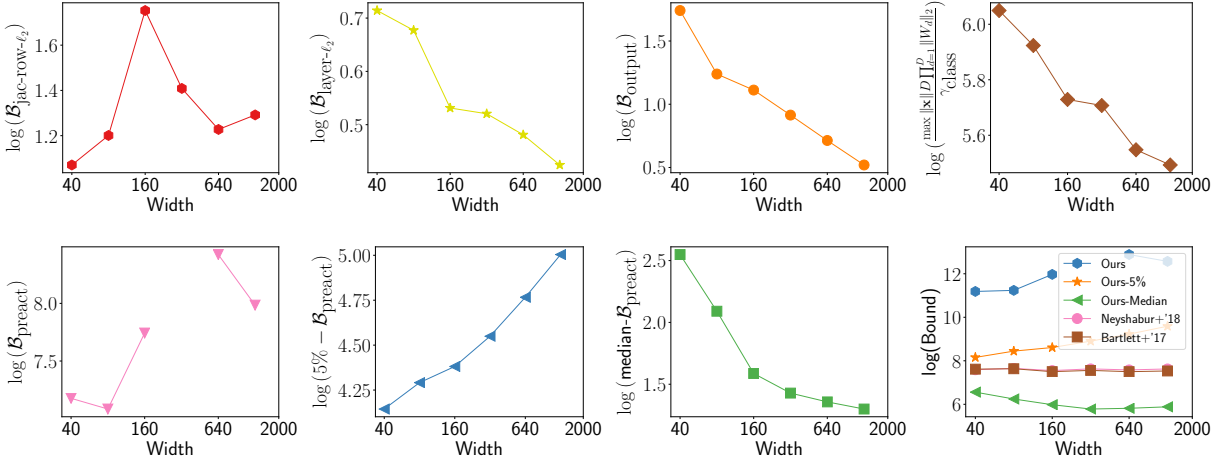

    \centering
    \begin{minipage}[t]{\textwidth}
    \centering
        \begin{minipage}{.25\textwidth}
        \centering
        \adjincludegraphics[width=1\textwidth,trim={0 {0} 0 0},clip,,valign=t]{Images/b_jac_vs_width_log_log_depth=13} %
    \end{minipage}%
            \begin{minipage}{.25\textwidth}
        \centering
        \adjincludegraphics[width=1\textwidth,trim={0 {0} 0 0},clip,valign=t]{Images/b_out_vs_width_log_log_depth=13} %
    \end{minipage}%
                \begin{minipage}{.25\textwidth}
        \centering
        \adjincludegraphics[width=1\textwidth,trim={0 {0} 0 0},clip,valign=t]{Images/b_output_preact_vs_width_log_log_depth=13} %
    \end{minipage}% 
               \begin{minipage}{.25\textwidth}
        \centering
        \adjincludegraphics[width=1\textwidth,trim={0 {0} 0 0},clip,valign=t]{Images/spectral_norm_term_vs_width_log_log_depth=13} %
    \end{minipage}%
    \\
    \vspace{5pt}
                    \begin{minipage}{.25\textwidth}
        \centering
        \adjincludegraphics[width=1\textwidth,trim={0 {0} 0 0},clip,valign=t]{Images/b_hidden_preact_vs_width_log_log_depth=13} %
    \end{minipage}%
                \begin{minipage}{.25\textwidth}
        \centering
        \adjincludegraphics[width=1\textwidth,trim={0 {0} 0 0},clip,valign=t]{Images/five_pc_b_hidden_preact_vs_width_log_log_depth=13} %
    \end{minipage}%
                    \begin{minipage}{.25\textwidth}
        \centering
        \adjincludegraphics[width=1\textwidth,trim={0 {0} 0 0},clip,valign=t]{Images/median_b_hidden_preact_vs_width_log_log_depth=13} %
    \end{minipage}%
                   \begin{minipage}{.25\textwidth}
        \centering
        \adjincludegraphics[width=1\textwidth,trim={0 {0} 0 0},clip,valign=t]{Images/overall_bound_vs_width_log_log_depth=13} %
    \end{minipage}%
    \end{minipage}
        \caption{Log-log plots of various terms in our bound for $D=14$ and varying width $H$.     
         } \label{fig:quantities-vs-width_13}
\end{figure}

In Figure~\ref{fig:quantities-vs-width_7} and Figure~\ref{fig:quantities-vs-width_13} we show log-log (note that here even the $x$-axis has been transformed logarithmically) plots of all the quantities for networks of varying width and $D=8$ and $D=14$ respectively.  (Note that if the slope of the $\log y$ vs $\log H$ plot is $c$, then $y \propto H^c$.)
Here, we observe that $\boundjacob$ is width-independent. On the other hand $\boundoutput$ and the product-of-spectral-norm term mildly decrease with width; $\boundoutputpreact$ {\em decreases} with width at the rate of $1/\sqrt{H}$.

As far as the term $\boundhiddenpreact$ is concerned, recall from our earlier discussions that the minimum pre-activation value $\trainpreact{d}$ of the network tends to be quite small in practice (and can be rounded to zero due to precision issues). Therefore the term $\boundhiddenpreact$ can be arbitrarily large and exhibit considerable variance across different widths/depths and different training runs.  On the other hand, interestingly, the hypothetical variation median-$\boundhiddenpreact$ {\em decreases} with width at the rate of $1/\sqrt{H}$, while  $5\%$-$\boundhiddenpreact$ increases with a $\sqrt{H}$ dependence on width.

Theoretically speaking, as far as the width-dependence is concerned, the best-case scenario for $\boundhiddenpreact$ can be realized when the preactivation values of each layer (which has a total $\ell_2$ norm that is width-independent in practice) are equally spread out across the hidden units. Then we will have that the smallest pre-activation value to be as large as $\Omega(1/\sqrt{H})$.

\section[Comparison with existing work]{Comparison of our noise-resilience conditions with existing conditions}
\label{app:comparison}

Recall from the discussion in the introduction to this chapter that prior works \citep{neyshabur17exploring,arora18compression}
have also characterized noise resilience in terms of conditions on the interactions between the activated weight matrices. Below, we discuss the conditions assumed by these works, which parallel the conditions we have studied in our paper (such as the bounded $\ell_2$ norm in each layer). 

There are two main high level similarities between the conditions studied across these works. First, these conditions --- all of which characterize the interactions between the activated weights matrices in the network --- are assumed only for the training inputs; such an assumption implies noise-resilience of the network on training inputs. Second, there are two kinds of conditions assumed. The first kind  allows one to bound the propagation of noise through the network under the assumption that the activation states do not flip; the second kind  allows one to bound the extent to which the activation states do flip.

\paragraph{Conditions in \cite{neyshabur17exploring}}
Using noise-resilience conditions assumed about the network on the training data,
\cite{neyshabur17exploring} derive a PAC-Bayes based generalization bound on a stochastic network. The first condition in \cite{neyshabur17exploring} characterizes how the Jacobians of different parts of the network interact with each other. Specifically, consider layers $d, d'$ and $d''$ such that $d'' \leq d' \leq d$. Then, consider the 
 Jacobian of layer $d'$ with respect to layer $d''$ and the Jacobian of layer $d$ with respect to $d'$. Then, they require that 
  $\frob{\jacobian{\W}{d'}{d}{}{\vec{x}}} \frob{\jacobian{\W}{d''}{d'-1}{}{\vec{x}}} = \mathcal{O}(\frob{\jacobian{\W}{d''}{d}{}{\vec{x}}})$. This specific condition allows one to bound how the noise injected into the parameters propagate through the network under the assumption that the activation states do not flip. In our paper, we pick an orthogonal approach by assuming an upper bound on 
the Jacobian $\ell_2$ norms and the layer output norms, which allows us to bound the propagation of noise under unchanged activation states.\\

The second condition in \cite{neyshabur17exploring} is that under a noise of variance $\sigma^2$, the number of units that flip their activation state in a particular layer must be bounded as $\mathcal{O}(H \sigma)$ i.e., smaller the noise, the smaller the proportion of units that flip their activation state. This condition is similar  to (although milder than) our lower bounds on the magnitudes of the pre-activation values (which allow us to pick a sufficiently large noise that does not flip the activation states).

Note that a bound on the Jacobian norms corresponds to a bound on the weights input to the active units in the network. However, since  \cite{neyshabur17exploring} allow a few units to flip activation states, they additionally require a bound on the weights input to the inactive units too. Specifically,  for every layer, the maximum row $\ell_2$ norm of the weight matrix $W_i$ is upper bounded in terms of the Frobenius norm of the Jacobian ${\jacobian{\W}{d-1}{d}{}{\vec{x}}}$.

\paragraph{Conditions in \cite{arora18compression}}

In contrast to our work and \cite{neyshabur17exploring}, \cite{arora18compression} use their assumed noise-resilience conditions to derive a bound on a compressed network. Another small technical difference here is that, the kind of noise analysed here is Gaussian noise injected into the activations of each layer of the network (and not exactly the weights).

 The first condition here characterizes the interaction between the Jacobian of layer $d$ with respect to $d'$ and the output of layer $d'$. Specifically, this is a lower bound on the so-called `interlayer cushion', which is evaluated as

\[
\frac{\elltwo{\jacobian{\W}{d'}{d}{}{\vec{x}} \nn{\W}{d'}{}{\vec{x}}}} 
{\frob{\jacobian{\W}{d'}{d}{}{\vec{x}}} \elltwo{\nn{\W}{d'}{}{\vec{x}}}}
\]

Essentially when the interlayer cushion is sufficiently large, it means that the output of layer $d'$ is well-aligned with the larger singular directions of the Jacobian matrix above it; as a result it can be shown that noise injected at/below layer $d'$ diminishes as it propagates through the weights above layer $d'$, assuming the activation states do not flip. Again, our analysis is technically orthogonal to this style of analysis as we bound the propogation of the noise under unchanged activation states assuming that the norms of the Jacobians and the layer outputs are bounded.\\

Another important condition in \cite{arora18compression} is that of ``interlayer smoothness'' which effectively captures how far the set of activation states between two layers, say $d'$ and $d$, flip under noise. Roughly speaking, the assumption made here is that when noise is injected into layer $d'$, there is not much difference between a) the output of the $d$th layer with the activation states of the units in layers $d'$ until $d$ frozen at their original state and b) the output of the $d$th layer with the activation states of the units in layers $d'$ to $d$ allowed to flip under the noise. As stated before, this condition is a relaxed version of our condition that essentially implies that none of the activation states flip.

\subsection{Note on dependence on pre-activation}

Finally, as noted before, we emphasize that the dependence of our bound on the pre-activation values is a limitation in how we characterize noise-resilience through our conditions rather than a drawback in our general PAC-Bayesian framework itself.  Specifically, using the assumed lower bound on the pre-activation magnitudes we can ensure that, under noise, the activation states of the units do not flip;  then the noise propagates through the network in a tractable, ``linear'' manner. Improving this analysis is an important direction for future work. For example, one could modify our analysis to allow perturbations large enough to flip a small proportion of the activation states; one could potentially formulate such realistic conditions by drawing inspiration from the conditions in \cite{neyshabur17exploring,arora18compression}.

However, we note that even though these prior approaches made more realistic assumptions about 
the magnitudes of the pre-activation values, the key limitation in these approaches is that even under our non-realistic assumption, their approaches would yield bounds only on stochastic/compressed networks. {\em Generalizing noise-resilience from training data to test data} is crucial to extending these bounds to the original network, which we accomplish.

\section{Conclusion}

In this chapter, we introduced a PAC-Bayesian framework for leveraging the noise-resilience of deep neural networks on training data, to derive a generalization bound on the original uncompressed, deterministic network. The main philosophy of our approach is to first generalize the noise-resilience from training data to test data using which we convert 
a PAC-Bayesian bound on a stochastic network to a standard margin-based generalization bound.
% We also build a fundamental technique for converting PAC-Bayesian bounds, that is more powerful than prior approaches. 
We apply our approach to ReLU based networks and derive a bound that scales with terms that capture the interactions between the weight matrices better than the product of spectral norms. 

For future work, the most important direction is that of removing the dependence on our strong assumption that  the magnitude of the pre-activation values of the network are not too small on training data. %Another interesting direction would be to tighten the dependence on $D^2$ in our bound in Theorem~\ref{thm:relu-generalization}. 
More generally, a better understanding of the source of noise-resilience in deep ReLU networks would help in applying our framework more carefully in these settings, leading to tighter guarantees on the original network.

\part{Moving away \\  from Uniform Convergence}
\graphicspath{{unif-conv/}}
\chapter{Norm-Based Complexity Measures vs. Training Set Size}

\label{chap:inc-norms}

% \begin{abstract}
% Aimed at explaining the surprisingly good generalization behavior of overparameterized deep networks, recent works have developed a variety of generalization bounds for deep learning,  all  based on the fundamental learning-theoretic technique of uniform convergence. While
% it is well-known that many of these existing bounds are numerically large, through numerous experiments, we bring to light a more concerning aspect of these bounds: 
% in practice,  these bounds can {\em increase} with the training dataset size. Guided by our observations,
% we then present examples of overparameterized linear classifiers and neural networks trained by  gradient descent (GD) where uniform convergence provably cannot ``explain generalization'' -- even if we take into account the implicit bias of GD {\em to the fullest extent possible}. More precisely, even if we consider only the set of classifiers output by GD, which have test errors less than some small $\epsilon$ in our settings, we show that applying (two-sided) uniform convergence on this set of classifiers will yield only a vacuous generalization guarantee larger than $1-\epsilon$. Through these findings,
% we cast doubt on the power of uniform convergence-based generalization bounds to provide a complete picture of why overparameterized deep networks generalize well. 
% \end{abstract}

% add comment on how compression and PAC-bayesian methods can help
\section{Introduction}

Let us take a step back and recall our high level objective in the previous few chapters.  \citet{neyshabur15inductive} and \citet{zhang17generalization} advocated a ``rethinking'' of conventional, algorithm-\textit{independent} techniques to explain generalization in overparameterized deep networks. Specifically, they argued that learning-theoretic approaches must be reformed by identifying and incorporating the implicit bias/regularization of stochastic gradient descent (SGD) \citep{brutzkus18sgd,soudry18sgd,neyshabur17exploring}. 

Subsequently, a huge variety of novel and refined, algorithm-\textit{dependent} generalization bounds for deep networks have been developed in many papers -- including the results we discussed in Chapter~\ref{chap:deterministic-pacbayes}.
 Notably, most of these bounds are based on \textit{uniform convergence}, which is also the most widely used tool in learning theory. The ultimate goal of this ongoing endeavor	
is to derive bounds on the generalization error that 
\begin{enumerate}[(a)]
 \item are small, ideally non-vacuous (i.e., $< 1$),
 \item  reflect the same width/depth dependence as the generalization error (e.g., become smaller with increasing width, as has been surprisingly observed in practice),
 \item  apply to the network learned by SGD (without any modification or explicit regularization) and 
 \item increase with the proportion of randomly flipped training labels (i.e., increase with memorization).\\
\end{enumerate}

While every bound meets some of these criteria (and sheds a valuable but partial insight into generalization in deep learning), there is no known bound that meets all of them simultaneously. While most bounds \citep{neyshabur15norm,bartlett17spectral,golowich17size,neyshabur18pacbayes,nagarajan18deterministic,neyshabur18unitwise} apply to the original network, they are neither numerically small for realistic dataset sizes, nor exhibit the desired width/depth dependencies (in fact, these bounds grow exponentially with the depth). The remaining bounds hold either only on a compressed network \citep{arora18compression} or a stochastic network \citep{langford01not} or a network that has been further modified via optimization or more than one of the above \citep{dziugaite17nonvacuous,zhou2018nonvacuous}. Extending these bounds to the original network was clearly non-trivial as we saw in our resulting from  \citet{nagarajan18deterministic}  in Chapter~\ref{chap:deterministic-pacbayes}.  While strong width-independent bounds have been derived for two-layer ReLU networks \citep{li18learning,zhu18beyond}, these rely on a carefully curated, small learning rate and/or large batch size. We refer the reader to Section~\ref{app:table} for a tabular summary of these bounds.\\

In this chapter, we bring to light another fundamental issue with existing bounds. We demonstrate that these bounds violate another natural but largely overlooked criterion for explaining generalization: 
\begin{enumerate}[(e)]
\item the bounds should decrease with the dataset size at the same rate as the generalization error. 
\end{enumerate}

A fundamental requirement from a generalization bound, however  numerically large the bound may be, is that it should vary inversely with the size of the training dataset size $(m)$ like the observed generalization error. Such a requirement is satisfied even by standard parameter-count-based VC-dimension bounds, like $\mathcal{O}(DH/\sqrt{m})$ for depth $D$, width $H$ ReLU networks \citep{harvey17vc}. Recent works have ``tightened'' the parameter-count-dependent terms in these bounds by replacing them with seemingly innocuous norm-based quantities. 
 However, we show in this chapter that this has also inadvertently introduced training-set-size-count dependencies in the numerator.  With these dependencies, the generalization bounds even {\em increase with training dataset size} for small batch sizes. This observation uncovers a conceptual gap in our understanding of the puzzle, by pointing 
 towards a source of vacuity unrelated to parameter count.

%While existing uniform convergence bounds provide partial explanations about generalization in GD-based trainining of overparameterized deep networks, through our arguments, we question their potential to explain the phenomenon fully. 

\subsection{Related Work}
\label{sec:inc-norms-related}

Prior works like \citet{neyshabur17exploring} and \citet{nagarajan17role}  have studied the behavior of weight norms in deep learning. Although these works do not explicitly study the dependence of these norms on training set size $m$, one can infer from their plots
 that weight norms of deep networks show some increase with $m$.  \citet{belkin18kernel} reported a similar paradox in kernel learning, observing that norms that appear in kernel generalization bounds increase with $m$, and that this is due to noise in the labels. \citet{kawaguchi17generalization} showed that there exist linear models with arbitrarily large weight norms that can generalize well, although such weights are not necessarily found by gradient descent. We crucially supplement these observations in three ways.
  First, we empirically and theoretically demonstrate how, even with {\em zero label noise} (unlike \citep{belkin18kernel}) and by gradient descent (unlike \citet{kawaguchi17generalization}), a significant level of $m$-dependence can arise in the weight norms -- significant enough to make even the generalization bound grow with $m$. Next, we identify uniform convergence as the root cause behind this issue, and thirdly and most importantly, we provably demonstrate this is so.

\subsection{Summary of existing generalization bounds.}
\label{app:table}

In this section, we provide an informal summary of the properties of (some of the) existing generalization bounds for ReLU networks in Table~\ref{tab:summary}.  We note that the analysis in \citet{li18learning} relies on a sufficiently small learning rate ($\approx \mathcal{O}(1/m^{1.2})$) and large batch size ($\approx \Omega(\sqrt{m})$). Hence, the resulting bound cannot describe how generalization varies with any other hyperparameter, like training set size or width, with everything else fixed. A similar analysis in \citet{zhu18beyond} requires fixing the learning rate to be inversely proportional to width. Their bound decreases only as ${\Omega}(1/m^{0.16})$, although, the actual generalization error is typically as small as $\mathcal{O}(1/m^{0.43})$.

\begin{table}[t!]
\begingroup
\fontsize{7.5pt}{12pt}\selectfont
\begin{center}
\begin{tabular}{|c|c|c|c|c|}
\hline
\textbf{Bound} & \shortstack{Norm \\dependencies} & \shortstack{Parameter-count\\ dependencies} & \shortstack{Numerical \\ value} & \shortstack{Holds on \\ original \\ network?} \\ \hline
\citet{harvey17vc} & - & depth $\times$ width & Large & Yes \\ \hline
\shortstack{\citet{bartlett17spectral} \\ \citet{neyshabur18pacbayes}}& \shortstack{Product of spectral norms \\ dist. from init. \\ (not necessarily $\ell_2$)} & \shortstack{poly(width) \\ exp(depth)} & Large & Yes \\ \hline
\shortstack{\citet{neyshabur15norm} \\ \citet{golowich17size}}& \shortstack{Product of Frobenius norms \\ $\ell_2$ dist. from init.} & $\sqrt{\text{width}}^\text{depth}$ & Very large & Yes \\ \hline
\shortstack{\citet{nagarajan18deterministic}}& \shortstack{Jacobian norms \\ $\ell_2$ dist. from init. \\ \text{Inverse pre-activations}} & \shortstack{poly(width) \\ poly(depth)} & \shortstack{Inverse \\ pre-activations \\ can be very large}& Yes \\ \hline
\shortstack{\citet{neyshabur18unitwise}\\ for two-layer networks } & \shortstack{Spectral norm ($1$st layer) \\ $\ell_2$ Dist. from init ($1$st layer) \\ Frobenius norm ($2$nd layer)} & $\sqrt{\text{width}}$ & {Small} & Yes \\ \hline
\citet{arora18compression} & \shortstack{Jacobian norms \\ dist. from init.} & \shortstack{poly(width) \\ poly(depth)}& Small & \shortstack{No. Holds on \\ compressed network} \\ \hline
\citet{dziugaite17nonvacuous} & \shortstack{dist. from init. \\ Noise-resilience of network} &  - & \shortstack{Non-vacuous\\ on MNIST} & \shortstack{No. Holds on an \\optimized, stochastic \\ network}\\ \hline
\citet{zhou2018nonvacuous} & \shortstack{Heuristic compressibility \& \\ noise-resilience of network} &  - &  \shortstack{Non-vacuous\\on ImageNet} & \shortstack{No. Holds on an \\ optimized, stochastic, \\ heuristically compressed, \\ network}\\ \hline
\citet{zhu18beyond} & \shortstack{$L_{2,4}$ norm ($1$st layer) \\ Frobenius norm ($2$nd layer)} & - & \shortstack{Small for carefully\\  scaled init.\\  and learning rate} & Yes\\ \hline
\citet{li18learning} & - & - & \shortstack{Small for carefully\\  scaled batch size \\ and learning rate} & Yes \\ \hline 
\end{tabular}
\end{center}
\vspace{5pt}
\caption{Summary of generalization bounds for ReLU networks. } \label{tab:summary}
\endgroup
\end{table}

\section{Norms grow with training set size}
\label{sec:uc-experiments}

\begin{figure}[t!]
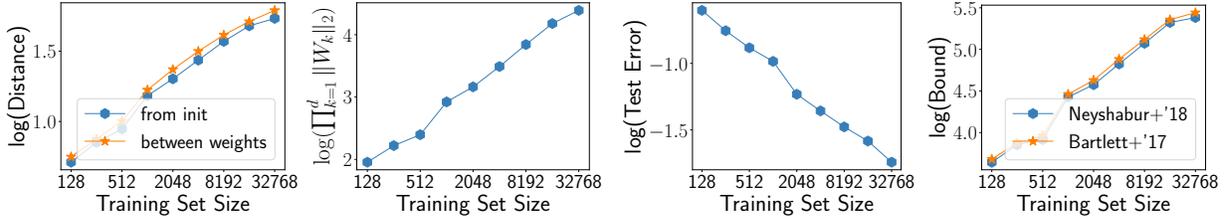

    \centering
    \begin{minipage}[t]{\textwidth}
           \begin{minipage}{.25\textwidth}
        \centering
        \adjincludegraphics[width=1\textwidth,trim={0 {0} 0 0},clip,,valign=t]{UniformConvergenceImages/distance_vs_m_width=1024_margin} %
    \end{minipage}%
            \begin{minipage}{.25\textwidth}
        \centering
        \adjincludegraphics[width=1\textwidth,trim={0 {0} 0 0},clip,valign=t]{UniformConvergenceImages/spectral_norm_vs_m_width=1024} %
    \end{minipage} 
               \begin{minipage}{.25\textwidth}
        \centering
        \adjincludegraphics[width=1\textwidth,trim={0 {0} 0 0},clip,,valign=t]{UniformConvergenceImages/test_error_m_width=1024_bs=1} %
    \end{minipage}%
               \begin{minipage}{.25\textwidth}
        \centering
        \adjincludegraphics[width=1\textwidth,trim={0 {0} 0 0},clip,,valign=t]{UniformConvergenceImages/bound_m_width=1024_bs=1_margin} %
    \end{minipage}%
    \end{minipage}
       \caption{Training set size vs. norms, test error and bounds.
       % Experiments in Section~\ref{sec:uc-experiments}:} In the \textbf{first} figure, we plot (i)  $\ell_2$ the distance of the network from the initialization and (ii) the $\ell_2$ distance between the weights learned on two random draws of training data starting from the same initialization.
       % % Both these quantities grow as $\Omega(m^{0.42})$.  
       % In the \textbf{second} figure we plot the product of spectral norms of the weights matrices.
       % % and observe that it grows as fast as $\Omega(m)$ for $D=5$. See Figure~\ref{fig:depth-wise}  in the appendix, for a layer-by-layer plot of these terms.       
       % In the \textbf{third} figure, we plot the  test error.
       % %which decreases with sample size as $\mathcal{O}(1/m^{0.43})$.  
       % In the \textbf{fourth} figure, we plot the bounds from \citet{neyshabur18pacbayes,bartlett17spectral}.
       % % and observe that they grow as $\Omega(m^{0.68})$. 
       % Note that we have presented log-log plots and the exponent of $m$ can be recovered from the slope of these plots. 
       } 
       \label{fig:terms}
\end{figure}

\paragraph{Experimental details.} We focus on fully connected networks of depth $D=5$, width $H=1024$ trained on MNIST, although we consider other settings in some later experiments. We use SGD with learning rate $0.1$ and batch size $1$ to minimize cross-entropy loss until $99\%$ of the training data are classified correctly by a {margin} of at least $\gamma^\star =10$. We emphasize that, from the perspective of generalization guarantees, this stopping criterion helps standardize training across different hyperparameter values, including different values of $m$ \citep{neyshabur17exploring}. Now, observe that for this particular stopping criterion, the test error empirically decreases with size $m$ as $1/m^{0.43}$ as seen in Figure~\ref{fig:terms} (third plot). However, we will see that the story is starkly different for the generalization bounds.

Before we examine the overall generalization bounds themselves, 
we first focus on two quantities that recur in the numerator of many recent bounds: the $\ell_2$ distance of the weights from their initialization \citep{dziugaite17nonvacuous,nagarajan17role} % TODO: cite our pac bayes paper and distance from initialization paper and role of overparametrization paper
 and the product of spectral norms of the weight matrices of the network \citep{neyshabur18pacbayes,bartlett17spectral}.

\paragraph{Observations.} 
 We observe in Figure~\ref{fig:terms} (first two plots, blue lines) that both these quantities grow at a polynomial rate with $m$: the former at the rate of at least $m^{0.4}$ and the latter at a rate of $m$. Our observation is a follow-up to our results from Chapter~\ref{chap:dist-from-init}
\citep{nagarajan17role} where we argued that while distance of the parameters from {the origin}
  grows with width as $\Omega(\sqrt{H})$, the distance from initialization is width-independent (and even decreases with width); hence, they concluded that incorporating the initialization would improve generalization bounds by a $\Omega(\sqrt{H})$ factor. However, our observations imply that, even though distance from initialization would help explain generalization better in terms of width, it conspicuously fails to help explain generalization in terms of its dependence on $m$.

\begin{figure}[t!]
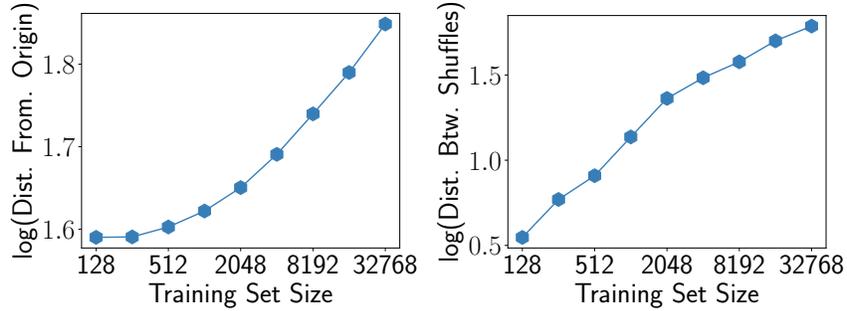

    \centering
        \begin{minipage}{.35\textwidth}
        \centering
                \adjincludegraphics[width=1\textwidth,trim={0 {0} 0 0},clip,valign=t]{UniformConvergenceImages/frobenius_vs_m_width=256_margin} %
    \end{minipage}%
            \begin{minipage}{.35\textwidth}
                    \adjincludegraphics[width=1\textwidth,trim={0 {0} 0 0},clip,,valign=t]{UniformConvergenceImages/within_distance_vs_m_width=1024} %
        \centering
    \end{minipage}
       \caption{Training setsize vs. distance from origin, and distance between two weights learned on shuffled datasets.
       %On the \textbf{left}, we plot the distance between the weights learned on the two different shuffles of the same dataset, and it grows as fast as the distance from initialization. On the \textbf{right}, we plot the distance of the weights from the origin, learned for a network of width $H=256$ and depth $d=6$;   for sufficiently large $m$, this grows as $\Omega(m^{0.42})$.  
       }      \label{fig:more-evidence-dist}
\end{figure}

\paragraph{Frobenius norms grow with $m$ when $m \gg H$.} Some bounds like \citep{golowich17size} depend on the Frobenius norms of the weight matrices (or the distance from the {\em origin}), which as noted in \cite{nagarajan17role} are in fact width-dependent, and grow as $\Omega(\sqrt{H})$. However, even these terms do grow with the number of samples in the regime where $m$ is larger than $H$. In Figure~\ref{fig:more-evidence-dist} (left), we report the total distance from origin of the learned parameters for a network with $H=256$ (we choose a smaller width to better emphasize the growth of this term with $m$); here, we see that for $m > 8192$, the distance from origin grows at a rate of $\Omega(m^{0.42})$ that is quite similar to what we observed for distance from initialization.

\paragraph{Diameter of explored parameter space.} We also examine another quantity as an alternative to distance from initialization:  the $\ell_2$ diameter of {\em the parameter space explored by SGD}. That is, for a fixed initialization and data distribution, we consider the set of all parameters learned by SGD across all draws of a dataset of size $m$; we then consider the diameter of the smallest ball enclosing this set. 
If this diameter exhibits a better behavior than the above quantities, one could then explain generalization better by replacing the distance from initialization with the distance from the center of this ball in existing bounds.  As a lower bound on this diameter, we consider the distance between the weights learned on two independently drawn datasets from the given initialization. Unfortunately, we observe that even this quantity shows a similar undesirable behavior with respect to $m$ like distance from initialization (see Figure~\ref{fig:terms}, first plot, orange line). 
In fact, in Figure~\ref{fig:more-evidence-dist} (right), we show that even the distance between the solutions learned on the same draw, but a different shuffling of the dataset grows substantially with $m$.

\begin{figure}[t!]
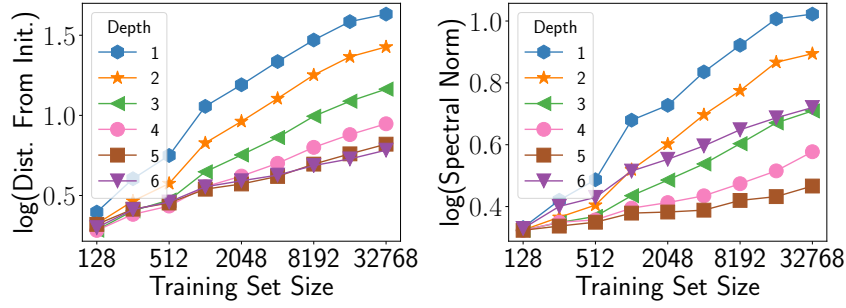

    \centering
            \begin{minipage}{.35\textwidth}
        \adjincludegraphics[width=1\textwidth,trim={0 {0} 0 0},clip,,valign=t]{UniformConvergenceImages/distance_from_init_vs_m_width=1024_bs=1_by_depth} %
    \end{minipage}%
                \begin{minipage}{.35\textwidth}
        \adjincludegraphics[width=1\textwidth,trim={0 {0} 0 0},clip,valign=t]{UniformConvergenceImages/spectral_norm_vs_m_width=1024_by_depth} %
    \end{minipage}% 
       \caption{Layerwise norms vs. training set size.   }      \label{fig:depth-wise}
\end{figure}

\paragraph{Layerwise dependence on $m$.}  In Figure~\ref{fig:depth-wise}, we show how the terms grow with sample size $m$ for each layer individually. Our main observation is that the first layer suffers from the largest dependence on $m$.

\paragraph{Effect of squared error loss.} It may be tempting to think that our observations are peculiar to the cross-entropy loss for which the optimization algorithm diverges to infinity.
Thus, one might suspect that our results are sensitive to the stopping criterion. It would therefore be useful to consider the squared error loss where the optimum on the training loss can be found in a finite distance away from the random initialization. Specifically, we consider the case where the squared error loss between the outputs of the network and the one-hot encoding of the true labels is minimized to a value of $0.05$ on average over the training data. 
 
\begin{figure}[t!]
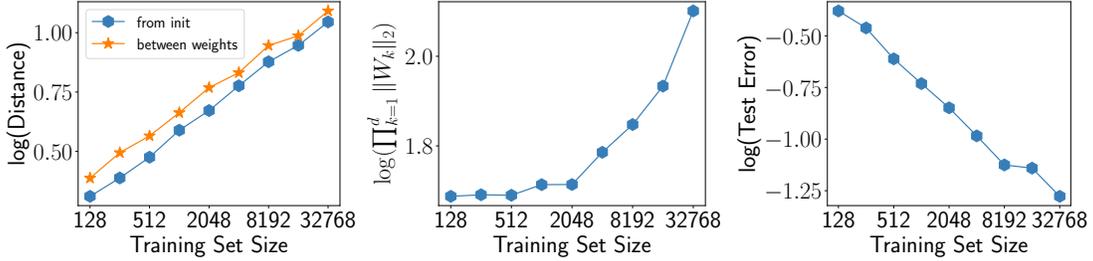

    \centering
            \begin{minipage}{.3\textwidth}
        \centering
        \adjincludegraphics[width=1\textwidth,trim={0 {0} 0 0},clip,,valign=t]{UniformConvergenceImages/squarederror_distance_vs_m_width=1024_margin} %
    \end{minipage}%
                \begin{minipage}{.3\textwidth}
        \centering
        \adjincludegraphics[width=1\textwidth,trim={0 {0} 0 0},clip,valign=t]{UniformConvergenceImages/squarederror_spectral_norm_vs_m_width=1024} %
    \end{minipage}% 
            \begin{minipage}{.3\textwidth}
        \adjincludegraphics[width=1\textwidth,trim={0 {0} 0 0},clip,,valign=t]{UniformConvergenceImages/squarederror_test_error_m_width=1024_bs=1} %
        \end{minipage}
       \caption{Training set size vs. norms, bounds and test error under squared error loss.
       % On the \textbf{left} we plot the distance from initialization and the distance between weights learned on two different random draws of the datasets, as a function of varying training set size $m$, when trained on the squared error loss. Both these quantities grow as $\Omega(m^{0.35})$. In the {\bf middle}, we show how the product of spectral norms grow as $\Omega(m^{0.315})$ for sufficiently large $m \geq 2048$. On the \textbf{right}, we observe that the test error (i.e., the averaged squared error loss on the test data) decreases with $m$ as $\mathcal{O}(m^{-0.38})$.
       }      \label{fig:squared-error}
\end{figure}

We observe in Figure~\ref{fig:squared-error} that even for this case, the distance from initialization and the spectral norms grow with the sample size at a rate of at least $m^{0.3}$. On the other hand, the test error decreases with sample size as $1/m^{0.38}$, indicating that even for the squared error loss, these terms hurt would hurt the generalization bound with respect to its dependence on $m$.

\section{Bounds grow with training set size}
We now turn to evaluating existing guarantees from \citet{neyshabur18pacbayes} and \citet{bartlett17spectral}, which involve the norms discussed in the previous section. As we note later, our observations apply to many other bounds too. % These guarantees can be written as follows, after incorporating the distance from initialization,
% as recommended by \citet{nagarajan17role}, 
% and after ignoring some log factors. 
%  Let $W_1, \hdots, W_d$ be the weights of the learned network (with $W_1$ being the weights adjacent to the inputs), $Z_1, \hdots, Z_d$  the random initialization, $\scrD$ the true data distribution and $S$ the training dataset. For all inputs $\vec{x}$, let $\| \vec{x}\|_2 \leq B$. Let $\|\cdot \|_2, \| \cdot\|_F, \| \cdot \|_{2,1}$ denote the spectral norm, the Frobenius norm and the matrix $(2,1)$-norm respectively; let $\mathbf{1}[\cdot]$ be the indicator function. Recall that 
% $\Gamma(f(\vec{x}),y) := f(\vec{x})[y] - \max_{y'\neq y} f(\vec{x})[y']$  denotes the margin of the network on a datapoint.
Recall from Section~\ref{sec:prelim-spec}, for any constant $\gamma > 0$, these generalization guarantees are written as follows, ignoring log factors:
\begin{align}
\scrL_{\scrD}(f_{\calW}) \leq \scrL^{(\gamma)}(f_{\calW})  + \mathcal{O}\left( \frac{Bd\sqrt{H}}{\gamma\sqrt{m}}\prod_{d=1}^{D} \| W_d\|_2 \times  \texttt{dist} \right), \label{eq:gen-error-bound}
\end{align}
 where $\texttt{dist}$ equals
$\sqrt{\sum_{d=1}^{D}\frac{\|\vecW_d - \vecZ_d \|_F^2}{\|\vecW_d\|^2_2}}$ in \citet{neyshabur18pacbayes}  and $\frac{1}{D\sqrt{H}} \left( {\sum_{d=1}^{D}\left(\frac{\|\vecW_d - \vecZ_d \|_{2,1}}{\|\vecW_d\|_2}\right)^{2/3}}\right)^{3/2} $ in \citet{bartlett17spectral}. Note that here we have modified the original bounds to incorporate the distance from initialization as discussed in Chapter~\ref{chap:dist-from-init}.

 In our experiments, since we train the networks to fit at least $99\%$ of the datapoints with a margin of $10$, in the above bounds, we set $\gamma=10$ so that the first train error term in the right hand side of Equation~\ref{eq:gen-error-bound} becomes a small value of at most $0.01$. We then plot in Figure~\ref{fig:terms} (fourth plot), the second term above, namely the generalization error bounds, and 
 %and also variations of it with the distance from initialization swapped with distance from a weight learned on another random draw of the dataset.  We 
 observe that all these bounds {\em grow with the sample size $m$} as $\Omega(m^{0.68})$, thanks to the fact that the terms in the numerator of these bounds grow with $m$. Note that, although we do not plot the bounds from \citep{nagarajan18deterministic,golowich17size}, these have nearly identical norms in their numerator, and so one would not expect these bounds to show  radically better behavior with respect to $m$. We report experiments conducted for other varied settings, and the neural network bound from \citep{neyshabur18unitwise} to in the upcoming sections.

\begin{figure}[t!]
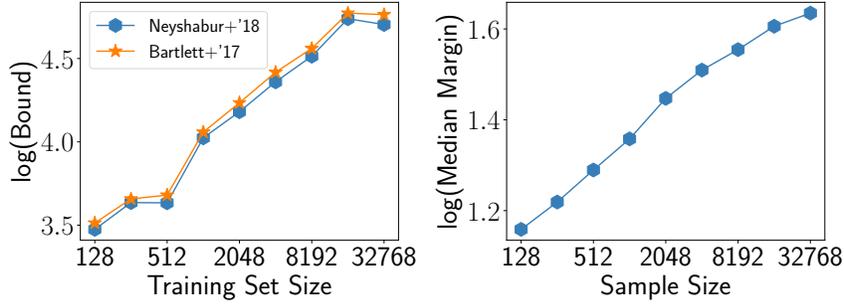

    \centering
       \begin{minipage}{.35\textwidth}
        \centering
        \adjincludegraphics[width=1\textwidth,trim={0 {0} 0 0},clip,valign=t]{UniformConvergenceImages/bound_m_width=1024_bs=1_median} %
    \end{minipage}%
        \begin{minipage}{.35\textwidth}
        \centering
                \adjincludegraphics[width=1\textwidth,trim={0 {0} 0 0},clip,,valign=t]{UniformConvergenceImages/median_margin_vs_m_width=1024} 
    \end{minipage}%
       \caption{A ``tighter'' bound computed with a median margin, and the median margin vs. the training set size.
        %In the \textbf{left} plot, we plot the bounds after setting $\gamma$ to be the median margin on the training data -- these bounds grow as $\Omega(m^{0.48})$.  In the \textbf{right} plot the median value of the margin $\Gamma(f(\vec{x}),y)$ on the training dataset and observe that it grows as $\mathcal{O}(m^{0.2})$. 
       } \label{fig:median-margin}
\end{figure}

\paragraph{Even a relaxed notion of margin does not address the $m$-dependency.}  Since we are free to plug in $\gamma$ in Equation~\ref{eq:gen-error-bound}, one may hope that there may exist a better choice of $\gamma$ for which we can observe a smaller increase on $m$ (since the plotted terms inversely depend on $\gamma$).  
%However, in Appendix Figure~\ref{fig:median-margin} we establish that even for larger values of $\gamma$, this $m$-dependence remains. Also 
 We consider this possibility by computing the  median margin of the network over the training set (instead of the $1\%$-percentile'th margin) and substituting this in the second term in the right hand side of the guarantee in Equation~\ref{eq:gen-error-bound}. By doing this, the first margin-based train error term in the right hand side of Equation~\ref{eq:gen-error-bound} would simplify to $0.5$ (as half the training data are misclassified by this large margin). Thereby we already forgo an explanation of half of the generalization behavior. At least we could hope that the second term no longer grows with $m$. Unfortunately, we observe in Figure~\ref{fig:median-margin} (left) that the bounds still grow with $m$ $\Omega(m^{0.48})$.  This is because, as shown in Figure~\ref{fig:median-margin} (right), the median margin value does not grow as fast with $m$ as the numerators of these bounds grow (it only grows as fast as $\mathcal{O}(m^{0.2})$).

\section{Relationship to flat minima}

\begin{figure}
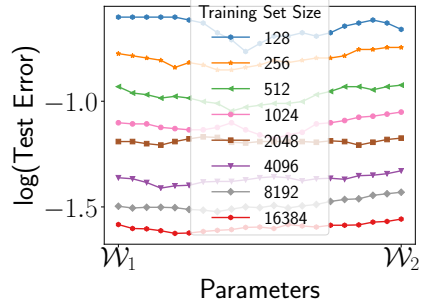

        \centering
        \adjincludegraphics[width=0.35\textwidth,trim={0 {0} 0 0},clip,,valign=t]{UniformConvergenceImages/interpolated_error_vs_m_width=1024_margin} %
        \caption{Networks learned on two different datasets are {\em linearly connected} by networks of similar error.
        %We plot the test errors of the networks that lie on the straight line  between two weights learned on two independent random draws of training data starting from the same initialization. We observe that all these intermediate networks have the same test error as the original networks themselves.
        }
                \label{fig:flat-minima}
\end{figure}

We also relate our observations regarding distance between two independently learned weights to the popular idea of ``flat minima''. Interestingly, Figure~\ref{fig:flat-minima} demonstrates that walking linearly from the weights learned on one dataset draw to that on another draw (from the same initialization) preserves the test error. Note that although a similar observation was made in \citet{felix18barriers,garipov18loss},  they show the existence of \textit{non-linear} paths of good solutions between parameters learned from \textit{different} initializations. Our observation on the other hand implies that for a fixed initialization, SGD explores the \textit{same} basin in the test loss minimum across different training sets. As discussed earlier, this explored basin/space has larger $\ell_2$-width for larger $m$ giving rise to a ``paradox'': on one hand, wider minima are believed to result in, or at least correlate with better generalization \citep{hochreiter97flat,hinton93mdl,keskar17largebatch}, but on the other, a larger $\ell_2$-width of the explored space results in larger uniform convergence bounds, making it harder to explain generalization. 
 % making it harder to explain generalization. On the other hand, wider minima are believed to result in (or at least, correlate with) better generalization \citep{hochreiter97flat,hinton93mdl,keskar17largebatch}, while they seem to hurt uniform convergence bounds (as we similarly noted about the paradoxical role of noise in Section~\ref{sec:related}).  

We note a similar kind of paradox concerning noise in training. Specifically,
it is intriguing that on one hand, generalization is aided by larger learning rates and smaller batch sizes \cite{jastrzebski18width,hoffer17longer,keskar17largebatch} due to increased noise in SGD. On the other, theoretical analyses benefit from the opposite;  \citet{zhu18beyond} even explicitly regularize SGD for their three-layer-network result  to help ``forget false information'' gathered by SGD. In other words, it seems that {noise aids generalization, yet hinders attempts at explaining generalization}. The intuition from our examples (such as the linear example) is that such ``false information'' could provably impair uniform convergence without affecting generalization.

\section{Effect of other hyperparameters}

We now report experiments were we examine the effect of other hyperparameters on the above observations..

\subsection{Depth}
\begin{figure}[t!]
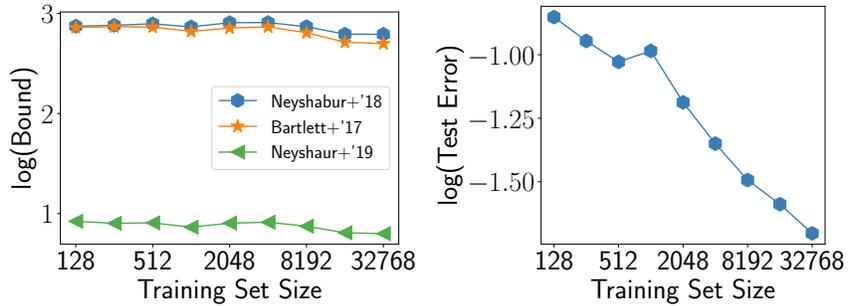

    \centering
        \begin{minipage}{.35\textwidth}
        \centering
        \adjincludegraphics[width=1\textwidth,trim={0 {0} 0 0},clip,,valign=t]{UniformConvergenceImages/onelayer_bound_m_width=50000_bs=1_margin} %
    \end{minipage}%
            \begin{minipage}{.35\textwidth}
        \centering
        \adjincludegraphics[width=1\textwidth,trim={0 {0} 0 0},clip,valign=t]{UniformConvergenceImages/onelayer_test_error_m_width=50000_bs=1} %
    \end{minipage}
        \caption{Bounds and test error for a 2-layer neural network.
       %On the \textbf{left}, we plot how the bounds vary with sample size for a single hidden layer network with $50k$ hidden units. 
       % We observe that these bounds are almost constant, and at best decrease at a meagre rate of $\Omega(m^{-0.066})$. On the \textbf{right}, we plot the test errors for this network and observe that it decreases with $m$ at the rate of at least $\mathcal{O}(m^{0.35})$.
       }      \label{fig:one-layer}
\end{figure}

It turns out that as the network gets shallower, the bounds do show better dependence with $m$. As an extreme case, we consider a network with only one hidden layer, and with $H=50000$. Here we also present a third bound, namely that of \citet{neyshabur18unitwise}, besides the two bounds discussed in the earlier sections. Specifically, if $\vecZ_1, \vecZ_2$ are the random initializations of the weight matrices in the network, the generalization error bound (the last term in Equation~\ref{eq:gen-error-bound}) here is of the following form, ignoring log factors:

\[
\frac{\|\vecW_2 \|_F (\|\vecW_1- \vecZ_1 \|_F + \| \vecZ_1\|_2)}{\gamma \sqrt{m}} + \frac{\sqrt{H}}{\sqrt{m}.
}\]
The first term here is meant to be width-independent, while the second term clearly depends on the width and does decrease with $m$ at the rate of $m^{-0.5}$. Hence, in our plots in Figure~\ref{fig:one-layer}, we only focus on the first term. We see that these bounds are almost constant and decrease at a minute rate of $\Omega(m^{-0.066})$ while the test errors decrease much faster, at the rate of $\mathcal{O}(m^{-0.35})$.

\subsection{Effect of width} 

\begin{figure}[t!]
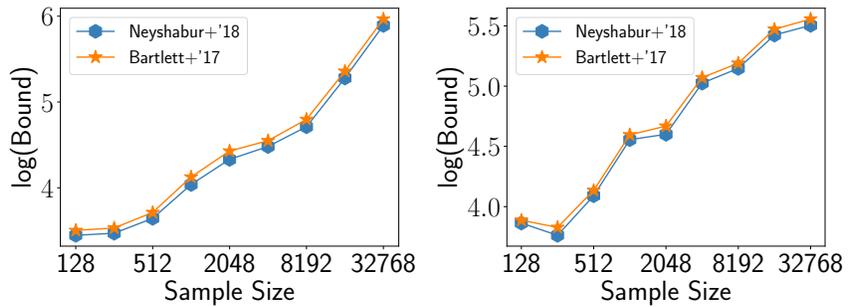

    \centering
        \begin{minipage}{.35\textwidth}
        \centering
        \adjincludegraphics[width=1\textwidth,trim={0 {0} 0 0},clip,,valign=t]{UniformConvergenceImages/bound_m_width=128_bs=1_margin} %
    \end{minipage}%
            \begin{minipage}{.35\textwidth}
        \centering
        \adjincludegraphics[width=1\textwidth,trim={0 {0} 0 0},clip,valign=t]{UniformConvergenceImages/bound_m_width=2000_bs=1_margin} %
    \end{minipage}
       \caption{Bounds vs. training set size for very thin ($H=128$, left) and very wide ($H=2000$, right) networks.
       %On the \textbf{left}, we plot the bounds for varying $m$ for $H=128$. All these bounds grow with $m$ as $\Omega(m^{0.94})$. On the \textbf{right}, we show a similar plot for $H=2000$ and observe that the bounds grow as $\Omega(m^{0.79})$.
       }      \label{fig:width}
\end{figure}

In Figure~\ref{fig:width}, we demonstrate that our observation that
the bounds increase with $m$ extends to widths $H=128$ and $H=2000$ too.

\subsection{Batch size}

\begin{figure}[t!]
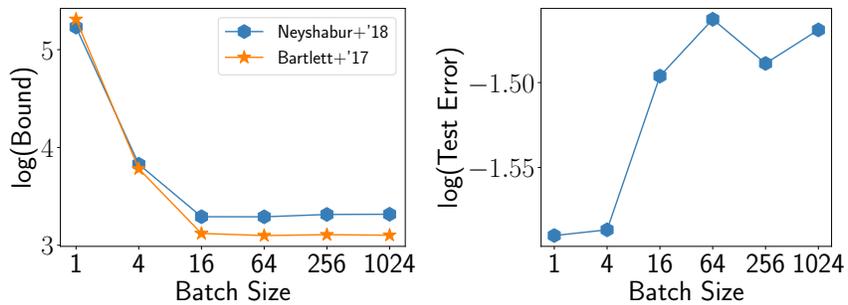

    \centering
        \begin{minipage}{.35\textwidth}
        \centering
        \adjincludegraphics[width=1\textwidth,trim={0 {0} 0 0},clip,,valign=t]{UniformConvergenceImages/batchsize_bound_m_width=1024_bs=1024_margin} %
    \end{minipage}%
            \begin{minipage}{.35\textwidth}
        \centering
        \adjincludegraphics[width=1\textwidth,trim={0 {0} 0 0},clip,valign=t]{UniformConvergenceImages/batchsize_test_error_m_width=1024_bs=1024} %
    \end{minipage}
       \caption{Bounds and test error vs. training set size.
       %On the \textbf{left}, we plot the bounds for varying batch sizes for $m=16384$ and observe that these bounds {\em decrease} by around $2$ orders of magnitude. On the \textbf{right}, we plot the test errors for varying batch sizes and observe that test error increases with batch size albeit slightly.
       }      \label{fig:bs-2}
\end{figure}

\paragraph{Bounds vs. batch size for fixed $m$.}
In Figure~\ref{fig:bs-2}, we show how the bounds vary with the batch size for a fixed sample size of $16384$. It turns out that even though the test error decreases with decreasing batch size (for our fixed stopping criterion), all these bounds {\em increase}  (by a couple of orders of magnitude)  with decreasing batch size. Again, this is because the terms like distance from initialization {\em increase} for smaller batch sizes (perhaps because of greater levels of noise in the updates). Overall, existing bounds do not reflect the same behavior as the actual generalization error in terms of their dependence on the batch size.

\begin{figure}[t!]
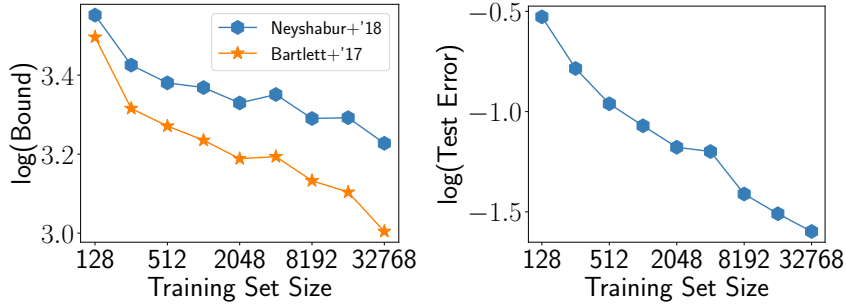

    \centering
        \begin{minipage}{.35\textwidth}
        \centering
        \adjincludegraphics[width=1\textwidth,trim={0 {0} 0 0},clip,,valign=t]{UniformConvergenceImages/bound_m_width=1024_bs=32_margin} %
    \end{minipage}%
            \begin{minipage}{.35\textwidth}
        \centering
        \adjincludegraphics[width=1\textwidth,trim={0 {0} 0 0},clip,valign=t]{UniformConvergenceImages/test_error_m_width=1024_bs=32} %
    \end{minipage}
       \caption{
       Bounds and test error vs. training set size for a relatively larger batch size of $32$.
       %On the \textbf{left}, we plot the bounds for varying $m$ for a batch size of $32$ and observe that these bounds do decrease with $m$ as $\mathcal{O}(1/m^{0.23})$. On the \textbf{right}, we plot the test errors for various $m$ for batch size $32$ and observe that test error varies as $\Omega(1/m^{0.44})$.
       }      \label{fig:bs}
\end{figure}

\paragraph{Bounds vs.  $m$ for batch size of $32$.}
So far, we have only dealt with a small batch size of $1$. In Figure~\ref{fig:bs}, we show bounds vs. sample size plots for a batch size of $32$. We observe that in this case, the bounds do decrease with sample size, although only at a rate of $\mathcal{O}(m^{-0.23})$ which is not as fast as the observed decrease in test error which is $\Omega(m^{-0.44})$. Our intuition as to why the bounds behave better (in terms of $m$-dependence) in the larger batch size regime is that here the amount of noise in the parameter updates is much less compared to smaller batch sizes (and as we discussed earlier, norm bounds find it challenging to explain away such noise).

\section{Pseudo-overfitting}
\label{app:pseudo-overfit}

While the bounds might show better $m$-dependence for many settings, we believe that the egregious break down of these bounds in many other settings must imply fundamental issues with the bounds themselves.
%In effect, our observations reveal that existing generalization bounds fail to meet some fundamental criterion or the other needed to explain generalization.
 While this may be addressed to some extent with a better understanding of implicit regularization in deep learning, we regard our observations as a call for taking a step back and clearly understanding any inherent limitations to our theoretical tools.

In this section, we hypothesize one possible limitation that might explain the failure of norm-based complexity measures (but we will eventually rule this out the reason). Specifically, we hypothesize that, for {\em some} (not all) existing bounds, the above problems could arise from what we term as {\em pseudo-overfitting}. Roughly speaking, a classifier pseudo-overfits when its decision boundary is simple but its {\em real-valued} output has large ``bumps'' around some or all of its training datapoint. \\ %As discussed in Appendix~\ref{app:pseudo-overfit}, deep networks pseudo-overfit only to a limited extent, and hence psuedo-overfitting does not provide a complete explanation for the issues faced by these bounds.% (although we note that it might still be worth studying pseudo-overfitting in isolation for future work).}

% Before, we proceed to our main contribution which identifies a serious limitation of  uniform convergence-based approaches to explaining generalization, we consider (and rule out) the possibility that existing approaches may be providing vacuous bounds due to a more superficial issue that does not involve uniform convergence. 

\paragraph{Pseudo-overfitting implies failure of bounds.} Our argument specifically applies to margin-based Rademacher complexity approaches (such as \citet{bartlett17spectral,neyshabur18unitwise}). These result in a bound like in Equation~\ref{eq:gen-error-bound} that we recall here:

\begin{align}
 \prsub{(x,y) \sim \scrD}{\Gamma(f(\vec{x}),y) \leq 0}  \leq \frac{1}{m}\sum_{(x,y)\in S}\mathbf{1}[\Gamma(f(\vec{x}),y) \leq \gamma]  + \texttt{generalization error bound}. \label{eq:margin-unif}
\end{align}

Recall from our discussion of Theorem~\ref{thm:rademacher-binary-class} that these methods upper bound generalization gap of the network in terms of the gap between the margins of the network.(see \cite{mohri12foundations} for more details about margin theory of Rademacher complexity). In particular, the ``generalization error bound'' above is also a bound on the difference between the test and training margins:
% particular upper bound on the generalization gap in the $\mathcal{L}^{(\gamma)}$ loss is also an upper bound on the generalization gap on the margins. That is, with high probability $1-\delta$ over the draws of $S$, the above bound is larger than the following term that corresponds to the difference in test/train margins:

% a uniform convergence bound on the margins of the network . Mathematically, the generalization error bound is essentially a quantity that satisfies the following:

% The resulting generalization error bound in Equation~\ref{eq:gen-error-bound} would take the following form, as per our notation from Definition~\ref{def:unif-alg}:
% \begin{align*}
% \sup_{S \in \mathcal{S}_{\delta}} \sup_{h \in \mathcal{H}_{\delta}}  \frac{1}{\gamma}  \left|   \mathbb{E}_{\scrD}[ \Gamma(h(\vec{x}),y) ] -  \frac{1}{m} \sum_{(x,y) \in S} \Gamma(h(\vec{x}),y)\right|. \numberthis\label{eq:margin-unif}
% \end{align*}

\begin{align}
 \frac{1}{\gamma} \left(    \mathbb{E}_{(x,y)\sim \scrD}[ \Gamma(h_{S}(\vec{x}),y) ] -  \frac{1}{m} \sum_{(x,y) \in S} \Gamma(h_{S}(\vec{x}),y)\right) \leq \texttt{generalization error bound}.
\numberthis \label{eq:margin-bound}
\end{align}
We argue that it is hypothetically possible for the actual generalization gap of the algorithm to decrease with $m$ (as roughly $m^{-0.5}$), but for the above quantity ``margin generalization gap'' to be independent of $m$. As a result, the upper bound bound in Equation~\ref{eq:margin-unif} will be non-decreasing in $m$, and even vacuous. Below we describe such a scenario.\\

Consider a network that first learns a simple hypothesis to fit the data, say, by learning a simple linear input-output mapping on linearly separable data. But subsequently, the classifier proceeds to {\em pseudo-overfit} to the samples by skewing up (down) the real-valued output of the network by some large constant $\Delta$ in a tiny neighborhood around the positive (negative) training inputs. Note that this would be possible if and only if the network is overparameterized. Now, even though the classifier's real-valued output is skewed around the training data, the decision boundary is still linear as the sign of the classifier's output has not changed on any input. Thus, the boundary is still simple and linear and the generalization error small. 

However, the training margins are at least a constant $\Delta$ larger than the test margins (which are not affected by the bumps created in tiny regions around the training data). Then, the LHS term in Equation~\ref{eq:margin-bound} would be larger than $\Delta/\gamma$. Therefore,
\begin{equation}
\frac{\Delta}{\gamma} \leq \texttt{generalization error bound}.
\end{equation}

 Now in the generalization guarantee of Equation~\ref{eq:margin-unif}, recall that we must pick a value of $\gamma$ such that the first term is low i.e., most of the training datapoints must be classified by at least $\gamma$ margin. In this case, we can at best let $\gamma \approx \Delta$ as any larger value of $\gamma$ would make the margin-based training error non-negligible; as a result of this choice of $\gamma$, the generalization error bound in Equation~\ref{eq:margin-unif} would be an $m$-independent constant close to $1$.\\ %The same would also hold for its upper bound in Equation~\ref{eq:margin-unif}, which is the generalization bound provided by the margin-based techniques.
% When the datapoints are classified only by a constant margin $\gamma$, then the bound we get is independent of $m$. 

\begin{figure}
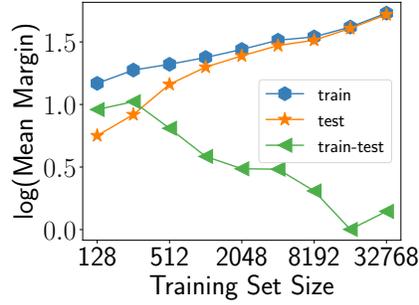

        \centering
        \adjincludegraphics[width=0.35\textwidth,trim={0 {0} 0 0},clip,,valign=t]{UniformConvergenceImages/margin_vs_samples__width=1024_margin} %
        \caption{Average test and train margins of a deep network.}
                \label{fig:margin-convergence}
\end{figure}

\paragraph{Psuedo-overfitting in practice.} Clearly, this is a potential fundamental limitation in existing approaches, and if deep networks were indeed {pseudo-overfitting} this way, we would have identified the reason why at least some existing bounds are vacuous. However, (un)fortunately, we rule this out by observing that the difference in the train and test margins in Equation~\ref{eq:margin-bound} does decrease with training dataset size $m$ (see Figure~\ref{fig:margin-convergence}) as $\mathcal{O}(m^{-0.33})$. Additionally, this difference is numerically much less than $\gamma^\star=10$ (which is the least margin by which $99\%$ of the training data is classified) as long as $m$ is large, implying that Equation~\ref{eq:margin-bound} is non-vacuous.

 It is worth noting that the generalization error decreases at a faster rate of $\mathcal{O}(m^{-0.43})$ implying that the upper bound in Equation~\ref{eq:margin-bound} which decreases only as $m^{-0.33}$,
 is loose.  This already indicates a partial weakness in this specific approach to deriving generalization guarantees. Nevertheless, even this upper bound decreases at a significant rate with $m$ which the subsequent uniform convergence-based upper bound in Equation~\ref{eq:margin-unif} is unable to capture, thus hinting at more fundamental weaknesses specific to uniform convergence.

\section{Conclusion}

A growing variety of uniform convergence based bounds
 \citep{neyshabur15norm,bartlett17spectral,golowich17size,arora18compression,
neyshabur18pacbayes,dziugaite17nonvacuous,zhou2018nonvacuous,
li18learning,zhu18beyond,nagarajan18deterministic,neyshabur18unitwise}
have sought to explain generalization in deep learning.
While these may provide partial intuition about the puzzle, we show that these bounds fail to fully explain the generalization behavior of deep networks. In particular, we show that these bounds can be bad at capturing a basic fact about generalization, namely that it improves with training set size. Why does this happen? We hypothesized that one possible cause for this might be the fact that the margins of a deep network may be skewed only around training data. However, we did not find strong empirical evidence suggesting this. Motivated by this, in the next chapter, we will delve deeper into these generalization bounds, and uncover a more fundamental source of weakness in these bounds, namely uniform convergence.
%point out that they can be fundamentally limited in the overparameterized regime.

\graphicspath{{unif-conv/}}
\chapter{Provable Failure of Uniform Convergence}

\label{chap:unif-conv}

% \begin{abstract}
% Aimed at explaining the surprisingly good generalization behavior of overparameterized deep networks, recent works have developed a variety of generalization bounds for deep learning,  all  based on the fundamental learning-theoretic technique of uniform convergence. While
% it is well-known that many of these existing bounds are numerically large, through numerous experiments, we bring to light a more concerning aspect of these bounds: 
% in practice,  these bounds can {\em increase} with the training dataset size. Guided by our observations,
% we then present examples of overparameterized linear classifiers and neural networks trained by  gradient descent (GD) where uniform convergence provably cannot ``explain generalization'' -- even if we take into account the implicit bias of GD {\em to the fullest extent possible}. More precisely, even if we consider only the set of classifiers output by GD, which have test errors less than some small $\epsilon$ in our settings, we show that applying (two-sided) uniform convergence on this set of classifiers will yield only a vacuous generalization guarantee larger than $1-\epsilon$. Through these findings,
% we cast doubt on the power of uniform convergence-based generalization bounds to provide a complete picture of why overparameterized deep networks generalize well. 
% \end{abstract}

% add comment on how compression and PAC-bayesian methods can help
\section{Introduction}

Motivated by the seemingly insurmountable hurdles towards developing generalization bounds satisfying all the desiderata, in this chapter, we take a step back to conduct a more fundamental investigation of these bounds. We examine how the technique underlying all these bounds --- uniform convergence --- may itself be inherently limited in the overparameterized regime. In particular, we present examples of overparameterized linear classifiers and neural networks trained by GD (or SGD) where uniform convergence can {\em provably} fail to explain generalization. 
%Notably, we will see that our example preserves our main empirical findings in deep learning, which also end up playing a crucial role in our theoretical analysis.  
Intuitively,  our examples highlight that overparameterized models trained by gradient descent can learn decision boundaries that are largely ``simple'' -- and hence generalize well -- but have ``microscopic complexities'' which cannot be explained away by uniform convergence. Thus our results call into question the active ongoing pursuit of using uniform convergence to fully explain generalization in deep learning.\\

%While existing uniform convergence bounds provide partial explanations about generalization in GD-based trainining of overparameterized deep networks, through our arguments, we question their potential to explain the phenomenon fully. 

More concretely,  we consider three example setups of overparameterized models trained by (stochastic) gradient descent -- a linear classifier, a sufficiently wide neural network with ReLUs and an infinite width neural network with exponential activations (with the hidden layer weights frozen) -- that learn some underlying data  distribution with small generalization error (say, at most $\epsilon$). These settings also simulate our observation that norms such as distance from initialization grow with dataset size $m$.
More importantly, we prove that, in these settings, {\em any} two-sided {uniform convergence bound} would yield a (nearly) vacuous generalization bound.  

Notably, this vacuity holds even if we ``aggressively'' take implicit regularization into account while applying uniform convergence --  described more concretely as follows. Recall that roughly speaking a uniform convergence bound essentially evaluates the complexity of a hypothesis class. 
%That is, recall that for a hypothesis class, two-sided uniform convergence demands that for most draws of a dataset $S$ of $m$ training datapoints from an underlying data distribution $\scrD$, and {\em for every hypothesis in the class} (and not just the hypothesis learned by the given algorithm $\mathcal{A}$ on $S$),  the expected error on $\scrD$ and the empirical error on $S$ must be close to each other . 
One {\em can} tighten uniform convergence bounds by pruning the hypothesis class to remove extraneous hypotheses never picked by the learning algorithm  for the data distribution of interest. But remarkably, in our setups, even if we apply uniform convergence on the set of {only those hypotheses picked by the learner whose test errors are all negligible (at most $\epsilon$)}, one can get no better than a nearly vacuous bound on the generalization error (that is at least $1-\epsilon$). In this sense, we say that uniform convergence provably cannot explain generalization in our settings. Finally, we note that while nearly all existing uniform convergence-based techniques are two-sided, we show that even PAC-Bayesian bounds, which are typically presented only as one-sided convergence, also boil down to nearly vacuous guarantees in our settings. 

% attempts at understanding generalization in deep learning using uniform convergence. % based tools for explaining generalization in deep learning.

\subsection{Related Work}
\label{sec:unif-conv-related}

Traditional wisdom is that uniform convergence bounds are a bad choice for complex classifiers like k-nearest neighbors because these hypotheses classes have infinite VC-dimension (which motivated the need for stability based generalization bounds in these cases \citep{rogerss78finite,bousquet02stability}). However, this sort of an argument against uniform convergence may still leave one with the faint hope that, by aggressively pruning the hypothesis class (depending on the algorithm and the data distribution), one can achieve meaningful uniform convergence. In contrast, we seek to {rigorously} and {thoroughly} rule out uniform convergence in the settings we study. We do this by first defining the tightest form of uniform convergence in Definition~\ref{def:unif-alg} -- one that lower bounds any uniform convergence bound -- and then showing that even this bound is vacuous in our settings. Additionally, we note that
we show this kind of failure of uniform convergence for linear classifiers, which is a much simpler model compared to k-nearest neighbors.
 For deep networks,
\citet{zhang17generalization} showed that applying uniform convergence on the whole hypothesis class fails, and that it should instead be applied in an algorithm-dependent way.  \textit{Ours is a much different claim} -- that uniform convergence is inherently problematic in that even the algorithm-dependent application would fail -- casting doubt on the rich line of post-\citet{zhang17generalization} algorithm-dependent approaches. At the same time, we must add the disclaimer that our results do not preclude the fact that uniform convergence may still work if GD is run with explicit regularization (such as weight decay). Such a regularized setting however, is not the main focus of the generalization puzzle \citep{zhang17generalization,neyshabur15inductive}.\\

%\paragraph{Learnability and Uniform Convergence.}
Prior works \citep{vapnik71uniform,shwartz10learnability} have also focused on understanding uniform convergence for {\em learnability of learning problems}. Roughly speaking, learnability is a strict notion that does not have to hold even though an algorithm may generalize well for simple distributions in a learning problem. While we defer the details of these works in Section~\ref{sec:learnability}, we emphasize here that these results are orthogonal to (i.e., neither imply nor contradict) our results.

%
%In fact, \citet{vapnik71uniform} showed in their seminal paper that learnability is \textit{equivalent} to {uniform convergence} in binary classification problems. 
% 

% as their result does not imply that uniform convergence may not be able to {\em explain generalization} of a particular algorithm for simple distributions in binary classification problems. %It is also worth noting that their result requires an extreme level of overparameterization ($\Omega(2^m)$ parameters), while we require only mild overparameterization ($\Omega(m)$).

\section[Tightest algorithm-dependent u.c.]{Tightest algorithm-dependent, distribution-dependent uniform convergence}
\label{sec:setup}
%\label{sec:preliminaries}
%In this section, we formally define the terms we will study theoretically: the generalization error of an algorithm, and the different flavors of uniform convergence based bounds.

For the rest of the discussion, we will consider a generic class of hypotheses $\mathcal{H}$ that is not necessarily a class of neural network functions. 
Let $\mathcal{A}$ be the learning algorithm and let
$h_{S}$ be the hypothesis output by the algorithm on a dataset $S$ (assume that any training-data-independent randomness, such as the initialization/data-shuffling is fixed). \\

Let us revisit some standard quantities from our discussion in Chapter~\ref{chap:prelim}.
For a given $\delta \in (0,1)$, the generalization error of the algorithm is essentially a bound on the difference between the error of the hypothesis $h_{S}$ learned on a training set $S$ and the expected error over $\scrD$, that holds with high probability of at least $1-\delta$ over the draws of $S$. More formally: \\

\begin{definition} 
\label{def:gen-error}
The\textbf{ generalization error}  of $\mathcal{A}$
%of $\mathcal{A}$ 
with respect to loss $\scrL$ is the smallest value $\epsilon_{\text{gen}}(m, \delta)$ such that:
\begin{equation}\mathbb{P}_{S \sim \scrD^m}\left[
 \scrL_{\scrD}(h_{S}) -
 \hat{\scrL}_{S}(h_{S}) 
\leq \epsilon_{\text{gen}}(m, \delta) \right] \geq 1- \delta.\end{equation}
%\numberthis \label{eq:generalization-error}
%\end{align*}
%\begin{align*}
%Pr_{S \sim \scrD^m}\left[ \scrL_{\scrD}(h_{S}) -
% \hat{\scrL}_{S}(h_{S}) 
%> \epsilon_{\text{gen}}(m, \delta)   \right] <  \delta
%\numberthis \label{eq:generalization-error}
%\end{align*}
\end{definition}

To theoretically bound the generalization error of the algorithm, the most common approach is to provide a two-sided uniform convergence bound on the hypothesis class used by the algorithm, where, for a given draw of $S$, we look at convergence for all the hypotheses in $\mathcal{H}$ instead of just $h_{S}$:\\ % \footnote{ \textbf{Note}: The VC-dimension is a distribution-{\em independent} (or distribution-free) variety of uniform convergence; it is an even stricter notion where the bound in Equation~\ref{eq:uniform-convergence} has to hold for {\em all} possible  $\scrD$. However, recent deep network bounds such as those based on Rademacher complexity are distribution-dependent as above. } \\

\begin{definition} 
\label{def:unif}
The \textbf{uniform convergence bound} 
with respect to loss $\scrL$ is the smallest value $\epsilon_{\text{unif}}(m, \delta)$ such that:
\begin{equation}\mathbb{P}_{S \sim \scrD^m}\left[
\sup_{h \in \mathcal{H}} \left| \scrL_{\scrD}(h) -
 \hat{\scrL}_{S}(h) \right| \leq \epsilon_{\text{unif}}(m, \delta)\right] \geq 1- \delta\end{equation}.
 %  \numberthis\label{eq:uniform-convergence}
%\end{align*}
\end{definition}

 The bound given by $\epsilon_{\text{unif}}$
can be tightened by ignoring many extraneous hypotheses in $\mathcal{H}$ never picked by $\mathcal{A}$ for a given simple distribution $\scrD$. This is typically done by focusing on a norm-bounded class of hypotheses that the algorithm $\mathcal{A}$ implicitly restricts itself to. Let
us take this to the extreme by applying uniform convergence on ``the smallest possible class'' of hypotheses, namely, {\em only} those hypotheses that are picked by $\mathcal{A}$ under $\scrD$, excluding everything else. Observe that pruning the hypothesis class any further would not imply a bound on the generalization error, and hence applying uniform convergence on this aggressively pruned hypothesis class would yield {the tightest possible uniform convergence bound}.
% (and hence would be smaller than $\epsilon_{\text{unif}}(m, \delta)$). 
 Recall that we care about this formulation because our goal is to rigorously and thoroughly rule out the possibility that no kind of uniform convergence bound, however cleverly applied, can explain generalization in our settings of interest (which we will describe later).\\

To formally capture this bound, it is helpful to first rephrase the above definition of $\epsilon_{\text{unif}}$: we can say that $\epsilon_{\text{unif}}(m,\delta)$ is  the smallest value for which there exists a set of sample sets $\mathcal{S}_{\delta} \subseteq (\mathcal{X} \times \{-1,1 \})^m$ for which $\prsub{S\sim \scrD^m}{S \in \mathcal{S}_{\delta}} \geq 1-\delta$ and furthermore, $\sup_{S \in \mathcal{S}_{\delta}} \sup_{h\in \mathcal{H}} | \scrL_{\scrD}(h) -
 \hat{\scrL}_{S}(h) | \leq \epsilon_{\text{unif}}(m,\delta)$. Observe that this definition is {\em equivalent} to Definition~\ref{def:unif}. Extending this rephrased definition, we can define the tightest uniform convergence bound by replacing $\mathcal{H}$ here with only those hypotheses that are explored by the algorithm $\mathcal{A}$ under the datasets belonging to $\mathcal{S}_{\delta}$:\\

\begin{definition} 
\label{def:unif-alg}
The \textbf{tightest algorithm-dependent, distribution-dependent uniform convergence bound}
%of (the hypothesis class $\mathcal{H}$, algorithm $\mathcal{A}$)-pair 
with respect to loss $\scrL$ is the smallest value $\epsilon_{\text{unif-alg}}(m, \delta)$ for which there exists a set of sample sets $\mathcal{S}_{\delta}$ such that \begin{equation}\prsub{S\sim \scrD^m}{S \in \mathcal{S}_{\delta}} \geq 1-\delta\end{equation} and 
 if we define the space of hypotheses explored  by $\mathcal{A}$ on $\mathcal{S}_{\delta}$ as \begin{equation}\mathcal{H}_{\delta} \coloneqq \bigcup_{S \in \mathcal{S}_{\delta}}  \{ h_{S} \} \subseteq \mathcal{H}, \end{equation} the following holds:
\begin{equation}
\sup_{S \in \mathcal{S}_{\delta}}
\sup_{h \in \mathcal{H}_{\delta}}  \left| 
\scrL_{\scrD}(h) -
 \hat{\scrL}_{S}(h)  
\right| \leq \epsilon_{\text{unif-alg}}(m,\delta). \end{equation}\\
\end{definition} 

%\begin{remark} Since $\forall S \in \mathcal{S}_{\delta}$, we have $h_{S} \in \mathcal{H}_{\delta}$, we also have $\epsilon_{\text{gen}}(m,\delta)  \leq \epsilon_{\text{unif-alg}}(m,\delta)  $. On  the other hand, since $\mathcal{H}_{\delta} \subseteq \mathcal{H}$, $\epsilon_{\text{unif-alg}}(m,\delta)  \leq \epsilon_{\text{unif}}(m,\delta)$. %More importantly,  $\epsilon_{\text{unif-alg}}$ is essentially a lower bound on all two-sided uniform convergence-based bounds because one cannot apply uniform convergence on a hypothesis smaller than $\mathcal{H}_{\delta}$ and claim to have explained generalization.
%\end{remark}

%\begin{remark} Sometimes uniform convergence bounds are written with an explicit dependence on the weights learned. In contrast, our definition $\epsilon_{\text{unif-alg}}(m,\delta)$ is evidently devoid of such dependence because our bound applies in supremum over all hypotheses picked by $\mathcal{A}$, which is what we ultimately care about in order to explain generalization e.g., if $\mathcal{A}$ always ensures $\|\vec{w} \|_2 \leq 1$, this bound would correspond to $1/\sqrt{m}$.  %Thus, our definition is not restrictive in any sense.  
%\end{remark}

%\textbf{Remark 3.} If the algorithm makes use of random bits independent of the training dataset (such as say the random initialization), one can fix these bits and then define $S$ as the hypothesis learned under these fixed bits. 

% I'm deleting this remark because it's already present in a different form later where I discuss this specifically for our setup.

%\section{Theoretical Model}

In the following sections, through examples of overparameterized models trained by GD (or SGD), we argue how even the above tightest algorithm-dependent uniform convergence can fail to explain generalization. i.e., in these settings, even though $\epsilon_{\text{gen}}$ is smaller than a negligible value $\epsilon$, we show that $\epsilon_{\text{unif-alg}}$ is large (specifically, at least $1-\epsilon$).   Before we delve into these examples, below we quickly outline the key mathematical idea by which uniform convergence is made to fail. \\

Consider a scenario where the algorithm generalizes well i.e., for every training set $\tilde{S}$, 
$h_{\tilde{S}}$ has zero error on $\tilde{S}$ and has small test error. While this means that $h_{\tilde{S}}$ has small error on \textit{random} draws of a test set, it may still be possible that for every such $h_{\tilde{S}}$, there exists a corresponding ``bad'' dataset $\tilde{S}'$ -- that is not random, but rather dependent on $\tilde{S}$ -- on which $h_{\tilde{S}}$ has a large empirical error (say $1$).  
Unfortunately, uniform convergence runs into trouble while dealing with such bad datasets. Specifically, as we can see from the above definition, uniform convergence demands that $|\scrL_{\scrD}(h_{\tilde{S}}) -
 \hat{\scrL}_{S}(h_{\tilde{S}})  |$ be small on all datasets in $\mathcal{S}_{\delta}$, which excludes a $\delta$ fraction of the datasets. While it may be tempting to think that we can somehow exclude the bad dataset as part of the  $\delta$ fraction, there is a significant catch here: we can not carve out a $\delta$ fraction specific to each hypothesis; we can ignore only a single chunk of $\delta$ mass common to all hypotheses in $\mathcal{H}_{\delta}$.  This restriction turns out to be a tremendous bottleneck: despite ignoring this $\delta$ fraction, for most $h_{\tilde{S}} \in \mathcal{H}_{\delta}$, the corresponding bad set $\tilde{S}'$ would still be left in  $\mathcal{S}_{\delta}$. Then, for all such $h_{\tilde{S}}$, $\scrL_{\scrD}(h_{\tilde{S}})$ would be small but $\hat{\scrL}_{S}(h_{\tilde{S}}) $ large; we can then set the $S$ inside the $\sup_{S \in \mathcal{S}_{\delta}}$ to be $\tilde{S}'$ to conclude that 
 $\epsilon_{\text{unif-alg}}$ is indeed vacuous. This is the kind of failure we will demonstrate in a high-dimensional linear classifier (Section~\ref{sec:uc-linear}), followed by a ReLU neural network (Section~\ref{sec:uc-hypersphere}), and then an infinitely wide exponential-activation neural network -- all trained by GD or SGD.    

\begin{remark} Our results about failure of uniform convergence holds even for bounds that output a different value for each hypothesis. In this case, the tightest uniform convergence bound for a given hypothesis would be at least as large as $\sup_{S \in \mathcal{S}_{\delta}} |\scrL_{\scrD}(h_{\tilde{S}}) - \hat{\scrL}_{S}(h_{\tilde{S}})|$ which by a similar argument would be vacuous for most draws of the training set $\tilde{S}$. We discuss this in more detail in Section~\ref{sec:weight-dependence}.
 \end{remark}
 
% 
%that this fraction must be ignored in common to all the hypotheses in $\mathcal{H}_{\delta}$ (as against ignoring a fraction specific to each hypothesis). 
%
%, it is easy to note that deriving a small uniform convergence bound boils down to picking a sample set space $\mathcal{S}_{\delta}$ of mass $1-\delta$ such that $\sup_{S \in \mathcal{S}_{\delta}}
%\sup_{h \in \mathcal{H}_{\delta}}  | 
%\scrL_{\scrD}(h) -
% \hat{\scrL}_{S}(h)  
%|$ is small. On one hand, it is possible that for every dataset $S \in \mathcal{S}_{\delta}$, the corresponding hypothesis $h_{S} \in \mathcal{H}_{\delta}$ has low empirical error on $S$ and low expected error on $\scrD$, resulting in low generalization error. On the other hand, two-sided uniform convergence demands that {\em every possible pair} of $S \in \mathcal{S}_{\delta}$ and $h \in \mathcal{H}_{\delta}$ have empirical error that is close to the test error of $h \in \mathcal{H}_{\delta}$, which is small. However, we will construct situations where there exist ``bad'' $(S,h)$ pairs {\em with large empirical error} and low test error that ruin uniform convergence.

\section[Warm-up]{Warm-up: An abstract setup}
\label{sec:warm-up}
As a warm up to our main results, we first present the failure of uniform convergence in an abstract setup. Although unconventional in some ways, our setup here conveys the essence behind how uniform convergence fails to explain generalization. \\

Let the underlying distribution over the inputs be a spherical Gaussian in $\mathbb{R}^N$ where $N$ can be however small or large as the reader desires. Note that our setup would apply to many other distributions, but a Gaussian would make our discussion easier. Let the labels of the inputs be determined by some $h^{\star}: \mathbb{R}^{N} \to \{-1, +1 \}$.
Consider a scenario where the learning algorithm outputs a very slightly modified version of $h^\star$. Specifically, let $S' = \{-\vec{x} \; | \; \vec{x} \in S\}$; then, the learner outputs
\begin{equation}
h_{S}(\vec{x}) = \begin{cases}
-h^\star(\vec{x}) & \text{if } \vec{x} \in S' \\
h^\star(\vec{x}) & \text{otherwise}.
\end{cases} 
\end{equation}
That is, the learner misclassifies inputs that correspond to the negations of the samples in the training data -- this would be possible if and only if the classifier is overparameterized with $\Omega(mN)$ parameters to store $S'$. We will show that uniform convergence fails to explain generalization for this learner. \\

First we establish that this learner generalizes well. A given $S$ has zero probability mass under $\scrD$, and so does $S'$. Then,
the training and test error are zero --- except for pathological draws of $S$ that intersect with $S'$, which are almost surely never drawn from $\scrD^m$ --- and hence, the generalization error of $\mathcal{A}$ is zero too.

It might thus seem reasonable to expect that one could explain this generalization using implicit-regularization-based uniform convergence by showing $\epsilon_{\text{unif-alg}}(m,\delta)=0$. Surprisingly, this is not the case as $\epsilon_{\text{unif-alg}}(m,\delta)$ is in fact $1$! \\

First it is easy to see why the looser bound $\epsilon_{\text{unif}}(m,\delta)$ equals 1, if we let $\mathcal{H}$ be the space of all hypotheses the algorithm could output: there must exist a non-pathological $S \in \mathcal{S}_{\delta}$, and we know that $h_{S'} \in \mathcal{H}$ misclassifies the negation of its training set, namely $S$. Then, $\sup_{h \in \mathcal{H}} |\scrL_{\scrD}(h) - \hat{\scrL}_{S}(h)| = |\scrL_{\scrD}(h_{S'}) - \hat{\scrL}_{S}(h_{S'})|  = |0-1|=1$.

One might hope that in the stronger bound of $\epsilon_{\text{unif-alg}}(m,\delta)$ since we truncate the hypothesis space, it is possible that the above adversarial situation would fall apart. However, with a more nuanced argument, we can similarly show that 
 $\epsilon_{\text{unif-alg}}(m,\delta)=1$. First, recall that any bound on $\epsilon_{\text{unif-alg}}(m,\delta)$,  would have to pick a truncated sample set space $\mathcal{S}_{\delta}$. Consider any choice of $\mathcal{S}_{\delta}$, and the corresponding set of explored hypotheses $\mathcal{H}_{\delta}$. We will show that for any choice of $\mathcal{S}_{\delta}$, there exists $S_{\star} \in \mathcal{S}_{\delta}$ such that (i) $h_{S_\star}$ has zero test error and (ii) the negated training set $S_{\star}'$ belongs to $\mathcal{S}_{\delta}$ and (iii) $h_{S_\star}$ has error $1$ on $S_\star$. Then, it follows that 
$\epsilon_{\text{unif-alg}}(m,\delta) = \sup_{S \in \mathcal{S}_{\delta}} \sup_{h \in \mathcal{H}_{\delta}} |{\scrL}_{\scrD}(h)- \hat{\scrL}_{S}(h)| \geq  |{\scrL}_{\scrD}(h_{S_\star})- \hat{\scrL}_{S_\star'}(h)|= |0-1| = 1 $.\\

% Since for all $h$ output by the algorithm, we have that $\scrL_{\scrD}(h) = 0$, we essentially have that $\epsilon_{\text{unif-alg}}(m,\delta) \geq \sup_{S \in \mathcal{S}_{\delta}} \sup_{h \in \mathcal{H}_{\delta}} |\hat{\scrL}_{S}(h)| $.

We can prove the existence of such an $S_{\star}$ by showing that the probability of picking one such set under $\scrD^{m}$ is non-zero for $\delta < 1/2$.  Specifically, under $S \sim \scrD^m$, we have by the union bound that
\begin{align*}
& \mathbb{P} \left[  \scrL_{\scrD}(h_S) = 0, \hat{\scrL}_{S'}(h_{S}) = 1, S \in \mathcal{S}_{\delta}, S' \in \mathcal{S}_{\delta}\right] \geq  \\
& 1- \mathbb{P}\left[  \scrL_{\scrD}(h_S) \neq 0,  \hat{\scrL}_{S'}(h_{S}) \neq 1\right] 
- \mathbb{P}\left[   S \notin \mathcal{S}_{\delta}\right] 
- \mathbb{P}\left[   S' \notin \mathcal{S}_{\delta}\right]. \numberthis
\end{align*}
Since the pathological draws have probability zero, the first probability term on the right hand side is zero. The second term is at most $\delta$ by definition of $\mathcal{S}_{\delta}$. Crucially, the last term too is at most $\delta$ because $S'$ (which is the negated version of $S$) obeys the same distribution as $S$ (since the isotropic Gaussian is invariant to a negation). Thus, the above probability is at least $1-2\delta > 0$, implying that there exist (many) $S_\star$, proving our main claim.

\begin{remark} While our particular learner might seem artificial, much of this artificiality is only required to make the argument simple. The crucial trait of the learner that we require is that the misclassified region in the input space (i) covers low probability and yet (ii) is complex and highly dependent on the training set draw. Our intuition is that SGD-trained deep networks possess these traits, which we will try to demonstrate in a later section.
\end{remark}
% Below, we provide an intuitive illustration of our conjecture.

\section{High-dimensional linear classifier}
\label{sec:uc-linear}

\subsection{Why a linear model?}  

Although we present a neural network example in the next section, we first emphasize why it is also important to understand how uniform convergence could fail for linear classifiers trained using GD. First, it is more natural to expect uniform convergence to yield poorer bounds in more complicated classifiers; linear models are arguably the simplest of classifiers, and hence showing failure of uniform convergence in these models is, in a sense, the most interesting.
Secondly, recent works (e.g., \citep{jacot18ntk}) have shown that as the width of a deep network goes to infinity, under some conditions, the network converges to a high-dimensional linear model (trained on a high-dimensional transformation of the data) -- thus making the study of high-dimensional linear models relevant to us. Note that our example is not aimed at modeling the setup of such linearized neural networks. However, 
it does provide valuable intuition about the mechanism by which uniform convergence fails, and we show how this extends to neural networks in the later sections.

%As we will see, the weight norms of this classifier grow with $m$ while the margins do not, resembling our observations in deep networks -- this disparity is in fact crucial to our lower bound on uniform convergence. We later present extensions of this idea to non-linear neural networks trained by SGD.

\subsection{Main result} 

Let each input be a $K+N$ dimensional vector (think of $K$  as a small constant and $N$ much larger than $m$). The value of any input $\vec{x}$ is denoted by $(\vec{x}_1, \vec{x}_2)$ where $\vec{x}_1 \in \mathbb{R}^K$ and $\vec{x}_2 \in \mathbb{R}^N$. Let the centers of the (two) classes be determined by an arbitrary vector $\vec{u} \in \mathbb{R}^K$ such that $\|\vec{u}\|_2=1/\sqrt{m}$.  Let $\scrD$ be such that the label $y$ has equal probability of being $+1$ and $-1$, and $\vec{x}_1 = 2\cdot y \cdot \vec{u}$ while $\vec{x}_2$ is sampled independently from a spherical Gaussian, $\mathcal{N}(0,\frac{32}{N}I)$.\footnote{As noted in Section~\ref{sec:remark}, it is easy to extend the discussion by assuming that $\vec{x}_1$ is spread out around $2y\vec{u}$.} Note that the distribution is linearly separable based on the first few ($K$) dimensions. 
%While some other aspects of our distribution may seem artificial (such as the fact that $\|\vec{u}\| = 1/\sqrt{m}$ or that there are at least $m$ noisy dimensions), we note that these requirements stem only in the linear model. When we look at examples of the more complicated neural network models, we can show failure of uniform convergence in realistic scenarios (where class separation is independent of $m$ and $N$ is small). 
For the learning algorithm $\mathcal{A}$, consider a linear classifier with weights $\vec{w} = (\vec{w}_1, \vec{w}_2)$ and whose output is $h(\vec{x}) = \vec{w}_1 \vec{x}_1 + \vec{w}_2 \vec{x}_2$. Assume the weights are initialized to the origin. Given a dataset $S$, $\mathcal{A}$ takes a gradient step of learning rate $1$ to maximize $y \cdot h(\vec{x})$ for each $(\vec{x},y) \in S$. Hence, regardless of the batch size, the learned weights would satisfy,  $\vec{w}_1 = 2m \vec{u}$ and $\vec{w}_2 = \sum_i y^{(i)} \vec{x}_2^{(i)}$.   Note that effectively $\vec{w}_1$ is aligned correctly along the class boundary while $\vec{w}_2$ is high-dimensional Gaussian noise.  It is fairly simple to show that this algorithm achieves {\em zero training error} for most draws of the training set. At the same time, for this setup, we have the following lower bound on uniform convergence for the $\scrL^{(\gamma)}$ loss:\footnote{While it is obvious from Theorem~\ref{thm:example} that the bound is nearly vacuous for any $\gamma \in [0,1]$, even for any $\gamma \geq 1$, the guarantee is nearly vacuous, although in a slightly different sense. We discuss this in Section~\ref{sec:any-gamma}. }

% \begin{theorem}
% For any $\epsilon,\delta > 0, \delta \leq 1/4$, when $N = \Omega\left(\max\left( m \ln \frac{m}{\delta}, m \ln\frac{1}{\epsilon}\right) \right)$, $\gamma \in [0,1]$,   the $\scrL^{(\gamma)}$ loss satisfies $\epsilon_{\text{gen}}(m,\delta) \leq \epsilon$, while $\epsilon_{\text{unif-alg}}(m,\delta) \geq 1- \epsilon$. Furthermore,
%  for all $\gamma \geq 0$, for the $\scrL^{(\gamma)}$ loss, $\epsilon_{\text{unif-alg}}(m,\delta) \geq 1- \epsilon_{\text{gen}}(m,\delta)$.\\
% \end{theorem}

Below, we state the precise theorem statement:
\begin{theorem}
\label{thm:example}
In the setup above, for any $\epsilon,\delta > 0$ and $\delta < 1/4$, let $N$ be sufficiently large that it satisfies
\begin{align*}
N & \geq \frac{1}{c_1} \ln \frac{6m}{\delta}, \numberthis\label{eq:d1}\\
{N} & \geq  {m} \left( \frac{4c_4 c_3}{c_2^2} \right)^2 \ln \frac{6m}{\delta},  \numberthis\label{eq:d2}\\
{N}  & \geq {m}\left(\frac{4c_4 c_3}{c_2^2}\right)^2 \cdot 2{\ln \frac{2}{\epsilon}},  \numberthis\label{eq:d3}
\end{align*}
where we've defined constants $c_1=1/2048$, $c_2 =\sqrt{15/16}$ and $c_3 = \sqrt{17/16}$ and $c_4 =\sqrt{2}$.
%$c_1 \coloneqq 1/32$, $c_2 \coloneqq 1/2$ and $c_3 \coloneqq 3/2$ and $c_4 \coloneqq \sqrt{2}$.

Then we have that  for all $\gamma \geq 0$, for the $\scrL^{(\gamma)}$ loss, 
\begin{equation}\epsilon_{\text{unif-alg}}(m,\delta) \geq 1- \epsilon_{\text{gen}}(m,\delta).\end{equation}

Specifically, for $\gamma \in [0,1]$, 
\begin{equation}\epsilon_{\text{gen}}(m,\delta) \leq \epsilon,
\end{equation} and so 
\begin{equation}
\epsilon_{\text{unif-alg}}(m,\delta) \geq 1- \epsilon.
\end{equation}
%then,  $\epsilon_{\text{gen}} (m,\delta)\leq \epsilon$  and $\epsilon_{\text{unif-alg}} \geq 1- \epsilon$ for both $\scrL^{(0)}$ and $\scrL_\gamma$
\end{theorem}

In other words, even the tightest uniform convergence bound is nearly vacuous despite good generalization. In order to better appreciate the implications of this statement, it will be helpful to look at the bound a standard technique would yield here. For example, the Rademacher complexity of the class of $\ell_2$-norm bounded linear classifiers would yield a bound of the form $\mathcal{O}(\|\vec{w}\|_2/(\gamma^\star\sqrt{m}))$  where $\gamma^\star$ is the margin on the training data. 
In this setup, the weight norm grows with dataset size as $\| \vec{w}\|_2 = \Theta(\sqrt{m})$ (which follows from the fact that $\vec{w}_2$ is a Gaussian with $m/D$ variance along each of the $N$ dimensions) and $\gamma^{\star} = \Theta(1)$. Hence, the Rademacher bound here would evaluate to a constant much larger than $\epsilon$.  One might persist and think that 
%the blame here still does not lie on uniform convergence; 
perhaps, the characterization of $\vec{w}$ to be bounded in $\ell_2$ norm does not  fully capture the implicit bias of the algorithm.\\

 Are there other properties of the Gaussian $\vec{w}_2$ that one could take into account to identify an even smaller class of hypotheses for which uniform convergence may work after all? Unfortunately, our statement rules this out:  even after fixing $\vec{w}_1$ to the learned value ($2m\vec{u}$) and for any possible $1-\delta$ truncation of the Gaussian $\vec{w}_2$, the resulting  pruned class of weights -- despite all of them having a test error less than $\epsilon$ -- would give only nearly vacuous uniform convergence bounds as $\epsilon_{\text{unif-alg}}(m,\delta) \geq 1-\epsilon$. 

% In summary, even in a mildly overparameterized setup, however hard one might try to incorporate the implicit bias of the algorithm, and whatever uniform convergence based tool one employs, one could grossly over-estimate the generalization error. % This reveals a crucial weakness in uniform convergence based approaches to explaining generalization in deep learning.

\paragraph{Proof outline.}
We now provide an outline of our argument for Theorem~\ref{thm:example}, deferring the proof to the upcoming subsection. First, the small generalization (and test) error arises from the fact that $\vec{w}_1$ is aligned correctly along the true boundary; at the same time, the noisy part of the classifier $\vec{w}_2$ is poorly aligned with at least $1-\epsilon$ mass of the test inputs, and hence does not dominate the output of the classifier on test data -- preserving the good fit of $\vec{w}_1$ on the test data. On the other hand, at a very high level, under the purview of uniform convergence, we can argue that the noise vector $\vec{w}_2$ is effectively stripped of its randomness. This misleads uniform convergence into believing that the $N$ noisy dimensions (where $N > m$) contribute meaningfully to
the representational complexity of the classifier, thereby giving nearly vacuous bounds. We describe this more concretely below.

As a key step in our argument, we show that w.h.p over draws of $S$, even though the learned classifier $h_S$ correctly classifies most of the randomly picked test data, it completely misclassifies a ``bad'' dataset, namely $S'= \{ ((\vec{x}_1, -\vec{x}_2),y) \; | \; (\vec{x},y) \in S \}$ which is the noise-negated version of $S$.  Now recall that to compute $\epsilon_{\text{unif-alg}}$ one has to begin by picking a sample set space $\mathcal{S}_{\delta}$ of mass $1-\delta$. We first argue that for {\em any} choice of $\mathcal{S}_{\delta}$, there must exist $S_\star$ such that all the following four events hold: 
\begin{enumerate}[(i)]
\item $S_\star \in \mathcal{S}_{\delta}$,
\item the noise-negated $S_\star' \in \mathcal{S}_{\delta}$,
\item $h_{S_\star}$ has test error less than $\epsilon$ and
\item $h_{S_\star}$ completely misclassifies $S_{\star}'$.
\end{enumerate}

 We prove the existence of such an $S_\star$ by arguing that over draws from $\scrD^m$, there is non-zero probability of picking a dataset that satisfies these four conditions. 
Note that our argument for this crucially makes use of the fact that we have designed the ``bad'' dataset in a way that it has the same distribution as the training set, namely $\scrD^m$.
 %First, over the draws of $S$, by construction of $\mathcal{S}_{\delta}$, (i) alone fails with probability at most $\delta$. We have established that (iii) and (iv) too fail with probability $\mathcal{O}(\delta)$.  As for (ii), note that under the draws of $S$, the noise-negated dataset $S'$ has the same distribution as that of $\scrD^m$ (since negating the Gaussian vector does not affect its distribution); hence, by construction of $\mathcal{S}_{\delta}$, even (ii) fails with probability $\delta$. 
Finally, for a given $\mathcal{S}_{\delta}$, if we have an $S_\star$ satisfying (i) to (iv), we can prove our claim as $\epsilon_{\text{unif-alg}}(m,\delta)  =
 \sup_{S \in \mathcal{S}_{\delta}} \sup_{h \in \mathcal{H}_{\delta}} |{\scrL}_{\scrD}(h)- \hat{\scrL}_{S}(h)| 
 \geq  |{\scrL}_{\scrD}(h_{S_\star})-\hat{\scrL}_{S_\star'}(h_{S_\star})|= |\epsilon-1| = 1-\epsilon$. \\

% Specifically, we have by the union bound that:
%\begin{align*}
%& Pr_{S \sim \scrD^m} \left[ S \in \mathcal{S}_{\delta}, S' \in \mathcal{S}_{\delta}, \scrL_{\scrD}(h_S) \leq \epsilon, \hat{\scrL}_{S'}(h_S) =1\right] \geq  \\
%& 1- Pr_{S \sim \scrD^m}\left[  \scrL_{\scrD}(h_S) > \epsilon\right] 
%- Pr_{S \sim \scrD^m}\left[   S \notin \mathcal{S}_{\delta}\right]  \\
%&
%- Pr_{S \sim \scrD^m}\left[   S' \notin \mathcal{S}_{\delta}\right] - Pr_{S \sim \scrD^m}\left[   \hat{\scrL}_{S'}(h_S) \neq 1\right] 
%\end{align*}

\begin{remark} Our analysis depends on the fact that $\epsilon_{\text{unif-alg}}$ is a two-sided convergence bound -- which is what existing techniques bound -- and our result would not apply for hypothetical one-sided uniform convergence bounds. 
%Existing uniform convergence based tools are only two-sided as it is more natural to bound the absolute value of the difference in the test/train errors. 
While PAC-Bayes based bounds are typically presented as one-sided bounds, we show in Section~\ref{sec:uc-pac-bayes} that even these are lower-bounded by the two-sided  $\epsilon_{\text{unif-alg}}$. To the best of our knowledge, it is non-trivial to make any of these tools purely one-sided. \\
\end{remark} 

\begin{remark} The classifier modified by  setting $\vec{w}_2 \gets 0$, has small test error
and also enjoys non-vacuous bounds as it has very few parameters.  However, such a bound would not fully explain why the original classifier generalizes well. One might then wonder if such a bound could be extended to the original classifier, like it was explored in \citet{nagarajan18deterministic} for deep networks. Our result implies that no such extension is possible in this particular example. \\
\end{remark}

\subsection{Proof for Theorem~\ref{thm:example}}
\label{sec:uc-linear-proof}

In this section, we prove the failure of uniform convergence for our linear model. We first recall the setup:

% \textbf{Distribution $\scrD$:} Each input $(\vec{x}_1, \vec{x}_2)$ is a $K+D$ dimensional vector where  $\vec{x}_1 \in \mathbb{R}^K$ and $\vec{x}_2 \in \mathbb{R}^N$. $\vec{u} \in \mathbb{R}^K$ determines the centers of the classes. The label $y$ is drawn uniformly from $\{ -1, +1\}$, and conditioned on $y$, we have $\vec{x}_1 = 2\cdot y \cdot \vec{u}$ while $\vec{x}_2$ is sampled independently from $\mathcal{N}(0,\frac{32}{N} \vecI)$. 

% \textbf{Learning algorithm $\mathcal{A}$:} We consider a linear classifier with weights $\vec{w} = (\vec{w}_1, \vec{w}_2)$. The output is computed as $h(\vec{x}) = \vec{w}_1 \vec{x}_1 + \vec{w}_2 \vec{x}_2$. Assume the weights are initialized to origin. 
% Given $S = \{(\vec{x}^{(1)},y^{(1)}), \hdots, (\vec{x}^{(m)},y^{(m)})  \}$, $\mathcal{A}$ takes a gradient step of learning rate $1$ to maximize $y \cdot h(\vec{x})$ for each $(\vec{x},y) \in S$. Regardless of the batch size, the learned weights would satisfy,  $\vec{w}_1 = 2m \vec{u}$ and $\vec{w}_2 = \sum_i y^{(i)} \vec{x}_2^{(i)}$.  

\begin{proof}
The proof follows from Lemma~\ref{lem:gen} stated below, where we upper bound the generalization error, and from Lemma~\ref{lem:uc-unif} stated after that, where we lower bound uniform convergence.
\end{proof}

We first prove that the above algorithm generalizes well with respect to the losses corresponding to $\gamma \in [0,1]$. First for the training data, we argue that both $\vec{w}_1$ and a small part of the noise vector $\vec{w}_2$ align along the correct direction, while the remaining part of the high-dimensional noise vector are orthogonal to the input; this leads to correct classification of the training set. Then, on the test data, we argue that $\vec{w}_1$ aligns well, while $\vec{w}_2$ contributes very little to the output of the classifier because it is high-dimensional noise. As a result, for most test data, the classification is correct, and hence the test and generalization error are both small.
\\

\begin{lemma}
\label{lem:gen}
In the setup of Section~\ref{sec:setup}, when $\gamma \in [0,1]$, for $\scrL^{(\gamma)}$, $\epsilon_{\text{gen}} (m,\delta)\leq \epsilon$.
\end{lemma}

\begin{proof}
The parameters learned by our algorithm satisfies $\vec{w}_1= 2 m\cdot \vec{u}$ and $\vec{w}_2  = \sum y^{(i)} \vec{x}_2^{(i)} \sim \mathcal{N}(0, \frac{8m}{c_2^2 N})$.  

First, we have from Corollary~\ref{cor:chi} that with probability $1-\frac{\delta}{3m}$ over the draws of $\vec{x}_2^{(i)}$, as long as $\frac{\delta}{3m} \geq 2e^{-c_1N}$ (which is given to hold by Equation~\ref{eq:d1}), \begin{equation}c_2 \leq  \frac{1}{2\sqrt{2}}{c_2} \|\vec{x}_2^{(i)}\| \leq c_3.
\label{eq:ce1}
 \end{equation}

Next, for a given $\vec{x}^{(i)}$, we have from Corollary~\ref{cor:gaussian}, with probability $1-\frac{\delta}{3m}$ over the draws of $\sum_{j\neq i} y^{(j)}\vec{x}_2^{(j)}$, \begin{equation}|\vec{x}_2^{(i)} \cdot \sum_{j\neq i} y^{(j)}\vec{x}_2^{(j)}| \leq  c_4 \|\vec{x}_2^{(i)} \| \frac{2\sqrt{2} \cdot \sqrt{m}}{c_2 \sqrt{N}} \sqrt{ \ln \frac{6m}{\delta
}}.
\label{eq:ce2}\end{equation}

Then, with probability $1-\frac{2}{3}\delta$ over the draws of the training dataset we have for all $i$, %, on the training input $\vec{x}_2^{(i)}$, the output of our hypothesis is:

\begin{align*}
y^{(i)} h(\vec{x}^{(i)}) &= y^{(i)} \vec{w}_1 \cdot \vec{x}^{(i)}_1 + y^{(i)} \cdot y^{(i)} \|\vec{x}_2^{(i)} \|^2 + y^{(i)} \cdot \vec{x}_2^{(i)} \cdot \sum_{j\neq i} y^{(j)}\vec{x}_2^{(j)}  \\
&=4 +  \underbrace{\|\vec{x}_2^{(i)} \|^2}_{\text{apply Equation~\ref{eq:ce1}}} +  \underbrace{y^{(i)} \vec{x}_2^{(i)} \cdot \sum_{j\neq i} y^{(j)}\vec{x}_2^{(j)}}_{\text{apply Equation~\ref{eq:ce2}}}  \\
& \geq 4 +4 \cdot 2   -  c_4 \frac{2 \sqrt{2} c_3}{c_2} \cdot \underbrace{\frac{2\sqrt{2} \cdot \sqrt{m}}{c_2 \sqrt{N}} \sqrt{\ln \frac{6m}{\delta}}}_{\text{apply Equation~\ref{eq:d2}}} \\
& \geq 4 + 8 - 2 = 10 > 1. \numberthis\label{eq:train-margin}
\end{align*}

Thus, for all $\gamma \in [0,1]$, the $\scrL^{(\gamma)}$ loss of this classifier on the training dataset $S$ is zero. 

Now, from Corollary~\ref{cor:chi}, with probability $1-\frac{\delta}{3}$ over the draws of the training data, we also have that, as long as $\frac{\delta}{3m} \geq 2 e^{-c_1 N}$ (which is given to hold by Equation~\ref{eq:d1}), \begin{equation}c_2 \sqrt{m} \leq \frac{1}{2\sqrt{2}} c_2 \|\sum y^{(i)} \vec{x}_2^{(i)}\| \leq c_3\sqrt{m}. \label{eq:ce3} \end{equation}

Next, conditioned on the draw of $S$ and the learned classifier, for any $\epsilon' > 0$, with probability $1-\epsilon'$ over the draws of a test data point, $(\vec{z},y)$, we have from Corollary~\ref{cor:gaussian} that

\begin{equation}|\vec{z}_2 \cdot \sum y^{(i)} \vec{x}_2^{(i)}| \leq c_4 \|\sum y^{(i)} \vec{x}_2^{(i)}\| \cdot \frac{2\sqrt{2}}{c_2 \sqrt{N}}  \cdot  \ln \frac{1}{\epsilon'}.
\label{eq:ce4}
\end{equation}

Using this, we have that with probability $1- 2 \exp \left(- \frac{1}{2}\left({\frac{c_2^2}{4 c_4 c_3}  \sqrt{\frac{N}{m}} }\right)^2\right)$ over the draws of a test data point, $(\vec{z},y)$,
\begin{align*}
 y h(\vec{x})  & =  y \vec{w}_1 \cdot \vec{z}_1 +\underbrace{  y \cdot \vec{z}_2 \cdot \sum_{j} y^{(j)}\vec{x}_2^{(j)} }_{\text{apply Equation~\ref{eq:ce4}} } \\
& \geq  4 -  c_4 \underbrace{\|\sum y^{(i)} \vec{x}_2^{(i)}\| }_{\text{apply Equation~\ref{eq:ce3}}}\cdot \frac{2\sqrt{2}}{c_2 \sqrt{N}} \frac{c_2^2}{4 c_4 c_3} \sqrt{\frac{N}{m}} \\
& \geq 4 - 2 \geq 2. \numberthis\label{eq:test-margin}
\end{align*}

%TODO Perhaps justify how we can modify the setup to include multiple datapoints

Thus, we have that for $\gamma \in [0,1]$, the $\scrL^{(\gamma)}$ loss of the classifier on the distribution $\scrD$ is $2\exp \left(-\frac{1}{2}\left({\frac{c_2^2}{4 c_4 c_3} \sqrt{\frac{N}{m}} }\right)^2\right)$ which is at most $\epsilon$ as assumed in Equation~\ref{eq:d3}. In other words, the absolute difference between the distribution loss and the train loss is at most $\epsilon$
 and this holds for at least $1-\delta$ draws of the samples $S$. Then, by the definition of $\epsilon_{\text{gen}}$ we have the result. 

\end{proof}

We next prove our uniform convergence lower bound. The main idea is that when the noise vectors in the training samples are negated, with high probability, the classifier misclassifies the training data. We can then show that for any choice of $\mathcal{S}_{\delta}$ 
as required by the definition of $\epsilon_{\text{unif-alg}}$, we can always find an $S_{\star}$ and its noise-negated version $S_{\star}'$ both of which belong to $\mathcal{S}_{\delta}$. Furthermore, we can show that $h_{S_\star}$ has small test error but high empirical error on $S_{\star}'$, and that this leads to a nearly vacuous uniform convergence bound.

\begin{lemma}
\label{lem:uc-unif}
In the setup of Section~\ref{sec:setup}, for any $\epsilon > 0$ and for any $\delta \leq 1/4$, and for the same lower bounds on $N$,  and for any $\gamma \geq 0$, we have that
\begin{equation}
\epsilon_{\text{unif-alg}}(m,\delta) \geq 1 - \epsilon_{\text{gen}}(m,\delta)
\end{equation}
for the $\scrL^{(\gamma)}$ loss.
\end{lemma}

\begin{proof}

For any $S$, let $S'$ denote the set of noise-negated samples $S' = \{ ((\vec{x}_1, - \vec{x}_2),u) \; | \;  ((\vec{x}_1,  \vec{x}_2),y) \in S \}$. We first show with high probability $1-2\delta/3$ over the draws of $S$, that the classifier learned on $S$,  misclassifies $S'$ completely. The proof for this is nearly identical to our proof for why the training loss is zero, except for certain sign changes. For any $\vec{x}_{\text{neg}}^{(i)} = (\vec{x}_1^{(i)}, -\vec{x}_2^{(i)})$, we have

\begin{align*}
 y^{(i)} h(\vec{x}_{\text{neg}}^{(i)})& = y^{(i)} \vec{w}_1 \cdot \vec{x}^{(i)}_1 - y^{(i)} \cdot y^{(i)} \|\vec{x}_2^{(i)} \|^2 - y^{(i)} \cdot \vec{x}_2^{(i)} \cdot \sum_{j\neq i} y^{(j)}\vec{x}_2^{(j)}  \\
&=4-  \underbrace{\|\vec{x}_2^{(i)} \|^2}_{\text{apply Equation~\ref{eq:ce1}}} -  \underbrace{y^{(i)} \vec{x}_2^{(i)} \cdot \sum_{j\neq i} y^{(j)}\vec{x}_2^{(j)}}_{\text{apply Equation~\ref{eq:ce2}}}  \\
& \leq  4 - 4 \cdot 2  +  c_4 \frac{2 \sqrt{2} c_3}{c_2} \cdot \underbrace{\frac{2\sqrt{2} \cdot \sqrt{m}}{c_2 \sqrt{N}} \ln \frac{3m}{\delta}}_{\text{apply Equation~\ref{eq:d2}}} \\
& \leq 4 - 8 + 2 = -2 < 0.
\end{align*}

Since the learned hypothesis misclassifies all of $S'$, it has loss of $1$ on $S'$.\\ % according to the $\scrL^{(\gamma)}$ loss.

 Now recall that, by definition, to compute $\epsilon_{\text{unif-alg}}$, one has to pick a sample set space $\mathcal{S}_{\delta}$ of mass $1-\delta$ i.e., $\prsub{S \sim \mathcal{S}^m}{S \in \mathcal{S}_{\delta}} \geq 1-\delta$. We first argue that for {\em any} choice of $\mathcal{S}_{\delta}$, there must exist a `bad' $S_\star$ such that (i) $S_\star \in \mathcal{S}_{\delta}$, (ii) $S_\star' \in \mathcal{S}_{\delta}$, (iii) $h_{S_\star}$ has test error less than $\epsilon_{\text{gen}}(m,\delta)$ and (iv) $h_{S_\star}$ completely misclassifies $S_{\star}'$. 

 We show the existence of such an $S_\star$, by arguing that over the draws of $S$, there is non-zero probability of picking an $S$ that satisfies all the above conditions. Specifically, we have by the union bound that
\begin{align*}
& \mathbb{P}_{S \sim \scrD^m} \big[ S \in \mathcal{S}_{\delta}, S' \in \mathcal{S}_{\delta}, \scrL_{\scrD}(h_S) \leq \epsilon_{\text{gen}}(m,\delta), \hat{\scrL}_{S'}(h_S) =1\big]   \\
 &\geq 1
- \mathbb{P}_{S \sim \scrD^m}\left[   S \notin \mathcal{S}_{\delta}\right] -  \mathbb{P}_{S \sim \scrD^m}\left[   S' \notin \mathcal{S}_{\delta}\right]\\
&- \mathbb{P}_{S \sim \scrD^m}\left[  \scrL_{\scrD}(h_S) > \epsilon_{\text{gen}}(m,\delta)\right] 
  - \mathbb{P}_{S \sim \scrD^m}\left[   \hat{\scrL}_{S'}(h_S) \neq 1\right]. \numberthis \label{eq:probability-term}
\end{align*}

By definition of $\mathcal{S}_{\delta}$, we know $\mathbb{P}_{S \sim \scrD^m}\left[   S \notin \mathcal{S}_{\delta}\right] \leq \delta$. Similarly, by definition of the generalization error, we know that $\mathbb{P}_{S \sim \scrD^m}\left[  \scrL_{\scrD}(h_S) > \epsilon_{\text{gen}}(m,\delta)\right]  \leq \delta$. We have also established above that  $\mathbb{P}_{S \sim \scrD^m}\left[   \hat{\scrL}_{S'}(h_S) \neq 1\right]  \leq 2\delta/3$. As for the term $\mathbb{P}_{S \sim \scrD^m}\left[   S' \notin \mathcal{S}_{\delta}\right]$, observe that under the draws of $S$, the distribution of the noise-negated dataset $S'$ is identical to $\scrD^m$. This is because the isotropic Gaussian noise vectors have the same distribution under negation. Hence, again by definition of $\mathcal{S}_{\delta}$, even this probability is at most $\delta$. Thus, we have that the probability in the left hand side of Equation~\ref{eq:probability-term} is at least $1-4\delta$, which is positive as long as $\delta < 1/4$.\\

This implies that for any given choice of $\mathcal{S}_{\delta}$, there exists $S_{\star}$ that satisfies our requirement.  Then, from the definition of $\epsilon_{\text{unif-alg}}(m,\delta)$, we essentially have that,
\begin{align}
\epsilon_{\text{unif-alg}}(m,\delta) &=  \sup_{S \in \mathcal{S}_{\delta}} \sup_{h \in \mathcal{H}_{\delta}} |{\scrL}_{\scrD}(h)- \hat{\scrL}_{S}(h)| \\
& \geq  |{\scrL}_{\scrD}(h_{S_\star})- \hat{\scrL}_{S_\star'}(h)|= |\epsilon-1| = 1-\epsilon.
\end{align}

\end{proof}

 \section{ReLU neural network}
  \label{sec:uc-hypersphere} 
We now design a non-linearly separable task (with no ``noisy'' dimensions) where a sufficiently wide ReLU network trained in the standard manner, like in the experiments of Section~\ref{sec:uc-experiments} leads to failure of uniform convergence.  For our argument, we will rely on a classifier trained {\em empirically}, in contrast to our linear examples where we rely on an analytically derived expression for the learned classifier. %That is, we theoretically argue how the learned decision boundary hurts uniform convergence, although we do not theoretically analyze how such a decision boundary is learned.
 Thus, this section illustrates that the effects we modeled theoretically in the linear classifier \emph{are} indeed reflected in typical training settings, even though here it is difficult to precisely analyze the learning process. % We also refer the reader to

\textbf{Setup.} We consider a distribution that was originally proposed in \cite{gilmer18adversarial} as the ``adversarial spheres'' dataset (although with slightly different hyperparameters) and was used to study the independent phenomenon of adversarial examples. Specifically, we consider 1000-dimensional data, where two classes are distributed uniformly over two origin-centered hyperspheres with radius $1$ and $1.1$ respectively.  We vary the number of training examples from $4k$ to $65k$ (thus ranging through typical dataset sizes like that of MNIST). Observe that compared to the linear example, this data distribution is more realistic in two ways. First, we do not have specific dimensions in the data that are noisy and second, the data dimensionality here as such is a constant less than $m$. Given samples from this distribution, we  train a two-layer ReLU network with $h=100k$ to minimize cross entropy loss using SGD with learning rate $0.1$ and batch size $64$. We train the network until $99\%$ of the data is classified by a margin of $10$.

\begin{figure}[t]
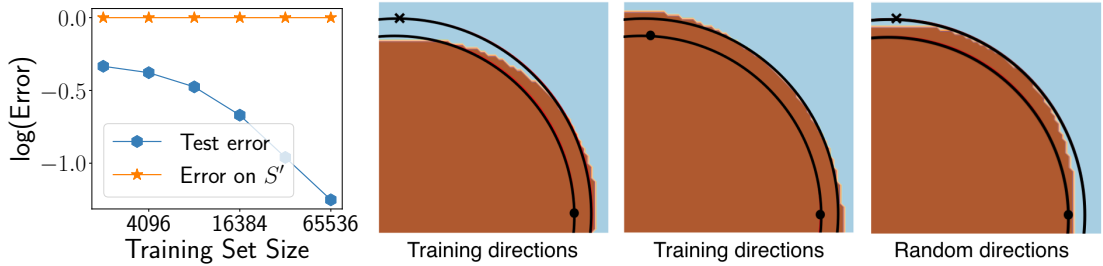

    \centering
    \begin{minipage}[t]{0.3\textwidth}
        \adjincludegraphics[width=1\textwidth,trim={0 {0} 0 0},clip,,valign=t]{UniformConvergenceImages/relu_error_vs_m} %
    \end{minipage}
    \begin{minipage}[t]{0.6\textwidth}
        \centering
        \adjincludegraphics[width=\textwidth,trim={0 {0} 0 0},clip,,valign=t]{UniformConvergenceImages/ReLU/all} %
    \end{minipage}
       \caption{Failure of uniform convergence for a ReLU network trained on a hypersphere classification data.} 
       \label{fig:hypersphere}
\end{figure}

\paragraph{Observations.} As shown in Figure~\ref{fig:hypersphere} (blue line), in this setup, the 0-1 error  (i.e., $\scrL^{(0)}$) as approximated by the test set, decreases with $m \in [2^{12}, 2^{16}]$ at the rate of  $O(m^{-0.5})$. 
Now, to prove failure of uniform convergence, we empirically show that a completely misclassified ``bad'' dataset $S'$ can be constructed in a manner similar to that of the previous example. In this setting, we pick $S'$ by simply projecting every training datapoint on the inner hypersphere onto the outer and vice versa, and then flipping the labels. Then, as shown in Figure~\ref{fig:hypersphere} (orange line), $S'$ is completely misclassified by the learned network. Furthermore, like in the previous example, we have  $S' \sim \scrD^m$ because the distributions are uniform over the hyperspheres. Having established these facts, the rest of the argument follows like in the previous setting, implying failure of uniform convergence as in Theorem~\ref{thm:example} here too.% in this setting too.  

In Figure~\ref{fig:hypersphere}  (right), we visualize how the learned boundaries are skewed around the training data in a way that $S'$ is misclassified.  The \textbf{second and third} images correspond to the decision boundary learned in this task, in the 2N quadrant containing two training datapoints (depicted as $\boldsymbol{\small \times}$ and $\bullet$). The black lines correspond to the two hyperspheres, while the brown and blue regions correspond to the class output by the classifier.  Here, we observe that  the boundaries are skewed around the training data in a way that it misclassifies the nearest point from the opposite class (corresponding to $S'$, that is not explicitly marked). The \textbf{fourth} image corresponds to two random (test) datapoints, where the boundaries are fairly random, and very likely to be located in between the hyperspheres (better confirmed by the low test error).

 Note that $S'$ is misclassified even when it has as many as $60k$ points, and even though the network was not explicitly trained to misclassify those points. Intuitively, this demonstrates that the boundary learned by the ReLU network has sufficient complexity that hurts uniform convergence while not affecting the generalization error, at least in this setting. We discuss the applicability of this observation to other hyperparameter settings in Section~\ref{sec:applicability}.

\subparagraph{Relationship to adversarial spheres \cite{gilmer18adversarial}.} While we use the same adversarial spheres distribution as \cite{gilmer18adversarial} and similarly show the existence a certain kind of an adversarial dataset, it is important to note that neither of our observations implies the other. Indeed, the observations in \cite{gilmer18adversarial} are insufficient to prove failure of uniform convergence. Specifically, \cite{gilmer18adversarial} show that in the adversarial spheres setting, it is possible to slightly perturb random {\em test} examples in some {\em arbitrary} direction to discover a misclassified example. However, to show failure of uniform convergence, we need to find a set of misclassified examples $S'$ corresponding to the {\em training} examples $S$, and furthermore, we do {\em not} want $S'$ to be arbitrary. We want $S'$ to have the same underlying distribution, $\mathcal{D}^m$.

\subsection{Deep learning conjecture} 
\label{sec:dl-conjecture}
Extending the above insights more generally, we conjecture that in overparameterized deep networks, SGD finds a fit that is simple at a macroscopic level (leading to good generalization) but also has many microscopic fluctuations (hurting uniform convergence). 
To make this more concrete, for illustration, consider the high-dimensional linear model that sufficiently wide networks have been shown to converge to \citep{jacot18ntk}. That is, roughly, these networks can be written as $h(\vec{x}) = \vec{w}^T \vec{\phi}(\vec{x})$ where $\vec{\phi}(\vec{x})$
is a rich high-dimensional representation of $\vec{x}$ computed from many random features (chosen independent of training data).

Inspired by our linear model  in Section~\ref{sec:uc-linear}, we conjecture that the weights $\vec{w}$ learned on a dataset $S$ can be expressed as $\vec{w}_1 + \vec{w}_2$, where $\vec{w}_1^T\vec{\phi}(\vec{x})$ dominates the output on most test inputs and induces a simple decision boundary. That is, it may be possible to apply uniform convergence on the function $\vec{w}_1^T \vec{\phi}(\vec{x})$ to obtain a small generalization bound.  On the other hand, $\vec{w}_2$ corresponds to meaningless signals that gradient descent gathered from the high-dimensional representation of the training set $S$. Crucially, these signals would be specific to $S$, and hence not likely to correlate with most of the test data i.e., 
$\vec{w}_2\phi(\vec{x})$  would be negligible on most test data, thereby not affecting the generalization error significantly. However, $\vec{w}_2\phi(\vec{x})$ can still create complex fluctuations on the boundary, in low-probability regions of the input space (whose locations would depend on $S$, like in our examples). As we argued, this can lead to failure of uniform convergence. Perhaps, existing works that have achieved strong uniform convergence bounds on modified networks, may have done so by implicitly suppressing  $\vec{w}_2$, either by compression, optimization or stochasticization. Revisiting these works may help verify our conjecture.

\section{Neural network with exponential activations}
\label{sec:uc-exp}
In this section, we prove the failure of uniform convergence for a neural network model with exponential activations. Unlike in the previous section, here we will analytically derive an expression for the weights learned by the network. 
We first define the setup. Furthermore, here we will prove failure of uniform convergence in a low-dimensional task. % 

\paragraph{Distribution} Let $\vec{u}$ be an arbitrary vector in $N$ dimensional space such that $\| \vec{u}\| = \sqrt{N}/2$. 
Consider an input distribution in $2N$ dimensional space such that, conditioned on the label $y$ drawn from uniform distribution over $\{-1,+1\}$, the first $N$ dimensions $\vec{x}_1$ of a random point is given by $y \vec{u}$ and the remaining $N$ dimensions are drawn from $\mathcal{N}(0,1)$. Note that in this section, we require $N$ to be only as large as $\ln m$, and not as large as $m$.

\paragraph{Architecture.} We consider an infinite width neural network with exponential activations, in which only the output layer weights are trainable. The hidden layer weights are frozen as initialized. Note that this is effectively a linear model with infinitely many randomized features. Indeed, recent work \citep{jacot18ntk} has shown that under some conditions on how deep networks are initialized and parameterized, they behave a linear models on randomized features.
Specifically, each hidden unit corresponds to a distinct (frozen) weight vector $\vec{w} \in \mathbb{R}^{2N}$ and an output weight $a_{\vec{w}}$ that is trainable. %We assume $a_{\vec{w}}$ are initialized to zero. 
%To simplify our analysis, we assume that the hidden layer weights are distributed uniformly on a $K+D$ hypersphere of radius $1$. 
We assume that the hidden layer weights are drawn from $\mathcal{N}(0,\vecI)$ and $a_{\vec{w}}$ initialized to zero. 
Note that the output of the network is determined as

\begin{equation}
h(\vec{x}) = \mathbb{E}_{\vec{w}}[a_{\vec{w}}\exp(\vec{w}\cdot \vec{x})].
\end{equation}

\paragraph{Algorithm} We consider an algorithm that takes a gradient descent step to maximize $y\cdot h(\vec{x})$ for each $(\vec{x},y)$ in the training dataset, with learning rate $\eta$. However, since, the function above is not a discrete sum of its hidden unit outputs, to define the gradient update on $a_{\vec{w}}$, we must think of $h$  as a functional whose input function maps every $\vec{w} \in \mathbb{R}^{2N}$ to $a_{\vec{w}} \in \mathbb{R}$. Then, by considering the functional derivative, one can conclude that the update on $a_{\vec{w}}$ can be written as

\begin{equation}
\label{eq:gradient-step}
a_{\vec{w}} \gets a_{\vec{w}} + \eta y \cdot \exp(\vec{w}\cdot \vec{x}) \cdot p(\vec{w}).
\end{equation}

where $p(\vec{w})$ equals the p.d.f of $\vec{w}$ under the distribution it is drawn from. In this case $p(\vec{w}) = \frac{1}{(2\pi)^N} \exp\left( -\frac{\| \vec{w} \|^2}{2} \right)$.

In order to simplify our calculations we will set $\eta = (4\pi)^N$, although our analysis would extend to other values of the learning rate too. Similarly, our results would only differ by constants if we consider the alternative update rule, $a_{\vec{w}} \gets a_{\vec{w}} + \eta y \cdot \exp(\vec{w}\cdot \vec{x})$.

We now state our main theorem: % (in terms of constants $c_1, c_2, c_3, c_4$ defined in Section~\ref{sec:lemma}).\\

\begin{theorem}
\label{thm:exp}
In the set up above, for any $\epsilon,\delta > 0$ and $\delta < 1/4$, let $N$ and $m$ be sufficiently large that it satisfies
\begin{align*}
N & \geq  \max \left(\frac{1}{c_2}, (16c_3 c_4)^2 \right) \cdot 2 \ln \frac{6m}{\epsilon} \numberthis\label{eq:d4}\\
{N} & \geq   \max \left(\frac{1}{c_2}, (16c_3 c_4)^2 \right) \cdot 2 \ln \frac{6m}{\delta}   \numberthis\label{eq:d5}\\
{N}  & \geq 6 \ln 2m \numberthis\label{eq:d6} \\
{m} & >  \max  8 \ln \frac{6}{\delta},  \numberthis\label{eq:m}
\end{align*}
where we've defined constants $c_1=1/2048$, $c_2 =\sqrt{15/16}$ and $c_3 = \sqrt{17/16}$ and $c_4 =\sqrt{2}$.
Then we have that  for all $\gamma \geq 0$, for the $\scrL^{(\gamma)}$ loss, 
\begin{equation}
\epsilon_{\text{unif-alg}}(m,\delta) \geq 1- \epsilon_{\text{gen}}(m,\delta).
\end{equation}

Specifically, for $\gamma \in [0,1]$, 
\begin{equation}\epsilon_{\text{gen}}(m,\delta) \leq \epsilon,\end{equation} and so 
\begin{equation}\epsilon_{\text{unif-alg}}(m,\delta) \geq 1- \epsilon.\end{equation}
\end{theorem}
%then,  $\epsilon_{\text{gen}} (m,\delta)\leq \epsilon$  and $\epsilon_{\text{unif-alg}} \geq 1- \epsilon$ for both $\scrL^{(0)}$ and $\scrL_\gamma$

\begin{proof}
The result follows from the following lemmas. First in Lemma~\ref{lem:exp-gradient}, we derive the closed form expression for the function computed by the learned network. In Lemma~\ref{lem:exp-gen}, we upper bound the generalization error and in Lemma~\ref{lem:exp-unif-conv}, we lower bound uniform convergence.
\end{proof}

% Also write an example where input dimension doesn't matter.

We first derive a closed form expression for how the output of the network changes under a gradient descent step on a particular datapoint.\\

\begin{lemma}
\label{lem:exp-gradient}
Let $h^{(0)}(\cdot)$ denote the function computed by the network before updating the weights. After updating the weights on a particular input $(\vec{x},y)$ according to Equation~\ref{eq:gradient-step}, the learned network corresponds to:
\begin{equation}
h(\vec{z}) = h^{(0)}(\vec{z}) +   y \exp\left( \left\|\frac{\vec{z}+\vec{x}}{2}\right\|^2\right).
\end{equation}

\end{lemma}

\begin{proof}
 From equation~\ref{eq:gradient-step}, we have that

 \begin{align}
 &\frac{h(\vec{z}) -h^{(0)}(\vec{z})}{\eta} \\ &= \int_{\vec{w}} \left( y \cdot \exp(\vec{w}\cdot \vec{x})p(\vec{w})\right) \cdot \exp(\vec{w}\cdot \vec{z})p(\vec{w}) d\vec{w} \\
 & = y \int_{\vec{w}} \exp(\vec{w}\cdot (\vec{x}+\vec{z})) \cdot \left( {\frac{1}{2\pi}}\right)^{2N} \exp(-\|\vec{w}\|^2) d\vec{w}  \\
 & = y \left( {\frac{1}{2\pi}}\right)^{2N}  \int_{\vec{w}} \exp(\vec{w}\cdot (\vec{x}+\vec{z})-\|\vec{w}\|^2) d\vec{w}  \\
 & = y \left( {\frac{1}{2\pi}}\right)^{2N} \exp\left( \left\|\frac{\vec{z}+\vec{x}}{2}\right\|^2\right) \times \int_{\vec{w}} \exp\left(-\left\|\vec{w} - \frac{\vec{z}+\vec{x}}{2}\right\|^2\right) d\vec{w}  \\
 & = y \left( {\frac{1}{4\pi}}\right)^{N} \exp\left( \left\|\frac{\vec{z}+\vec{x}}{2}\right\|^2\right) \times   \left( \frac{1}{\sqrt{2\pi (0.5)}}\right)^{2N} \int_{\vec{w}} \exp\left(-\left\|\vec{w} - \frac{\vec{z}+\vec{x}}{2}\right\|^2\right) d\vec{w}  \\
  & = y \left( {\frac{1}{4\pi}}\right)^{N} \exp\left( \left\|\frac{\vec{z}+\vec{x}}{2}\right\|^2\right).   
 \end{align}

 In the last equality above, we make use of the fact that the second term corresponds to the integral of the p.d.f of $\mathcal{N}(\frac{\vec{z}+\vec{x}}{2}, 0.5 \vecI)$ over $\mathbb{R}^{2N}$.  Since we set $\eta = (4\pi)^N$ gives us the final answer.
\end{proof}

Next, we argue that the generalization error of the algorithm is small. From Lemma~\ref{lem:exp-gradient}, we have that the output of the network is essentially determined by a summation of contributions from every training point.
To show that the training error is zero, we argue that on any training point, the contribution from that training point dominates all other contributions, thus leading to correct classification.  On any test point, we similarly show that the contribution of training points of the same class as that test point dominates the output of the network. Note that our result requires $N$ to scale only logarithmically with training samples $m$. \\

\begin{lemma}
\label{lem:exp-gen}
In the setup of Section~\ref{sec:setup}, when $\gamma \in [0,1]$, for $\scrL^{(\gamma)}$, $\epsilon_{\text{gen}} (m,\delta)\leq \epsilon$.
\end{lemma}

\begin{proof}
We first establish a few facts that hold with high probability over the draws of the training set $S$. First, from Corollary~\ref{cor:chi} we have that, since $N \geq \frac{1}{c_2} \ln \frac{3m}{\delta}$ (from Equation~\ref{eq:d4}), with probability at least $1-\delta/3$ over the draws of $S$, for all $i$, the noisy part of each training input can be bounded as

\begin{align*}
c_2 \sqrt{N} \leq \| \vec{x}_2^{(i)}\| \leq c_3 \sqrt{N}. \numberthis\label{eq:exp-norm-bound}
\end{align*}

Next, from Corollary~\ref{cor:gaussian}, we have that with probability at least $1-\frac{\delta}{3m^2}$ over the draws of $\vec{x}_2^{(i)}$ and $\vec{x}_2^{(j)}$ for  $i \neq j$,

\begin{align*}
|\vec{x}_2^{(i)} \cdot \vec{x}_2^{(j)}| \leq \| \vec{x}_2^{(i)} \| \cdot c_4 \sqrt{2\ln \frac{6m}{\delta}}.   \numberthis\label{eq:exp-dot-prod-bound}
\end{align*} 

Then, by a union bound, the above two equations hold for all $i\neq j$ with probability at least $1-\delta/2$. 

Next, since each $y^{(i)}$ is essentially an independent sub-Gaussian with mean $0$ and sub-Gaussian parameter $\sigma = 1$, we can apply Hoeffding's bound (Lemma~\ref{lem:uc-hoeffding}) to conclude that with probability at least $1-\delta/3$ over the draws of $S$,
\begin{align*}
\left|\sum_{j=1}^{m} y^{(j)}\right| \leq \underbrace{\sqrt{2m\ln \frac{6}{\delta} }}_{Eq~\ref{eq:m}} < \frac{m}{2}  \numberthis\label{eq:class-count}.
\end{align*}

Note that this means that there must exist at least one training data in each class.

%In other words, this means that the number of training data in any particular class deviates from $m/2$ by at most $\sqrt{2m\ln \frac{6}{\delta}}$.

Given these facts, we first show that the training error is zero by showing that for all $i$, $y^{(i)} h(\vec{x}^{(i)})$ is sufficiently large. On any training input $(\vec{x}^{(i)},y^{(i)})$, using Lemma~\ref{lem:exp-gradient}, we can write

\begin{align}
{y^{(i)} h(\vec{x}^{(i)})} =  & \exp \left( \left\| \vec{x}^{(i)}  \right\|^2 \right) +\sum_{j \neq i} y^{(i)}y^{(j)} \exp \left( \left\| \frac{\vec{x}^{(i)} + \vec{x}^{(j)}}{2} \right\|^2 \right)\\
\geq& \exp \left( \left\| \vec{x}^{(i)}  \right\|^2 \right) 
- \sum_{\substack{j \neq i \\ y^{(i)} \neq y^{(j)}} }  \exp \left( \left\| \frac{\vec{x}^{(i)} + \vec{x}^{(j)}}{2} \right\|^2 \right) \\
\geq& \exp \left( \left\| \vec{x}^{(i)}  \right\|^2 \right) \times  \left(1 - \sum_{\substack{j \neq i \\ y^{(i)} \neq y^{(j)}} } \exp \left(  \frac{\|\vec{x}^{(i)} + \vec{x}^{(j)}\|^2 - 4 \| \vec{x}^{(i)}\|^2}{4} \right) \right).
\end{align}

% +\sum_{\substack{j \neq i \\ y^{(i)} = y^{(j)}} }  \exp \left( \left\| \frac{\vec{x}^{(i)} + \vec{x}^{(j)}}{2} \right\|^2 \right)\\

Now, for any  $j$ such that $y^{(j)} \neq y^{(i)}$, we have that % from  and Equation~\ref{eq:exp-dot-prod-bound} that:

\begin{align}
\|\vec{x}^{(i)}  + \vec{x}^{(j)}\|^2 - 4\| \vec{x}^{(i)} \|^2 &=  -3\|\vec{x}^{(i)} \|^2 + \|\vec{x}_1^{(j)} \|^2+ 2 \vec{x}_1^{(i)} \cdot \vec{x}_1^{(j)} \\
& \; \; \; + \underbrace{\|\vec{x}_2^{(j)}\|^2}_{Eq~\ref{eq:exp-norm-bound}}  + \underbrace{2 \vec{x}_2^{(i)} \cdot \vec{x}_2^{(j)}}_{Eq~\ref{eq:exp-dot-prod-bound}} \\
& \leq  -3\|\vec{u}\|^2 - 3\underbrace{\| \vec{x}_2^{(i)}\|^2}_{Eq~\ref{eq:exp-norm-bound}}  + \| \vec{u}\|^2 - 2\|\vec{u}\|^2  +c_3^2 N + \underbrace{\| \vec{x}_2^{(i)}\|}_{Eq~\ref{eq:exp-norm-bound}} \cdot 2 c_4\sqrt{2 \ln \frac{6m}{\delta}} \\
& \leq  -4 \| \vec{u} \|^2 -3c_2^2N +c_3^2 N + \underbrace{\sqrt{N} \cdot c_3 c_4\sqrt{2 \ln \frac{6m}{\delta}}}_{Eq~\ref{eq:d5}} \\
& \leq  -1 - \frac{45}{16}N + \frac{17}{16} N + \frac{1}{16}N  = -\frac{43}{16}N.
\end{align}

Plugging this back in the previous equation we have that 

\begin{align}
{y^{(i)} h(\vec{x}^{(i)})} \geq  & \geq \exp \left( \underbrace{\left\| \vec{x}^{(i)}  \right\|^2}_{Eq~\ref{eq:exp-norm-bound}} \right) \left(1-m \underbrace{\exp\left(-\frac{43}{64}N\right)}_{Eq~\ref{eq:d6}}\right) \\
& \geq \underbrace{\exp \left( \frac{15}{16}N\right)}_{Eq~\ref{eq:d6}} \cdot \frac{1}{2} \geq 1.
\end{align}

Hence, $\vec{x}^{(i)}$ is correctly classified by a margin of $1$ for every $i$.

Now consider any test data point $(\vec{z},y)$. Since $N \geq \frac{1}{c_2} \ln \frac{2}{\epsilon}$ (Equation~\ref{eq:d5}), we have that with probability at least $1-\epsilon/2$ over the draws of $\vec{z}_2$,  by Corollary~\ref{cor:chi}

\begin{align*}
c_2 \sqrt{N} \leq \|\vec{z}_2 \| \leq c_3 \sqrt{N}. \numberthis \label{eq:z-exp-norm-bound}
\end{align*}

Similarly, for each $i$, we have that with probability at least $1-\epsilon/2m$ over the draws of $\vec{z}$, the following holds good by Corollary~\ref{cor:gaussian}
\begin{align*}
|\vec{x}_2^{(i)} \cdot \vec{z}_2| \leq \| \vec{x}_2^{(i)} \| \cdot c_4 \sqrt{2\ln \frac{6m}{\epsilon}}.   \numberthis\label{eq:z-exp-dot-prod-bound}
\end{align*} 

Hence, the above holds over at least $1-\epsilon/2$ draws of $\vec{z}$, and by extension, both the above equations hold over at least $1-\epsilon$ draws of $\vec{z}$.

Now, for any $i$ such that $y^{(i)} = y$, we have that 

\begin{align}
\| \vec{x}^{(i)} + \vec{z}\|^2 =  &\|\vec{x}^{(i)}_1\|^2 + \|\vec{z}_1 \|^2 +  2 \vec{x}^{(i)}_1 \cdot \vec{z}_1 + \underbrace{\|\vec{x}^{(i)}_2\|^2}_{Eq~\ref{eq:exp-norm-bound}} + \underbrace{\|\vec{z}^{(i)}_2\|^2}_{Eq~\ref{eq:z-exp-norm-bound}}+  2 \underbrace{\vec{x}^{(i)}_2 \cdot \vec{z}_2}_{Eq~\ref{eq:z-exp-dot-prod-bound},~\ref{eq:exp-norm-bound}} \\ 
& \geq 4\|\vec{u}\|^2 + 2c_2^2 N - \underbrace{\sqrt{N} \cdot 2 c_3 c_4 \sqrt{2 \ln \frac{6m}{\epsilon}}}_{Eq~\ref{eq:d4}} \\
& \geq N + \frac{30}{16}N - \frac{1}{16} N = \frac{45}{16} N.
\end{align}

Similarly, for any $i$ such that $y^{(i)} \neq y$, we have that 

\begin{align}
\| \vec{x}^{(i)} + \vec{z}\|^2 =  &\|\vec{x}^{(i)}_1\|^2 + \|\vec{z}_1 \|^2 +  2 \vec{x}^{(i)}_1 \cdot \vec{z}_1 + \underbrace{\|\vec{x}^{(i)}_2\|^2}_{Eq~\ref{eq:exp-norm-bound}} + \underbrace{\|\vec{z}^{(i)}_2\|^2}_{Eq~\ref{eq:z-exp-norm-bound}} +  2 \underbrace{\vec{x}^{(i)}_2 \cdot \vec{z}_2}_{Eq~\ref{eq:exp-dot-prod-bound},~\ref{eq:exp-norm-bound}} \\ 
& \leq 2\|\vec{u}\|^2- 2\|\vec{u}\|^2 + 2c_3^2 N + \underbrace{\sqrt{N} \cdot 2 c_3 c_4 \sqrt{2 \ln \frac{6m}{\delta}}}_{Eq~\ref{eq:d5}} \\
& \leq \frac{34}{16}N - \frac{1}{16} N = \frac{33}{16}N.
\end{align}

Since from Equation~\ref{eq:class-count} we know  there exists at least one training sample with a given label, we have that

\begin{align}
{y^{(i)} h(\vec{x}^{(i)})} =  &\sum_{i: y^{(i)} = y}  \exp \left( \left\| \frac{\vec{x}^{(i)} + \vec{z}}{2} \right\|^2 \right) - \sum_{i: y^{(i)} \neq y}  \exp \left( \left\| \frac{\vec{x}^{(i)} + \vec{z}}{2} \right\|^2 \right) \\
&\geq \exp \left( \frac{45}{64}N \right) - m \exp \left( \frac{33}{64}N \right) \\
& \geq \exp \left( \frac{45}{64}N \right) \cdot \left(1 - m \underbrace{\exp \left( -\frac{12}{64}N \right)}_{Eq~\ref{eq:d6}} \right) \\
& \geq \underbrace{\exp \left( \frac{45}{64}N \right)}_{Eq~\ref{eq:d6}} \cdot \frac{1}{2} \geq 1.
\end{align}

Thus, at least $1-\epsilon$ of the test datapoints are classified correctly. 

\end{proof}

We next show that the uniform convergence bound is nearly vacuous. In order to do this, we create a set $S'$ from $S$ by negating all values but the noise vector. We then show that for every point in $S'$, the contribution from the corresponding point in $S$ dominates over the contribution from all other points. (This is because of how the non-negated noise vector in the point from $S'$ aligns adversarially with the noise vector from the corresponding point in $S$).
 As a result, the points in $S'$ are all labeled like in $S$, implying that $S'$ is completely misclassified. Then, similar to our previous arguments, we can show that uniform convergence is nearly vacuous.\\

\begin{lemma}
\label{lem:exp-unif-conv}
In the setup of Section~\ref{sec:uc-exp}, for any $\epsilon > 0$ and for any $\delta \leq 1/4$, and for the same lower bounds on $N$ and $m$ as in Theorem~\ref{thm:exp},  and for any $\gamma \geq 0$, we have that
\begin{equation}
\epsilon_{\text{unif-alg}}(m,\delta) \geq 1 - \epsilon_{\text{gen}}(m,\delta)
\end{equation}
for the $\scrL^{(\gamma)}$ loss.
\end{lemma}

\begin{proof}
Let $S'$ be a modified version of the training set where all values are negated except that of the noise vectors i.e., $S' = \{ ((-\vec{x}_1, \vec{x}_2),-y) \; | \;  ((\vec{x}_1,  \vec{x}_2),y) \in S \}$.  First we show that with probability at least $1-2\delta/3$ over the draws of $S$, $S'$ is completely misclassified. First, we have that with probability $1-2\delta/3$, Equations~\ref{eq:exp-norm-bound} and ~\ref{eq:exp-dot-prod-bound} hold good. Let $(\vec{x}_{\textrm{neg}}^{(i)},y_{\textrm{neg}}^{(i)})$ denote  the $i$th sample from $S'$.
Then,
 we have that 

\begin{align}
{y_{\textrm{neg}}^{(i)} h(\vec{x}_{\textrm{neg}}^{(i)})} = & -\exp \left( \left\| \frac{\vec{x}^{(i)} + \vec{x}_{\textrm{neg}}^{(i)} }{2} \right\|^2 \right) + \sum_{j \neq i} y_{\textrm{neg}}^{(i)}y^{(j)} \exp \left( \left\| \frac{\vec{x}_{\textrm{neg}}^{(i)} + \vec{x}^{(j)}}{2} \right\|^2 \right)\\
\leq& -\exp \left( \left\| \vec{x}_2^{(i)}  \right\|^2 \right) 
+ \sum_{\substack{j \neq i \\ y_{\textrm{neg}}^{(i)} = y^{(j)}} }  \exp \left( \left\| \frac{\vec{x}_{\textrm{neg}}^{(i)} + \vec{x}^{(j)}}{2} \right\|^2 \right) \\
\leq& \exp \left( \left\| \vec{x}_2^{(i)}  \right\|^2 \right) \times  \left(-1 + \sum_{\substack{j \neq i \\ y_{\textrm{neg}}^{(i)} = y^{(j)}} } \exp \left(  \frac{\|\vec{x}_{\textrm{neg}}^{(i)} + \vec{x}^{(j)}\|^2 - 4 \| \vec{x}_2^{(i)}\|^2}{4} \right) \right). \numberthis{\label{eq:exp-bad-margin}}
\end{align}

Now, consider  $j$ such that $y^{(j)} = y_{\textrm{neg}}^{(i)}$.  we have that % from  and Equation~\ref{eq:exp-dot-prod-bound} that:

\begin{align}
\|\vec{x}_{\textrm{neg}}^{(i)}  + \vec{x}^{(j)}\|^2 - 4\| \vec{x}_2^{(i)} \|^2 &=  \|\vec{x}_1^{(i)}\|^2 + \|\vec{x}_1^{(j)} \|^2 - 2 \vec{x}_1^{(i)} \cdot \vec{x}_1^{(j)}  - \underbrace{3\|\vec{x}_2^{(i)} \|^2}_{Eq~\ref{eq:exp-norm-bound}}  + \underbrace{\|\vec{x}_2^{(j)}\|^2}_{Eq~\ref{eq:exp-norm-bound}}  - \underbrace{2 \vec{x}_2^{(i)} \cdot \vec{x}_2^{(j)}}_{Eq~\ref{eq:exp-dot-prod-bound}} \\
& \leq  4\|\vec{u}\|^2 -  3c_2^2 N  +c_3 N + \underbrace{\| \vec{x}_2^{(i)}\|}_{Eq~\ref{eq:exp-norm-bound}} \cdot 2 c_4\sqrt{2 \ln \frac{6m}{\delta}} \\
& \leq  4 \| \vec{u} \|^2 -3c_2^2N +c_3^2 N + \underbrace{\sqrt{N} \cdot c_3 c_4\sqrt{2 \ln \frac{6m}{\delta}}}_{Eq~\ref{eq:d5}} \\
& \leq  N  - \frac{45}{16}N  + \frac{17}{16} N + \frac{1}{16}N = \frac{-11}{16} N. 
\end{align}

Plugging the above back in Equation~\ref{eq:exp-bad-margin}, we have
\begin{align}
\frac{y_{\textrm{neg}}^{(i)} h(\vec{x}_{\textrm{neg}}^{(i)})}{ \exp \left( \left\| \vec{x}_2^{(i)}  \right\|^2 \right) }  \leq  -1 + m \underbrace{\exp \left( -11D/64\right)}_{Eq~\ref{eq:d6}} \leq -1/2,
\end{align}
implying that $\vec{x}_{\textrm{neg}}^{(i)}$ is misclassified. This holds simultaneously for all $i$, implying that $S'$ is misclassified with high probability $1-2\delta/3$ over the draws of $S$. Furthermore, $S'$ has the same distribution as $\scrD^m$. Then, by the same argument as that of Lemma~\ref{lem:uc-unif}, we can prove our final claim.

\end{proof}

\section{Further Remarks.}

In this section, we make some clarifying remarks about our theoretical results.

\subsection{Nearly vacuous bounds for any $\gamma > 0$.}
\label{sec:any-gamma}
Typically, like in \citet{mohri12foundations,bartlett17spectral}, the 0-1 test error is upper bounded in terms of the $\scrL^{(\gamma)}$ test error for some optimal choice of $\gamma > 0$ (as it is easier to apply uniform convergence for $\gamma > 0$). From the main theoretical results (such as Theorem~\ref{thm:example}), it is obvious that for $\gamma \leq 1$, this approach would yield vacuous bounds. We now establish that this is the case even for $\gamma > 1$.

To help state this more clearly, for the scope of this particular section,
 let $\epsilon^{(\gamma)}_{\textrm{unif-alg}}, \epsilon^{(\gamma)}_{\textrm{gen}}$ denote the uniform convergence and generalization error for $\scrL^{(\gamma)}$ loss. Then, 
 the following inequality is used to derive a bound on the 0-1 error:
\begin{align*}
\scrL^{(0)}_{\scrD}(h_S) \leq \scrL^{(\gamma)}_{\scrD}(h_S) \leq \hat{\scrL}^{(\gamma)}_{S}(h_S) + \epsilon^{(\gamma)}_{\textrm{unif-alg}}(m,\delta) \numberthis\label{eq:margin-upper-bound}
\end{align*}
where the second inequality above holds with probability at least $1-\delta$ over the draws of $S$, while the first holds for all $S$ (which follows by definition of $\scrL^{(\gamma)}$ and $\scrL^{(0)}$).

To establish that uniform convergence is nearly vacuous in any setting of $\gamma$, we must show that the right hand side of the above bound is nearly vacuous for any choice of $\gamma \geq 0$ (despite the fact that $\scrL_{\scrD}^{(0)}(S) \leq \epsilon$). In our results, we explicitly showed this to be true for only small values of $\gamma$, by arguing that the second term in the R.H.S, namely $\epsilon^{(\gamma)}_{\textrm{unif-alg}}(m,\delta)$, is nearly vacuous.\\

Below, we show that the above bound is indeed nearly vacuous for any value of $\gamma$, when we have that $\epsilon_{\text{unif-alg}}^{(\gamma)}(m,\delta) \geq 1- \epsilon_{\text{gen}}^{(\gamma)}(m,\delta)$. Note that we established the relation $\epsilon_{\text{unif-alg}}^{(\gamma)}(m,\delta) \geq 1- \epsilon_{\text{gen}}^{(\gamma)}(m,\delta)$ to be true in all of our setups. \\

\begin{proposition}
Given that for all $\gamma \geq 0$,
$\epsilon_{\text{unif-alg}}^{(\gamma)}(m,\delta) \geq 1- \epsilon_{\text{gen}}^{(\gamma)}(m,\delta)$
then, we then have that for all $\gamma \geq 0$,
\begin{equation}
\mathbb{P}_{S \sim \scrD^m}\left[ \hat{\scrL}^{(\gamma)}_{S}(h_S) + \epsilon^{(\gamma)}_{\textrm{unif-alg}}(m,\delta) \geq \frac{1}{2} \right] > \delta
\end{equation}
or in other words, 
the guarantee from the right hand side of Equation~\ref{eq:margin-upper-bound} is nearly vacuous.
\end{proposition}

\begin{proof}
Assume on the contrary that for some choice of $\gamma$, we are able to show that with probability at least $1-\delta$ over the draws of $S$, the right hand side of Equation~\ref{eq:margin-upper-bound} is less than $1/2$. This means that $\epsilon^{(\gamma)}_{\textrm{unif-alg}}(m,\delta) < 1/2$. Furthermore, this also means that with probability at least $1-\delta$ over the draws of $S$, $\hat{\scrL}^{(\gamma)}_{S}(h_S) < 1/2$ and  $\scrL^{(\gamma)}_{\scrD}(h_S) < 1/2$ (which follows from the second inequality in Equation~\ref{eq:margin-upper-bound}). 

As a result, we have that with probability at least $1-\delta$, $\scrL^{(\gamma)}_{\scrD}(h_S) - \hat{\scrL}^{(\gamma)}_{S}(h_S) < 1/2$. In other words, $\epsilon_{\textrm{gen}}^{(\gamma)}(m,\delta) < 1/2$. 
Since we are given that $\epsilon_{\text{unif-alg}}^{(\gamma)}(m,\delta) \geq 1- \epsilon_{\text{gen}}^{(\gamma)}(m,\delta)$, by our upper bound on the generalization error, we have $\epsilon_{\text{unif-alg}}^{(\gamma)}(m,\delta) \geq 1/2$, which is a contradiction to our earlier inference that $\epsilon_{\text{unif-alg}}^{(\gamma)}(m,\delta) < 1/2$. Hence, our assumption is wrong. 

\end{proof}

\subsection{Generality of the failure of u.c. from Section~\ref{sec:uc-hypersphere}}
\label{sec:applicability}
Recall that in Section~\ref{sec:uc-hypersphere} we discussed a setup where two hyperspheres of radius $1$ and $1.1$ respectively are classified by a sufficiently overparameterized ReLU network. We saw that even when the number of training examples was as large as $65536$, we could project all of these examples on to the other corresponding hypersphere, to create a completely misclassified set $S'$. How well does this observation extend to other hyperparameter settings?\\

First, we note that in order to achieve full misclassification of $S'$, the network would have to be sufficiently overparameterized i.e., either the width or the input dimension must be larger. When the training set size $m$ is too large, one would observe that $S'$ is not as significantly misclassified as observed. (Note that on the other hand, increasing the parameter count would not hurt the generalization error. In fact it would improve it.)

Second, we note that our observation is sensitive to the choice of the difference in the radii between the hyperspheres (and potentially to other hyperparameters too). For example, when the outer sphere has radius $2$, SGD learns to classify these spheres perfectly, resulting in zero error on both test data and on $S'$.  As a result, our lower bound on $\epsilon_{\text{unif-alg}}$ would not hold in this setting. \\

However, here we sketch a (very) informal argument as to why there is reason to believe that our lower bound can still hold on a weaker notion of uniform convergence, a notion that is always applied in practice (in the main results we focus on a strong notion of uniform convergence as a negative result about it is more powerful). More concretely, in reality, uniform convergence is computed without much knowledge about the data distribution, save a few weakly informative assumptions such as those bounding its support. Such a uniform convergence bound is effectively computed uniformly in supremum over a class of distributions.

Going back to the hypersphere example, the intuition is that even when the radii of the spheres are far apart, and hence, the classification perfect, the decision boundary learned by the network could still be microscopically complex -- however these complexities are not exaggerated enough to misclassify $S'$. Now, for this given decision boundary, one would be able to construct an $S''$ which corresponds to projecting $S$ on two concentric hyperspheres that fall within these skews. Such an $S''$ would have a distribution that comes from some $\scrD'$ which, although not equal to $\scrD$, still obeys our assumptions about the underlying distribution. The uniform convergence bound which also holds for $\scrD'$ would thus have to be vacuous.

\subsection{On the dependence of $\epsilon_{\textrm{gen}}$ on $m$ in our examples.}
\label{sec:remark}
% In our particular example, the noise in the learning algorithm can be explained by the distribution of the input examples i.e., many of the features in the input points are noise, and this noise gets absorbed into the classifier. However, this is not critical for our analysis -- our analysis would apply even to a situation where  the noise in the learned classifier does not necessarily come from directly from the inputs, but from the way it is trained (say using SGD, to optimize a non-convex, non-smooth loss). That is, we could think of $\vec{w}$ being a noise vector picked independent of the training set.  The main property that we want from the noise is that it should not align adversarially with most of the test points and as long as the learning algorithm involves high-dimensional noise, this would reasonably hold.

As seen in the proof of Lemma~\ref{lem:gen}, the generalization error $\epsilon$ depends on dataset size $m$ and input dimensionality (and also the parameter count) $N$ as $\mathcal{O}(e^{-N/m})$ ignoring some constants in the exponent. Clearly, this error decreases with the parameter count $N$. % However, we note that this decrease would only be mild when $N \gg m$. For example, if we double $N$, the factor by which the generalization error would decrease would equal $e^{-N/m}/e^{-2N/m} = e^{D/m}$. 

On the other hand, one may also observe that this generalization error grows with the number of samples $m$, which might at first make this model seem inconsistent with our real world observations. However, we emphasize that this is a minor artefact of the simplifications in our setup, rather than a conceptual issue. With a small modification to our setup, we can make the generalization error decrease with $m$, mirroring our empirical observations. Specifically, in the current setup, we learn the true boundary along the first $K$ dimensions exactly. We can however modify it to a more standard learning setup where the boundary is not exactly recoverable and needs to be estimated from the examples. This would lead to an additional generalization error that scales as $\mathcal{O}(\sqrt{\frac{K}{m}})$ that is non-vacuous as long as $K \ll m$. Thus, the overall generalization error would be $\mathcal{O}(e^{-N/m} + \sqrt{\frac{K}{m}})$. \\

What about the overall dependence on $m$? Now, assume we have an overparameterization level of $N \gg m \ln (m/K)$, so that $e^{-N/m} \ll \sqrt{K/m}$. Hence, in the sufficiently overparameterized regime, the generalization error $\mathcal{O}(e^{-N/m})$ that comes from the noise we have modeled, pales in comparison with the generalization error that would stem from estimating the low-complexity boundary. Overall, as a function of $m$, the resulting error would behave like $\mathcal{O}(\sqrt{\frac{K}{m}})$ and hence show a decrease with increasing $m$ (as long the increase in $m$ is within the overparameterized regime). 

\subsection{Failure of hypothesis-dependent uniform convergence bounds.}

\label{sec:weight-dependence}
Often, uniform convergence bounds are written as a bound on the generalization error of a specific hypothesis rather than the algorithm. These bounds have an explicit dependence on the weights learned. As an example, a bound may be of the form that, with high probability over draws of training set $\tilde{S}$, for any hypothesis $h$ with weights $\vec{w}$,

\begin{equation}
 \scrL_{\scrD}(h) -
 \hat{\scrL}_{\tilde{S}}(h)  \leq \frac{\|\vec{w}\|_2}{\sqrt{m}}.
\end{equation}

Below we argue why even these kinds of hypothesis-dependent bounds fail in our setting. \\

We can informally define the tightest hypothesis-dependent uniform convergence bound as follows, in a manner similar to Definition~\ref{def:unif-alg} of the tightest uniform convergence bound. 
Recall that we first pick a set of datasets $\mathcal{S}_{\delta}$ such that $\mathbb{P}_{\tilde{S} \sim \scrD^m}[\tilde{S} \notin \mathcal{S}_{\delta}] \leq \delta$. Then, for all $\tilde{S} \in S_{\mathcal{\delta}}$, we denote
the upper bound on the generalization gap of $h_{\tilde{S}}$ by $\epsilon_{\text{unif-alg}}(h_{\tilde{S}},m,\delta)$, where:

\begin{equation}
\epsilon_{\text{unif-alg}}(h_{\tilde{S}},m,\delta) := \sup_{\tilde{S} \in \mathcal{S}_{\delta}} |\scrL_D(h_{\tilde{S}}) - \hat{\scrL}_{S}(h_{\tilde{S}})|.
\end{equation}

In other words, the tightest upper bound here corresponds to the difference between the test and empirical error of the specific hypothesis $h_{\tilde{S}}$ but computed across nearly all datasets $S$ in $\mathcal{S}_{\delta}$.

To show failure of the above bound, recall from all our other proofs of failure of uniform convergence, we have that for at least $1-O(\delta)$ draws of the sample set ${\tilde{S}}$, four key conditions are satisfied: (i) ${\tilde{S}} \in \mathcal{S}_{\delta}$, (ii) the corresponding bad dataset  ${\tilde{S}}' \in \mathcal{S}_{\delta}$, (iii) the error on the bad set $\hat{\scrL}_{{\tilde{S}}'}(h_{\tilde{S}})=1$ and (iv) the test error $\scrL_\scrD(h_{\tilde{S}}) \leq \epsilon_{\text{gen}}(m,\delta)$.  For all such ${\tilde{S}}$, in the definition of $\epsilon_{\text{unif-alg}}(h_{\tilde{S}},m,\delta)$, let us set $S$ to be ${\tilde{S}}'$. Then, we would get $\epsilon_{\text{unif-alg}}(h_{\tilde{S}},m,\delta) \geq |\scrL_D(h_{\tilde{S}}) - \hat{\scrL}_{{\tilde{S}}'}(h_{\tilde{S}})| \geq 1-\epsilon_{\text{gen}}(m,\delta)$. In other words, with probability at least $1-O(\delta)$ over the draw of the training set, even a hypothesis-specific generalization bound fails to explain generalization of the corresponding hypothesis.

\subsection{Learnability and Uniform Convergence}

\label{sec:learnability}

Below, we provide a detailed discussion on learnability, uniform convergence and generalization. Specifically, we argue why the fact that uniform convergence is necessary for learnability does not preclude the fact that uniform convergence maybe unable to {\em explain} generalization of a particular algorithm for a particular distribution.

We first recall the notion of learnability. First, formally, a binary classification problem consists of a hypothesis class $\mathcal{H}$ and an instance space $\mathcal{X} \times \{-1,1\}$. %The loss of $h \in \mathcal{H}$ at an instance $(x,y)$ is $\scrL(h(x),y) = \mathbf{1}[h(x) \neq y]$. 
The problem is said to be {\em learnable} if there exists a learning rule $\mathcal{A}': \bigcup\limits_{m=1}^{\infty} \mathcal{Z}^m \to \mathcal{H}$ and a monotonically decreasing sequence $\epsilon_{\text{lnblty}}(m)$ such that $\epsilon_{\text{lnblty}}(m) \xrightarrow{m\to\infty} 0$ and
\begin{align*}
\forall \scrD' \;
\mathbb{E}_{S \sim \scrD'^m}\left[ 
\scrL^{(0)}_{\scrD'}(\mathcal{A}'(S)) -
\min_{h \in \mathcal{H}} \scrL^{(0)}_{\scrD'}(h)  \right] \leq \epsilon_{\text{lnblty}}(m).
\numberthis\label{eq:learnability}
\end{align*}\\

 \citet{vapnik71uniform} showed that finite VC dimension  of the hypothesis class is  necessary and sufficient for {learnability} in binary classification problems. As \citet{shwartz10learnability} note, since finite VC dimension is equivalent to uniform convergence, it can thus be concluded that uniform convergence is necessary and sufficient for learnability binary classification problems.

However, learnability is a strong notion that does not necessarily have to hold for a particular learning algorithm to generalize well for a particular underlying distribution. Roughly speaking, this is because learnability evaluates the algorithm under all possible distributions, including many complex distributions; while a learning algorithm may generalize well for a particular distribution under a given hypothesis class, it may fail to do so on more complex distributions under the same hypothesis class.\\

For more intuition, we present a more concrete but informal argument below. However, this argument is technically redundant because learnability is equivalent to uniform convergence for binary classification, and since we established the lack of necessity of uniform convergence, we effectively established the same for learnability too. However, we still provide the following informal argument as it provides a different insight into why learnability and uniform convergence are not necessary to explain generalization.

Our goal is to establish that in the set up of Section~\ref{sec:setup}, even if we considered the binary classification problem corresponding to $\mathcal{H}_{\delta}$ (the class consisting of only those hypotheses explored by the algorithm $\mathcal{A}$ under a distribution $\scrD$), the corresponding binary classification problem is not learnable i.e., Equation~\ref{eq:learnability} does not hold when we plug in $\mathcal{H}_{\delta}$ in place of $\mathcal{H}$. \\

First consider distributions of the following form that is more complex than the linearly separable $\scrD$: for any dataset $S'$, let $\scrD_{S'}$ be the distribution that has half its mass on the part of the linearly separable distribution $\scrD$ excluding $S'$,
and half its mass on the distribution that is uniformly distributed over $S'$. Now let $S'$ be a random dataset drawn from $\scrD$ {\em but with all its labels flipped}; consider the corresponding complex distribution $\scrD_{S'}$.

%Let $\mathcal{H}_{\delta}$ correspond to any choice of a small class of hypothesis that our algorithm picks from with high probability over draws of $S$ from the simple distribution $\scrD$. 
We first show that there exists $h \in \mathcal{H}_{\delta}$ that fits this distribution well.
Now, for most draws of the ``wrongly'' labeled $S'$, we can show that the hypothesis $h$ for which $\vec{w}_1 = 2 \cdot \vec{u}$ and $\vec{w}_2 = \sum_{(x,y) \in S'} y \cdot \vec{x}_2$ fits the ``wrong'' labels of $S'$ perfectly; this is because, just as argued in Lemma~\ref{lem:uc-unif}, $\vec{w}_2$ dominates the output on all these inputs, although $\vec{w}_1$ would be aligned incorrectly with these inputs. Furthermore, since $\vec{w}_2$ does not align with most inputs from $\scrD$, by an argument similar to Lemma~\ref{lem:gen}, we can also show that this hypothesis has at most $\epsilon $ error on $\scrD$, and that this hypothesis belongs to $\mathcal{H}_{\delta}$.  Overall this means that, w.h.p over the choice of $S'$, there exists a hypothesis $h \in \mathcal{H}_{\delta}$ for which the error on the complex distribution $\scrD_{S'}$ is at most $\epsilon/2$ i.e., 

\begin{equation}
\min_{h \in \mathcal{H}} \mathbb{E}_{(x,y) \sim \scrD_{S'}}[\scrL(h(x),y)]  \leq \epsilon/2.
\end{equation}

On the other hand, let $\mathcal{A}'$ be any learning rule which outputs a hypothesis given $S \sim \scrD_{S'}$. With high probability over the draws of $S \sim \scrD_{S'}$, only at most, say $3/4$th of $S$ (i.e., $0.75 m$ examples) will be sampled from  $S'$ (and the rest from $\scrD$). Since the learning rule which has access only to $S$, has not seen at least a quarter of $S'$, with high probability over the random draws of $S'$, the learning rule will fail to classify roughly half of the unseen 
examples from $S'$ correctly (which would be about $(m/4) \cdot 1/2 = m/8$). Then, the error on $\scrD_{S'}$ will be at least $1/16$.  From the above arguments, we have that $\epsilon_{\text{learnability}}(m) \geq 1/16 - \epsilon/2$, which is a non-negligible constant that is independent of $m$.

\subsection{Do our example setups suffer from pseudo-overfitting?}

Before we wrap up this section, we discuss a question brought up by an anonymous reviewer, which we believe is worth addressing. Recall that in Section~\ref{sec:uc-linear} and Section~\ref{sec:uc-hypersphere}, we presented a linear and hypersphere classification task where we showed that uniform convergence provably fails. In light of the above discussion, one may be tempted to ask: do these two models fail to obey uniform convergence because of pseudo-overfitting (the phenomenon described in Section~\ref{app:pseudo-overfit})? 

The answer to this is that our proof for failure of uniform convergence in both these examples did not rely on any kind of pseudo-overfitting -- had our proof relied on it, then we would have been able to show failure of only specific kinds of uniform convergence bounds (as discussed above). More formally, pseudo-overfitting in itself does not imply the lower bounds on $\epsilon_{{\textrm{\tiny unif-alg}}}$ that we have shown in these settings. \\

One may still be curious to understand the level of pseudo-overfitting in these examples, to get a sense of the similarity of this scenario with that of the MNIST setup. To this end, we note that our linear setup does indeed suffer from significant pseudo-overfitting -- the classifier's output does indeed have bumps around each training point (which can be concluded from our proof). 

\begin{figure}
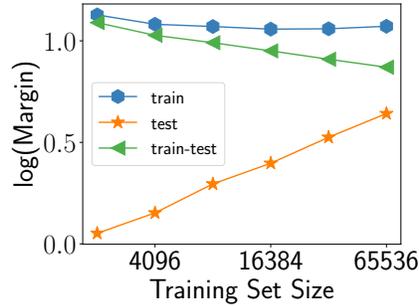

        \centering
        \adjincludegraphics[width=0.35\textwidth,trim={0 {0} 0 0},clip,,valign=t]{UniformConvergenceImages/margin_vs_m_hypersphere} %
        \caption{In the hypersphere example of Section~\ref{sec:uc-hypersphere}, we plot the average margin of the network on the train and test data, and the difference between the two. We observe the train and test margins do converge to each other.}
                \label{fig:margin-convergence-hypersphere}
\end{figure}

In the case of the hypersphere example, we present Figure~\ref{fig:margin-convergence-hypersphere}, where we plot of the average margins in this setup like in Figure~\ref{fig:margin-convergence}. Here, we observe that, the mean margins on the test data (orange line) and on training data (blue line) {do converge to each other} with more training data size $m$ i.e., {the gap in the mean test and training margins (green line) {\em does} decrease with $m$}. Thus our setup exhibits a behavior similar to deep networks on MNIST in Figure~\ref{fig:margin-convergence}. 
As noted in our earlier discussion, since the rate of decrease of the mean margin gap in MNIST is not as large as the decrease in test error itself,  there should be ``a small amount'' of psuedo-overfitting in MNIST. The same holds in this setting, although, here we observe an even milder decrease, implying a larger amount of pseudo-overfitting. Nevertheless, we emphasize that, our proof shows that uniform convergence cannot capture even this decrease with $m$.

To conclude, pseudo-overfitting is certainly a phenomenon worth exploring better; however, our examples elucidate that there is a phenomenon beyond pseudo-overfitting that is at play in deep learning.

\section[Deterministic PAC-Bayes bounds]{Deterministic PAC-Bayes bounds are two-sided uniform convergence bounds}
\label{sec:uc-pac-bayes}

By definition, VC-dimension, Rademacher complexity and other covering number based bounds are known to upper bound the term $\epsilon_{\text{unif-alg}}$ and therefore our negative result immediately applies to all these bounds. However, it may not be immediately clear if bounds derived through the PAC-Bayesian approach fall under this category too. In this discussion, we show that existing deterministic PAC-Bayes based bounds are in fact two-sided in that they are lower bounded by $\epsilon_{\text{unif-alg}}$ too. \\

Recall from Section~\ref{sec:prelim-pac-bayes} that for a given prior distribution $P$ over the  parameters, a PAC-Bayesian bound is of the following form: with high probability $1-\delta$ over the draws of the data $S$, we have that {\em for all distributions} $Q$ over the hypotheses space:
\begin{align*}
& KL\left( \left. \mathbb{E}_{\tilde{h} \sim Q}[\hat{\scrL}_{S}(\tilde{h}) ]\right \| \mathbb{E}_{\tilde{h} \sim Q}[\scrL_{\scrD}(\tilde{h}) ] \right) \leq  \underbrace{\frac{KL(Q \| P) + \ln \frac{2m}{\delta}}{m-1}}_{:= \epsilon_{\textrm{pb}}(P,Q,m,\delta)}. \numberthis \label{eq:pb}
\end{align*}

Note that here for any $a,b \in [0,1]$, $KL(a\| b) = a \ln \frac{a}{b} + (1-a) \ln \frac{1-a}{1-b}$. Since the precise form of the PAC-Bayesian bound on the right hand side is not relevant for the rest of the discussion, we will concisely refer to it as $\epsilon_{\textrm{pb}}(P,Q,m,\delta)$. What is of interest to us is the fact that the above bound holds for all $Q$ for most draws of $S$ and that the KL-divergence on the right-hand side is in itself two-sided, in some sense.

Typically, the above  bound is simplified to derive the following one-sided bound on the difference between the expected and empirical errors of a stochastic network (see \cite{mcallester03simplified} for example):
\begin{align*}
 &\mathbb{E}_{\tilde{h} \sim Q}[\scrL_{\scrD}(\tilde{h}) ] -  \mathbb{E}_{\tilde{h} \sim Q}[\hat{\scrL}_{S}(\tilde{h}) ] \leq \sqrt{2\epsilon_{\textrm{pb}}(P,Q,m,\delta)} + 2\epsilon_{\textrm{pb}}(P,Q,m,\delta).
 \numberthis \label{eq:stochastic-pac-bayes}
\end{align*}

This bound is then manipulated in different ways to obtain bounds on the deterministic network. In the rest of this discussion, we focus on the two major such derandomizing techniques and argue that both these techniques boil down to two-sided convergence. While, we do not formally establish that there may exist other techniques which ensure that the resulting deterministic bound is strictly one-sided, we suspect that no such techniques may exist. This is because the KL-divergence bound in Equation~\ref{eq:pb} is in itself two-sided in the sense that for the right hand side bound to be small, both the stochastic test and train errors must be close to each other; it is not sufficient if the stochastic test error is smaller than the stochastic train error.

\subsection{Deterministic PAC-Bayesian Bounds of Type A}
To derive a deterministic generalization bound, one approach is to add extra terms that account for the perturbation in the loss of the network. For example, this was the style of derandomization that we took in Chapter~\ref{chap:derandomized} \citep{nagarajan18deterministic} and also in other papers \citep{neyshabur17exploring,mcallester03simplified,}. That is, define:
\begin{align}
\Delta(h,Q,\scrD) &= | \scrL_{\scrD}(h) -  \mathbb{E}_{\tilde{h} \sim Q}[\scrL_{\scrD}(\tilde{h}) ]|, \\
\Delta(h,Q,S) &= \left|  \hat{\scrL}_{S}(h) -  \mathbb{E}_{\tilde{h} \sim Q}[\hat{\scrL}_{S}(\tilde{h}) ]\right|. 
\end{align}
Then, one can get a deterministic upper bound as:
\begin{align}
  &\scrL_{\scrD}(h)-   \hat{\scrL}_{S}(h) \leq\sqrt{2\epsilon_{\text{pb}}(P,Q_h,m,\delta)} +  2\epsilon_{\text{pb}}(P,Q_h,m,\delta)  + { \Delta(h,Q,\scrD) + \Delta(h,Q,S)}.
\end{align}

Note that while applying this technique, for any hypothesis $h$, one picks a posterior $Q_h$ specific to that hypothesis (typically, centered at that hypothesis). \\

We formally define the deterministic bound resulting from this technique below. We consider the algorithm-dependent version and furthermore, we consider a bound that results from the best possible choice of $Q_h$ for all $h$. 
We define this deterministic bound in the format of $\epsilon_{\text{unif-alg}}$ as follows: \\

\begin{definition} 
The distribution-dependent, algorithm-dependent, deterministic PAC-Bayesian bound of (the hypothesis class $\mathcal{H}$, algorithm $\mathcal{A}$)-pair with respect to $\scrL$ is defined to be the smallest value $\epsilon_{\text{pb-det-A}}(m, \delta)$ such that the following holds:
\begin{enumerate}
\item there exists a set of $m$-sized samples $\mathcal{S}_{\delta} \subseteq (\mathcal{X} \times \{-1, +1 \})^m$ for which:
\begin{equation}
\mathbb{P}_{S \sim \scrD^m} [S \notin \mathcal{S}_{\delta}]  \leq \delta,
\end{equation} 
\item and if we define $\mathcal{H}_{\delta} = \bigcup_{S \in \mathcal{S}_{\delta}}  \{ h_{S} \}$ to be the space of hypotheses explored only on these samples, then there must exist a prior $P$ and for each $h \in \mathcal{H}_{\delta}$, a distribution $Q_h$, such that uniform convergence must hold as follows:
\begin{align*}
& \sup_{S \in \mathcal{S}_{\delta}}\ \sup_{h \in \mathcal{H}_{\delta}}  \sqrt{2\epsilon_{\text{pb}}(P,Q_h,m,\delta)} +  2\epsilon_{\text{pb}}(P,Q_h,m,\delta)\\
& + \Delta(h,Q_h,\scrD) + \Delta(h,Q_h,S) < \epsilon_{\text{pb-det-A}}(m,\delta), 
\numberthis \label{eq:pb-det-A}
\end{align*}
%\begin{equation}
% \mathbb{E}_{\tilde{h} \sim Q_h}[ {\scrL}_{\scrD}(\tilde{h}) ] - \mathbb{E}_{\tilde{h} \sim Q_h}[ \hat{\scrL}_{S}(\tilde{h}) ] + \Delta(h,Q_h,\scrD) + \Delta(h,Q_h,S) < \epsilon_{\text{pb-det-A}}(m,\delta)  
%\end{equation}
as a result of which, by Equation~\ref{eq:stochastic-pac-bayes}, the following one-sided uniform convergence also holds:
\begin{equation}
\label{eq:pb-one-side}
\sup_{S \in \mathcal{S}_{\delta}}\sup_{h \in \mathcal{H}_{\delta}}  \scrL_{\scrD}(h)-   \hat{\scrL}_{S}(h) < \epsilon_{\text{pb-det-A}}(m,\delta).
\end{equation}
\end{enumerate}

\end{definition}

%TODO on delta being small

Now, recall that $\epsilon_{\text{unif-alg}}(m,\delta)$ is a two-sided bound, and in fact our main proof crucially depended on this fact in order to lower bound   $\epsilon_{\text{unif-alg}}(m,\delta)$. Hence, to extend our lower bound to $\epsilon_{\text{pb-det-A}}(m,\delta) $ we need to show that it is also two-sided in that it is lower bounded by 
 $\epsilon_{\text{unif-alg}}(m,\delta)$.  The following result establishes this:

\begin{theorem}
\label{thm:pb-det-A}
Let $\mathcal{A}$ be an algorithm such that on at least $1-\delta$ draws of the training dataset $S$, the algorithm outputs a hypothesis $h_S$ that has $\hat{\epsilon}(m,\delta)$ loss on the training data $S$. Then

\begin{align}
&e^{-3/2} \cdot \epsilon_{\text{unif-alg}}(m,3\delta) - (1-e^{-3/2})(\hat{\epsilon}(m,\delta) + \epsilon_{\text{gen}}(m,\delta)) \leq \epsilon_{\textrm{pb-det-A}} (m,\delta).
\end{align}
 
\end{theorem}

\begin{proof}

% Fix a $h, Q_h$ and $S$. 

First, by the definition of the generalization error, we know that with probability at least $1-\delta$ over the draws of $S$,
 \begin{equation}{\scrL}_{N}(h_S)\leq \hat{\scrL}_{S}(h_S)  + \epsilon_{\text{gen}}(m,\delta).  \end{equation}

Furthermore since the training loss it at most $\hat{\epsilon}(m,\delta)$ on at least $1-\delta$ draws %, and the difference between test and train error is at most $\epsilon_{\text{gen}}(m,\delta) $ on at least $1-\delta$ draws,
 we have that on at least $1-2\delta$ draws of the dataset,
 \begin{equation}{\scrL}_{N}(h_S)\leq \hat{\epsilon}(m,\delta) + \epsilon_{\text{gen}}(m,\delta).  \end{equation}
 Let $\mathcal{H}_{\delta}$
 and $\mathcal{S}_{\delta}$ be the subset of hypotheses and sample sets as in the definition of $\epsilon_{\text{pb-det-A}}$. Then, from the above,
 there exist $\mathcal{H}_{3\delta} \subseteq \mathcal{H}_{\delta}$ and $\mathcal{S}_{3\delta} \subseteq \mathcal{S}_{\delta}$ such that

\begin{equation}
\mathbb{P}_{S \sim \scrD^m} [S \notin \mathcal{S}_{3\delta}]  \leq 3\delta
\end{equation}
and $\mathcal{H}_{3\delta} = \bigcup_{S \in \mathcal{S}_{3\delta}}  \{ h_{S} \}$, and furthermore, 
 \begin{equation}\sup_{h\in\mathcal{H}_{3\delta}}{\scrL}_{\scrD}(h) \leq \hat{\epsilon}(m,\delta) + \epsilon_{\text{gen}}(m,\delta).
\end{equation}.

Using the above, and the definition of $\Delta$, we have for all $h\in\mathcal{H}_{3\delta}$, the following upper bound on its stochastic test error:

 \begin{align*} &  \mathbb{E}_{\tilde{h} \sim Q_h}[ {\scrL}_{\scrD}(\tilde{h}) ] \leq {\scrL}_{\scrD}(h)  +\Delta(h,Q_h,\scrD)  \leq \hat{\epsilon}(m,\delta) + \epsilon_{\text{gen}}(m,\delta) + \underbrace{\Delta(h,Q_h,\scrD)}_{\text{applying Equation}~\ref{eq:pb-det-A}}\\
 &  \leq {\hat{\epsilon}(m,\delta) + \epsilon_{\text{gen}}(m,\delta) + \epsilon_{\text{pb-det-A}}(m,\delta)}. \numberthis \label{eq:stochastic-test-ub}
\end{align*}

Now, for each pair of $h\in\mathcal{H}_{3\delta}$ and $S \in \mathcal{S}_{3\delta}$, we will bound its empirical error minus the expected error in terms of $\epsilon_{\text{pb-det-A}}(m,\delta)$. For convenience, let us denote by $a:=\mathbb{E}_{\tilde{h} \sim Q_h}[ \hat{\scrL}_{S}(\tilde{h}) ]$ and  $b := \mathbb{E}_{\tilde{h} \sim Q_h}[ {\scrL}_{\scrD}(\tilde{h}) ]$ (note that $a$ and $b$ are terms that depend on a hypothesis $h$ and a sample set $S$).  \\
 
We consider two cases. 
First, for some $h\in\mathcal{H}_{3\delta}$ and $S \in \mathcal{S}_{3\delta}$, consider the case that $e^{3/2} b > a$. Then, we have
\begin{align*}
  \hat{\scrL}_{S}(h)-  {\scrL}_{\scrD}(h)   \leq & 
a-b + \underbrace{\Delta(h,Q_h,\scrD) + \Delta(h,Q_h,S)}_{\text{applying Equation}~\ref{eq:pb-det-A}}\\
\leq & (e^{3/2} - 1)\underbrace{b}_{\text{apply Equation}~\ref{eq:stochastic-test-ub}} + \epsilon_{\text{pb-det-A}}(m,\delta)  \\
\leq & (e^{3/2} - 1) ({\hat{\epsilon}}(m,\delta) 
+ \epsilon_{\text{gen}}(m,\delta) + \epsilon_{\text{pb-det-A}}(m,\delta)) \\
&+ \epsilon_{\text{pb-det-A}}(m,\delta)  \\
\leq & (e^{3/2} - 1) ({\hat{\epsilon}}(m,\delta) + \epsilon_{\text{gen}}(m,\delta))+
 e^{3/2} \cdot \epsilon_{\text{pb-det-A}}(m,\delta). \numberthis \label{eq:case-1} \\
\end{align*}

Now consider the case where $a > e^{3/2} b$. This means that $(1-a) < (1-b)$. Then, if we consider the PAC-Bayesian bound of Equation~\ref{eq:pb},

\begin{equation}
\label{eq:pb-a-b}
 a \ln \frac{a}{b} + (1-a) \ln \frac{1-a}{1-b} \leq \epsilon_{\textrm{pb}}(P,Q_h,m,\delta),
\end{equation}
on the second term, 
we can apply the inequality
$\ln x \geq \frac{(x-1)(x+1)}{2x} = \frac{1}{2}\left( x - \frac{1}{x} \right)$ which holds for $x \in [0,1]$ to get:

\begin{align}
 (1-a) \ln \frac{1-a}{1-b} \geq\frac{1}{2} (1-a) \left( \frac{1-a}{1-b} - \frac{1-b}{1-a}\right) &=\left(  \frac{(b-a)(2-a-b)}{2(1-b)}\right) \\
&\geq - (a-b)\left(  \frac{(2-a-b)}{2(1-b)}\right) \\
& \geq  - (a-b)\left(  \frac{(2-b)}{2(1-b)}\right)\\
& \geq  - \frac{(a-b)}{2}\left(  \frac{1}{(1-b)} + 1\right).
\end{align}
Plugging this back in Equation~\ref{eq:pb-a-b}, we have,

\begin{align}
 \epsilon_{\textrm{pb}}(P,Q_h,m,\delta)  & \geq a \underbrace{\ln \frac{a}{b}}_{\geq 3/2}   - \frac{(a-b)}{2}\left(  \frac{1}{(1-b)} + 1\right) \\
& \geq \frac{2a(1-b)-(a-b)}{2(1-b)} + \frac{b}{2} \\
& \geq \frac{2a(1-b)-(a-b)}{2(1-b)} \geq \frac{a-2ab+b}{2(1-b)} \\
& \geq \frac{a-2ab+ab}{2(1-b)} \geq \frac{a}{2} \geq \frac{a-b}{2}\\
 & \geq \frac{1}{2} \left(  \hat{\scrL}_{S}(h)-  {\scrL}_{\scrD}(h)   - (\Delta(h, Q_h,\scrD) + \Delta(h,Q_h, S) )\right).\\
\end{align}

Rearranging, we get:
\begin{align}
 \hat{\scrL}_{S}(h)-  {\scrL}_{\scrD}(h) &\leq \underbrace{2 \epsilon_{\textrm{pb}}(P,Q_h,m,\delta)  +  (\Delta(h, Q_h,\scrD) + \Delta(h,Q_h, S) )}_{\text{Applying Equation}~\ref{eq:pb-det-A} } \\
 & \leq \epsilon_{\textrm{pb-det-A}}(m,\delta).  \numberthis \label{eq:case-2}
\end{align}

Since, for all $h \in \mathcal{H}_{3\delta}$ and $S \in \mathcal{S}_{3\delta}$, one of Equations~\ref{eq:case-1} and ~\ref{eq:case-2}
hold, we have that:

\begin{align}
&\frac{1}{e^{3/2}}\left(\sup_{h \in \mathcal{H}_{3\delta}} \sup_{S \in \mathcal{S}_{3\delta}}   \hat{\scrL}_{S}(h) -
 {\scrL}_{\scrD}(h)\right)  - \frac{(e^{3/2} - 1)}{e^{3/2}}  (\hat{\epsilon}(m,\delta) + \epsilon_{\text{gen}}(m,\delta)) \leq \epsilon_{\textrm{pb-det-A}}(m,\delta).
\end{align}

It follows from Equation~\ref{eq:pb-one-side} that the above bound holds good even after we take the absolute value of the first term in the left hand side. However, the absolute value is lower-bounded by $\epsilon_{\text{unif-alg}}(m,3\delta)$ (which follows from how $\epsilon_{\text{unif-alg}}(m,3\delta)$ is defined to be the smallest possible value over the choices of $\mathcal{H}_{3\delta}, \mathcal{S}_{3\delta}$).  % that the term on the RHS is lower bounded by $\epsilon_{\text{unif-alg}}(m,3\delta)$.

\end{proof}%

As a result of the above theorem, we can show   that $\epsilon_{\textrm{pb-det-A}}(m,\delta) = {\Omega}(1) - \mathcal{O}(\epsilon)$, thus establishing that, for sufficiently large  $N$, even though the generalization error would be negligibly small, the PAC-Bayes based bound would be as large as a constant. \\

\begin{corollary}
In the setup of Section~\ref{sec:setup}, for any $\epsilon,\delta > 0, \delta < 1/12$, when
 \begin{equation}N = \Omega\left(\max\left( m \ln \frac{3}{\delta}, m \ln\frac{1}{\epsilon}\right) \right),\end{equation}we have,
\begin{equation}e^{-3/2} \cdot (1-\epsilon) - (1-e^{-3/2})(\epsilon) \leq \epsilon_{\textrm{pb-det-A}} (m,\delta).\end{equation}
\end{corollary}

\begin{proof}
The fact that $\epsilon_{\text{gen}}(m,\delta) \leq \epsilon$ follows from  Theorem~\ref{thm:example}. 
Additionally, $\hat{\epsilon}(m,\delta)=0$ follows from the proof of Theorem~\ref{thm:example}. Now, as long as $3\delta < 1/4$, and $N$ is sufficiently large (i.e., in the lower bounds on $N$ in Theorem~\ref{thm:example}, if we replace $\delta$ by $3\delta$), we have from Theorem~\ref{thm:example}
 that $\epsilon_{\text{unif-alg}}(m,3\delta) > 1-\epsilon$. Plugging these in Theorem~\ref{thm:pb-det-A}, we get the result in the above corollary.
 \end{proof}

\subsection{Deterministic PAC-Bayesian Bounds of Type B}

In this section, we consider another standard approach to making PAC-Bayesian bounds deterministic \citep{neyshabur18pacbayes,langford02pacbayes}. Here, the idea is to pick for each $h$ a distribution $Q_h$ such that for all $\vec{x}$:
\begin{equation}
\scrL^{(0)}(h(\vec{x}),y) \leq \mathbb{E}_{\tilde{h} \sim Q_h} [\scrL'^{(\gamma/2)}(\tilde{h}(\vec{x}),y)] \leq  \scrL'^{(\gamma)}(h(\vec{x}),y),
\end{equation}

where 
\begin{equation}
\scrL'^{(\gamma)}(y,y') = \begin{cases} 0 & y \cdot y' \geq \gamma \\
1 & \text{else}.
\end{cases}
\end{equation}

Then, by applying the PAC-Bayesian bound of Equation~\ref{eq:stochastic-pac-bayes} for the loss $\scrL'_{\gamma/2}$, one can get a deterministic upper bound as follows, without having to introduce the extra $\Delta$ terms,

\begin{align}
 {\scrL}^{(0)}_{\scrD}(h)  -  \hat{\scrL}^{(\gamma)}_{S}(h)  \leq 
&  \mathbb{E}_{\tilde{h} \sim Q_h} [\mathcal{L'}^{(\gamma/2)}(\tilde{h})] -  \mathbb{E}_{\tilde{h} \sim Q_h} [\hat{\scrL}_{S}'^{(\gamma/2)}(\tilde{h})]   \\
& \leq \sqrt{2\epsilon_{\text{pb}}(P,Q_h,m,\delta)} +  2\epsilon_{\text{pb}}(P,Q_h,m,\delta).
 \end{align}

 Recall that we touched upon this style of derandomization in Section~\ref{sec:derandomized-advantage}.
We define this derandomization technique formally:

\begin{definition} 
The distribution-dependent, algorithm-dependent, deterministic PAC-Bayesian bound of (the hypothesis class $\mathcal{H}$, algorithm $\mathcal{A}$)-pair is defined to be the smallest value $\epsilon_{\text{pb-det-B}}(m, \delta)$ such that the following holds:
\begin{enumerate}
\item there exists a set of $m$-sized samples $\mathcal{S}_{\delta} \subseteq (\mathcal{X} \times \{-1, +1 \})^m$ for which:
\begin{equation}
\mathbb{P}_{S \sim \scrD^m} [S \notin \mathcal{S}_{\delta}]  \leq \delta.\end{equation}
\item and if we define $\mathcal{H}_{\delta} = \bigcup_{S \in \mathcal{S}_{\delta}}  \{ h_{S} \}$ to be the space of hypotheses explored only on these samples, then there must exist a prior $P$ and  for each $h$ a distribution $Q_h$, such that uniform convergence must hold as follows: for all $S \in \mathcal{S}_{\delta}$ and for all $h \in \mathcal{H}_{\delta}$,

\begin{align*}
 \sqrt{2\epsilon_{\text{pb}}(P,Q_h,m,\delta)} +  2\epsilon_{\text{pb}}(P,Q_h,m,\delta) < \epsilon_{\text{pb-det-B}}(m,\delta).
 \numberthis\label{eq:pb-det-B} 
\end{align*}

and for all $\vec{x}$: 
\begin{align*}
\scrL^{(0)}(h(\vec{x}),y) \leq \mathbb{E}_{\tilde{h} \sim Q_h} [\scrL'^{(\gamma/2)}(\tilde{h}(\vec{x}),y)] 
\leq  \scrL'^{(\gamma)}(h(\vec{x}),y)
\numberthis \label{eq:losses}
\end{align*}
as a result of which the following one-sided uniform convergence also holds:
\begin{align}
\label{eq:pb-one-side-B}
&\sup_{S \in \mathcal{S}_{\delta}}\sup_{h \in \mathcal{H}_{\delta}}{\scrL}^{(0)}_{\scrD}(h)  -  \hat{\scrL}_{S}'^{(\gamma)}(h)  < \epsilon_{\text{pb-det-B}}(m,\delta).
\end{align}
\end{enumerate}

\end{definition}

We can similarly show that $\epsilon_{\text{pb-det-B}}(m,\delta)$ is lower-bounded by the uniform convergence bound of $\epsilon_{\text{unif-alg}}$ too.

\begin{theorem}
\label{thm:pb-det-B}
Let $\mathcal{A}$ be an algorithm such that on at least $1-\delta$ draws of the training dataset $S$, the algorithm outputs a hypothesis $h_S$ such that the margin-based training loss can be bounded as:

\begin{equation}
\hat{\scrL}_{S}'^{(\gamma)}(h_S) \leq \hat{\epsilon}(m,\delta)
\end{equation}
and with high probability $1-\delta$ over the draws of $S$, the generalization error can be bounded as:
\begin{equation}
 \scrL'^{(\gamma)}_{\scrD}(h_{S})    - \scrL'^{(\gamma)}_{S}(h_{S}) \leq \epsilon_{\text{gen}}(m, \delta).
\end{equation}

Then there exists a set of samples $\mathcal{S}_{3\delta}$ of mass at least $1-3\delta$, and a corresponding set of hypothesis $\mathcal{H}_{3\delta}$ learned on these sample sets such that:
\begin{align}
&\left(\sup_{h \in \mathcal{H}_{3\delta}} \sup_{S \in \mathcal{S}_{3\delta}}  \scrL^{(0)}_{S}(h) -  \scrL'^{(\gamma)}_{\scrD}(h)\right)  - (e^{3/2}-1) (\hat{\epsilon}(m,\delta) + \epsilon_{\text{gen}}(m,\delta)) \leq \epsilon_{\textrm{pb-det-B}} (m,\delta).
\end{align} 
\end{theorem}

Note that the above statement is slightly different from how Theorem~\ref{thm:pb-det-A} is stated as it is not expressed in terms of $\epsilon_{\text{unif-alg}}$. In the corollary that follows the proof of this statement, we will see how it can be reduced in terms of $\epsilon_{\text{unif-alg}}$.

\begin{proof}
Most of the proof is similar to the proof of Theorem~\ref{thm:pb-det-A}. Like in the proof of Theorem~\ref{thm:pb-det-A}, we can argue that there exists $\mathcal{S}_{3\delta}$ and $\mathcal{H}_{3\delta}$
 for which the test error can be bounded as,

 \begin{equation}
 %\label{eq:stochastic-test-ub-B}
\mathbb{E}_{\tilde{h} \sim Q_h}[ \scrL'^{(\gamma/2)}_{\scrD}(\tilde{h})]
\leq \scrL'^{(\gamma)}_{\scrD}(h)
 \leq \hat{\epsilon}(m,\delta) + \epsilon_{\text{gen}}(m,\delta),
\end{equation}
where we have used $\epsilon_{\text{gen}}(m,\delta)$ to denote the generalization error of $\scrL'^{(\gamma)}$ and not the 0-1 error (we note that this is ambiguous notation, but we keep it this way for simplicity).

 For convenience, let us denote by $a:=\mathbb{E}_{\tilde{h} \sim Q_h}[ \hat{\scrL}'^{(\gamma/2)}_{S}(\tilde{h})]$ and  $b := \mathbb{E}_{\tilde{h} \sim Q_h}[ \scrL'^{(\gamma/2)}_{\scrD}(\tilde{h})]$. Again, let us consider, for some $h\in\mathcal{H}_{3\delta}$ and $S \in \mathcal{S}_{3\delta}$, the case that $e^{3/2} b \geq a$. Then, we have, using the above equation,

\begin{align}
\hat{\scrL}^{(0)}_{S}({h}) -  \hat{\scrL}^{(\gamma)}_{\scrD}({h})
\leq &  a-b \\
\leq & (e^{3/2} - 1) b \\
\leq & (e^{3/2} - 1) (\hat{\epsilon}(m,\delta) + \epsilon_{\text{gen}}(m,\delta))   \\
\leq & (e^{3/2} - 1) (\hat{\epsilon}(m,\delta) + \epsilon_{\text{gen}}(m,\delta) + \epsilon_{\text{pb-det-B}}(m,\delta)).  \label{eq:case-B-1} 
\end{align}

Now consider the case where $a  > e^{3/2} b$. Again, by similar arithmetic manipulation in the PAC-Bayesian bound of Equation~\ref{eq:stochastic-pac-bayes} applied on $\scrL'^{(\gamma/2)}$, we get,

\begin{align}
 \epsilon_{\textrm{pb}}(P,Q_h,m,\delta) & \geq 
a \underbrace{\ln \frac{a}{b}}_{\geq 3/2}   - \frac{(a-b)}{2}\left(  \frac{1}{(1-b)} + 1\right)\\
&  \geq \frac{a-b}{2}\\
 & \geq \frac{1}{2} \left( \scrL^{(0)}_{S}(h) -  \scrL'^{(\gamma)}_{\scrD}(h)  \right). 
\end{align}

Rearranging, we get:
\begin{align}
 & \scrL^{(0)}_{S}(h) -  \scrL'^{(\gamma)}_{\scrD}(h) \leq \underbrace{2 \epsilon_{\textrm{pb}}(P,Q_h,m,\delta)}_{\text{Applying Equation}~\ref{eq:pb-det-B}}\\
 & \leq  \epsilon_{\textrm{pb-det-B}}(m,\delta).   \label{eq:case-B-2}
\end{align}

Since, for all $h \in \mathcal{H}_{3\delta}$ and $S \in \mathcal{S}_{3\delta}$, one of Equations~\ref{eq:case-B-1} and ~\ref{eq:case-B-2}
hold, we have the claimed result.

\end{proof}%
Similarly,  as a result of the above theorem, we can show   that $\epsilon_{\textrm{pb-det-B}}(m,\delta) = {\Omega}(1) - \mathcal{O}(\epsilon)$, thus establishing that, for sufficiently large  $N$, even though the generalization error would be negligibly small, the PAC-Bayes based bound would be as large as a constant and hence cannot explain generalization. \\

\begin{corollary}
In the setup of Section~\ref{sec:setup}, for any $\epsilon,\delta > 0, \delta < 1/12$, when
 \begin{equation}N = \Omega\left(\max\left( m \ln \frac{3}{\delta}, m \ln\frac{1}{\epsilon}\right) \right),\end{equation} we have,
\begin{equation}1-(e^{3/2}-1)\epsilon \leq \epsilon_{\textrm{pb-det-B}} (m,\delta).\end{equation}
\end{corollary}

\begin{proof}
It follows from the proof of Theorem~\ref{thm:example} that $\hat{\epsilon}(m,\delta)=0$, since all training points are classified by a margin of $\gamma$ (see Equation~\ref{eq:train-margin}). Similarly, from Equation~\ref{eq:test-margin} in that proof,  since most test points are classified by a margin of $\gamma$,  $\epsilon_{\text{gen}}(m,\delta) \leq \epsilon$. Now, as long as $3\delta < 1/4$, and $N$ is sufficiently large (i.e., in the lower bounds on $N$ in Theorem~\ref{thm:example}, if we replace $\delta$ by $3\delta$), we will get that there exists $S \in \mathcal{S}_{3\delta}$ and $h \in \mathcal{H}_{3\delta}$ for which the empirical loss  $\scrL^{(0)}$ loss is $1$. Then, by Theorem~\ref{thm:pb-det-B}, we get the result in the above corollary.
 \end{proof}

\section{Conclusion}

Research on the generalization puzzle has been dominated by uniform convergence-based generalization bounds. Against this backdrop, we ask a critical, high level question: by pursuing this broad direction, is it possible to achieve the grand goal of a small generalization bound that shows appropriate dependence on the sample size, width, depth, label noise, and batch size? We cast doubt on this in the previous chapter, empirically showing that existing bounds can surprisingly increase with training set size for small batch sizes. We then presented example setups, including that of a ReLU neural network, for which uniform convergence provably fails to explain generalization, even after taking implicit bias into account.

% has tried to explain why deep networks generalize While uniform convergence bounds may provide partial intuition for why deep networks generalize well,  through empirical and theoretical evidences, we cast doubt on their potential to achieve the grand goal: a small bound that shows appropriate dependence on the sample size, width, depth, label noise, and batch size. First, we empirically showed that existing uniform convergence bounds can surprisingly increase with training set size for small batch sizes. We then presented example setups, including that of a ReLU neural network, for which uniform convergence provably fails to explain generalization even after taking implicit bias into account.  % We conjecture that deep network parameters consist of a low-complexity component and noise, where the latter hurts uniform convergence. Proving this conjecture would be an interesting direction for future work.
%We conjecture that strong bounds have been achieved on modified networks because these effectively strip away noise in the parameters that uniform convergence fails to explain away -- 

Future work in understanding implicit regularization in deep learning may be better guided with our knowledge of the sample-size-dependence in the weight norms. To understand generalization, it may also be promising to explore other learning-theoretic techniques like, say, algorithmic stability \citep{feldman18uniformly,hardt16stability,bousquet02stability,shwartz10learnability};  our linear setup might also inspire new tools. Overall,  through our work, we call for going beyond uniform convergence to fully explain generalization in deep learning.

% Overall, while the recent, growing line of works in 

 %through our work, we call for going beyond uniform convergence to fully explain generalization in deep learning.

%We envision the following: assume we manage to explicitly characterize the distribution of noise in deep networks; then, by applying uniform convergence on the low-complexity component (e.g., a compressed network), followed by standard tail bounds to argue that the noisy component does not affect the output, one would have a complete story for why deep networks generalize.

% \mytodo{Replicate this argum

 \graphicspath{{disagreement/}}

\chapter{Predicting Generalization via Disagreement on Unlabeled Data}

\label{chap:disagreement}

% We empirically show that the test error of deep networks can be estimated by simply training the same architecture on the same training set but with a different run of Stochastic Gradient Descent (SGD), and measuring the disagreement rate between the two networks on unlabeled test data. This builds on -- and is a stronger version of -- the observation in \citet{nakkiran20distributional}, which requires the second run to be on an altogether fresh training set.
% We further theoretically show that this peculiar phenomenon arises from the \emph{well-calibrated} nature of \emph{ensembles} of SGD-trained models. This finding not only provides a simple empirical measure to directly predict the test error using unlabeled test data, but also establishes a new conceptual connection between generalization and calibration. 

\section{Introduction}

In the last couple of chapters we took a pessimistic view of the existing {\em de facto} technique for deriving generalization bounds via uniform convergence. In this chapter and the next, we will take a promising approach towards generalization from a radically different lens. In particular, rather than trying to {\em explain} generalization, we will provide a surprisingly simple empirical technique to accurately {\em predict} generalization. Furthermore, our technique will leverage {\em fresh unlabeled data} that was not used during training\footnote{In the conclusion chapter however, we will provide a defense of why using unlabeled data to predict generalization, need not necessarily be a bad idea even from an ``explanatory theory'' point of view.}.  We will then provide a theory for why our technique can predict generalization so well in practice. Neither our technique nor our theory relies on tools like uniform convergence.

 \begin{figure}[t]
     \centering
     \begin{minipage}{.23\textwidth}
         \centering
        \includegraphics[width=\textwidth]{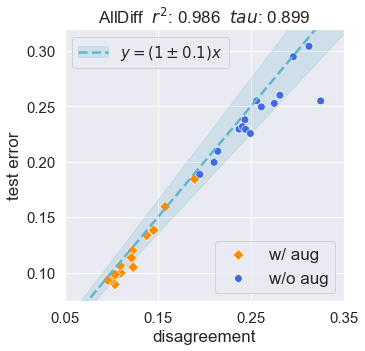}
        %  \caption{All sources of stochasticity are on.}
         \label{fig:disag-alldiff_scatter}
     \end{minipage}
    %  \hfill
     \begin{minipage}{.23\textwidth}
         \centering
                  \includegraphics[width=\textwidth]{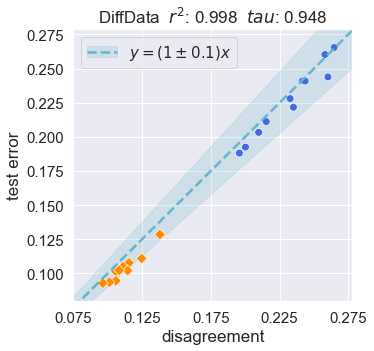}
        %  \caption{Pairs share the same dataset but have independent data order and initialization.}
         \label{fig:disag-sameinit_scatter}
     \end{minipage}
    %  \hfill
     \begin{minipage}{.23\textwidth}
         \centering                  
         \includegraphics[width=\textwidth]{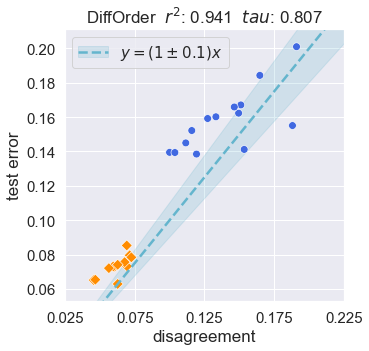}
        %  \caption{Pairs share see the same data in same order but have different initializations.}
         \label{fig:disag-samedata_scatter}
     \end{minipage}
    \begin{minipage}{.23\textwidth}
         \centering
            \includegraphics[width=\textwidth]{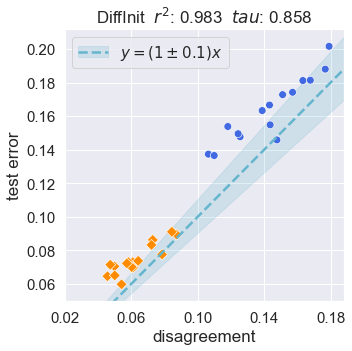}
        %  \caption{Pairs share see the same data in same order but have different initializations.}
         \label{fig:disag-sameorder_scatter}
     \end{minipage}
        \caption{Generalization-disagreement equality for ResNet18 on CIFAR-10.
        %\textbf{GDE on CIFAR-10:} The scatter plots of pair-wise model disagreement (x-axis) vs the test error (y-axis) of the different ResNet18 trained on CIFAR10. The dashed line is the diagonal line where disagreement equals the test error. Orange dots represent models that use data augmentation. The first two plots correspond to pairs of networks trained on independent datasets, and in the last two plots, on the same dataset. The details are described in Sec~\ref{sec:main-observation}.
        }
        \label{fig:disag-scatter}
\end{figure}

Our result builds on the following intriguing observation made in \citet{nakkiran20distributional}. Train two networks of the same architecture to zero training error on two independently drawn datasets $S_1$ and $S_2$ of the same size. Both networks would achieve a test error (or equivalently, a generalization gap) of about the same value, denoted by $\epsilon$. Now, take a fresh unlabeled dataset $U$ and measure the rate of disagreement of the predicted label between these two networks on $U$. Based on a triangle inequality, one can quickly surmise that this disagreement
 rate could lie anywhere between $0$ and $2\epsilon$. However, across various training set sizes and for various models like neural networks, kernel SVMs and decision trees,  \citet{nakkiran20distributional} (or \citetalias{nakkiran20distributional} in short) report that the disagreement rate not only linearly correlates with the test error $\epsilon$, but nearly {\em equals} $\epsilon$ (see first two plots in Fig~\ref{fig:disag-scatter}).
 What brings about this unusual equality?
 Resolving this open question from \citetalias{nakkiran20distributional} could help us identify fundamental patterns in how neural networks make errors. That might further shed insight into  generalization and other poorly understood empirical phenomena in deep learning. \\

In this work, we first identify a stronger observation. 
Consider two neural networks trained with the same hyperparameters and {\em the same dataset}, but with different random seeds (this could take the form e.g., of the data being presented in different random orders and/or by using a different random initialization of the network weights). We would expect the disagreement rate in this setting to be much smaller than in \citetalias{nakkiran20distributional}, since both models see the same data. Yet, this is not the case: we observe on the SVHN~\citep{netzer2011reading}, CIFAR-10/100~\citep{krizhevsky2009learning} datasets, and for variants of Residual Networks \citep{he2016deep} and Convolutional Networks, that the disagreement rate is still approximately equal to the test error  (see last two plots in Fig~\ref{fig:disag-scatter}), only slightly deviating from the behavior in \citetalias{nakkiran20distributional}.

In fact, while \citetalias{nakkiran20distributional} show that the disagreement rate captures significant changes in test error with varying training set sizes, we highlight a much stronger behavior:  the disagreement rate is able to capture even {\em minute variations in the test error under varying hyperparameters like width and depth}.  Furthermore, we show that under certain training conditions, these properties even hold on many kinds of \emph{out-of-distribution} data in the PACS dataset \citep{li17deeper} (albeit not on all kinds).\\

The above observations --- which we will refer to as the {\em Generalization Disagreement Equality} \footnote{\citet{nakkiran20distributional} refer to this as the {\em Agreement Property}, but we use the term Generalization Disagreement Equality to be more explicit and to avoid confusion regarding certain technical differences.}
 --- not only raise deeper conceptual questions but also crucially yields a practical benefit. In particular, our disagreement rate is a meaningful estimator of test accuracy as calculating it does not require a fresh labeled dataset (unlike the rate in \citetalias{nakkiran20distributional}) but rather only requires a fresh {\em unlabeled} dataset. In addition, unlike many other generalization measures \citep{jiang2018predicting,Jiang2020Fantastic,jiang2020neurips,natekar20representation} that merely correlate with the generalization gap or provide an overly conservative upper bound, this evidently gives us a direct estimate of the generalization error, \textit{without} requiring us to carefully compute proportionality constants and other multiplicative factors. 
Further, unlike these measures, our estimator shows promise even under certain kinds of distribution shift.

The results of this chapter have been previously published in \citet{jiang21assessing}.

\section{Related work}

\paragraph{Unconventional approaches to the generalization puzzle.} 
%Generalization in deep learning has attracted extensive interest because the behaviors of overparameterized neural network defy the conventional wisdom of machine learning. Conventionally, generalization in deep learning has been studied through the lens of PAC learning \citep{valiant1984theory}. Under this framework, generalization is roughly equivalent to bounding the size of the search space of a learning algorithm.  Representative works in this large area of research include \citet{neyshabur2014search,neyshabur17exploring, neyshabur18pacbayes, dziugaite2017computing, bartlett17spectral, nagarajan2019deterministic, nagarajan2019generalization,generalizationworkshop,natekar20representation}. 

Several works have in their own ways questioned whether the dominant approaches towards the generalization puzzle over the last few years are truly making progress \citep{belkin18understand,nagarajan19uniform,Jiang2020Fantastic,dziugaite2020search}.  Subsequently, recent works have proposed unconventional ways to derive generalization bounds \citep{negrea20defense,zhou20uc,garg21unlabeled}. Indeed, even our disagreement-based estimate of the test error marks a significant departure from complexity-based approaches to generalization bounds and measures. Of particular relevance to our work is \citet{garg21unlabeled}, who leverage unlabeled data to derive their bound. Their computation, however, requires modifying the original training set and then performing a careful early stopping during training. Thus, their bound is inapplicable to (and becomes vacuous for) interpolating networks. On the other hand, our estimate of the test error applies to the original training process without modifications. However, as we will see in the next chapter, our estimate comes with a guarantee only if we know {\em a priori} that the stochastic training procedure results in well-calibrated ensembles.

\paragraph{Empirical phenomena in deep learning.} Broadly, our work falls in the area of research on identifying \& understanding empirical phenomena in deep learning \citep{deepphenomena}, especially in the context of overparameterized models that interpolate (i.e., fit to zero training error). Some example phenomena include the generalization puzzle \citep{zhang2016understanding,neyshabur2014search}, double descent  \citep{belkin19reconciling,nakkiran20deep}, and simplicity bias \citep{kalimeris19sgg,arpit17closer}. 

As stated earlier, we build on \citetalias{nakkiran20distributional}'s empirical observation of the Generalization Disagreement Equality (GDE) in pairs of models trained on independently drawn datasets. Here we provide a detailed discussion of how our results are distinct from and/or complement  their other relevant findings. First, \citetalias{nakkiran20distributional} formally prove GDE for 1-Nearest Neighbor models. Their proof is however specific to 1-Nearest Neighbors, and relies on the two models being trained on two independent datasets. Our result on the other hand does not restrict the hypothesis class, the algorithm or its stochasticity.

Finally, \citetalias{nakkiran20distributional} in their Appendix D.7.1, do report connections to deep ensembles but in an independent context. In particular they show that ensembles of varied random seeds and ensembles of varied data both act as approximate pointwise conditional density estimators.
This phenomenon is however orthogonal to GDE. Furthermore, the GDE-related experiments in \citetalias{nakkiran20distributional} are all reported only on ensembles trained on different data. Hence, overall, their empirical results do not imply our GDE results in the context of ensembles trained on the same data.

\section{Main setup}
%%%%
% TODO: change for later
%%%%
\label{sec:main-setup}

In this section, we demonstrate on various datasets and architectures that the test error can be estimated directly by training two stochastic runs of SGD and measuring their disagreement on an unlabeled dataset. Importantly, we show that the disagreement rate can track even minute variations in the test error brought about by varying hyperparameters, besides larger variations brought about by varying training set size. Remarkably, this estimate of the test error does not require an independent labeled dataset.

\paragraph{Notations.} Let $h: \calX \to [K]$ denote a hypothesis from a hypothesis space $\calH$, where $[K]$ denotes the set of $K$ labels $\{0, 1, \hdots, K-1\}$ \footnote{Note that this notation is different from the rest of the thesis since the output of the hypothesis is not a real-valued vector. }. Let $\mathscr{D}$ be a distribution over $\calX \times [K]$. We will use $(X,Y)$ to denote the random variable with the distribution $\scrD$, and $(x,y)$ to denote specific values it can take. Let $\calA$ be a stochastic training algorithm that induces a distribution $\scrH_\calA$ over hypotheses in $\calH$. Let
%For convenience, let $h_k(x)$ denote $\mathbb{1}[h(x) = k]$.
 $h, h' \sim \scrH_\calA$ denote random hypotheses output by two independent runs of the training procedure.   We note that the stochasticity in $\calA$ could arise from any arbitrary source. This may arise from either the fact that each $h$ is trained on a random dataset drawn from the (test) distribution $\scrD$ or even a completely different training distribution $\scrD'$. The stochasticity could also arise from merely a different random initialization or data ordering. Next, we denote the test error and disagreement rate for hypotheses $h, h' \sim \scrH_\calA$ by:
 \begin{equation}
 \begin{split}
     \testerr_{\scrD}(h) & \coloneqq \Esub{\scrD}{\mathbb{1}[h(X) \neq Y]} \\
     \disag_{\scrD}(h, h') & \coloneqq \Esub{\scrD}{\mathbb{1}[h(X) \neq h'(X) ]}. 
     \end{split}
     \end{equation}
 %Note that the stochasticity could come from one or more of many different sources (such as the data, its ordering or the random initialization).
% Where needed, we will use the subscripts $\scrD$ or $\calA$ to denote probability/expectation over either one of these 
Let $\Tilde{h}$ denote the ``ensemble'' corresponding to $h \sim \scrH_\calA$. 
  In particular, define
  \begin{align}
  \Tilde{h}_k(x) \coloneqq \Esub{\scrH_\calA}{\mathbb{1}[h(x) = k]}
  \end{align}
  to be the probability value (between $[0,1]$) given by the ensemble $\Tilde{h}$ for the $k^{th}$ class. Note that the output of $\tilde{h}$ is {\em not} a one-hot value based on plurality vote.

\paragraph{Main experimental setup.} We report our main observations on variants of Residual Networks \citep{he2016deep}, convolutional neural networks \citep{lin2013network} and fully connected networks trained with Momentum SGD on  CIFAR-10/100~\citep{krizhevsky2009learning}, and SVHN~\citep{netzer2011reading}. Each variation of the ResNet has a unique hyperparameter configuration such as width or learning rate (discussed shortly) and all models are {interpolating}. 

For each hyperparameter setting, we train two copies of models which experience two independent draws from one or more sources of stochasticity, namely 
\begin{enumerate}
\item \textit{random initialization} (denoted by \texttt{Init})  and/or
\item \textit{ordering of a fixed training dataset} (\texttt{Order}) and/or 
\item \textit{different (disjoint) training data} (\texttt{{Data}}).
\end{enumerate}

We will use the term \texttt{Diff}
to denote whether a source of stochasticity is ``on''. 
%For simplicity, we will assume that all stochasticity are turned off unless stated otherwise.
For example, \texttt{DiffInit} means that the two models have different initialization but see the same data in the same order. In \texttt{DiffOrder}, models share the same initialization and see the same data, but in different orders. In \texttt{DiffData}, the models share the initialization, but see different data.
% In \texttt{SameOrder}, they share the data ordering (and naturally, the data too), but differ in initializations. 
% In \texttt{SameData}, they share the data, but differ in ordering and in initialization. 
In \texttt{AllDiff}, the two models differ in both data and in initialization\footnote{If the two copies of models differ in data, the training dataset is split into two disjoint halves to ensure no overlap.}.  The disagreement rate between a pair of models is computed as the proportion of the test data on which the (one-hot) predictions of the two models do not match. 
%%%%%%%%%
% Add table
%%%%%%%%%
\paragraph{Hyperparameter details.} The main architectures we used are ResNet18 with the following hyperparameter configurations:
\begin{enumerate}
    \item width multiplier: \{$1\times$, $2\times$\}
    \item initial learning rate: \{$0.1$, $0.05$\}
    \item weight decay: \{$0.0$, $0.0001$\}
    \item minibatch size: \{$100$, $200$\}
    \item data augmentation: \{No, Yes\}
\end{enumerate}
Width multiplier refers to how much wider the model is than the architecture presented in \citet{he2016deep} (i.e. every filter width is multiplied by the width multiplier). All models are trained with SGD with momentum of $0.9$. The learning rate decays $10\times$ every 50 epochs. The training stops when the training accuracy reaches $100\%$.

For Convolutional Neural Network experiments, we use architectures similar to Network-in-Network \citep{lin2013network}. On a high level, the architecture contains blocks of $3\times3$ convolution followed by two $1\times 1$ convolution (3 layers in total). Each block has the same width and the final layer is projected to output class number with another $1\times 1$ convolution followed by a global average pooling layer to yield the final logits. Other differences from the original implementation are that we do not use dropout and add batch normalization layer is added after every layer. The hyperparameters are:
\begin{enumerate}
    \item depth: \{7, 10, 13\}
    \item width: \{128, 256, 384\}
    \item weight decay: \{$0.0$, $0.001$\}
    \item minibatch size: \{$100$, $200$, 300\}
\end{enumerate}
All models are optimized with momentum of 0.9 and uses the same learning rate schedule as ResNet18.

For Fully Connected Networks, we use:
\begin{enumerate}
    \item depth: \{1,2,3,4\}
    \item width: \{128, 256, 384, 512\}
    \item weight decay: \{$0.0$, $0.001$\}
    \item minibatch size: \{$100$, $200$, 300\}
\end{enumerate}
All models are optimized with momentum of 0.9 and uses the same learning rate schedule as ResNet18.\\

\section{Main observation: disagreement tracks generalization}

\begin{figure}[t]
     \centering
     \begin{minipage}{.23\textwidth}
         \centering
        \includegraphics[width=\textwidth]{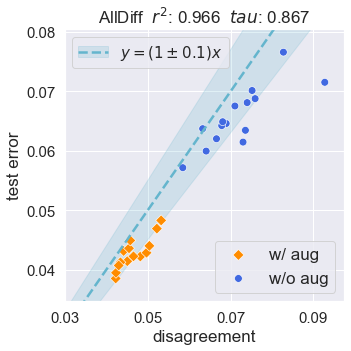}
        %  \caption{All sources of stochasticity are on.}
         \label{fig:disag-svhn_alldiff_scatter}
     \end{minipage}
    %  \hfill
     \begin{minipage}{.23\textwidth}
         \centering
                  \includegraphics[width=\textwidth]{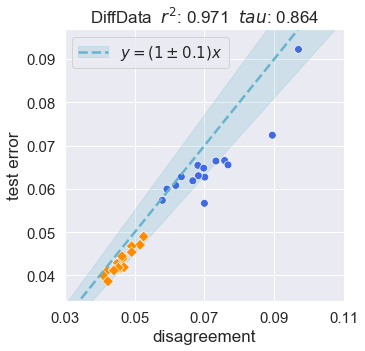}
        %  \caption{Pairs share the same dataset but have independent data order and initialization.}
         \label{fig:disag-svhn_sameinit_scatter}
     \end{minipage}
    %  \hfill
     \begin{minipage}{.23\textwidth}
         \centering                  
         \includegraphics[width=\textwidth]{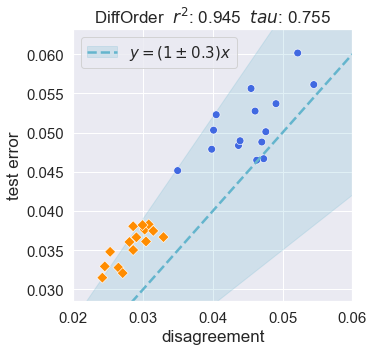}
        %  \caption{Pairs share see the same data in same order but have different initializations.}
         \label{fig:disag-svhn_samedata_scatter}
     \end{minipage}
    \begin{minipage}{.23\textwidth}
         \centering
            \includegraphics[width=\textwidth]{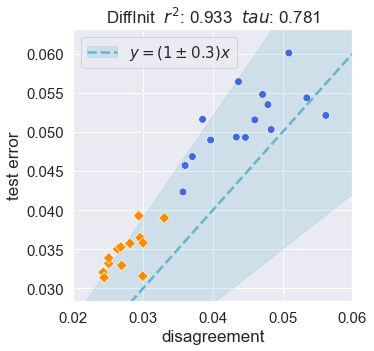}
        %  \caption{Pairs share see the same data in same order but have different initializations.}
         \label{fig:disag-svhn_sameorder_scatter}
     \end{minipage}
        \caption{\textbf{GDE on SVHN:} The scatter plots of pair-wise model disagreement (x-axis) vs the test error (y-axis) of the different ResNet18 trained on SVHN.}
        \label{fig:disag-svhn-scatter}\bigskip 
%\end{figure}
% \vaishnavh{The SVHN plots could also be zoomed in}
 %\begin{figure}[t!]
     \centering
     \begin{minipage}{.23\textwidth}
         \centering
        \includegraphics[width=\textwidth]{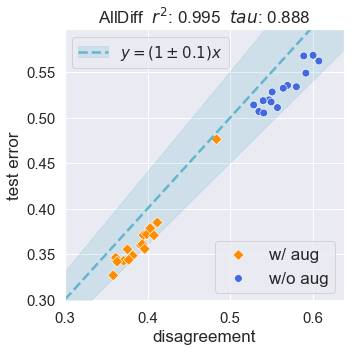}
     \end{minipage}
    %  \hfill
     \begin{minipage}{.23\textwidth}
        \centering
        \includegraphics[width=\textwidth]{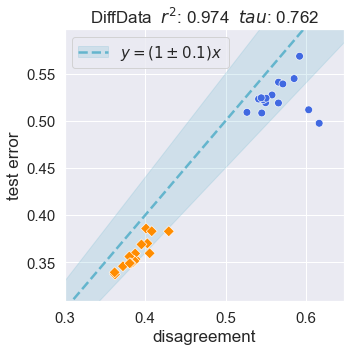}
     \end{minipage}
    %  \hfill
     \begin{minipage}{.23\textwidth}
         \centering                  
         \includegraphics[width=\textwidth]{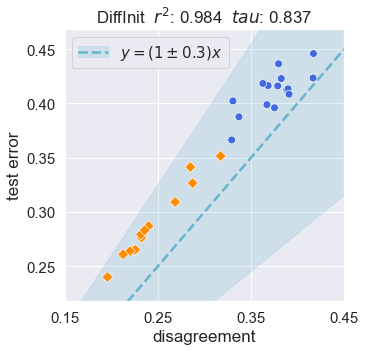}
     \end{minipage}
    \begin{minipage}{.23\textwidth}
        \centering
        \includegraphics[width=\textwidth]{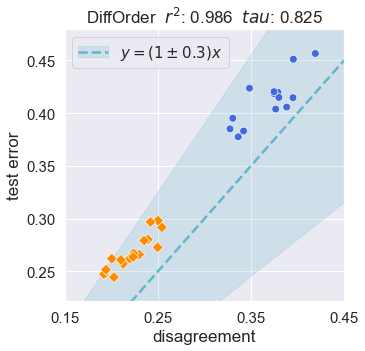}
     \end{minipage}
    \caption{\textbf{GDE on CIFAR-100:}  The scatter plots of pair-wise model disagreement (x-axis) vs the test error (y-axis) of the different ResNet18 trained on CIFAR100.}
    \label{fig:disag-cifar100-scatter}
    %\bigskip
\end{figure}

We illustrate test error ($y$) vs disagreement error ($x$) scatter plots for CIFAR-10, SVHN and CIFAR-100 in Figures \ref{fig:disag-scatter}, \ref{fig:disag-svhn-scatter} and \ref{fig:disag-cifar100-scatter} respectively (and for CNNs on CIFAR-10 in Fig~\ref{fig:disag-other-datasets}). Naively, we would expect these scatter plots to be arbitrarily distributed anywhere between $y = 0.5 x$ (if the errors of the two models are disjoint) and $x = 0$ (if the errors are identical).
However, in all these scatter plots, we observe that test error and disagreement error lie very close to the diagonal line $y=x$  across different sources of stochasticity, while only slightly deviating  in \texttt{DiffInit/Order}. In particular, in \texttt{AllDiff} and \texttt{DiffData}, the points typically lie between $y =x$ and $y = 0.9 x$ while in \texttt{DiffInit} and \texttt{DiffOrder}, the disagreement rate drops slightly (since the models are trained on the same data) and so the points typically lie between $y = x$ and $y=1.3x$. We further quantify correlation  via  the $R^2$ coefficient and Kendall's Ranking coefficient (\texttt{tau}) reported on top of each scatter plot. Both metrics range from $0$ to $1$ with $1$ being perfect correlation. Indeed, we observe that these quantities are high in all the settings. \\

The positive observations about \texttt{DiffInit} and \texttt{DiffOrder} are surprising for two reasons. First, when the second network is trained on the same dataset rather than a fresh dataset, we would expect its predictions to be largely aligned with the original network. Naturally, we would think that the disagreement rate would be negligible, and that the equality observed in \citetalias{nakkiran20distributional} would no longer hold. Furthermore, since we calculate the disagreement rate without using a fresh labeled dataset, we would expect disagreement to be much less predictive of test error when compared to \citetalias{nakkiran20distributional}.  Our observations defy both these expectations. \\

\begin{figure}[t!]
     \centering
     \begin{minipage}{.23\textwidth}
         \centering
        \includegraphics[width=\textwidth]{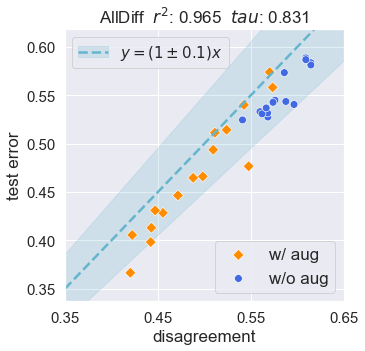}
     \end{minipage}
    %  \hfill
     \begin{minipage}{.23\textwidth}
        \centering
        \includegraphics[width=\textwidth]{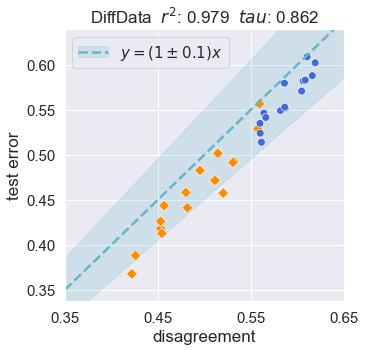}
     \end{minipage}
    %  \hfill
     \begin{minipage}{.23\textwidth}
         \centering                  
         \includegraphics[width=\textwidth]{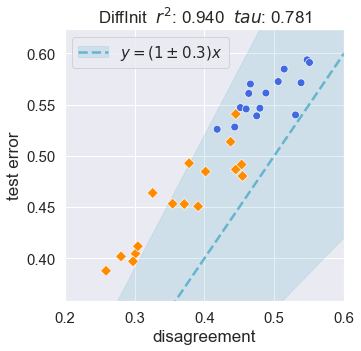}
     \end{minipage}
    \begin{minipage}{.23\textwidth}
        \centering
        \includegraphics[width=\textwidth]{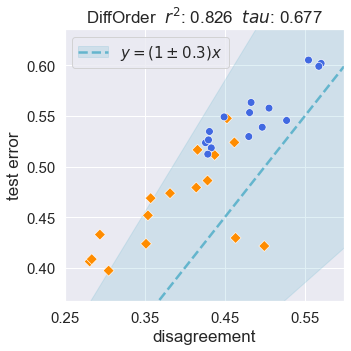}
     \end{minipage}
    \caption{\textbf{GDE on 2k subset of CIFAR-10:} The scatter plots of pair-wise model disagreement (x-axis) vs the test error (y-axis) of the different ResNet18 trained on CIFAR10 with only 2000 training points.}
    \label{fig:disag-cifar10-2k-scatter}
\end{figure}

There are a few more noteworthy aspects about these scatter plots. In the low data regime where the test error is high, we would expect the models to be much less well-behaved. However, consider the CIFAR-100 plots (Fig~\ref{fig:disag-cifar100-scatter}), and additionally, the plots in
Fig~\ref{fig:disag-cifar10-2k-scatter} where we train on CIFAR-10 with just $2000$ training points. In both these settings the network achieves an error as high as $0.5$ to $0.6$. Yet, we observe a 
behavior similar to the other settings (albeit with some deviations) -- the scatter plot lies in $y = (1\pm 0.1)x$ (for \texttt{AllDiff} and \texttt{DiffData}) and in $y=(1\pm 0.3)x$ (for \texttt{DiffInit/Order}), and the correlation metrics are high. Similar positive results were established in \citetalias{nakkiran20distributional} for \texttt{AllDiff} and \texttt{DiffData}. 

Finally, it is important to highlight that each scatter plot here corresponds to varying certain hyperparameters that cause only mild variations in the test error. Yet, the disagreement rate is able to capture those variations in the test error. This is a stronger version of the finding in \citetalias{nakkiran20distributional} that disagreement captures larger variations under varying dataset size.


\subsection{Other datasets and architectures}

\begin{figure}[h]
     \centering
     \begin{minipage}{0.23\textwidth}
         \centering
         \includegraphics[width=\textwidth]{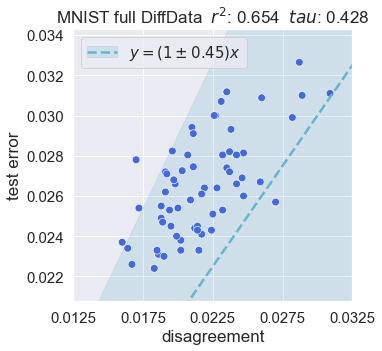}
         %\caption{MNIST FCN} %, 96 models
         \caption{}
        %  \label{fig:disag-cali1}
     \end{minipage}
    % \hspace{0.01\textwidth}
    %  \begin{minipage}{0.23\textwidth}
    %      \centering
    %     %  \includegraphics[width=\textwidth]{plots/fcn_mnist_calibration.png}
    %     \includegraphics[width=\textwidth]{plots/mnist_full_calibration.png}
    %      \caption{$ECE = 0.002, err(\tildeh)=0.055, \mu=0.049, \sigma=0.005$}
    %     %  \label{fig:disag-cali2}
    %  \end{minipage}
    %  \hspace{0.01\textwidth}
     \begin{minipage}{0.23\textwidth}
         \centering
         \includegraphics[width=\textwidth]{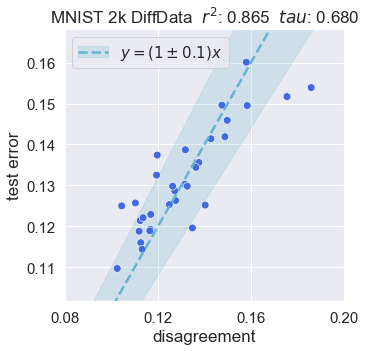}
         %\caption{MNIST FCN, 2k datapoints.} %32 models,
          \caption{}
          \label{fig:disag-scatter-mnist-2k}
     \end{minipage}
     \begin{minipage}{0.23\textwidth}
         \centering
         \includegraphics[width=\textwidth]{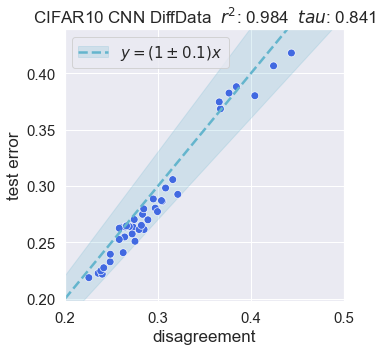}
         \caption{}
        %  \label{fig:disag-cali1}
     \end{minipage}
     \hspace{0.05\textwidth}
    %  \hspace{0.01\textwidth}
    %  \begin{minipage}{0.23\textwidth}
    %      \centering
    %      \includegraphics[width=\textwidth]{plots/mnist_2k_calibration.png}
    %      \caption{Calibration for FCN trained on MNIST with 2000 data points.}
    %  \end{minipage}
     
    \caption{Scatter plots for fully-connected networks on MNIST, 2k subset of MNIST and convolutional networks on CIFAR-10.}
        \label{fig:disag-other-datasets}
\end{figure}

In Fig~\ref{fig:disag-other-datasets}, we provide scatter plots for fully-connected networks (FCN) on MNIST, and convolutional networks (CNN) on CIFAR10. We observe that when trained on the whole MNIST dataset, there is larger deviation from the $x=y$ behavior (see left-most image). But when we reduce the dataset size to $2000$, we recover the GDE observation on MNIST. We observe that the CNN settings satisfies GDE too. 

\subsection{Effect of distribution shift and pre-training}

We also explore how these observations vary under the effect of distribution shift and for models where the initialization is pre-trained. We analyze this in the context of the PACS dataset \citep{li17deeper}, a popular domain generalization benchmark dataset consisting of data from four distinct distributions, \texttt{Photo} (\texttt{P} in short), \texttt{Art} (\texttt{A}), \texttt{Cartoon} (\texttt{C}) and \texttt{Sketch} (\texttt{S}).

\paragraph{Experimental setup.}
All domains consist of the same seven classes. On each of these domains, we train 45 pairs of ResNet50 with a linear layer on top, varying the random seeds (keeping hyperparameters constant). Both models in a pair are trained on the same $80\%$ of the data, along with data augmentation, and only differ in their initialization, data ordering and the augmentation on the data. We then evaluate the test error and disagreement rate of all pairs on each of the four domains.  We consider both randomly initialized models and ImageNet pre-trained models~\citep{deng2009imagenet}. For pre-trained models, only the linear layer is initialized differently between the two models in a pair. 

On all our experiments on the PACS dataset, we use ResNet50 (with Batch Normalization layers frozen and the final fully-connected layer removed) as our featurizer and one linear layer as our classifier. All our models are trained until 3000 steps after reaching 0.995 training accuracy with the following hyperparameter configurations:

\begin{enumerate}
    \item learning rate: $0.00005$
    \item weight decay: 0.0
    \item learning rate decay: None
    \item minibatch size: 100
    \item data augmentation: Yes
\end{enumerate}

\paragraph{Observations.}

\begin{figure}[t]
    \centering
    \includegraphics[scale=0.3]{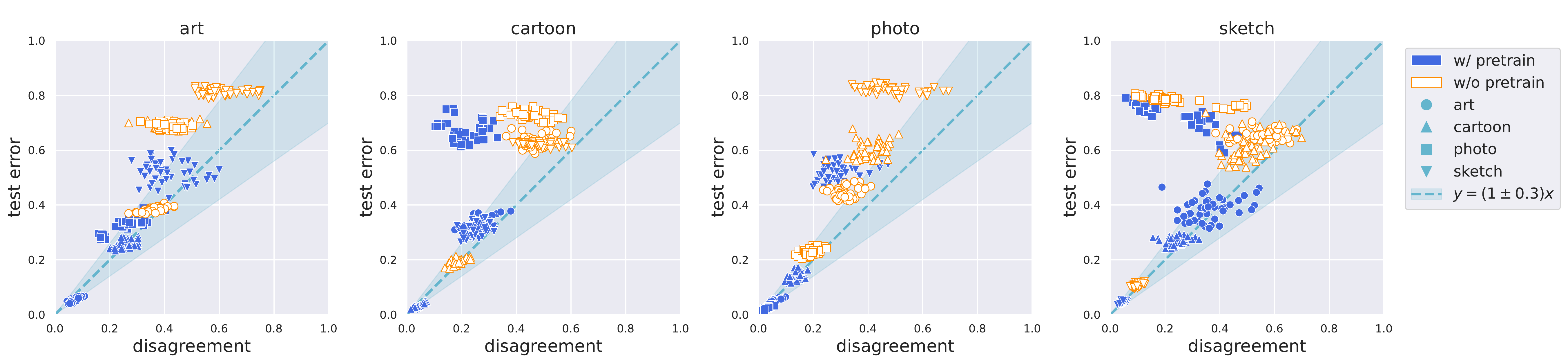}
    \caption{\textbf{GDE under distribution shift:}  The scatter plots of pair-wise model disagreement (x-axis) vs the test error (y-axis) of the different ResNet50 trained on PACS. Each plot corresponds to models evaluated on the domain specified in the title. The source/training domain is indicated by different marker shapes.
    }
    \label{fig:disag-distributionshift}
\end{figure}

We report our observations in Fig~\ref{fig:disag-distributionshift}. The surprising phenomenon here is the fact that there {\em are} many pairs of source-target domains where GDE is
approximately satisfied despite the distribution shift. Especially, for pre-trained models, we find that with the exception of three pairs of (source, target) domains (namely $(\texttt{P},\texttt{C})$, $(\texttt{P},\texttt{S})$, $(\texttt{S},\texttt{P})$), the remaining 9 pairs of domains where the target differs from the source satisfy GDE approximately. The other notable observation is the fact that pre-trained models can satisfy GDE, and often better than randomly initialized models, under distribution shift. 
This is counter-intuitive, since we would expect pre-trained models to be strongly predisposed towards specific kinds of features, resulting in models that disagree rarely. Yet, pre-trained models do disagree non-trivially to a similar extent as they do with the ground truth.

 \graphicspath{{disagreement/}}

\chapter{Calibration Implies The Generalization-Disagreement Equality}

\label{chap:calibration}

Why is the disagreement rate unusually precise in being able to predict generalization? Why are the stochasticity
in data and the stochasticity in random seed both equally effective in being able to predict the performance on the network on unseen data? These phenomena suggest that there are strong patterns underlying the errors that neural networks make. Understanding the patterns may give us valuable insights into generalization in deep learning.\\

In this chapter, we provide a theoretical investigation of these questions. Informally stated, we prove that
%, by connecting disagreement rates to {\em calibration}. A well-calibrated model is a model with output probabilities that are neither ``over-confident'' nor ``under-confident'', Informally stated, we prove that
\begin{quote} for any stochastic learning algorithm, {\em if} the algorithm leads to a {\em well-calibrated} ensemble (the ensemble's output probabilities are neither over-confident nor under-confident), \textit{then} the ensemble satisfies the {Generalization Disagreement Equality} (GDE) {\em in expectation over the stochasticity}. 
\end{quote}

 Indeed, ensembles of networks learned from different stochastic runs of the training algorithm (e.g., across different random seeds) are well-known to be calibrated accurately in practice \citep{lakshmi17ensembles}, and thus our theory  offers a valuable insight into the practical generalization properties of deep networks. Our theory is also general as it makes no restrictions on the hypothesis class, the algorithm, the source of stochasticity, or the test distributions (which may be different from the training distribution). Overall, our work establishes a new connection between generalization and calibration. This connection has both theoretical and practical implications in understanding the generalization gap of deep networks, and in understanding the effect of stochasticity in SGD.

The results of this chapter have been previously published in \citet{jiang21assessing}.

\section{Related work}

\paragraph{Calibration.} Calibration of a statistical model is the property that the probability obtained by the model reflects the true likelihood of the ground truth \citep{murphy1967verification, dawid1982well}. A well-calibrated model provides an accurate confidence on its prediction which is paramount for high-stake decision making and interpretability. In the context of deep learning, several works \citep{guo2017calibration, lakshmi17ensembles, fort2019deep, wu2021should, bai2021don, mukhoti2021deterministic} have found that while individual neural networks are usually over-confident about their predictions, ensembles of several independently and stochastically trained models tend to be naturally well-calibrated. In particular, there are two types of ensembles that have typically been studied in literature: (a) ensembles where each member is trained by independently sampling training data (with replacement) from a particular pool of data, also called as {\em bagging} \citep{breiman96bagging} and (b) ensembles where each member is trained on the same pool of data, but with different random seeds (e.g., different random initialization and data ordering), also called as {\em deep ensembles} \citep{lakshmi17ensembles}. It is worth noting that deep ensembles typically achieve much better accuracy and calibration than bagging \citep{why20nixon}.

On the theoretical side, \citet{zhu20ensemble} have studied why deep ensembles outperform individual models in terms of accuracy. Work in calibration has studied different post-processing methods of calibration \citep{kumar19verified}, established relationships to confidence intervals \citep{gupta20distribution}, and derived upper bounds on calibration error either in terms of sample complexity or in terms of the accuracy of the model \citep{bai2021don,ji21early,liu19implicit,jung20moment,shabat20sample}. \\

The discussion in our works complements the above in multiple ways. First, most work within the machine learning literature focuses on {\em top-class calibration}, which is concerned only with the confidence level of the top predicted class for each point. The theory in our work, however, requires looking at the confidence level of the model aggregated over all the classes. We then empirically show that SGD ensembles are well-calibrated even in this class-aggregated sense. Furthermore, we carefully investigate what sources of stochasticity result in well-calibrated ensembles.
Finally, we provide an exact formal relationship between generalization and calibration via the notion of disagreement, which is fundamentally different from existing theoretical calibration bounds.

\paragraph{Feature calibration in \citetalias{nakkiran20distributional}.}   \citetalias{nakkiran20distributional} also identify an independent set of properties
they term as ``feature calibration''. While the standard notion of calibration can be intuitively thought of as a specific instantiation of feature calibration, the instantiations of feature calibration that are empirically studied in \citetalias{nakkiran20distributional} are significantly different from standard calibration. \citetalias{nakkiran20distributional} also argue that feature calibration and GDE can all be roughly generalized under an umbrella phenomenon called ``indistinguishability''. Nevertheless, they treat GDE and feature calibration as independent phenomena. Conversely, we show that calibration in the standard sense {\em implies} GDE.

\section{Theoretical setup}

We now formalize our main observation. In particular, we define ``the Generalization Disagreement Equality'' as the phenomenon that the test error equals the disagreement rate {\em in expectation over $h \sim \scrH_\calA$}. This phenomenon was formalized with slight differences as the {\em Agreement Property} in \citet{nakkiran20distributional},

\begin{definition}
\label{def:disagreement-property}
We say that the stochastic algorithm $\calA$ satisfies \textbf{the Generalization Disagreement Equality  (GDE)} on $\scrD$ if,
% \begin{center} \scalebox{0.9}{%
% $
% %\mathbb{E}_{h, h' \sim \scrH_\calA}[{\underbrace{\Esub{\scrD}{\mathbb{1}[h(X) \neq h'(X)]}}_{\text{disagreement rate}}}] =  \mathbb{E}_{h \sim \scrH_\calA}[{\underbrace{\mathbb{P}_{\scrD}[h(X) \ne Y]}_{\text{test error}}}].
% \mathbb{E}_{h, h' \sim \scrH_\calA}[\disag_\scrD(h,h')] =  \mathbb{E}_{h \sim \scrH_\calA}[\testerr_\scrD(h)].
% $} \end{center}
\begin{equation}
\mathbb{E}_{h, h' \sim \scrH_\calA}[\disag_\scrD(h,h')] =  \mathbb{E}_{h \sim \scrH_\calA}[\testerr_\scrD(h)].
\end{equation}
\end{definition}

Note that the definition in itself does not imply that the equality holds for each pair of $h, h'$ (which is what we observed empirically). However, for simplicity, we will stick to the above ``equality in expectation'' as it captures the essence of the underlying phenomenon while also being easier to analyze.
%In particular, the above equality can be rewritten purely in terms of the ensemble $\tilde{h}$ by interchanging the expectation terms. 
For example, for binary classification, both sides of the equality can be simplified in terms of the ensemble $\tilde{h}$ to get:
\begin{align}
{\Esub{\scrD}{2 \tilde{h}_0(X) \tilde{h}_1(X)}} =  \Esub{\scrD}{\tilde{h}_{1-y}(X)}.
\label{eq:simplified-dp}
 \end{align}

Therefore, for binary classification, explaining why GDE holds boils down to explaining why the ensemble satisfies the above equality.

\begin{remark} All our our results hold more generally for any probabilistic classifier $\tilde{h}$ that is not necessarily an ensemble. For example, if $\tilde{h}$ was an individual neural network whose predictions are given by softmax probabilities (rather than a one-hot vector), and if those softmax predictions are well-calibrated, then one can state that GDE holds for the neural network itself i.e., the disagreement rate between two independently sampled one-hot predictions from that network would equal the test error of the softmax predictions.
\end{remark}

\subsection{An incorrect explanation: The easy-hard model}

To motivate why proving GDE is technically non-trivial, as a warm-up, let us look at the most natural hypothesis that \citetalias{nakkiran20distributional} identify (and rule out).
 Imagine that all datapoints $(x,y)$ are one of two types: (a) the datapoint is so ``easy'' that w.p. 1 over $h \sim \scrH_\calA$, $h(x) = y$  (b) the datapoint is so ``hard'' that $h(x)$ corresponds to picking a label uniformly at random. In this case, both sides of Eq~\ref{eq:simplified-dp} can be simplified to the same quantity\footnote{Note that we will use the probability function $p(\cdot)$ rather than $\mathbb{P}$ since we will be dealing with joint distributions of continuous and discrete variables.}, $\frac{1}{2}p((X,Y) \text{ is hard})$.

\begin{figure}[t!]
     \centering
     \begin{minipage}{0.3\textwidth}
         \centering
         \includegraphics[width=\textwidth]{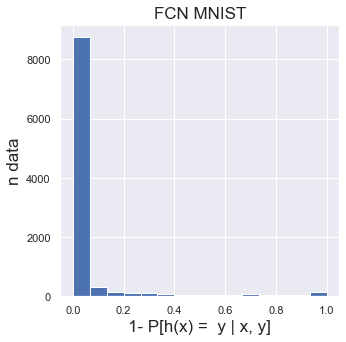}
        % \caption{Error distribution for theexperiment}
        %  \label{fig:cal-cali1}
        %  with \texttt{SameInit}
     \end{minipage}
     \hfill
     \begin{minipage}{0.3\textwidth}
         \centering
         \includegraphics[width=\textwidth]{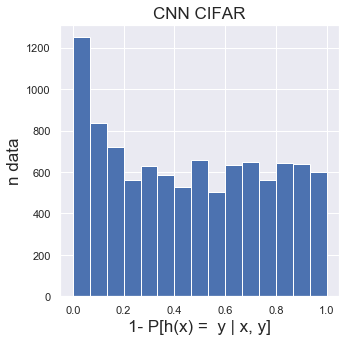}
        % \caption{Error distribution for the CIFAR10 CNN experiment}
        %  \label{fig:cal-cali2}
        %  with \texttt{AllDiff}
     \end{minipage}
     \hfill
     \begin{minipage}{0.3\textwidth}
         \centering
         \includegraphics[width=\textwidth]{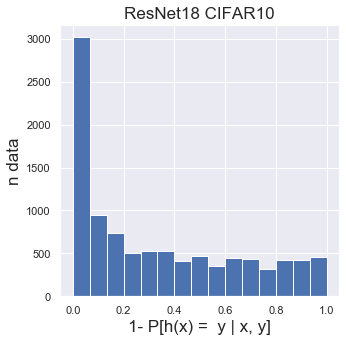}
        % \caption{Error distribution for the CIFAR10 ResNet18 experiment with \texttt{SameInit}}
        %  \label{fig:cal-cali2}
     \end{minipage}
    \caption{Error distributions of  MNIST+FCN, CIFAR10+CNN, CIFAR10+ResNet19.}
    \label{fig:cal-err_dist}
\end{figure}

 Unfortunately, \citetalias{nakkiran20distributional} also proceed to argue that the easy-hard condition is not true in practice. For completeness, we provide empirical results verify that this is indeed the case. In Fig~\ref{fig:cal-err_dist}, we show the error distribution\footnote{As a side note, we observe that all these error distributions can be fit well by a Beta distribution.
} of the ensemble similar to \citetalias{nakkiran20distributional}. The x-axis of these plots represent $1-\tildeh_y(X)$ in the context of our work. As \citetalias{nakkiran20distributional} note, these plots are {\em not} bimodally distributed on zero error and random-classification-level error (of $\frac{K-1}{K}$ where $K$ is the number of classes). This disproves the easy-hard hypothesis. \\

The main trouble with the above possible explanation is that for it to hold, GDE must hold for each individual point i.e., for each point $x$,  the disagreement rate in expectation over $h \sim \scrH_\calA$ (i.e., $2\tilde{h}_0(x) \tilde{h}_1(x)$) must equal the error in expectation over $h \sim \scrH_\calA$ (i.e., $\tilde{h}_{1-y}(x)$). But in practice, for a significant fraction of the points, the expected disagreement rate dominates the expected error rate, and for another significant fraction, the error rate dominates the expected disagreement rate. What is however surprising is that there is somehow a delicate balance between these two types of points such that overall these disparities cancel out each other giving rise to the equality in Equation~\ref{eq:simplified-dp}.  

What could create this delicate balance? We identify that this can arise from the fact that {\em the ensemble $\tilde{h}$ is well-calibrated}. 

\section{Class-wise calibration}

Informally, a well-calibrated model is one whose output probability for a particular class (i.e., the model's ``confidence'') is indicative of the probability that the ground truth class is indeed that class (i.e., the model's ``accuracy''). There are many ways in which calibration can be formalized. Below, we provide a particular formalism called as class-wise calibration.
%That is, for any given class, if the ensemble model assigns a probability of $q$ to that class, then it is indeed the case that with probability $q$, the true label is the given class \footnote{It is worth noting that the easy-hard setting is a specific case of calibrated setting.}.
%Formally,

\begin{definition}
\label{def:calibration}
The ensemble model $\tildeh$ 
satisfies \textbf{class-wise calibration}
%exhibits \textbf{well-calibration} 
on $\scrD$ if for any confidence value $q \in [0,1]$ and for any class $k \in [K]$,
\begin{equation}
    p ( Y = k \mid \tildeh_k(X) = q  ) = q.
\end{equation}
\end{definition}

Next, we show that if the ensemble is class-wise calibrated on the distribution $\scrD$, then GDE does hold on $\scrD$. Note however that shortly we show a more general result where even a weaker notion of calibration is sufficient to prove GDE. But since this stronger notion of calibration is  easier to understand, and the proof sketch for this captures the key intuition of the general case, we will focus on this first in detail. It is worth emphasizing that besides requiring well-calibration on the (test) distribution, all our theoretical results are general. We do not restrict the hypothesis class (it need not necessarily be neural networks), or the test/training distribution (they can be different) or where the stochasticity comes from (it need not necessarily come from the random seed or the data).

\begin{theorem}
\label{thm:main} 
Given a stochastic learning algorithm $\calA$, if its corresponding ensemble $\tildeh$ satisfies class-wise calibration on $\scrD$, then $\calA$ satisfies the Generalization Disagreement Equality on $\scrD$.
\end{theorem}
% The property doesn't hold pointwise, but rather holds in expectation in a given
% calibration level set or whatever
\begin{proof}(\textbf{Proof sketch for binary classification.} The actual proof follows as a corollary of a later more general Theorem~\ref{thm:agg-main}. See Section~\ref{sec:main_corrol}) 
Let $\scrD_q$ correspond to a ``confidence level set'' of the ensemble, in that it is the distribution of $X$ conditioned on $\tilde{h}_0(X) = q$. {\em Our key idea is to show that for a class-wise calibrated model, GDE holds
within each confidence level set} i.e., for each $q \in [0,1]$, the (expected) disagreement rate equals test error on $\scrD_q$. Since $\scrD$ is a combination of these level sets,  it automatically follows that GDE holds over $\scrD$. It is worth contrasting this proof idea with the easy-hard explanation which requires showing that GDE holds point-wise, rather than confidence-level-set-wise.\\

Now, let us calculate the disagreement on $\scrD_q$. For any fixed $x$ in the support of $\scrD_q$, the disagreement rate in expectation over $h,h' \sim \scrH_\calA$ corresponds to $q(1-q) + (1-q)q = 2q(1-q)$. This is nothing but the probability of the event that $h$ predicts $0$ and $h'$ predicts $1$ summed with the probability that the both predictions are reversed. Hence, the expected disagreement rate on $\scrD_q$ equals $2q(1-q)$. 

Next, we calculate the expected error of $h \sim \scrH_\calA$ on $\scrD_q$. At any $x$, the expected error equals $\tilde{h}_{1-y}(x)$. Now, from calibration, we have that exactly $q$ fraction of $\scrD_q$ has the true label $0$. On these points, the error rate is $\tilde{h}_{1}(x) = 1-q$. On the remaining $1-q$ fraction of $\scrD_q$, the true label is $1$, and hence the error rate on those is $\tilde{h}_{0}(x) = q$. The total error rate across both the class $0$ and class $1$ points is therefore $q(1-q) + (1-q)q = 2q(1-q)$. 
\end{proof}

\section[Class-aggregated calibration]{A more general result: class-aggregated calibration}

For an ensemble to be class-wise calibrated, we would require the equality in Definition~\ref{def:calibration} to hold for each class. When there are many classes however, like in the case of CIFAR-100, it is less likely that all the 100 corresponding equalities would hold well. To this end, in this section we will show that GDE holds under a more relaxed notion of calibration, which holds ``on average'' over the classes rather than individually for each class. Indeed, we demonstrate in a later section (see Section~\ref{sec:individual-class}) that this averaged notion of calibration holds more gracefully than class-wise calibration in practice. \\

Formally, we define {\em class-aggregated calibration} below. Recall that in class-wise calibration we look at the the conditional probability $\frac{p(Y=k, \tilde{h}_k(X)=q)}{p(\tilde{h}_k(X)=q)}$ for each $k$. Here, we will take an average of these conditional probabilities by weighting the $k^{th}$ conditional probability by $p(\tilde{h}_k(X)=q)$.  The result is the following definition:

\begin{definition}
\label{def:agg-calibration}
We say that the ensemble $\tilde{h}$ 
satisfies \textbf{class-aggregated calibration}
%is %\textbf{well-calibrated in aggregate} 
on $\scrD$ if for each $q \in [0,1]$,
\begin{equation}
    \frac{\sum_{k=0}^{K-1}  p(Y=k, \tilde{h}_k(X) = q)}{\sum_{k=0}^{K-1}  p(\tilde{h}_k(X) = q)} = q.
\end{equation}
\end{definition}

Intuitively, the denominator here corresponds to the proportion of points where {\em some} class gets confidence value $q$; the numerator corresponds to the proportion of points where some class gets confidence value $q$ {\em and} that class also happens to be the ground truth. 
Note however both the proportions involve counting a point $x$ multiple times if $\tilde{h}_k(x)=q$ for multiple classes $k$. 
 %, hence the class-aggregated calibration is the weaker notion.
%compared to perfect calibration.
% This definition is also closely related to the definition of calibration is Sec 4.3 of \citet{nixon2019measuring}.

We now formally state that this weaker notion of calibration is sufficient to show GDE. 
We prove this in Section~\ref{sec:cal-proof}.
From from this result, Theorem~\ref{thm:main} automatically follows as corollary since class-wise calibration implies class-aggregated calibration.

\begin{theorem}
\label{thm:agg-main}
Given a stochastic learning algorithm $\calA$, if its corresponding ensemble $\tildeh$ satisfies class-aggregated calibration on $\scrD$, then $\calA$ satisfies GDE on $\scrD$.
\end{theorem}

\paragraph{Comparison to existing notions of calibration.} Calibration in machine learning literature \citep{guo2017calibration,nixon2019measuring} is often concerned only with the confidence level of the top predicted class for each point. While top-class calibration is weaker than class-wise calibration, it is neither stronger nor weaker than class-aggregated calibration. 
 Class-wise calibration is a notion of calibration that has appeared originally under different names in \citet{zadrozny01obtaining,wu2021should}. On the other hand, the only closest existing notion to class-aggregated calibration seems to be that of {\em static calibration} in \citet{nixon2019measuring}, where it is only indirectly defined. Another existing notion of calibration for the multi-class setting  is that of {\em strong calibration} \citep{vaicenavicius19evaluation,widmann19calibration} which evaluates the accuracy of the model conditioned on $\tilde{h}(X)$ taking a particular value in the $K$-simplex. This is significantly stronger than class-wise calibration since this would require about $\exp(K)$ many equalities to hold rather than just the $K$ equalities in Definition~\ref{def:calibration}.

\subsection{Calibration is sufficient but not necessary for GDE} 

Theorem \ref{thm:agg-main} shows that calibration implies GDE. Below, we show that the converse is not true. That is, if the ensemble satisfies GDE, it is not necessarily the case that it satisfies class-aggregated calibration. 
This means that calibration and GDE are not equivalent 
phenomena, but rather only that calibration may lead to the latter.

\begin{proposition}
For a stochastic algorithm $\calA$ to satisfy GDE, it is not necessary that its corresponding ensemble $\tilde{h}$ satisfies class-aggregated calibration.
\end{proposition}

\begin{proof}
Consider an example where $\tilde{h}$ assigns a probability of either $0.1$ or $0.2$ to class $0$.  In particular, assume that with $0.5$ probability over the draws of  $(x,y) \sim \scrD$, $\tilde{h}_0(x)=0.1$ and with $0.5$ probability, $\tilde{h}_0(x)=0.2$. The expected disagreement rate (\texttt{EDR}) of this classifier is given by $\Esub{\scrD}{2\tilde{h}_0(x)\tilde{h}_1(x)} = 2\left(\frac{0.1 \cdot 0.9 +  0.2 \cdot 0.8}{2} \right)= 0.25$.

Now, it can be verified that the binary classification setting, class-aggregated and class-wise calibration are identical. Therefore, letting $p( Y = 0 \mid \tilde{h}_0(X)=0.1) \coloneqq \epsilon_1$ and $p( Y = 0 \mid \tilde{h}_0(X)=0.2) \coloneqq \epsilon_2$, our goal is to show that it is possible for $\epsilon_1 \neq 0.1$ or $\epsilon_2 \neq 0.2$ and still  
have the expected test error (\texttt{ETE}) equal the \texttt{EDR} of $0.25$. Now, the \texttt{ETE}  on $\scrD$ conditioned on $\tilde{h}_0(x)=0.1$ is given by $(0.1 (1- \epsilon_1) + 0.9 \epsilon_1)$ and on $\tilde{h}_0(x)=0.2$ is given by $(0.2 (1- \epsilon_2) + 0.8 \epsilon_1)$. Thus, the \texttt{ETE} on $\scrD$ is given by $0.15 + 0.5(0.8 \epsilon_1 + 0.6 \epsilon_2)$. We want $0.15 + 0.5(0.8 \epsilon_1 + 0.6 \epsilon_2) = 0.25$ or in other words, $0.8 \epsilon_1 + 0.6 \epsilon_2 = 0.2$.

Observe that while $\epsilon_1 = 0.1$ and $\epsilon_2 = 0.2$ is one possible solution where $\tilde{h}$ would satisfy class-wise calibration/class-aggregated calibration, there are also infinitely many other solutions for this equality to hold (such as say $\epsilon_1 = 0.25$ and $\epsilon_2 = 0$) where calibration does not hold. Thus, class-aggregated/class-wise calibration is just one out of infinitely many possible ways in which $\tilde{h}$ could be configured to satisfy GDE. 
\end{proof}

\section{Proof of Theorem~\ref{thm:agg-main}}
\label{sec:cal-proof}

We will now prove Theorem~\ref{thm:agg-main} which states that if the ensemble $\tilde{h}$ satisfies class-aggregated calibration, then the expected test error equals the expected disagreement rate.

\begin{proof}
We'll first simplify the expected test error and then proceed to simplifying the expected disagreement rate to the same quantity. %Throughout the rest of the discussion we will use $p()$ to denote the probability density function corresponding to $\scrD$. 

\paragraph{Test Error} Recall that the expected test error (which we will denote as $\texttt{ETE}$ for short) corresponds to $\Esub{ \scrH_\calA}{p(h(X) \ne Y \mid h)}$.

\begin{align}
    \texttt{ETE}&:=  \Esub{ h \sim \scrH_\calA}{p(h(X) \ne Y \mid h)} \\
    &= \Esub{h \sim \scrH_\calA}{\Esub{(X,Y) \sim \scrD}{\mathbb{1}[h(X) \ne Y]}}     \\
    &= \Esub{(X,Y) \sim \scrD}{\Esub{ \scrH_\calA}{\mathbb{1}[ h(X) \ne Y]}} && \text{(exchanging expectations by Fubini's theorem)}\\
    &= \Esub{(X,Y) \sim \scrD}{1-\tildeh_Y(X)}.
\end{align}
For our further simplifications, we'll explicitly deal with integrals rather than expectations, so we get,
\begin{align}
    \texttt{ETE}& =  \sum_{k=0}^{K-1}\int_{x} (1-\tildeh_k(x)) p(X = x, Y = k)dx. \\
\intertext{We'll also introduce $\tilde{h}(X)$ as a r.v. as,}
    \texttt{ETE}& =  \int_{\vecq \in \Delta^K}\sum_{k=0}^{K-1}\int_{x} (1-\tildeh_k(x)) p(X = x, Y = k, \tilde{h}(X) = \vecq)dx d\vecq. \\
\intertext{Over the next few steps, we'll get rid of the integral over $x$. First, splitting the joint distribution over the three r.v.s by conditioning on the latter two,}
    \texttt{ETE} & = \int_{\vecq \in \Delta^K} \sum_{k=0}^{K-1} p(Y=k, \tilde{h}(X)=\vec{q}) \int_{x}  (1-\underbrace{\tildeh_k(x)}_{=q_k}) p(X=x \mid Y=k , \tilde{h}(X)=\vec{q}) dx d\vecq \\
     & = \int_{\vec{q}\in \Delta^K} \sum_{k=0}^{K-1}  p(Y=k, \tildeh(X) = \vecq) \int_{x} \underbrace{(1-q_k)}_{\text{constant w.r.t }\int_x} p(X = x \mid \tildeh(X) = \vecq, Y=k)dx d\vecq\\
     & = {\int_{\vec{q}\in \Delta^K} \sum_{k=0}^{K-1}}  p(Y=k, \tildeh(X) = \vecq) (1-q_k) 
\underbrace{\int_{x}  p(X = x \mid \tildeh(X) = \vecq, Y=k)dx}_{=1} d\vecq \\
 & = \underbrace{\int_{\vec{q}\in \Delta^K} \sum_{k=0}^{K-1}}_{\text{swap}}  p(Y=k, \tildeh(X) = \vecq) (1-q_k) d\vecq.  \\
 & =  \sum_{k=0}^{K-1} \int_{\vec{q}\in \Delta^K}  p(Y=k, \tildeh(X) = \vecq) (1-q_k) d\vecq. \\
\end{align}
In the next few steps, we'll simplify the integral over $\vecq$ by marginalizing over all but the $k$th dimension. First, we rewrite the joint distribution of $\tilde{h}(X)$ in terms of its $K$ components. For any $k$, let $\tilde{h}_{-k}(X)$ and $\vecq_{-k}$ denote the $K-1$ dimensions of both vectors excluding their $k$th dimension. Then, $\texttt{ETE} =$
    \begin{align}
&= \sum_{k=0}^{K-1}\int_{q_k} \int_{\vec{q}_{-k}}  p(\tildeh_{-k}(X) = \vecq_{-k} \mid Y=k, \tildeh_k(X) = q_k) \underbrace{p(Y=k, \tildeh_k(X) = q_k) (1-q_k)}_{\text{constant w.r.t } \int_{\vecq_{-k}}} d\vecq_{-k} dq_k \\
&= \sum_{k=0}^{K-1}\int_{q_k} p(Y=k, \tildeh_k(X) = q_k) (1-q_k) \underbrace{\int_{\vec{q}_{-k}}  p(\tildeh_{-k}(X) = \vecq_{-k} \mid Y=k, \tildeh_k(X) = q_k)  d\vecq_{-k}}_{=1} dq_k \\
&=  \sum_{k=0}^{K-1}\int_{q_k} p(Y=k, \tildeh_k(X) = q_k) (1-q_k) dq_k. 
\end{align}
 Rewriting $q_k$ as just $q$,
 \begin{align}
  \texttt{ETE} & =  \underbrace{\sum_{k=0}^{K-1} \int_{q \in [0,1]}}_{\text{swap}}  p(Y=k, \tildeh_k(X) = q) (1-q)  dq \\
  &=  \int_{q \in [0,1]}  \sum_{k=0}^{K-1} p(Y=k, \tildeh_k(X) = q) (1-q)  dq. \\
  \intertext{Finally, we have from the calibration in aggregate assumption that $\sum_{k=0}^{K-1} p(Y=k, \tildeh_k(X) = q) = q \sum_{k=0}^{K-1} p(\tildeh_k(X) = q)$ (Definition \ref{def:agg-calibration}). So, applying this, we get}
  &=  \int_{q \in [0,1]}  q \sum_{k=0}^{K-1} p(\tildeh_k(X) = q)  (1-q)  dq. \\
  \intertext{Rearranging,}
 \texttt{ETE} &=  \int_{q \in [0,1]}  q (1-q) \sum_{k=0}^{K-1} p(\tildeh_k(X) = q)    dq.  
    \label{eq:test-error-simplified}
\end{align}

\paragraph{Disagreement Rate} The expected disagreement rate (denoted by $\texttt{EDR}$ in short) is given by the probability that two i.i.d samples from $\tildeh$ disagree with each other over draws of input from $\scrD$, taken in expectation over draws from $\scrH_\calA$. That is,
\begin{align}
   \texttt{EDR} &\coloneqq \Esub{h, h' \sim \scrH_\calA}{p(h(X) \neq h'(X) \mid h, h')} \\
   &= \Esub{h, h' \sim \scrH_\calA}{\Esub{(X,Y) \sim \scrD}{\mathbb{1}[h(X) \neq h'(X)]}} \\
   &= \Esub{(X,Y) \sim \scrD}{\Esub{h, h' \sim \scrH_\calA}{\mathbb{1}[h(X) \neq h'(X)]}}.
\end{align}
In the last step, we have exchanged expectations by Fubini's Theorem.
Over the next few steps, we'll write this in terms of $\tilde{h}$ rather than $h$ and $h'$.
\begin{align}
\texttt{EDR}    = & \Esub{(X,Y) \sim \scrD}{\Esub{h, h' \sim \scrH_\calA}{ \sum_{k=0}^{K-1}\mathbb{1}[h(X) = k] \left( 1-  \mathbb{1}[h'(X) = k]\right)}}.  \\
\intertext{Swapping the expectation and the summation,}
\texttt{EDR}    = & \Esub{(X,Y) \sim \scrD}{\sum_{k=0}^{K-1}\Esub{h, h' \sim \scrH_\calA}{\mathbb{1}[h(X) = k] \left( 1-  \mathbb{1}[h'(X) = k]\right)}}. \\
\intertext{Since $h$ and $h'$ are independent samples from $\scrH_\calA$,}
 \texttt{EDR}    = & \Esub{(X,Y) \sim \scrD}{\sum_{k=0}^{K-1}{p(h(X) = k \mid X) \left( 1-  p(h'(X) = k \mid X)\right)}} \\
    = & \Esub{(X,Y) \sim \scrD}{\sum_{k=0}^{K-1}\tildeh_k(X)(1-\tildeh_k(X))}.
    \end{align}

From here, we'll deal with integrals instead of expectations.
\begin{align}
\texttt{EDR} = & \int_x \sum_{k=0}^{K-1}\tildeh_k(x)(1-\tildeh_k(x)) p(X=x) dx. \\
 \intertext{Let us introduce the random variable $\tilde{h}(X)$ as,}
 \texttt{EDR} = & \int_{\vecq \in \Delta^K} \int_x \sum_{k=0}^{K-1}\tildeh_k(x)(1-\tildeh_k(x)) p(X=x, \tilde{h}(X) = \vecq) dx d\vecq.
 \end{align}
 
In the next few steps, we'll get rid of the integral over $x$. First, we split the joint distribution as, 
 \begin{align}
\texttt{EDR} = & \int_{\vecq \in \Delta^K} p(\tilde{h}(X) = \vecq) \int_x \sum_{k=0}^{K-1}\underbrace{\tildeh_k(x)(1-\tildeh_k(x))}_{\text{apply } \tilde{h}_k(x)=q_k} p(X=x \mid \tilde{h}(X) = \vecq) dx d\vecq. \\
 = & \int_{\vecq \in \Delta^K} p(\tilde{h}(X) = \vecq) \int_x \underbrace{\sum_{k=0}^{K-1}}_{\text{bring to the front}} q_k(1-q_k) p(X=x \mid \tilde{h}(X) = \vecq) dx d\vecq. \\
 = & \sum_{k=0}^{K-1} \int_{\vecq \in \Delta^K} p(\tilde{h}(X) = \vecq) \int_x  \underbrace{q_k(1-q_k)}_{\text{constant w.r.t }\int_x} p(X=x \mid \tilde{h}(X) = \vecq) dx d\vecq. \\
  = & \sum_{k=0}^{K-1} \int_{\vecq \in \Delta^K} p(\tilde{h}(X) = \vecq)  q_k(1-q_k)\underbrace{ \int_x  p(X=x \mid \tilde{h}(X) = \vecq) dx}_{1} d\vecq. \\
    = & \sum_{k=0}^{K-1} \int_{\vecq \in \Delta^K} p(\tilde{h}(X) = \vecq)  q_k(1-q_k) d\vecq.
    \end{align}
    
Next, we'll simplify the integral over $\vecq$ by marginalizing over all but the $k$th dimension.
\begin{align}
 \texttt{EDR}= & \sum_{k=0}^{K-1} \int_{q_k}\int_{\vecq_{-k}}  p(\tilde{h}_{-k}(X) = \vecq_{-k} \mid \tilde{h}_k(X) = q_k)  \underbrace{p(\tilde{h}_k(X) = q_k)q_k(1-q_k)}_{\text{constant w.r.t. }\int_{\vecq_{-k}}} d\vecq_{-k} dq_{k} \\
 = & \sum_{k=0}^{K-1} \int_{q_k} p(\tilde{h}_k(X) = q_k)q_k(1-q_k) \underbrace{\int_{\vecq_{-k}}  p(\tilde{h}_{-k}(X) = \vecq_{-k} \mid \tilde{h}_k(X) = q_k)  d\vecq_{-k}}_{=1} dq_{k} \\
 =& \sum_{k=0}^{K-1} \int_{q_k} p(\tilde{h}_k(X) = q_k)q_k(1-q_k) dq_{k}. \\
\intertext{Rewriting $q_k$ as just $q$,}
 \texttt{EDR} = & \underbrace{\sum_{k=0}^{K-1} \int_{q \in [0,1]}}_{\text{swap}} p(\tilde{h}_k(X) = q) q(1-q) dq \\
  = & \int_{q \in [0,1]} q(1-q) \sum_{k=0}^{K-1}  p(\tilde{h}_k(X) = q)  dq.
\end{align}

This is indeed the same term as Eq~\ref{eq:test-error-simplified}, thus completing the proof.

\end{proof}

\subsection{Proof of Theorem~\ref{thm:main}}
\label{sec:main_corrol}
\begin{proof}
Observe that if $\tilde{h}$ satisfies the class-wise calibration condition as in Definition~\ref{def:calibration}, it must also satisfy class-aggregated calibration. Then, we can invoke Theorem~\ref{thm:agg-main} to claim that Disagreement Property holds.
\end{proof}

\section{Deviation from calibration}  
\label{sec:deviation}

In practice, an ensemble does not need to satisfy class-aggregated calibration precisely. How much can a deviation from calibration hurt GDE? To answer this question, we quantify calibration error as follows:

\begin{definition}
\label{def:deviation}
We define the \textbf{Class Aggregated Calibration Error} (CACE)
of an ensemble $\tilde{h}$ on $\scrD$ as 

\begin{equation}
\text{CACE}_{\scrD}(\tilde{h}) \coloneqq
  {\displaystyle\int_{q \in [0,1]}} \left\vert \frac{\sum_{k}  p(Y=k, \tilde{h}_k(X) = q)}{\sum_{k}  p(\tilde{h}_k(X) = q)} - q
 \right\vert \cdot \sum_{k}  p(\tilde{h}_k(X) = q) dq.
\end{equation}
\end{definition}

In other words, for each confidence value $q$, we look at the absolute difference between the left and right hand sides of Definition~\ref{def:agg-calibration}, and then weight the difference by the proportion of instances where a confidence value of $q$ is achieved. We integrate this over all possible values of $q$. It is worth keeping in mind that, while the absolute difference term lies in $[0,1]$, the weight terms alone would integrate to a value of $K$. Therefore, $\text{CACE}_{\scrD}(\tilde{h})$ can lie anywhere in the range $[0,K]$.

Using the definition of CACE, we show below that GDE holds approximately when the calibration error is low. 
This result is a further generalized version of Theorem~\ref{thm:agg-main} since
when $\tilde{h}$ satisfies class-aggregated calibration on $\scrD$, $\text{CACE}_{\scrD}(\tilde{h})$ will be  zero, therefore recovering Theorem~\ref{thm:agg-main}.

\begin{theorem}
\label{thm:deviation}
For any stochastic learning algorithm $\calA$:
\begin{align*}
   \left\vert \mathbb{E}_{h, h' \sim \scrH_\calA}[\disag_\scrD(h,h')] - \mathbb{E}_{h \sim \scrH_\calA}[\testerr_\scrD(h)] \right| \leq \text{CACE}_{\scrD}(\tilde{h}).
\end{align*}

\end{theorem}

%in Theorem \ref{thm:deviation} in  that GDE holds approximately when the calibration error is low. Concretely, we show that the expected disagreement error and` test error are within $\text{CACE}_{\scrD}(\tilde{h})$ of each other. 

Note that CACE is different from the ``expected calibration error (ECE)'' \citep{naeini15obtaining,guo2017calibration} commonly used in the machine learning literature, which applies only to  top-class calibration. As is the case with many other metrics of calibration, it is difficult to establish formal and succinct relation  between different metrics and ECE. Nonetheless, we show later in Table \ref{tab: calibration-sample-size} that CACE empirically follows the same trend  as ECE which suggests that in practice it is not too different from existing notions of calibration. \\

\begin{proof}\textbf{(Proof of Theorem~\ref{thm:deviation})}
Recall from the proof of Theorem~\ref{thm:agg-main} that the expected test error (\texttt{ETE}) satisfies:
\begin{align*}
    \texttt{ETE} &=  \int_{q \in [0,1]}    \underbrace{\sum_{k=0}^{K-1} p(Y=k, \tildeh_k(X) = q)}_{\text{subtract and add a $q \sum_{k=0}^{K-1} p( \tildeh_k(X) = q)$}}  (1-q)  dq \numberthis \\
     &=  \int_{q \in [0,1]}    \left(\sum_{k=0}^{K-1} {p(Y=k, \tildeh_k(X) = q) -q \sum_{k=0}^{K-1} p( \tildeh_k(X) = q)}\right)  (1-q)  dq \\
     & \;\;\; + \int_{q \in [0,1]}    q \sum_{k=0}^{K-1} p( \tildeh_k(X) = q) (1-q)  dq. \numberthis \\
\intertext{Recall that the second term on R.H.S is equal to the expected disagreement rate \texttt{EDR}. Therefore,}
\left|\texttt{ETE} - \texttt{EDR}\right| &=  \int_{q \in [0,1]}    \left(\sum_{k=0}^{K-1} {p(Y=k, \tildeh_k(X) = q) -q \sum_{k=0}^{K-1} p( \tildeh_k(X) = q)}\right)  (1-q)  dq. \numberthis\\
\intertext{Multiplying and dividing the inner term by $ \sum_{k=0}^{K-1} p( \tildeh_k(X) = q)$,}
\left|\texttt{ETE} - \texttt{EDR}\right| &=  \left|\int_{q \in [0,1]}    \left(\frac{\sum_{k=0}^{K-1} {p(Y=k, \tildeh_k(X) = q)}}{\sum_{k=0}^{K-1} p( \tildeh_k(X) = q)} -q\right) \sum_{k=0}^{K-1} p( \tildeh_k(X) = q) (1-q)  dq \right| \numberthis\\
 &\leq  \int_{q \in [0,1]}    \left|\frac{\sum_{k=0}^{K-1} {p(Y=k, \tildeh_k(X) = q)}}{\sum_{k=0}^{K-1} p( \tildeh_k(X) = q)} -q\right| \sum_{k=0}^{K-1} p( \tildeh_k(X) = q) \underbrace{(1-q)}_{\leq 1}  dq \numberthis\\
 &\leq  \int_{q \in [0,1]}    \left|\frac{\sum_{k=0}^{K-1} {p(Y=k, \tildeh_k(X) = q)}}{\sum_{k=0}^{K-1} p( \tildeh_k(X) = q)} -q\right| \sum_{k=0}^{K-1} p( \tildeh_k(X) = q)  dq \numberthis\\
& = \text{CACE}(\tilde{h}). \numberthis
\end{align*}

\end{proof}

Note that it is possible to consider a more refined definition of CACE that yields a tighter bound on the gap.  In particular, in the last series of equations, we can leave the $1-q$ as it is, without upper bounding by $1$. In practice, this tightens CACE by upto a value of $2$. We however avoid considering the refined definition as it is less intuitive as an error metric.

\section{Empirical analysis of calibration}

As stated in the introduction, it is a well-established observation that ensembles of SGD trained models provide good confidence estimates \citep{lakshmi17ensembles}. However, typically the output of these ensembles correspond to the average {\em softmax probabilities} of the individual models, rather than an average of the top-class predictions. Our theory is however based upon the latter type of ensembles. Furthermore, there exists many different evaluation metrics for calibration in literature, while we are particularly interested in the precise definition we have in Definition~\ref{def:agg-calibration}. We report our observations keeping these requirements in mind.

\subsection{Experimental details}

%For ensembles, unless specified otherwise, we use the combination of the first option for each hyperparameters in Section \ref{sec:exp_details}.
For every ensemble experiment, we train a standard ResNet 18 model (width multiplier $1\times$, initial learning rate $0.1$, weight decay $0.0001$, minibatch size $200$ and no data augmentation). Below, we discuss how we empirically compute CACE.

\subsubsection{Finite-Sample Approximation of CACE}
\label{sec:ace_approx}

To estimate the calibration, we use the testset $\calD_{test}$. We split $[0,1]$ into $10$ equally sized bins. For a class $k$, we can group all $(x,y) \in \calD_{test}$ into different bins $\calB_{i}^k$ according to $\tildeh_k(x)$ (all bins have boundaries that do not overlap with other bins). In total, there are $10 \times K$ bins.
\begin{align}
    \calB_{i}^k = \{(x, y) \mid \text{lower}(\calB_{i}^k) \leq \tildeh_k(x) < \text{upper}(\calB_{i}^k) \,\, \text{and}\,\, (x,y) \in \calD_{test}\}
\end{align}
Where \textit{upper} and \textit{lower} are the boundaries of the bin.
% Then the accuracy of each bin is computed as $ \text{acc}(\calB^k_i) = |\calB^k_i|^{-1}\sum_{(x,y) \in \calB^k_i} \mathbb{1}[y=k]$. 
To mitigate the effect of insufficient samples for some of the middling confidence value in the middle (e.g. $p=0.5$), we further aggregate the calibration accuracy over the classes into a single bin $\calB_i = \bigcup_{k=1}^K \calB_i^k$ in a weighted manner. Concretely, for each bin, we sum over all the classes when computing the accuracy:
\begin{align}
    \text{acc}(\calB_i) = \frac{1}{\sum_{k=1}^K |\calB^k_i|} \sum_{k=1}^K\sum_{(x,y) \in \calB^k_i} \mathbb{1}[y=k] = \frac{1}{|\calB_i|} \sum_{k=1}^K\sum_{(x,y) \in \calB^k_i} \mathbb{1}[y=k]
\end{align}

To quantify how ``far'' the ensemble is from the ideal calibration level, we use the Class Aggregated Calibration Error (CACE) which is an average of how much each bin deviates from $y=x$ weighted by the number of samples in the bin:

\begin{align}
    \widehat{CACE} =  \sum_{i=1}^{N_\calB} \frac{|\calB_i|}{|\calD_{test}|} \left| \text{acc}(\calB_i) - \text{conf}(\calB_i)  \right|
\end{align}
% \yiding{f}

where $N_\calB$ is number of bins (usually 10 unless specified otherwise), $\text{conf}(\calB_i)$ is the ideal confidence level of the bin, which we set to the average confidence of all data points in the bin. This is the sample-based approximation of Definition \ref{def:deviation}.

\subsubsection{Finite-Sample Approximation of ECE}
ECE is a widely used metric for measuing calibration of the top predicted class. For completeness, we will reproduce its approximation here. Let $\hat{Y}$ be the class with highest probability under $\tildeh$ (we are omitting the dependency on $X$ in the notation since it is clear):
\begin{align}
    \hat{Y} = \argmax {k \in [K]} \tildeh_k(X)
\end{align}
We once again split $[0,1]$ into $10$ equally sized bins but do not divide further into $K$ classes. Each bin is constructed as:
\begin{align}
    \calB_{i} = \{(x, y) \mid \text{lower}(\calB_{i}) \leq \tildeh_{{\hat{y}}}(x) < \text{upper}(\calB_{i}) \,\, \text{and}\,\, (x,y) \in \calD_{test}\}
\end{align}
With the same notation used for CACE, the accuracy is computed as:
\begin{align}
    \text{acc}(\calB_i) = \frac{1}{|\calB_i|} \sum_{(x,y) \in \calB_i} \mathbb{1}[y=\hat{y}] 
\end{align}
Finally, the approximation of ECE is computed as the following:
\begin{align}
    \widehat{ECE} =  \sum_{i=1}^{N_\calB} \frac{|\calB_i|}{|\calD_{test}|} \left| \text{acc}(\calB_i) - \text{conf}(\calB_i)  \right|
\end{align}

\subsection{Empirical evidence for theory.}

%\vaishnavh{Replace ``perfect calibration'' with top-class calibration.}

\begin{figure}[t]
     \centering
     \begin{minipage}{.23\textwidth}
         \centering
         \includegraphics[width=\textwidth]{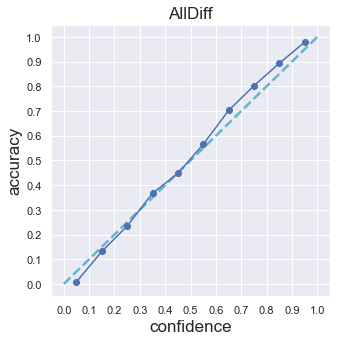}
        %  \caption{$\widehat{CACE} = 0.0045$}
        % \label{fig:cal-cali1}
     \end{minipage}
    %  \hfill
         \begin{minipage}{.23\textwidth}
         \centering
         \includegraphics[width=\textwidth]{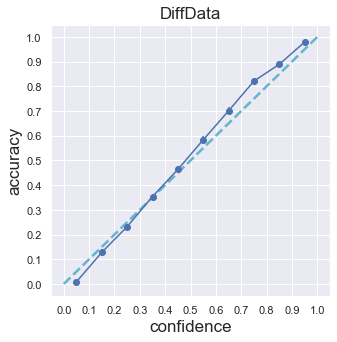}
        %  \caption{$\widehat{CACE} = 0.0049$}
        % \label{fig:cal-cali2}
     \end{minipage}
     \centering
     \begin{minipage}{.23\textwidth}
         \centering
         \includegraphics[width=\textwidth]{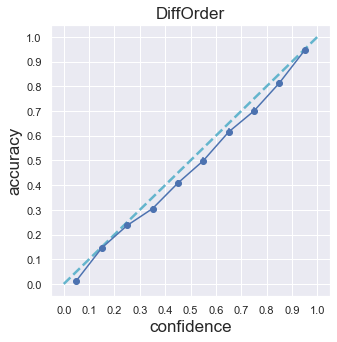}
        %  \caption{$\widehat{CACE} = 0.0079$}
        % \label{fig:cal-cali1}
     \end{minipage}
    %  \hfill
 \begin{minipage}{.23\textwidth}
         \centering
         \includegraphics[width=\textwidth]{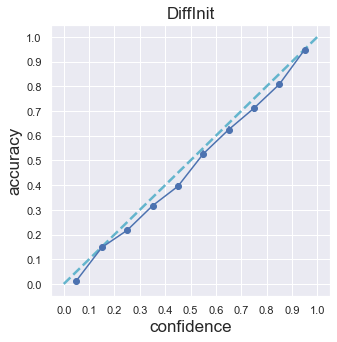}
        %  \caption{$\widehat{CACE} = 0.0071$}
       %  \label{fig:cal-cali2}
     \end{minipage}
    \caption{\textbf{Calibration on CIFAR10}: Calibration plot of different ensembles of 100 ResNet18 trained on CIFAR10. The error bar represents one bootstrapping standard deviation (most are extremely small). The estimated CACE for each scenario is shown in Table \ref{tab: calibration-sample-size}.}
        \label{fig:cal-calibration-cifar10}
        \bigskip
        %\vspace{-4mm}
%\end{figure}
%\begin{figure}[t]
     \centering
     \begin{minipage}{.23\textwidth}
         \centering
         \includegraphics[width=\textwidth]{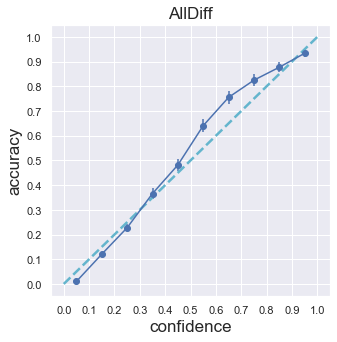}
     \end{minipage}
    %  \hfill
         \begin{minipage}{.23\textwidth}
         \centering
         \includegraphics[width=\textwidth]{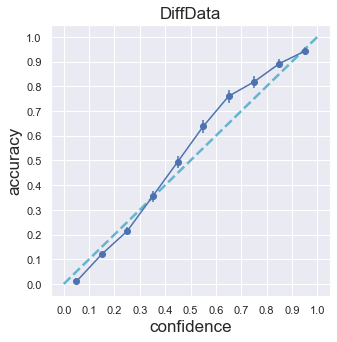}
     \end{minipage}
     \centering
     \begin{minipage}{.23\textwidth}
         \centering
         \includegraphics[width=\textwidth]{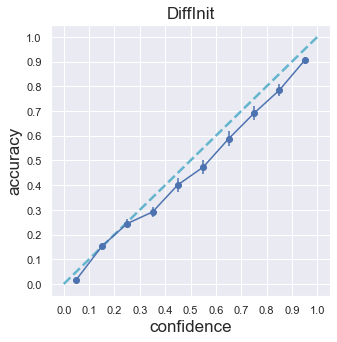}
     \end{minipage}
    %  \hfill
 \begin{minipage}{.23\textwidth}
         \centering
         \includegraphics[width=\textwidth]{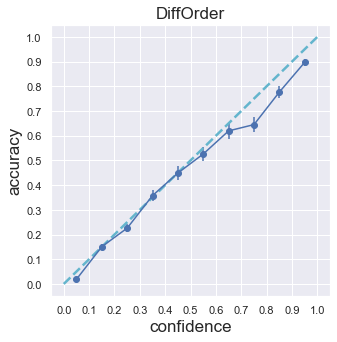}
     \end{minipage}
    \caption{\textbf{Calibration on 2k subset of CIFAR10}: Calibration plot of different ensembles of 100 ResNet18 trained on CIFAR10 with 2000 training points.} \label{fig:cal-calibration-cifar10-2k} 
       % \vspace{-4mm}
\end{figure}

\begin{figure}[t!]
     \centering
     \begin{minipage}{.23\textwidth}
         \centering
         \includegraphics[width=\textwidth]{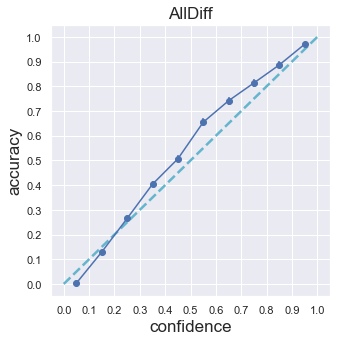}
     \end{minipage}
    %  \hfill
         \begin{minipage}{.23\textwidth}
         \centering
         \includegraphics[width=\textwidth]{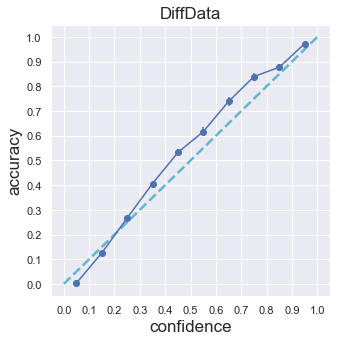}
     \end{minipage}
     \centering
     \begin{minipage}{.23\textwidth}
         \centering
         \includegraphics[width=\textwidth]{plots/2k_diff_init_calibration.png}
     \end{minipage}
    %  \hfill
 \begin{minipage}{.23\textwidth}
         \centering
         \includegraphics[width=\textwidth]{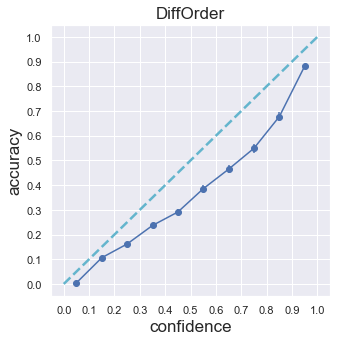}
     \end{minipage}
    \caption{\textbf{Calibration on CIFAR100}: Calibration plot of different ensembles of 100 ResNet18 trained on CIFAR100 with 10000 data points.}
        \label{fig:cal-calibration-cifar100}
\end{figure}

In Figure~\ref{fig:cal-calibration-cifar10},~\ref{fig:cal-calibration-cifar10-2k} and ~\ref{fig:cal-calibration-cifar100}, we show plots demonstrating that SGD ensembles \textit{do} nearly satisfy class-aggregated calibration {\em for all the sources of stochasticity we have considered}. In each plot, we report the conditional probability in the L.H.S of Definition~\ref{def:agg-calibration} along the $y$ axis and the confidence value $q$ along the $x$ axis. We observe that the plot closely follows the $x=y$ line. 

\begin{figure}[t!]
    \centering
    \includegraphics[scale=0.3]{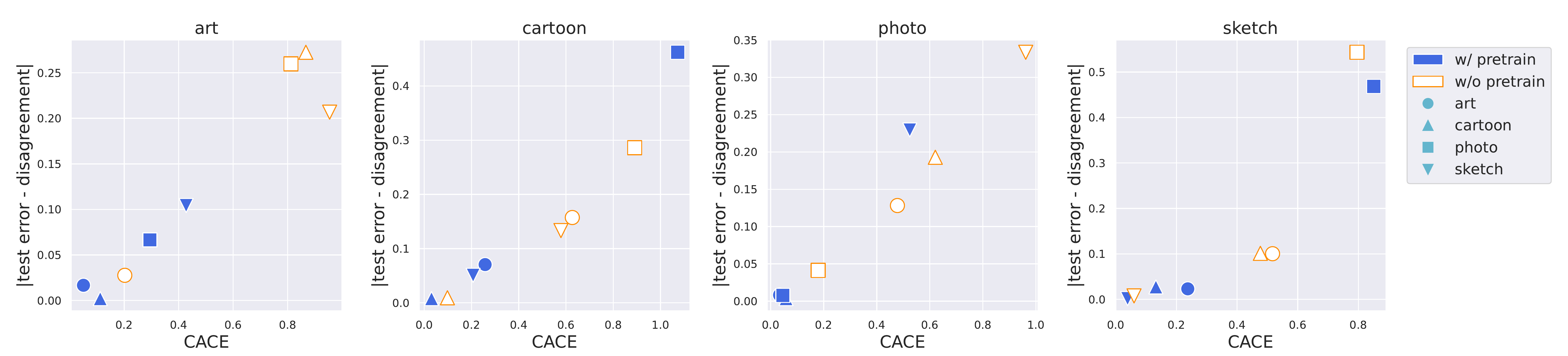}
    \caption{\textbf{Calibration error vs. deviation from GDE under distribution shift}: The scatter plots of CACE (x-axis) vs the gap between the test error and disagreement rate (y-axis) averaged over an ensemble of 10 ResNet50 models trained on PACS. Each plot corresponds to models evaluated on the domain specified in the title. The source/training domain is indicated by different marker shapes.}
    \label{fig:cal-distributionshift_cace}
\end{figure}

For a more precise quantification of how well calibration captures GDE, we also look at our notion of calibration error, namely CACE, which also acts as an upper bound on the difference between the test error and the disagreement rate. We estimate CACE over 100 models and report its values in Table \ref{tab: calibration-sample-size} (for CIFAR-10) and Table~\ref{tab: calibration-sample-size-cifar100} (for CIFAR-100). Most importantly, we observe that the CACE across different stochasticity settings correlates with the actual gap between the test error and the disagreement rate. In particular, CACE for \texttt{AllDiff/DiffData} are about two to three times smaller than that for \texttt{DiffInit/DiffOrder}, paralleling the behavior of the gap between test error and disagreement in these settings.  Even in the case of distribution shifts in the PACS dataset, we observe in Fig~\ref{fig:cal-distributionshift_cace}, that CACE estimated from a 10-model ensemble correlates with deviation from GDE.

While CACE correlates well with deviation from GDE, we however note that magnitude-wise, it is about three to ten times larger than the actual gap. We believe there are a couple of reasons for this. First, the definition of CACE can be tightened by roughly a factor of $2$, although the refined definition is less intuitive (see discussion in Section ~\ref{sec:deviation}). Secondly, we suspect that if we estimated CACE over a larger ensemble of models, it could bring the value of CACE down, especially in the case of CIFAR-100 since there are many classes involved.

\begin{table}[t!]
\centering
\resizebox{\textwidth}{!}{%
\begin{tabular}{c|ccc|ccc|c}
\toprule
       & \textbf{Test Error} & \textbf{Disagreement} & \textbf{Gap}& $\textbf{CACE}^{(\textbf{100})}$ & $\textbf{CACE}^{(\textbf{5})}$ & $\textbf{CACE}^{(\textbf{2})}$  & \textbf{ECE}\\ \hline
\texttt{AllDiff} &  $0.336 \pm 0.015$ & $0.348\pm 0.015$ & 0.012& 0.0437 & 0.2064  & 0.4244  & 0.0197\\
\texttt{DiffData} & $0.341 \pm 0.020$ & $0.354 \pm 0.020$ & 0.013& 0.0491 & 0.2242 & 0.4411  & 0.0267\\
\texttt{DiffInit} &  $0.337\pm 0.017$ & $0.307 \pm 0.022$ & 0.030 & 0.0979 & 0.2776 & 0.4495  & 0.0360\\  
\texttt{DiffOrder} &  $0.335\pm 0.017$ & $0.302 \pm 0.020$ & 0.033&0.1014  & 0.2782 & 0.4594  & 0.0410\\
\bottomrule
\end{tabular}}
\vspace{1 em}
\caption{\textbf{Calibration error vs. deviation from GDE for a 10k subset of CIFAR10 for ResNet18 ensembles.} For calibration error, size of ensemble denoted in the superscript. Test Error, Disagreement statistics and ECE are averaged over 100 models.}
\label{tab: calibration-sample-size}
\end{table}

\begin{table}[t!]
\centering
\begin{tabular}{c|ccc|c|c}
\toprule
   &  \textbf{Test Error} & \textbf{Disagreement} & \textbf{Gap}   &$\textbf{CACE}^{(\textbf{100})}$ & \textbf{ECE}        \\ \hline
\texttt{AllDiff} &  $0.679 \pm 0.0098$ & $0.6947 \pm 0.0076$ & 0.0157  & 0.1300 & 0.0469 \\
\texttt{DiffData}  &  $0.682 \pm 0.011$ & $0.6976 \pm 0.0074$ & 0.015  &0.1354 &  0.0503  \\
\texttt{DiffInit}  &   $0.681\pm 0.010$ & $0.5945 \pm 0.0127$ & 0.0865 &0.3816 & 0.1400 \\  
\texttt{DiffOrder} &  $0.679\pm 0.0097$ & $0.588 \pm 0.0100$ &  0.091 &0.3926 & 0.1449  \\
\bottomrule
\end{tabular}
\vspace{1 em}
\caption{\textbf{Calibration error vs. deviation from GDE for CIFAR100:} Test error, disagreement rate, the gap between the two, and ECE and CACE for ResNet18 on CIFAR100 with 10k training examples computed over 100 models.}
\label{tab: calibration-sample-size-cifar100}
\end{table}

\begin{figure}[t!]
     \centering
     \begin{minipage}{0.3\textwidth}
         \centering
         \includegraphics[width=\textwidth]{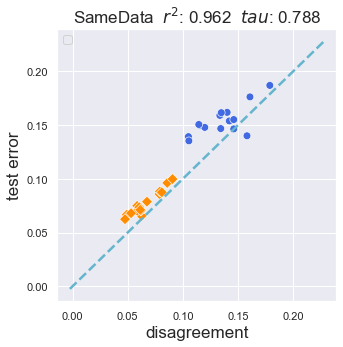}
         %\caption{Scatter plot}
         \label{fig:cal-cali1}
     \end{minipage}
    %  \hfill
     \begin{minipage}{0.3\textwidth}
         \centering
         \includegraphics[width=\textwidth]{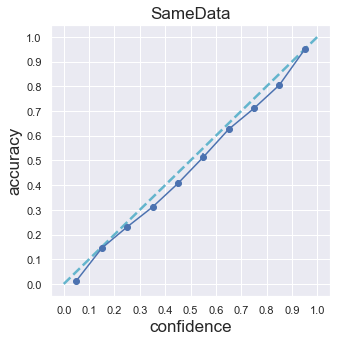}
         \label{fig:cal-cali2}
     \end{minipage}
    \caption{The scatter plot and calibration plot for model pairs that use  different initialization and different data ordering.}
    \label{fig:cal-same-data}
\end{figure}

\paragraph{Combining Stochasticity.} In Fig~\ref{fig:cal-same-data}, for the sake of completeness, we consider a setting where both the random initialization and the data ordering varies between two runs. We call this setting the \texttt{SameData} setting. We observe that this setting behaves similar to \texttt{DiffData} and \texttt{DiffInit}.

\begin{figure}[!t]
      \centering
     \begin{minipage}{0.23\textwidth}
         \centering
         \includegraphics[width=\textwidth]{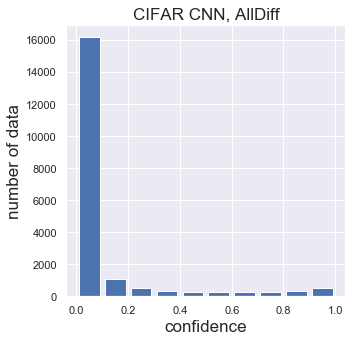}
         %\caption{CIFAR10 + CNN}
        %  \label{fig:cal-cali1}
     \end{minipage}
     \centering
     \begin{minipage}{0.23\textwidth}
         \centering
         \includegraphics[width=\textwidth]{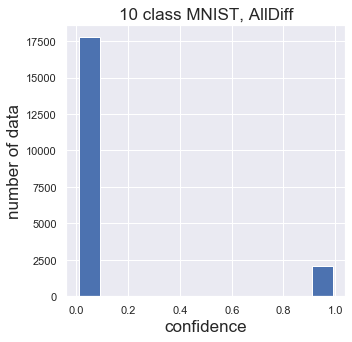}
         %\caption{MNIST + 2 layer FCN}
        %  \label{fig:cal-cali1}
     \end{minipage}
    %  \hfill
     \begin{minipage}{0.23\textwidth}
         \centering
         \includegraphics[width=\textwidth]{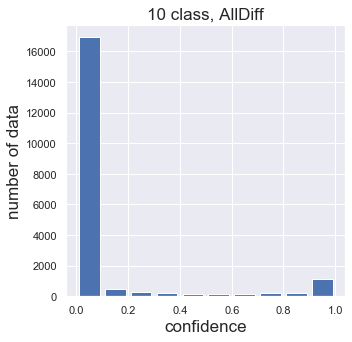}
         %\caption{Full Cifar10 + ResNet18}
        %  \label{fig:cal-cali2}
     \end{minipage}
    % \begin{minipage}{0.23\textwidth}
    %      \centering
    %      \includegraphics[width=\textwidth]{plots/cifar10_2class_calibration_hist.png}
    %      \caption{2 class Cifar10 + ResNet18}
    %     %  \label{fig:cal-cali2}
    %  \end{minipage}
    % \begin{minipage}{0.23\textwidth}
    %      \centering
    %      \includegraphics[width=\textwidth]{plots/cifar10_2class_samedata_calibration_hist.png}
    %      \caption{2 class Cifar10 + ResNet18}
    %     %  \label{fig:cal-cali2}
    %  \end{minipage}
    % \begin{minipage}{0.23\textwidth}
    %      \centering
    %      \includegraphics[width=\textwidth]{plots/cifar10_2class_sameorder_calibration_hist.png}
    %      \caption{2 class Cifar10 + ResNet18}
    %     %  \label{fig:cal-cali2}
    %  \end{minipage}
    \caption{Histogram of calibration confidence for CIFAR-10+CNN, MNIST+2-layer FCN, ResNet18+CIFAR-10.}
        \label{fig:cal-histogram}
\end{figure}

\paragraph{Calibration Confidence Histogram} For the sake of completeness, in Fig~\ref{fig:cal-histogram}, we report the number of points that fall into each bin in calibration plots. In other words, for each value of $p$, we report the number of times the ensemble $\tilde{h}$ satisfies $\tilde{h}_k(x) \approx p$ for some $k$ and some $x$.

\subsection{Class-wise Calibration vs Class-aggregated Calibration}
\label{sec:individual-class}
\begin{figure}[!t]
     \centering
    \begin{minipage}{0.23\textwidth}
         \centering
         \includegraphics[width=\textwidth]{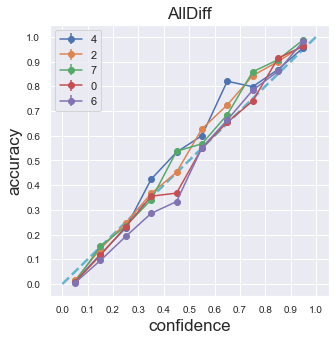}
         \caption{CIFAR10}
     \end{minipage}
    \begin{minipage}{0.23\textwidth}
         \centering
         \includegraphics[width=\textwidth]{plots/calibration_all_diff.png}
         \caption{CIFAR10's calibration plot}
     \end{minipage}
     \begin{minipage}{0.23\textwidth}
         \centering
         \includegraphics[width=\textwidth]{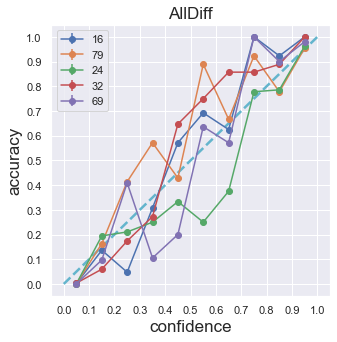}
         \caption{CIFAR100}
     \end{minipage}
          \begin{minipage}{0.23\textwidth}
         \centering
         \includegraphics[width=\textwidth]{plots/c100_all_diff_calibration.png}
         \caption{CIFAR100's calibration plot}
     \end{minipage}
    \caption{The calibration plot for 5 randomly selected individual classes vs the aggregated calibration plot for ResNet18 trained on CIFAR10 and CIFAR100.}
    \label{fig:cal-individual-class}
\end{figure}

In Fig~\ref{fig:cal-individual-class}, we report the calibration plots for a few random classes in the CIFAR10 and CIFAR100 setup and compare it with the class-aggregated calibration plots. We observe that the class-wise plots have a lot more variance, indicating that calibration within each class may not always be perfect. However, when aggregating across classes, calibration becomes much more well-behaved. This suggests that the calibration is smoothed over all the classes. It is worth noting that a similar effect also happens for ECE, although not reported here.

\section{Further remarks}
\paragraph{GDE for a single $(h,h')$ pair.} Our theory justifies the equality between disagreement and test error in expectation over $h, h'$ drawn from $\scrH_\calA$. But why does GDE hold for even individual $(h,h')$ pairs in practice? One possible justification for this could be the fact that 
even an ensemble of a handful of networks (say $5$ to $10$) is known to exhibit reasonable levels of calibration, as shown in \citet{lakshmi17ensembles}. However, if we look at the precise calibration levels in Table~\ref{tab: calibration-sample-size}, ensembles of 3 or 5 models result in CACE that are an order of magnitude larger than the 100-model counterpart. For an alternative justification, we could directly look at the test error and the disagreement rate and evaluate their standard deviation from their expectations. Indeed, we see in Tables~\ref{tab: calibration-sample-size} and ~\ref{tab: calibration-sample-size-cifar100} that both these quantities have negligible standard deviation. Future work could examine why these standard deviations (especially that of disagreement) are low.

\sloppy  \paragraph{The effect of different sources of stochasticity.}
%  While \texttt{DiffInit/Order} are slightly less calibrated (and also slightly deviate from GDE)
  Compared to \texttt{AllDiff/DiffData}, \texttt{DiffInit/Order} is still well-calibrated, with only slight deviations. 
  Why is varying the training data almost as effective in calibration as varying the random seed?
   One might propose the following natural hypothesis in the context of \texttt{DiffOrder} vs \texttt{DiffData}. In the first few steps of SGD, the data seen under two different reorderings are likely to not intersect at all, and hence the two trajectories would \emph{initially} behave as though being trained on two independent datasets. Further, if the first few steps largely determine the kind of minimum that the training falls into, then it is reasonable to expect that the stochasticity in data and in ordering both have the same effect on calibration. 
  
  However, this hypothesis falls apart
when we try to understand why two runs with the {\em same ordering} and {\em different initialization} (\texttt{DiffInit}) exhibits the same effect as \texttt{DiffData}. Indeed, \citet{fort2019deep} have empirically shown that such two such SGD runs explore diverse regions in the function space. 
Hence, we believe that there is a more nuanced reason behind why different types of stochasticity have a similar effect on ensemble calibration. One promising hypothesis for this could be the multi-view hypothesis from \citet{zhu20ensemble}. They theoretically show that   different random initializations could encourage the network to latch on to different predictive features of the image (even when exposed to the same training set), and thus result in ensembles with better test accuracy than the original model. Extending their study to understand similar effects on calibration would be a useful direction for future research.

% \begin{table}[t]
% \centering
% \begin{tabular}{c|cccc|ccc}
% \toprule
%       & $\textbf{CACE}^{(\textbf{2})}$ & $\textbf{CACE}^{(\textbf{3})}$ & $\textbf{CACE}^{(\textbf{5})}$ & $\textbf{CACE}^{(\textbf{100})}$ & \textbf{Test Error} & \textbf{Disagreement} & \textbf{ECE}\\ \hline
% \texttt{AllDiff} & 0.4244 & 0.3076 & 0.2064  &  0.0437 & $0.336 \pm 0.015$ & $0.348\pm 0.038$ & 0.0197\\
% \texttt{DiffData} & 0.4411 & 0.3121 & 0.2242 &  0.0491 & $0.341 \pm 0.020$ & $0.354 \pm 0.041$ & 0.0267\\
% \texttt{DiffInit} & 0.4495 & 0.3624 & 0.2776 &  0.0979 & $0.337\pm 0.017$ & $0.307 \pm 0.037$ & 0.0360\\  
% \texttt{DiffOrder} & 0.4594 & 0.3854 & 0.2782 &  0.1014 & $0.335\pm 0.017$ & $0.302 \pm 0.037$ & 0.0410\\
% \bottomrule
% \end{tabular}
% \vspace{1 em}
% \caption{Estimated CACE for ensembles with different number of models (denoted in the superscript) for ResNet18 on CIFAR10 with 10000 training examples. Test Error, Disagreement statistics and ECE are averaged over 100 models. Here ECE is the standard measure of top-class calibration error, provided for completeness.}
% \label{tab: calibration-sample-size}
% \end{table}

%%%%%%%%%%%%%%%%%%%%

\paragraph{Caveats.} While we believe our work provides a simple theoretical insight into how calibration leads to GDE, there are a few gaps that we do not address. First, we do not provide a theoretical characterization of \textit{when} we can expect good calibration (and hence, when we can expect GDE to hold). In our empirical results, we do see that for a variety of datasets and architectures, calibration/GDE hold approximately. But we also find other situations where this deteriorates or simply does not hold. This includes scenarios where there are certain kinds of distribution shift and/or low test accuracy, or as shown in Fig~\ref{fig:disag-other-datasets} in the previous chapter, for high accuracy FCNs trained on MNIST (while counter-intuitively, GDE holds for low accuracy FCNs in the same setting). It is important for future work in uncertainty estimation and calibration to develop a precise and exhaustive characterization of when calibration and GDE would hold.

Next, our theory sheds insight into why GDE holds in expectation over the stochasticity in training. However, we lack an explanation as to why the disagreement rate (and the test error) for a single pair of models lies close to this expectation. Finally, recall that our estimate of calibration error, CACE, provides an upper bound on how much the model would deviate from GDE. However, we observed that CACE is looser than the actual gap between the test error and disagreement rate. While this could simply be due to a lack of data/models in our estimation, it could also imply that our theory can be further refined.

\section{Conclusion}

 Building on \citet{nakkiran20distributional}, over the course of the last two chapters, we have observed that remarkably, two networks trained on the same dataset, tend to disagree with each other on unlabeled data nearly as much as they disagree with the ground truth. This gives us a surprisingly simple and accurate procedure for predicting the generalization performance of deep networks in practice. Besides, even when it comes to estimating out-of-distribution accuracy -- which is often the challenge in the real-world -- our technique shows promise.
Our empirical predictor of generalization is not a heuristic. We have also theoretically shown that it works well because of the fact that SGD ensembles are well-calibrated. 

  Broadly, these findings contribute to the larger pursuit of identifying and understanding empirical phenomena in deep learning.   Future work could shed light on why different sources of stochasticity surprisingly have a similar effect on calibration. Our work could also inspire other novel ways to leverage unlabeled data to estimate generalization. We also hope that our finding motivates a new cross-pollination of ideas between research in generalization and calibration.

\part{Conclusion}
\chapter{Summary}
\label{chap:summary}
%\addcontentsline{toc}{chapter}{\nameref{chap:conclusion}}

Understanding and predicting the generalization behavior of overparameterized models has become a central challenge in deep learning theory. In this dissertation, we took a winding journey towards this goal.  We began by taking the direction of empirically understanding implicit bias and we discovered that distance from initialization is one such strong form of bias. We then further pushed along this direction towards a uniform-convergence-based generalization bound that incorporates distance from initialization and more importantly, certain data-dependent notions of complexity. We arrived at a bound that is devoid of the exponential depth-dependence of existing bounds while also applying to the original network learned by SGD.

We then took a step back and realized that the broad direction pursued so far --- not just in this thesis, but also in most other existing works at that point --- could possibly lead us to a dead-end. In particular, we showed both simple linear counter-examples and neural network based counter-examples where uniform convergence bounds become vacuous despite the model's good generalization. We questioned whether ``simple models generalize well'' is the right way to think about generalization in deep learning, or even overparameterized models at large.

With this realization in mind, we veered around towards empirically predicting generalization by using unlabeled data. We proposed a technique based on disagreement that predicts generalization remarkably well, and we understood why it works. 
Overall, our estimate of the generalization gap does not fall into the conventional setup of training-data-based generalization bounds, leave alone the uniform convergence setup.

\chapter{The explanatory power of distribution-dependent bounds}

An arguably unconventional idea that we have barely scratched the surface of is that of using unlabeled (test) data to develop generalization bounds. This idea leads us to some interesting philosophical dilemmas about what it means to explain generalization. This is best laid out in the form of a conversation between {\em the Optimist}, who sees great promise in this idea in terms of its explanatory power, and {\em the Pessimist} who thinks otherwise.\\

\begin{dialogue}

\speak{The Optimist} I think the idea of developing {\em distribution-dependent} generalization bounds --- such as ones that depend on unlabeled test data --- can open doors to a wide variety of bounds we could have never thought of!

\speak{The Pessimist} Don't we already have well-developed distribution-dependent bounds? For example, unlike VC dimension bounds, margin-based bounds are not agnostic to the underlying distribution. They depend on the margin of the classifier on the data drawn from the distribution. 

\speak{The Optimist} Margin-based bounds depend on the distribution only via the training data. Let us call these bounds as {\em data}-dependent bounds. But what I am referring to are a class of bounds that depend on other information about the underlying distribution, not necessarily available through the training data. For example, say, information you can get from unlabeled test data.\\

\speak{The Pessimist} I appreciate the point in such bounds, but only to some extent. If one is a practitioner who wants to {\em predict} generalization, this idea would be promising. For example, they could use the technique from Chapter~\ref{chap:disagreement} to get a precise estimate for generalization.  But, if they really cared about {\em explaining} generalization, the idea seems questionable. A bound that uses unlabeled data can, at best, only {\em partially} explain generalization.

\speak{Optimist} Why do you say so?

\speak{Pessimist}  Existing theories of generalization --- at least within the ``indirect approach'' in Section~\ref{sec:indirect-approach} --- religiously adhere to the rule of using only the training data in deriving their generalization bounds. This choice comes from the (unwritten) philosophy that {\em in order to explain why the algorithm generalizes well, the explainer should have only as much information about the training procedure as the learner does}. There's a reason behind this philosophy: it guarantees that the bound does not ``cheat'' by being an empty hold-out bound in disguise. In a similar vein, any bound that uses unlabaled data that wasn't available to the learner seems suspect --- perhaps not as suspicious as a bound that uses extra {\em labeled} data, but somewhat suspicious.

\speak{Optimist} I agree that the idea of using ``extra information'' does seem to be contentious at first sight. But perhaps, the philosophy of not using extra information is overly cautious! I believe it is possible to develop an explanatory theory of generalization while also using extra information.

\speak{Pessimist} I am afraid no such theory exists!

\speak{Optimist} In fact, one doesn't need to look far to find such kinds of theory.  The direct approaches to bounding generalization (Section~\ref{sec:direct-approach})  indeed use extra information! These analyses assume a particular distribution beforehand, such as a Gaussian distribution with covariance $\vec{\Sigma}$, and liberally use that information to arrive at an almost-precise bound that would depend on many properties of $\vec{\Sigma}$, such as its rank. Neither the value of $\vec{\Sigma}$ nor any of its properties was privy to the learner. \\

\speak{Pessimist} It is not clear to me how explanatory those theories are either! What if, beneath all the layers of theorems and lemmas, the ``proofs'' are essentially doing what a computer does: compute the error on a hold-out dataset, perhaps not numerically, but analytically?

\speak{Optimist} Maybe, but there {\em can} be a significant difference. The direct-approach-based results can still yield {\em insights} that are general, at least to some extent. For example, they could tell us that the learner happens to generalize well because {\em fortunately}, the distribution being learned is ``nice'' --- for example, the underlying $\vec{\Sigma}$ satisfies some favorable properties which make the distribution easy to learn. It makes sense that these {\em fortuitous distributional properties} cause good generalization even if the learner was not aware of those properties in order to deliberately take advantage of them.
  Therefore, the explainer {\em can} have access to that information and still produce a valid explanation!\\

\speak{Pessimist} Indeed, that's a possibility I've not considered. But here's a possibility that you have not considered either: perhaps the niceness in the distribution {\em was} reflected in the training data, and the learner {\em had} to ``learn'' that fact that from the training data, so the learner did not just ``get lucky''. Then the explainer would have to explain why the learner was successful in learning that fact. Let me try to illustrate this in a few different ways:

\begin{enumerate}
	\item \textbf{Clustering example:} Imagine that the distribution is nice in that it consists of a few tight clusters, and each cluster corresponds to some class. Your explainer would say that ``fortunately, there is good ground-truth clustering, and because the learner was able to learn the class of each cluster, the learner was able to generalize''. However,  why was the learner able to successfully recover the underlying clustering given only finite data?

\item \textbf{Gaussian example:} Assume $\vec{\Sigma}$ satisfies some properties which give us some clue as to what the ground truth classifier is.  It's likely that these properties are also {\em approximately} satisfied by $\hat{\vec{\Sigma}}$, and hence the learner is able to use $\hat{\vec{\Sigma}}$ cleverly to recover the ground truth classifier approximately. 
%If this was really the case, an explainer who tries to  utilize the niceness of $\vec{\Sigma}$  seems to be cheating by terming the generalization fortuitous. 
Why did the learner's deliberate attempt at adapting to $\hat{\vec{\Sigma}}$ also generalize to $\vec{\Sigma}$?

\item \textbf{Flat minima example:} From Chapter~\ref{chap:deterministic-pacbayes}, we know that deep learning seems to find solutions that lie in flat minima in the {\em training loss}. Importantly, these solutions also happen to be flat in the test loss. Can the explainer simply assume that the test loss minimum is flat by blaming it on some kind of niceness of the data distribution? Or should the explainer also try to infer the flatness of the test loss minimum from the flatness of the training loss minimum?
\end{enumerate}

\speak{Optimist} I agree --- if it was really the case that the learner inferred the niceness of the distribution via the training data, the explainer is indeed cheating if they blame the success on the niceness of the distribution. But I strongly suspect that {\em there are many forms of distributional niceness that are simply not inferrable from the training data}. I am not sure what they are yet, but for example, in the case of high-dimensional linear regression, one cannot tell anything about how the distribution behaves in the dimensions that are not spanned by the training data --- unless of course, they make certain distributional assumptions such as the Gaussian assumption. When these properties cannot be inferred, clearly, it cannot be inferred by the learner either. Despite that, if the explainer forcefully tries to infer such properties from the training data, they would end up with vacuous bounds. The only way out is to simply declare that the learner was fortunate enough to face a nice distribution.\\

\speak{Pessimist} But the Gaussian assumption seems restrictive. Insights such as the ones about the covariance matrix do not tell us much about why overparameterized models work well on many kinds of nice distributions, including real-world distributions like CIFAR-10 and ImageNet.

\speak{Optimist} That is right. To get there, perhaps, we could try to combine the best of both the indirect and direct approaches.  Indirect approaches provide us with many abstract tools, such as those based on uniform convergence, which allow us to derive bounds that apply generally to any distribution.  However, so far these bounds have been based only on {\em training-data-dependent} notions of complexity. Direct approaches, while they typically apply only to Gaussian distributions, tell us that it is okay to infer extra information about the distribution. So, a healthy combination could be to use abstract tools like uniform convergence to derive bounds for generic distributions, and to make sure that the analysis is based on {\em distribution-dependent} notions of complexity. 

\speak{Pessimist} How would we derive distribution-dependent notions of complexity without making Gaussian assumptions about the data?

\speak{Optimist} If we are clever enough to identify what structural assumptions CIFAR-10 satisfy, then we are done. Another way out could be to use information from extra labeled unlabeled data or even better, unlabeled data. Of course, we should be careful not to use that extra data to simply produce what is essentially a hold-out bound.\\

\speak{Pessimist} Okay, I understand that. To summarize, uniform convergence bounds based on distribution-dependent notions of complexity may be promising... But wait, didn't we see in Chapter~\ref{chap:unif-conv} that even distribution-dependent notions of uniform convergence bounds could yield only vacuous bounds in some learning tasks?

\speak{Optimist} Yes, that is correct. Distribution-dependent uniform convergence would fail in those examples {\em if it was applied on the ``whole model''}. But the chapter did make a conjecture in Section~\ref{sec:dl-conjecture} that the weights of an overparameterized model can be decomposed into two parts: a  ``simple'' model that primarily determines the output of the model, and a ``noisy'' model that only minorly changes the output. The conjecture was that we could bound the error of the simple model via uniform convergence and then extend it to the whole network by arguing that the noisy component is irrelevant. This application of uniform convergence on the simple model could be distribution-dependent. The insight this would give us is that the simple function generalizes well because the function and the underlying distribution are {\em in conjunction} simple enough.

\speak{Pessimist} How would we extend that distribution-dependent uniform convergence bound on the simple model to the original model?

\speak{Optimist} Here again, we might want a distribution-dependent analysis. For example, by having access to unlabeled data, we would be able to derive a tight Hoeffding-inequality-based bound on the difference between the error of the original model and the simple model, without having to rely on uniform convergence. This is just a rudimentary example, but there may be more insightful ways to do it. We could identify certain structural assumptions that are satisfied by CIFAR-10 that might help us tackle this step differently.

\speak{Pessimist} What is the insight that we would potentially get from the approach in this step?

\speak{Optimist} Through this step we want to explain why a part of the function learned by the learner does not affect its predictions on most inputs from the distribution. Instead of taking the above route, the explainer {\em could}  have tried to infer this from the training data. In a way, this is what the derandomized PAC-Bayesian bound did in Chapter~\ref{chap:deterministic-pacbayes}. The bound tried to generalize the noise-resilience of the network from training data to test data. 

But as we discussed before, it's likely that this is simply impossible to infer from the data alone! The above distribution-dependent approach gives us an alternative explanation. It tells us that the learner was fortunate enough to face a distribution where much of the weights it happened to learn do not play a significant role in its predictions. The learner was not aware of this, nor did it cleverly learn such weights. It was sheer luck!\\

\speak{Pessimist} To make sure I understood all of this, let me try to summarize what you've told me in my own words. Basically, there are two aspects to explaining generalization, one that is deliberate and one that is fortuitous:

\begin{enumerate}
 \item \textbf{Deliberate aspects}: We want to identify every nice property that the learner ``deliberately'' satisfies to bring about good generalization. This could be any bias induced by the training algorithm (both in the form of explicit and implicit capacity control), or any nice properties about the distribution that the learner inferred from the training data, and took advantage of. 
 \item \textbf{Fortuitous aspects}: There may also be nice properties about the distribution that the learner did not infer from the training data. We should convert these into distributional assumptions that fortunately hold for the learner to generalize well. These assumptions may feature in the bound as distribution-dependent notions of complexity. Hopefully these assumptions also hold in real-world distributions, so we get an insight into why deep learning generalizes well in practice.

\end{enumerate} 
\speak{Optimist} Yes, that's right! 

\speak{Pessimist} Taking off our ``explainer hat'' and wearing the ``practitioner hat'', is there any point to worrying about this categorization in practice?

\speak{Optimist} Yes. If we can understand deliberate generalization, we might get actionable insights into how to improve the learner, especially to take full advantage of niceness in the underlying distribution. On the other hand, there is not much we can do to improve fortuitious generalization since that is an immutable aspect of the learning task. Hence, it would be valuable to disentangle these two effects in the current generalization behavior of deep networks.

\speak{Pessimist} Here is a hypothetical list of properties that are satisfied by the ImageNet distribution. How do we know which ones to assume away for free as part of the fortuitous aspects of generaliation and which ones to tackle as deliberate aspects of generalization?

\speak{Optimist} That is a question best addressed by future work!

\end{dialogue}

%-
\appendix
\backmatter

%\renewcommand{\baselinestretch}{1.0}\normalsize

% By default \bibsection is \chapter*, but we really want this to show
% up in the table of contents and pdf bookmarks.
%\renewcommand{\bibsection}{\chapter{\bibname}}
%\newcommand{\bibpreamble}{This text goes between the ``Bibliography''
%  header and the actual list of references}
%\bibliographystyle{plainnat}
%\bibliography{refs} %your bib file
%\printbibliography

%\newcommand{\bibpreamble}{This text goes between the ``Bibliography''
%  header and the actual list of references}
\setcitestyle{numbers} % Doing this here will output a sorted list, while preserving author year citation in the main text.
\bibliographystyle{plainnat}
\bibliography{refs} %your bib file

\end{document}